\documentclass{article}

\PassOptionsToPackage{numbers, compress}{natbib}

\usepackage[preprint]{neurips_2025}

\usepackage[utf8]{inputenc} 
\usepackage[T1]{fontenc}    
\usepackage{hyperref}       
\usepackage{url}            
\usepackage{booktabs}       
\usepackage{amsfonts}       
\usepackage{nicefrac}       
\usepackage{microtype}      
\usepackage{xcolor}         

\usepackage{graphicx}
\usepackage{subcaption}
\usepackage{amsmath}
\usepackage{amssymb}
\usepackage{mathtools}
\usepackage{amsthm}
\usepackage{bm}              
\usepackage{bbm}             
\usepackage{colortbl}        
\usepackage{multirow}        
\usepackage{makecell}        
\usepackage{array}           
\usepackage{enumitem}        
\usepackage{pifont}          
\usepackage{fontawesome5}    
\usepackage{tikz}
\usetikzlibrary{shapes.geometric, arrows, positioning}
\usepackage{wrapfig}         

\usepackage{algorithm}       
\usepackage{algorithmic}     

\usepackage{amsmath,amsfonts,bm}

\def\figref#1{Fig.~\ref{#1}}
\def\Figref#1{Fig.~\ref{#1}}

\def\eqref#1{equation~\ref{#1}}

\def\tabref#1{Tab.~\ref{#1}}
\def\Tabref#1{Tab.~\ref{#1}}

\def\1{\bm{1}}

\def\rvepsilon{{\mathbf{\epsilon}}}

\def\vtheta{{\bm{\theta}}}

\def\mI{{\bm{I}}}

\DeclareMathAlphabet{\mathsfit}{\encodingdefault}{\sfdefault}{m}{sl}
\SetMathAlphabet{\mathsfit}{bold}{\encodingdefault}{\sfdefault}{bx}{n}

\def\gN{{\mathcal{N}}}

\newcommand{\Ls}{\mathcal{L}}
\newcommand{\R}{\mathbb{R}}

\usepackage[capitalize,noabbrev]{cleveref}

\theoremstyle{plain}

\theoremstyle{definition}

\theoremstyle{remark}

\usepackage[textsize=tiny]{todonotes}

\definecolor{best}{rgb}{1.0, 0.65, 0.65}
\definecolor{best2}{rgb}{1.0, 0.75, 0.75}
\definecolor{best3}{rgb}{1.0, 0.90, 0.90}
\definecolor{bestlora}{rgb}{1.0, 0.80, 0.80}

\newcommand{\TODO}[1][]{%
  \noindent\fcolorbox{red!80!black}{red!15}{%
    \parbox{\dimexpr\linewidth-2\fboxsep-2\fboxrule}{%
      \textcolor{red!80!black}{\textbf{[TODO]}}%
      \ifx&#1&\else\space\textit{#1}\fi%
    }%
  }\par\vspace{2pt}%
}

\newcommand{\teaserbox}[3]{%
    \begin{tikzpicture}
        \node[inner sep=0pt] (img) {\includegraphics[width=0.14\textwidth]{#1}};
        \draw[red, thick] ([xshift=#2, yshift=#3]img.center) ++(-0.3cm, -0.3cm) rectangle ++(0.6cm, 0.6cm);
    \end{tikzpicture}%
}

\title{TransNormal: Dense Visual Semantics for Diffusion-based Transparent Object Normal Estimation}

\author{%
  Mingwei Li$^{1,2}$ \quad
  Hehe Fan$^{1}$ \quad
  Yi Yang$^{1,\text{\faEnvelope}}$ \\
  $^{1}$Zhejiang University, Hangzhou, China \quad
  $^{2}$Zhongguancun Academy, Beijing, China \\[0.5ex]
  \texttt{\{mingweili, hehefan, yangyics\}@zju.edu.cn} \\[0.5ex]
  Project page: \href{https://longxiang-ai.github.io/TransNormal}{https://longxiang-ai.github.io/TransNormal}
}

\begin{document}

\maketitle
\begin{figure}[h!]
    \centering
    \setlength{\tabcolsep}{1pt}
    \renewcommand{\teaserbox}[3]{%
        \begin{tikzpicture}
            \node[inner sep=0pt] (img) {\includegraphics[width=0.14\columnwidth]{#1}};
            \draw[red, thick] ([xshift=#2, yshift=#3]img.center) ++(-0.3cm, -0.3cm) rectangle ++(0.6cm, 0.6cm);
        \end{tikzpicture}%
    }
    \begin{tabular}{@{}c@{\hspace{1pt}}c@{\hspace{1pt}}c@{\hspace{1pt}}c@{\hspace{1pt}}c@{\hspace{1pt}}c@{\hspace{1pt}}c@{}}
        \includegraphics[width=0.14\columnwidth]{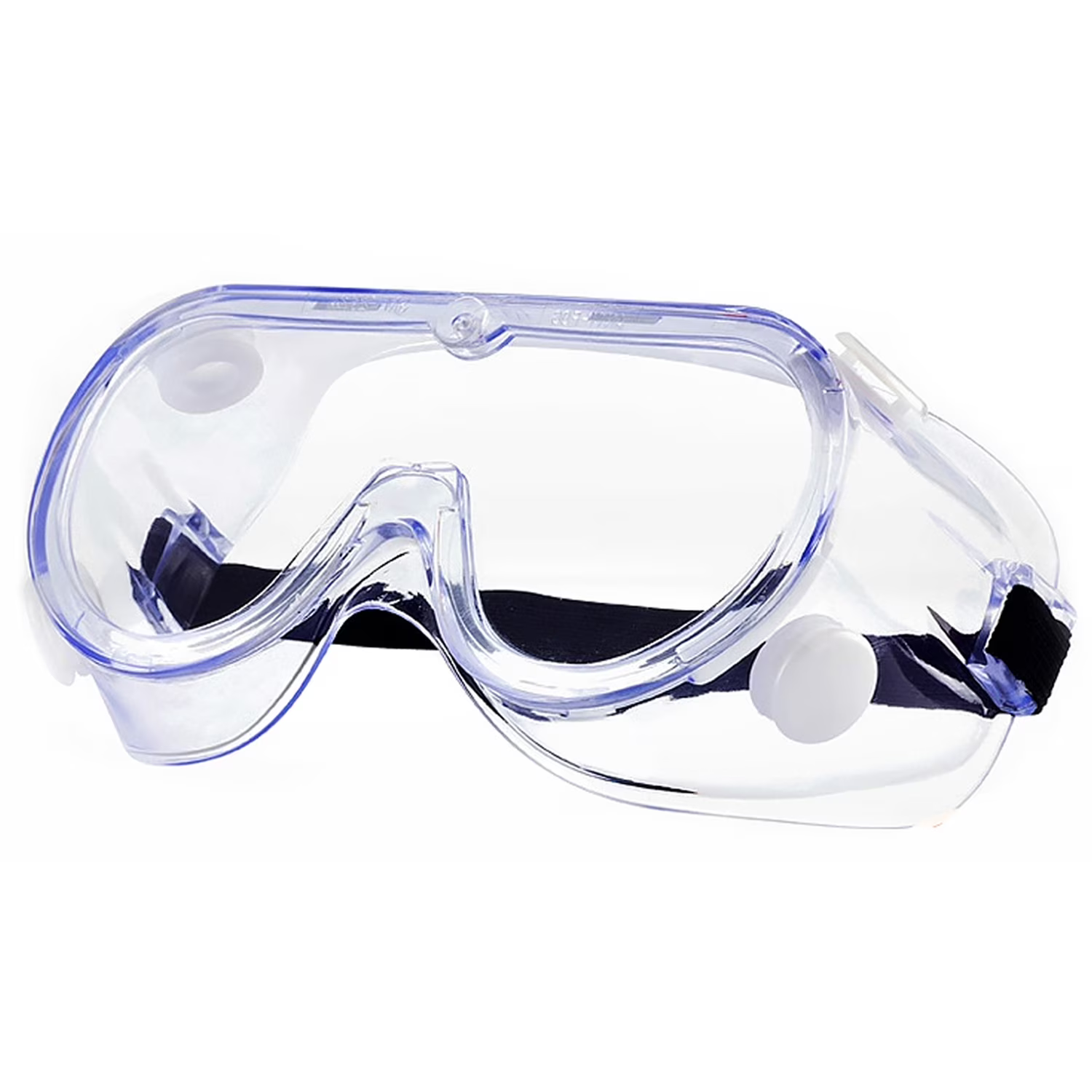} &
        \teaserbox{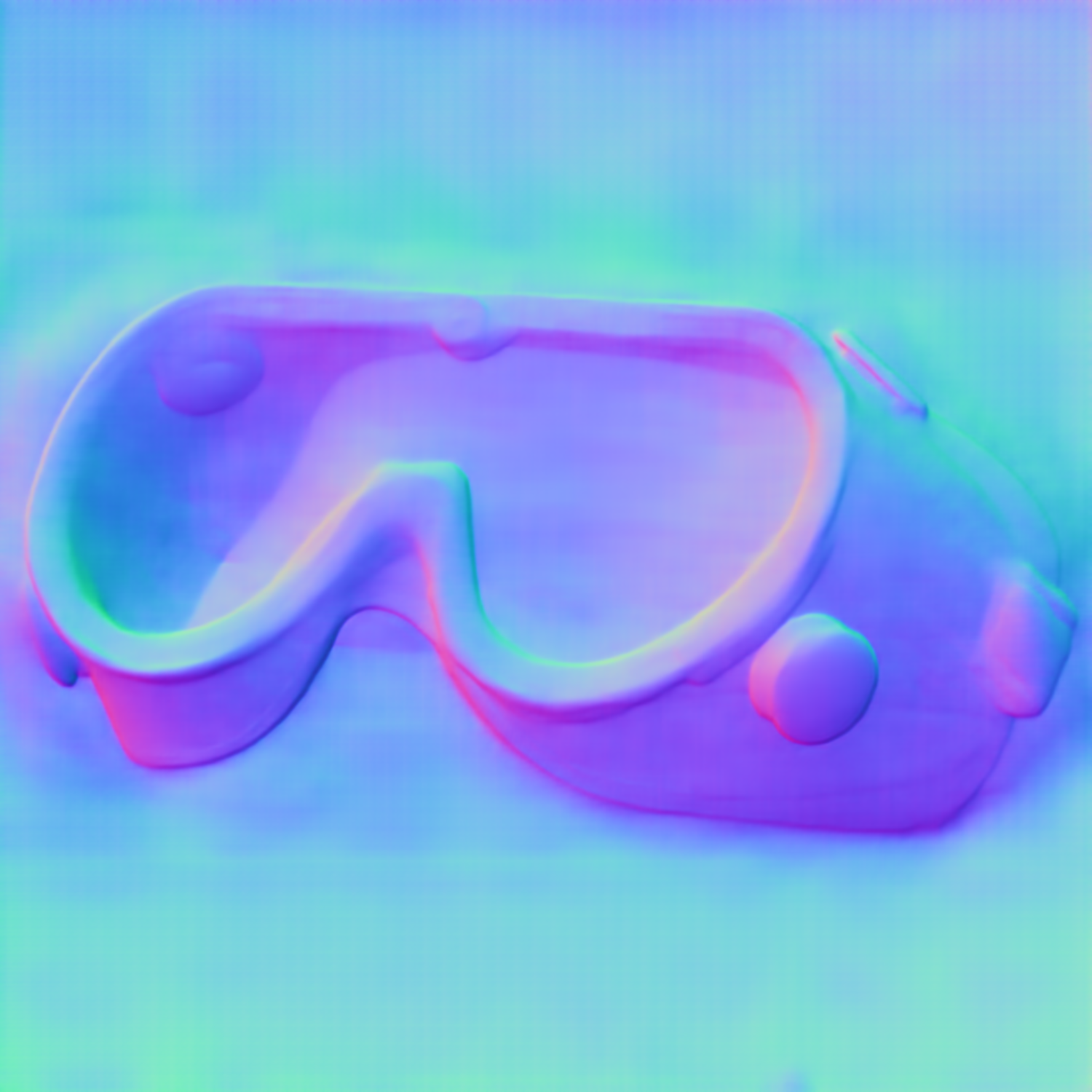}{-0.5cm}{0.3cm} &
        \teaserbox{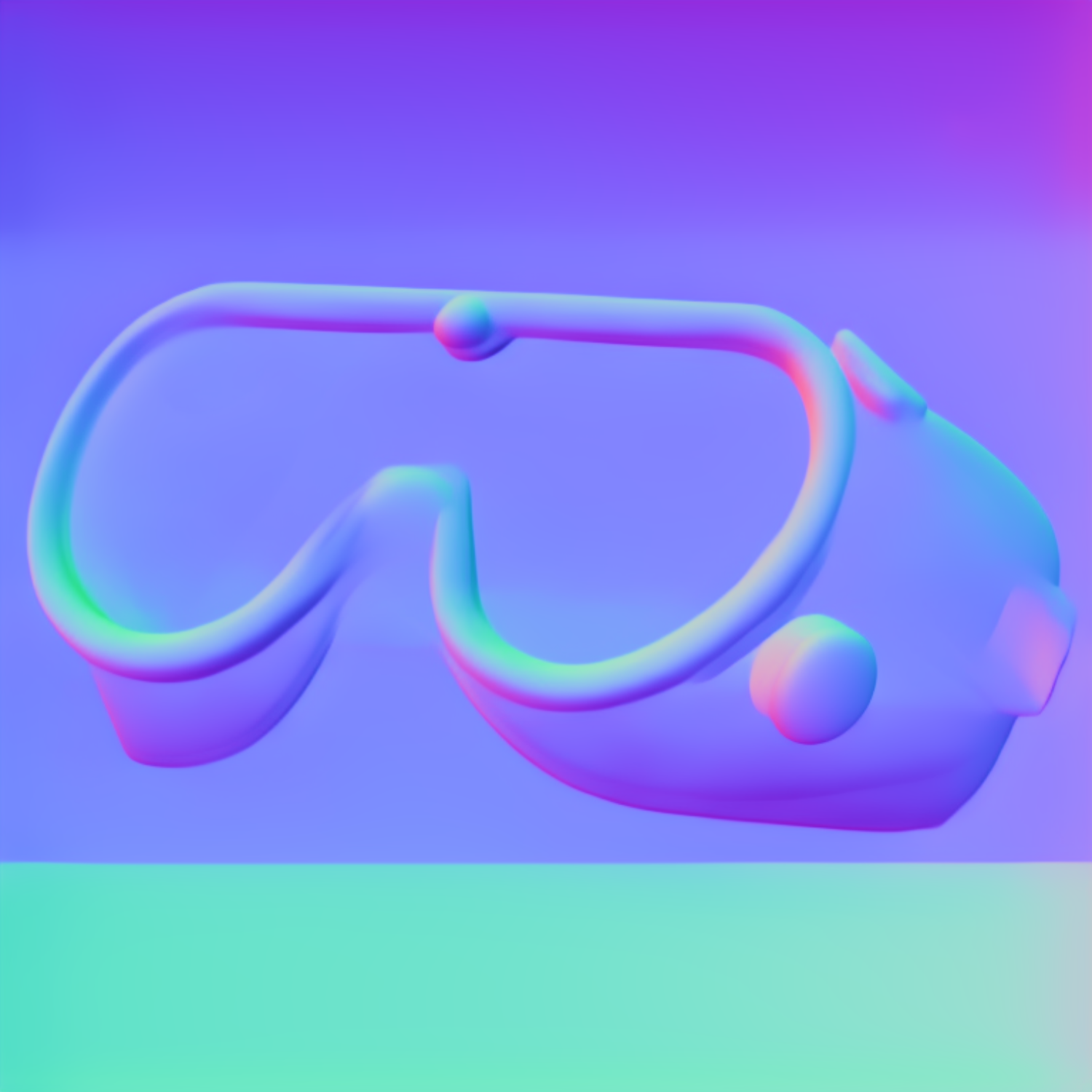}{-0.5cm}{0.3cm} &
        \teaserbox{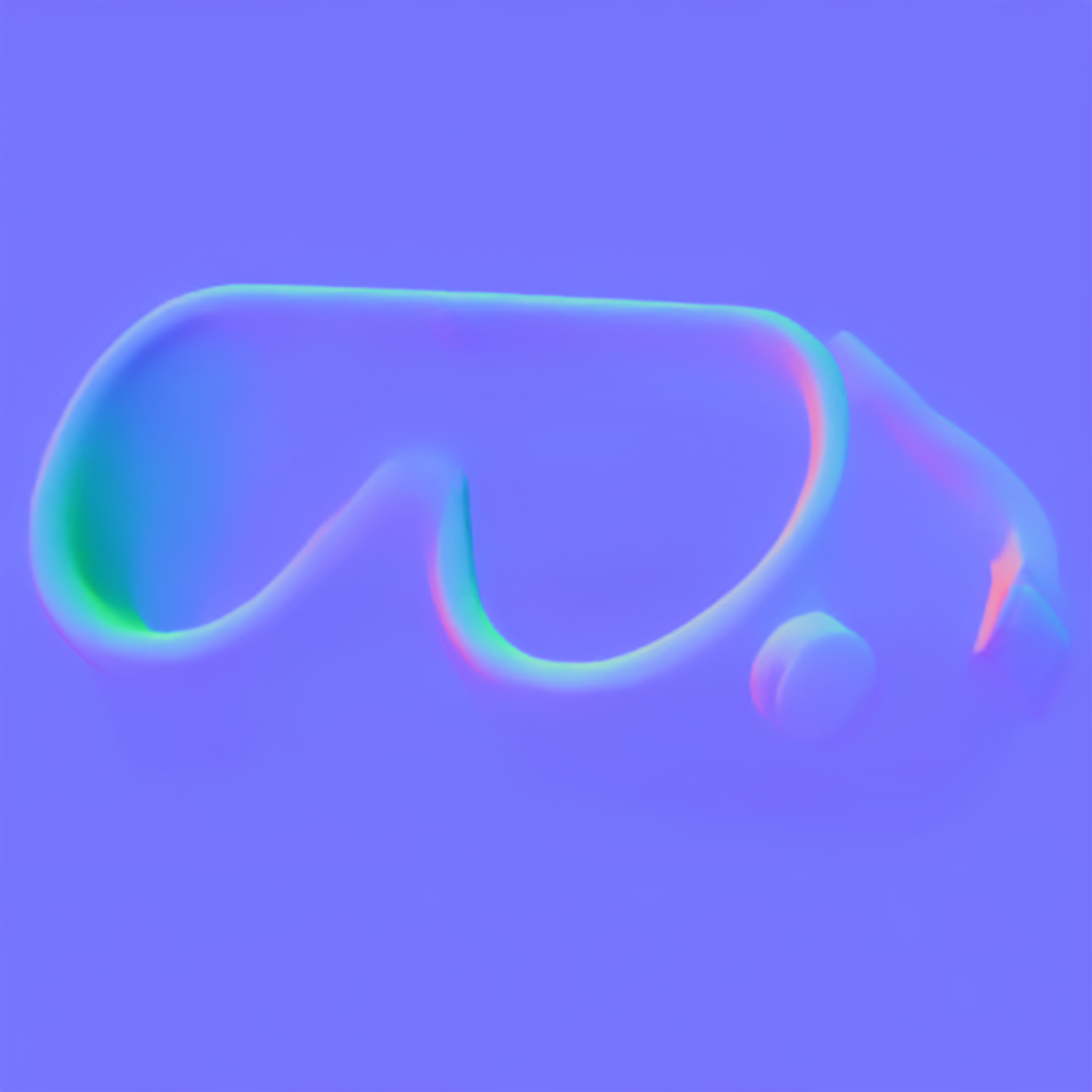}{-0.5cm}{0.3cm} &
        \teaserbox{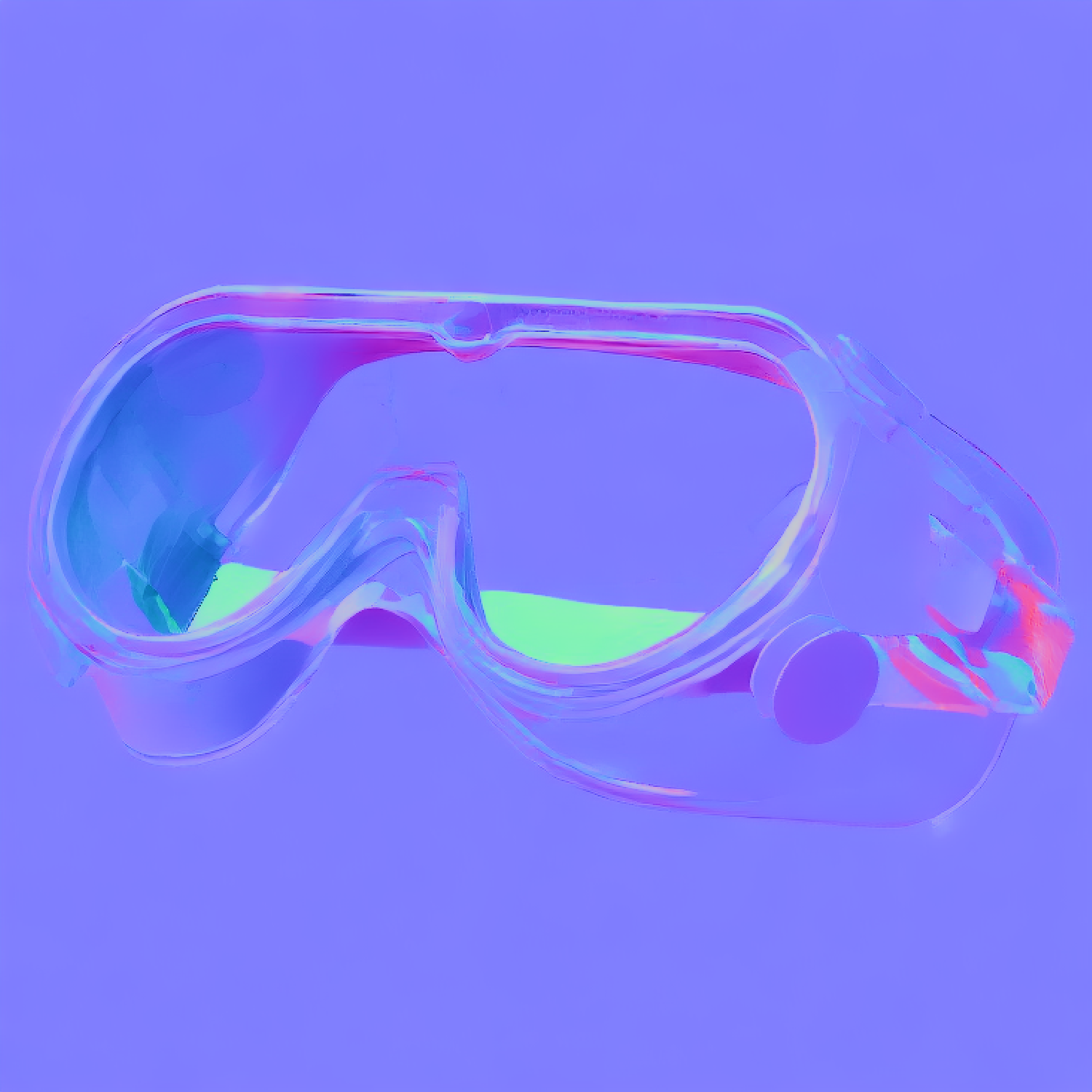}{-0.5cm}{0.3cm} &
        \teaserbox{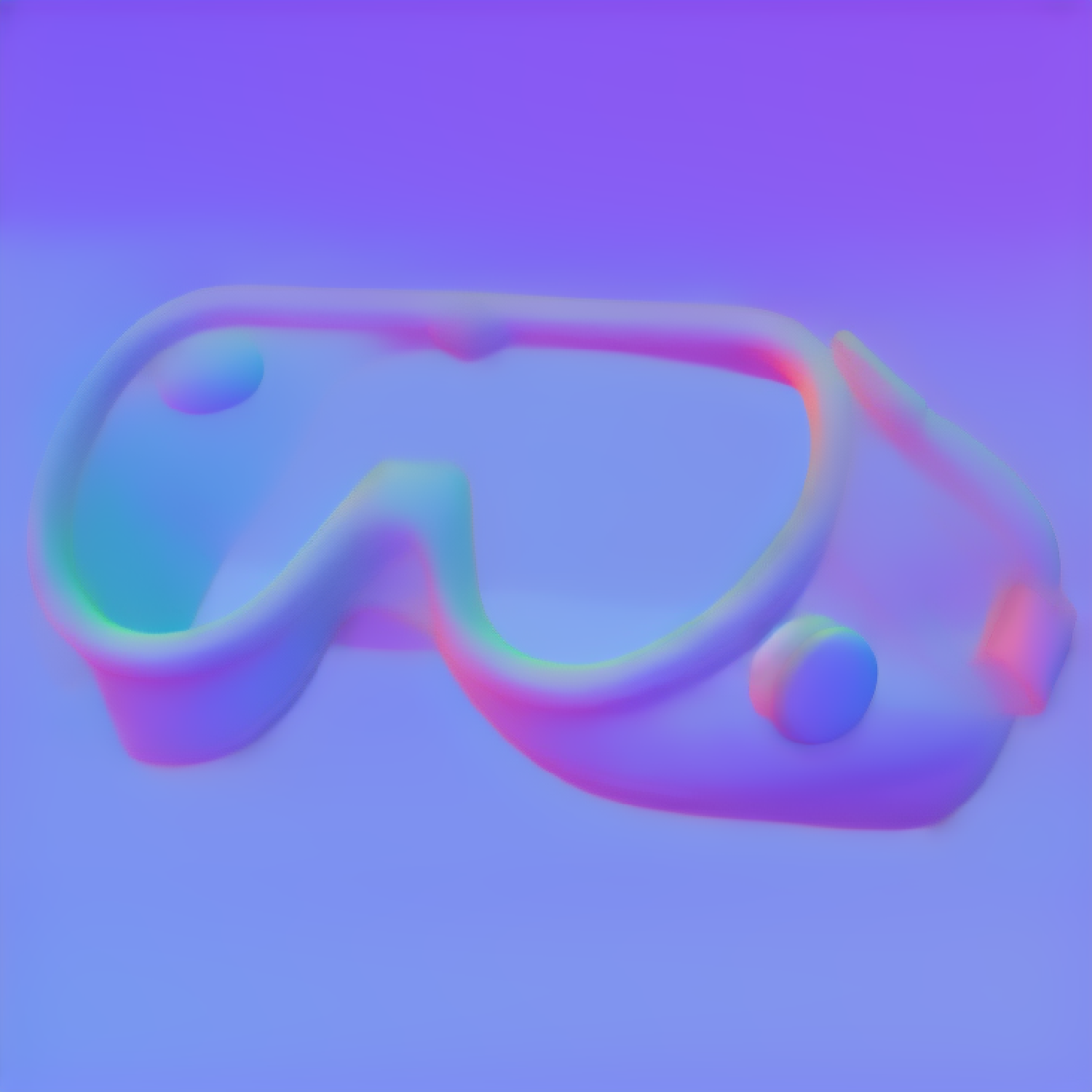}{-0.5cm}{0.3cm} &
        \teaserbox{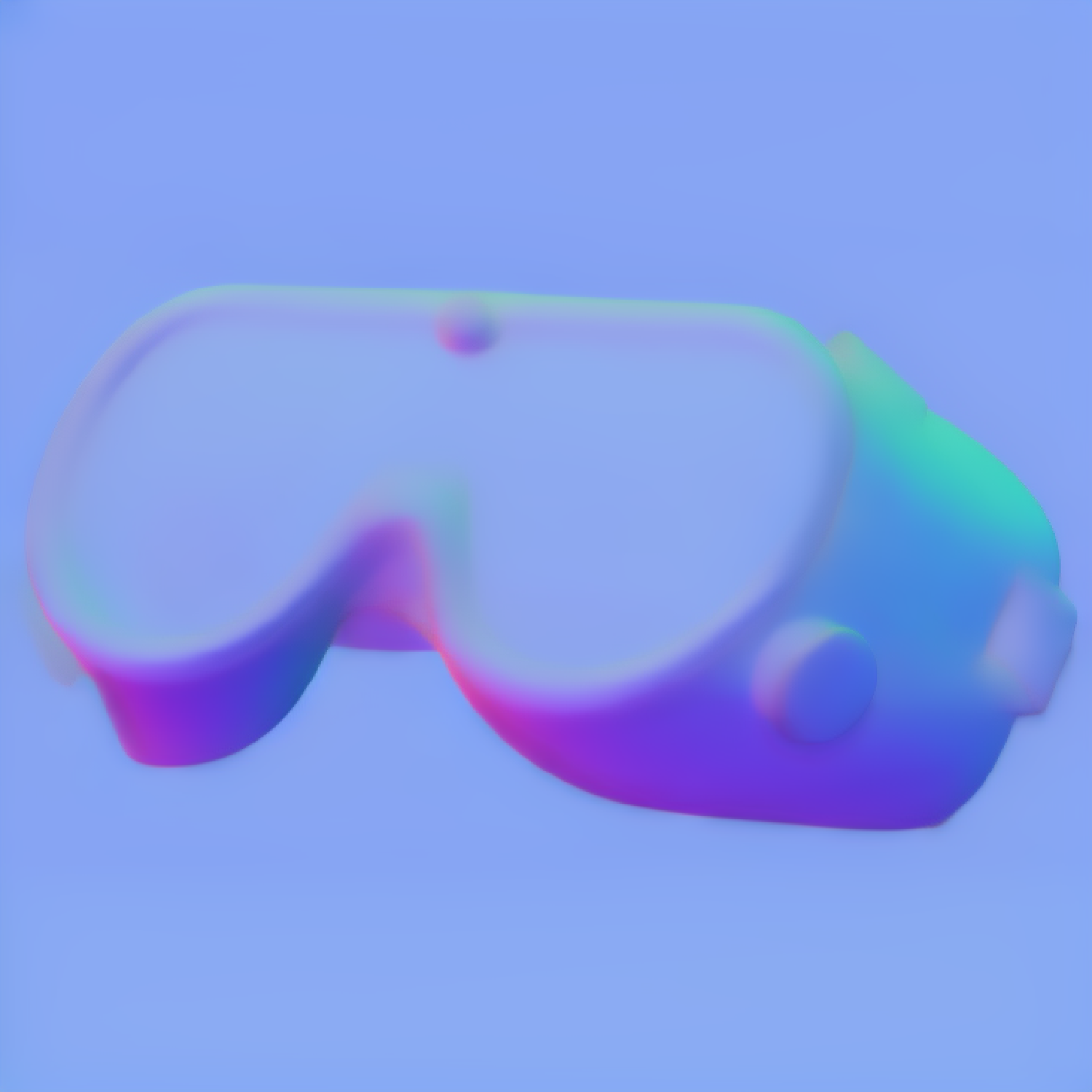}{-0.5cm}{0.3cm} \\[1pt]
        \includegraphics[width=0.14\columnwidth]{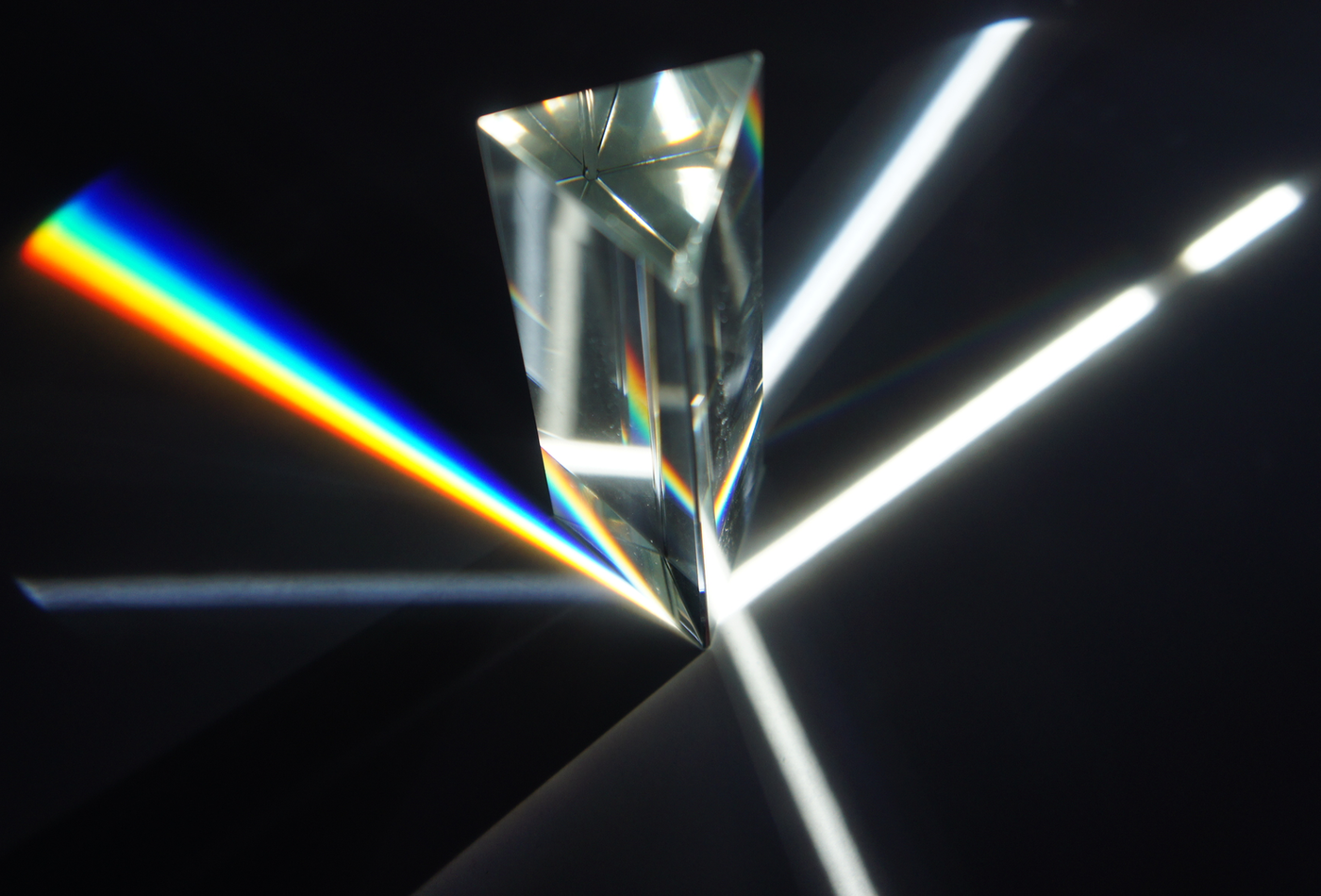} &
        \teaserbox{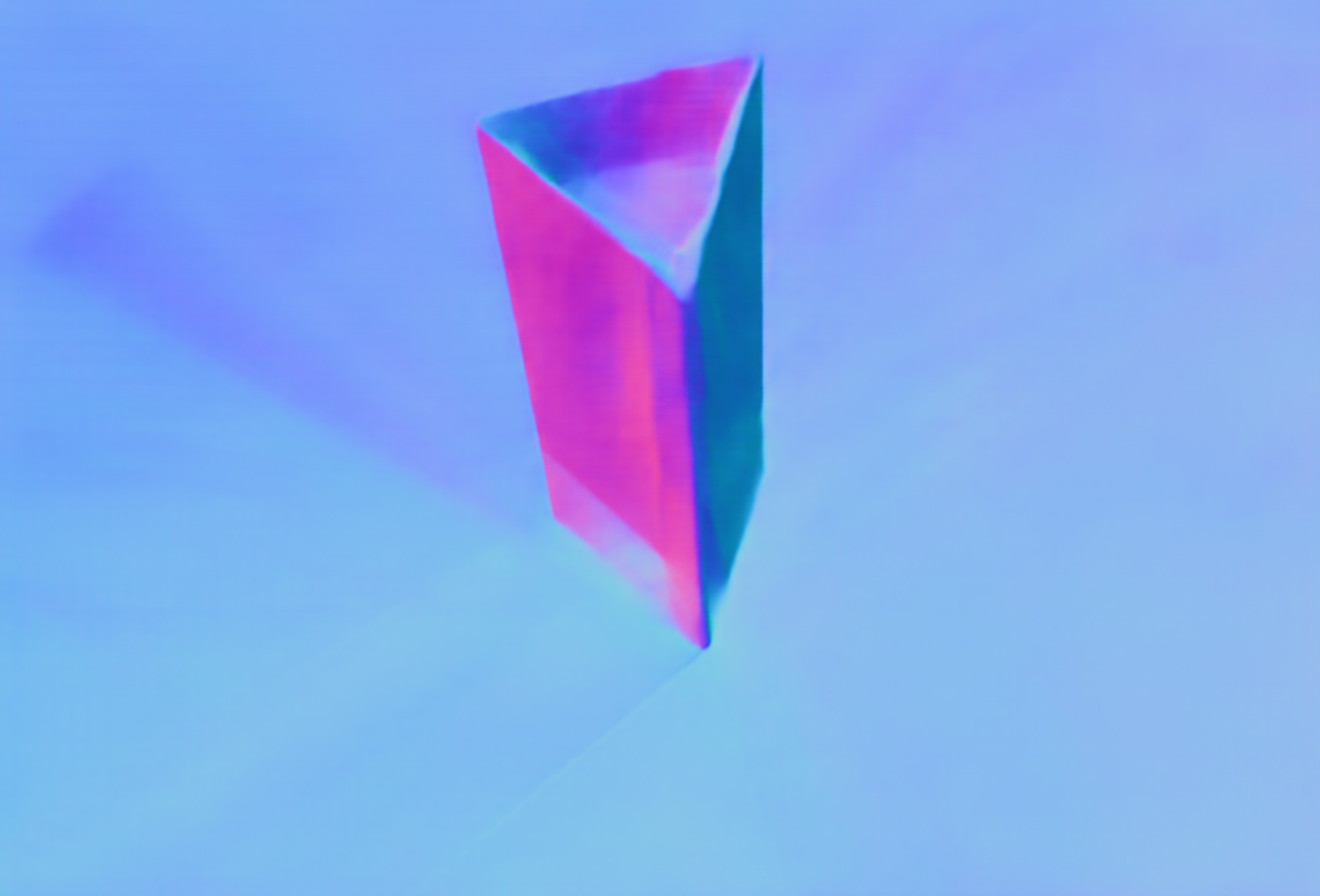}{-0.025cm}{0.35cm} &
        \teaserbox{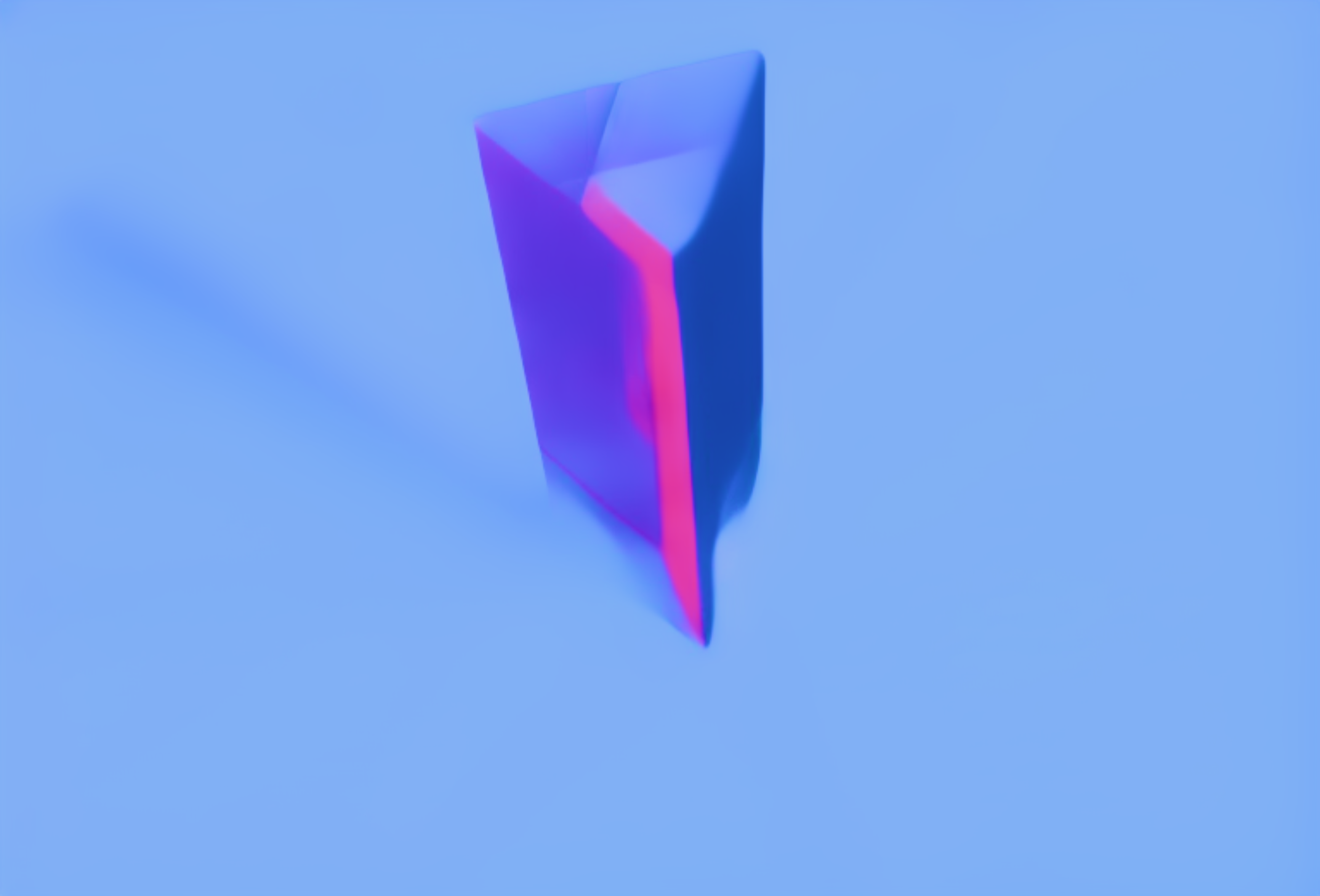}{-0.025cm}{0.35cm} &
        \teaserbox{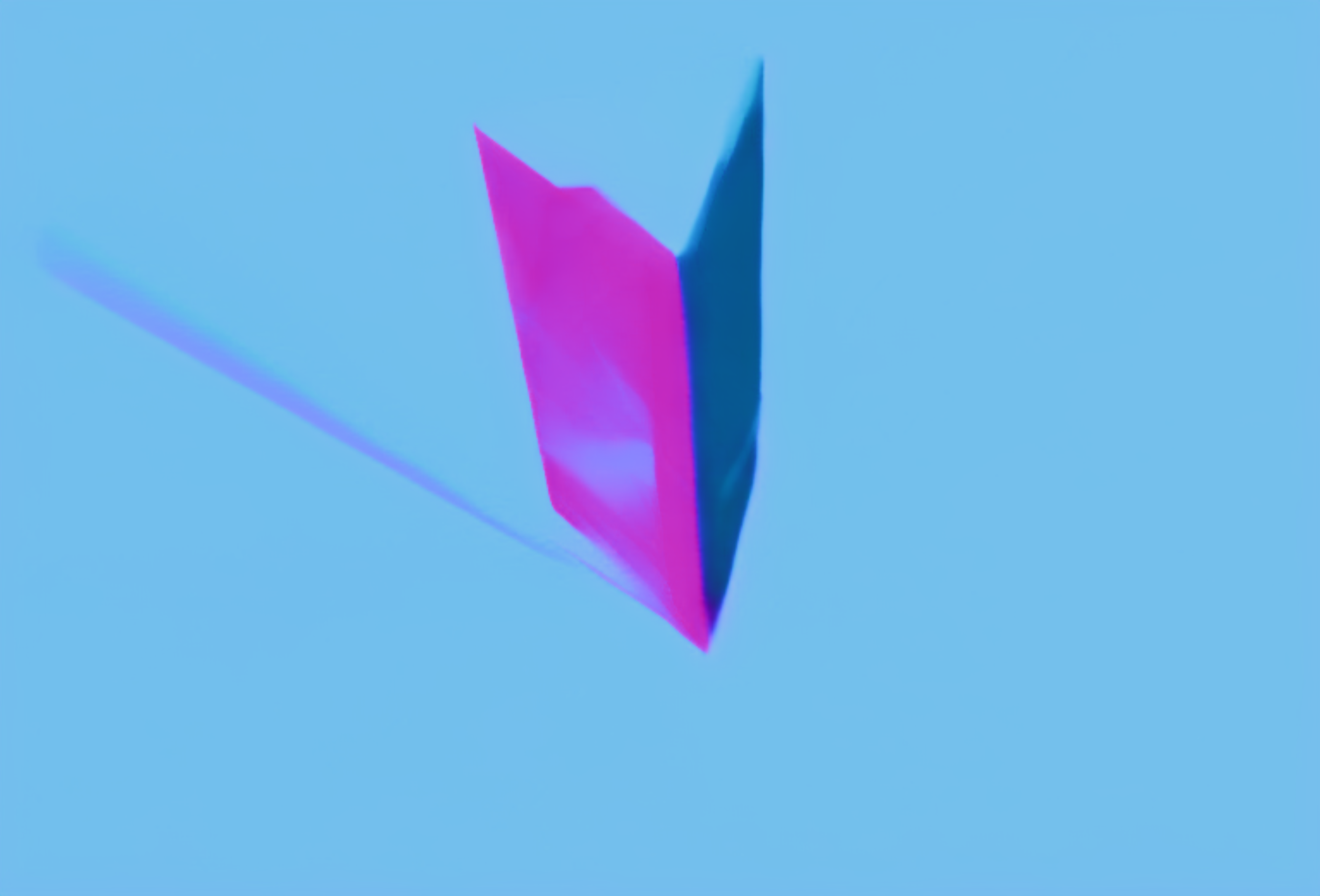}{-0.025cm}{0.35cm} &
        \teaserbox{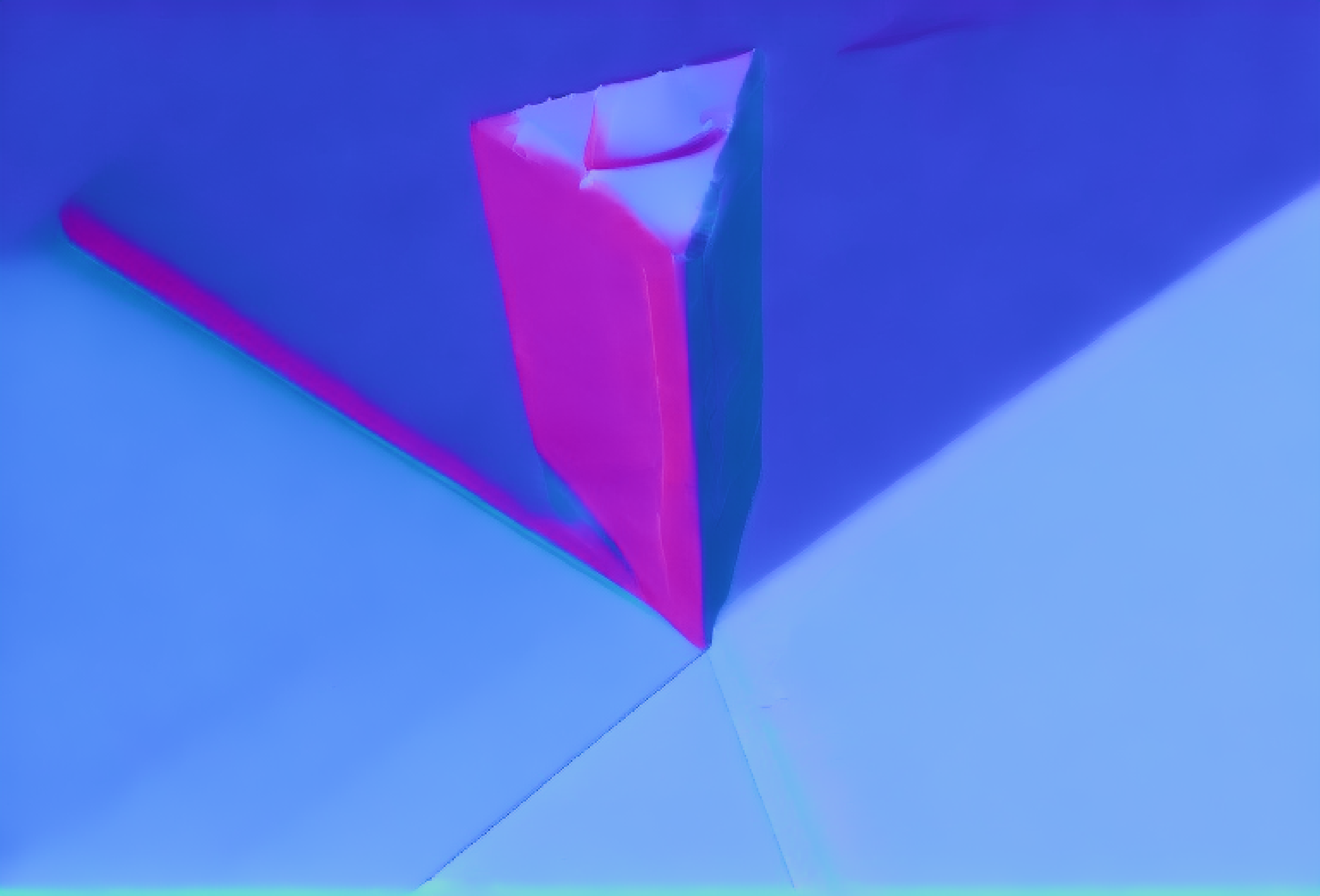}{-0.025cm}{0.35cm} &
        \teaserbox{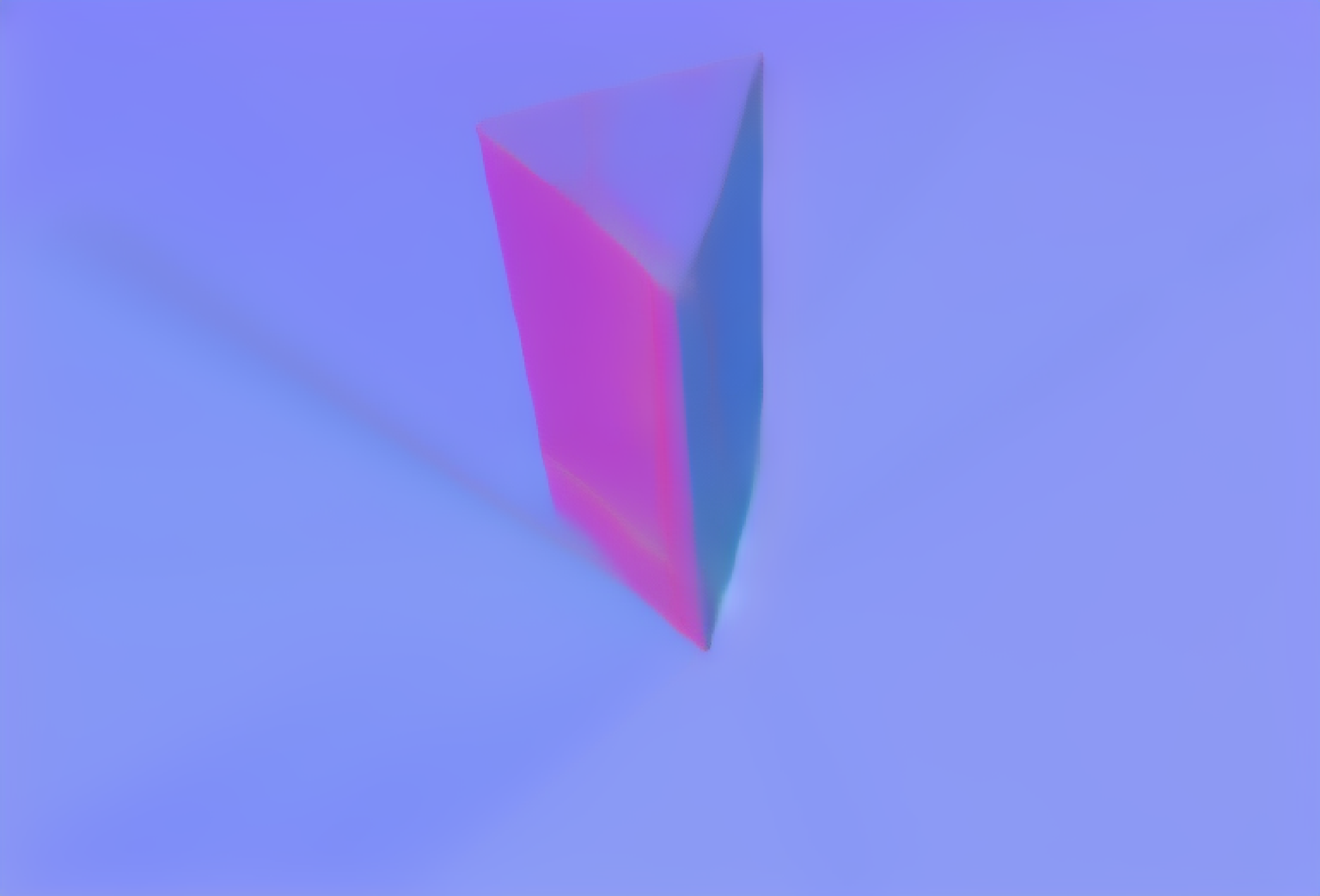}{-0.025cm}{0.35cm} &
        \teaserbox{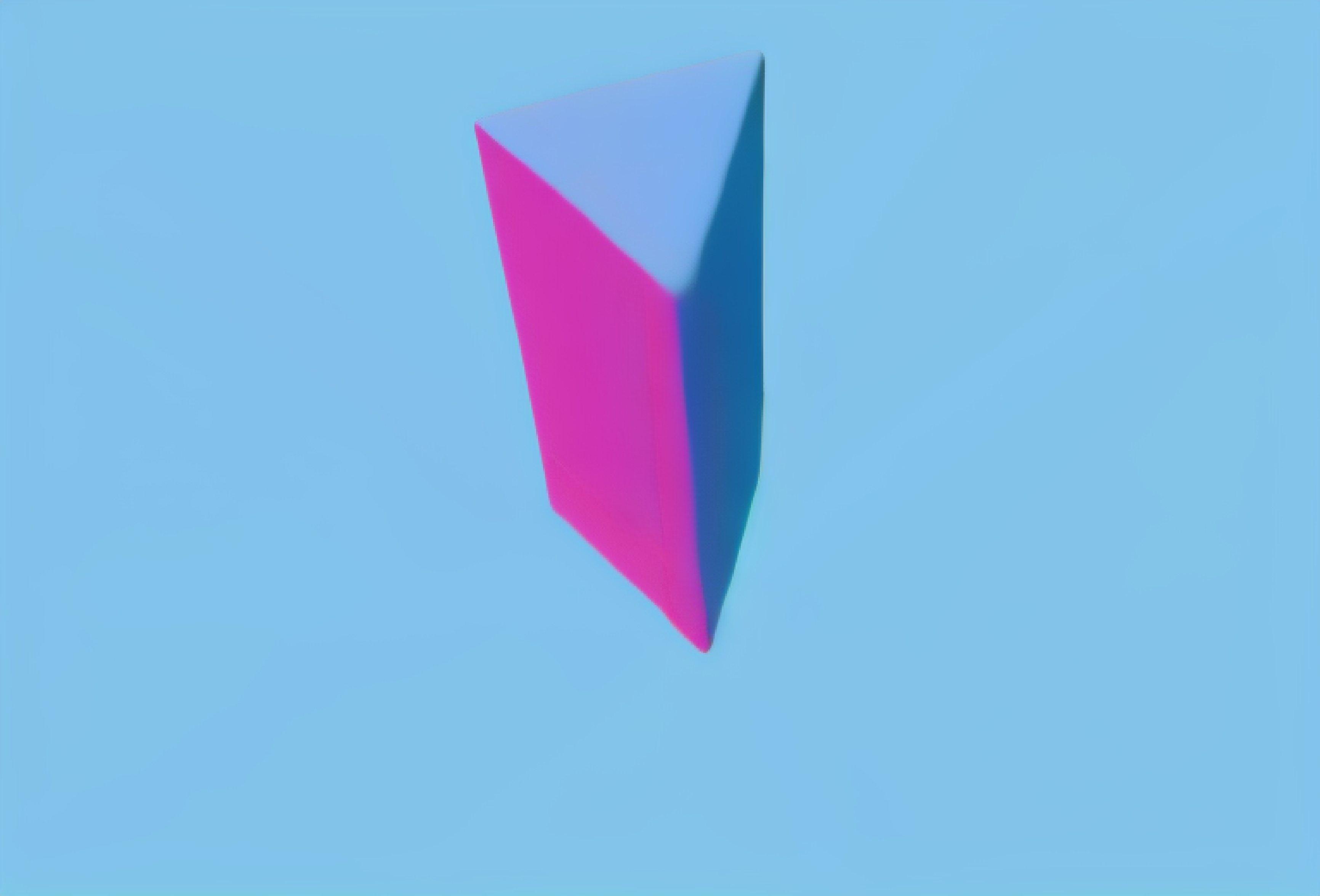}{-0.025cm}{0.35cm} \\[1pt]
        \includegraphics[width=0.14\columnwidth]{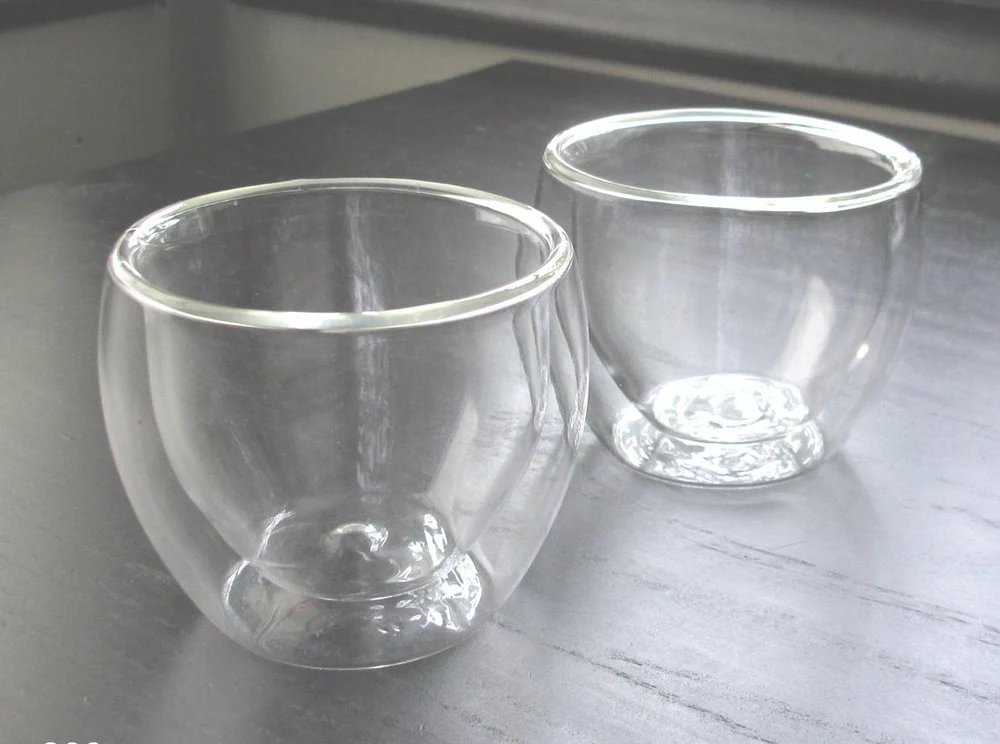} &
        \teaserbox{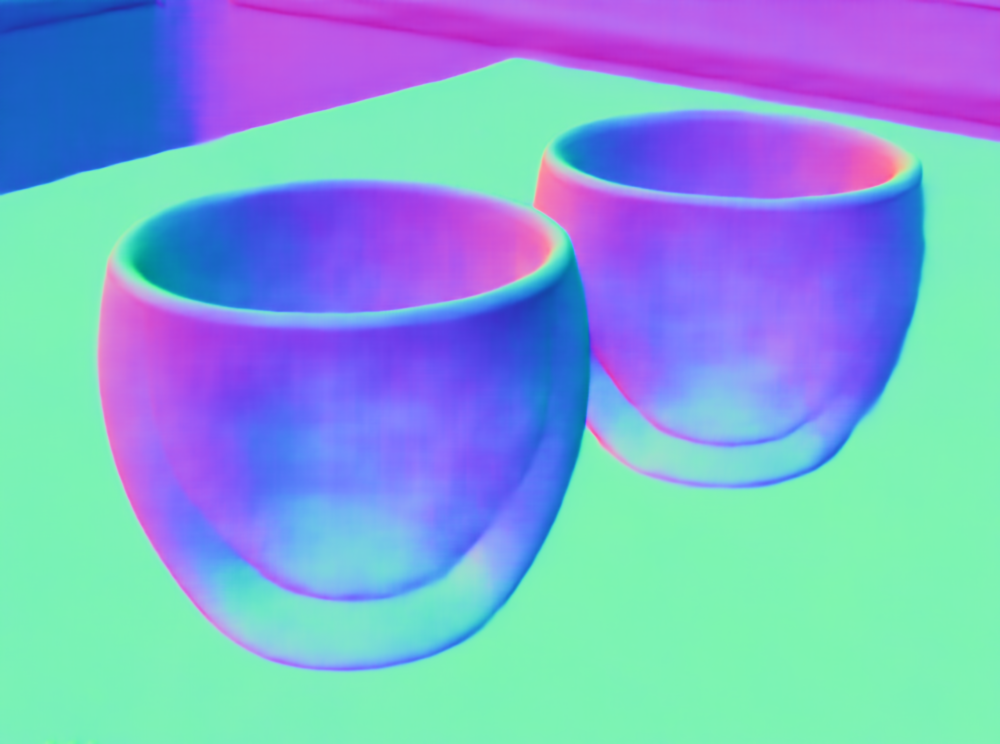}{-0.3cm}{-0.25cm} &
        \teaserbox{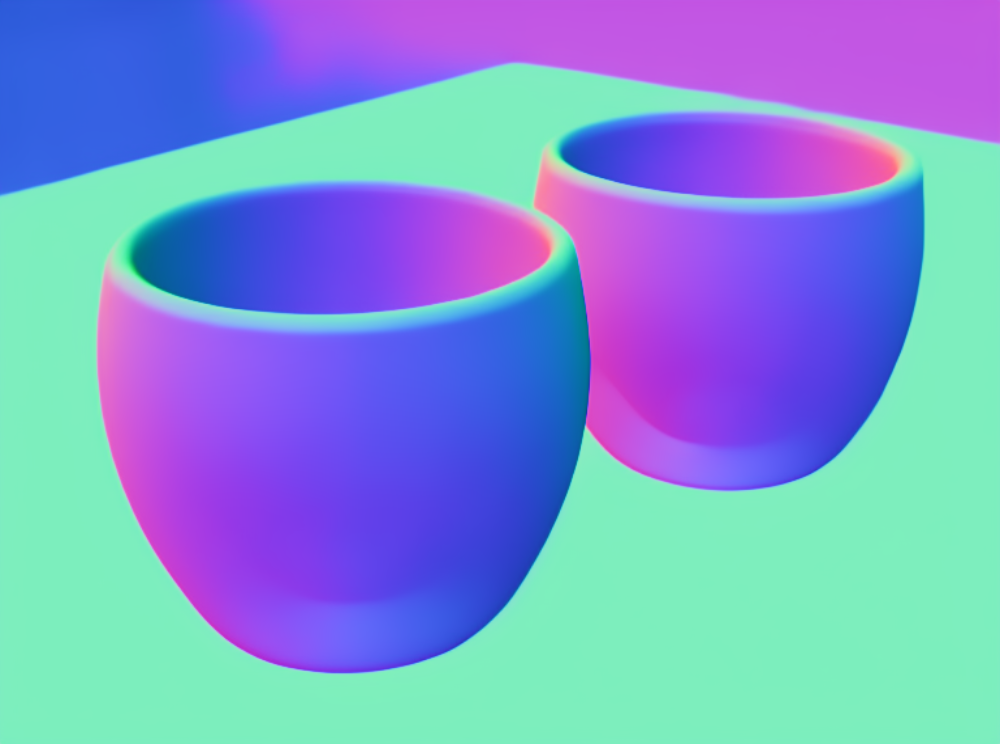}{-0.3cm}{-0.25cm} &
        \teaserbox{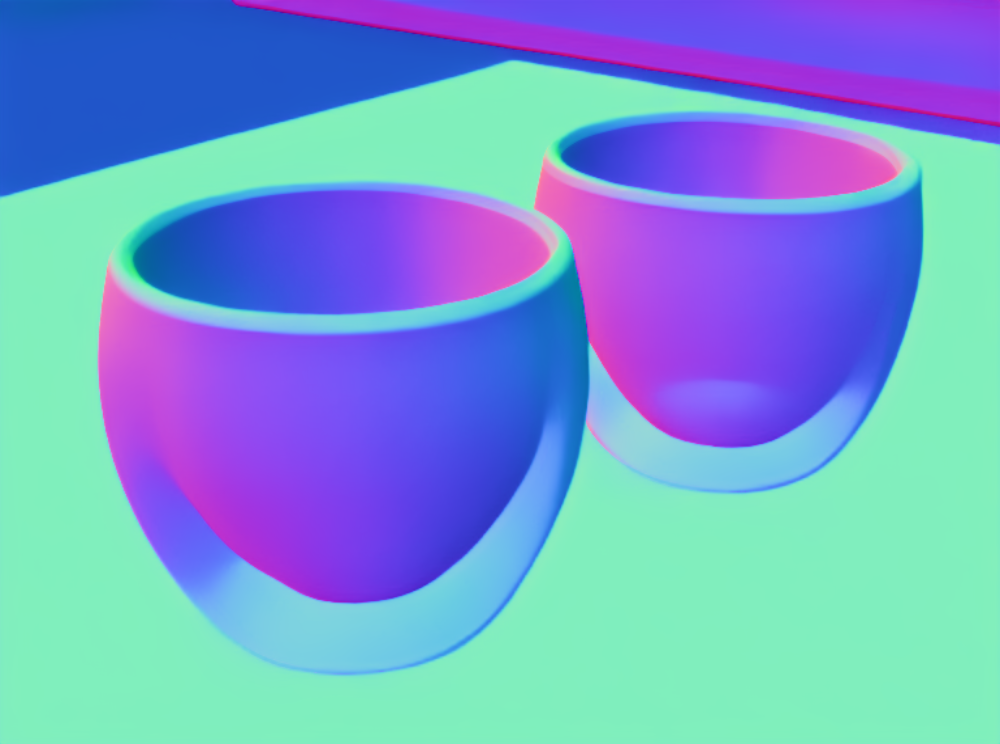}{-0.3cm}{-0.25cm} &
        \teaserbox{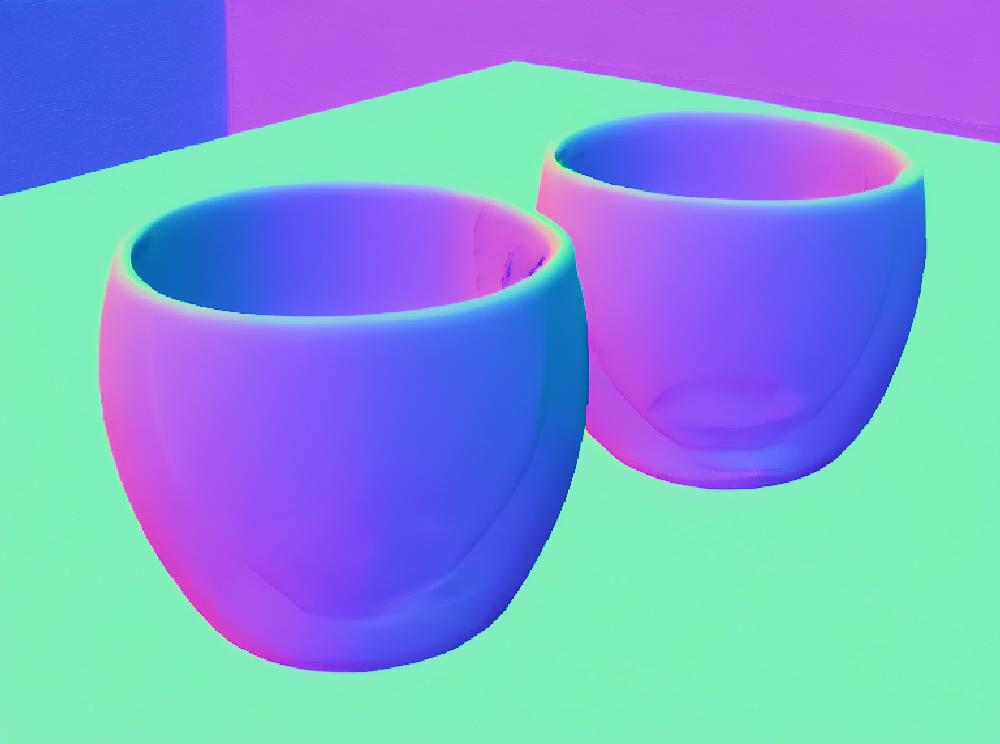}{-0.3cm}{-0.25cm} &
        \teaserbox{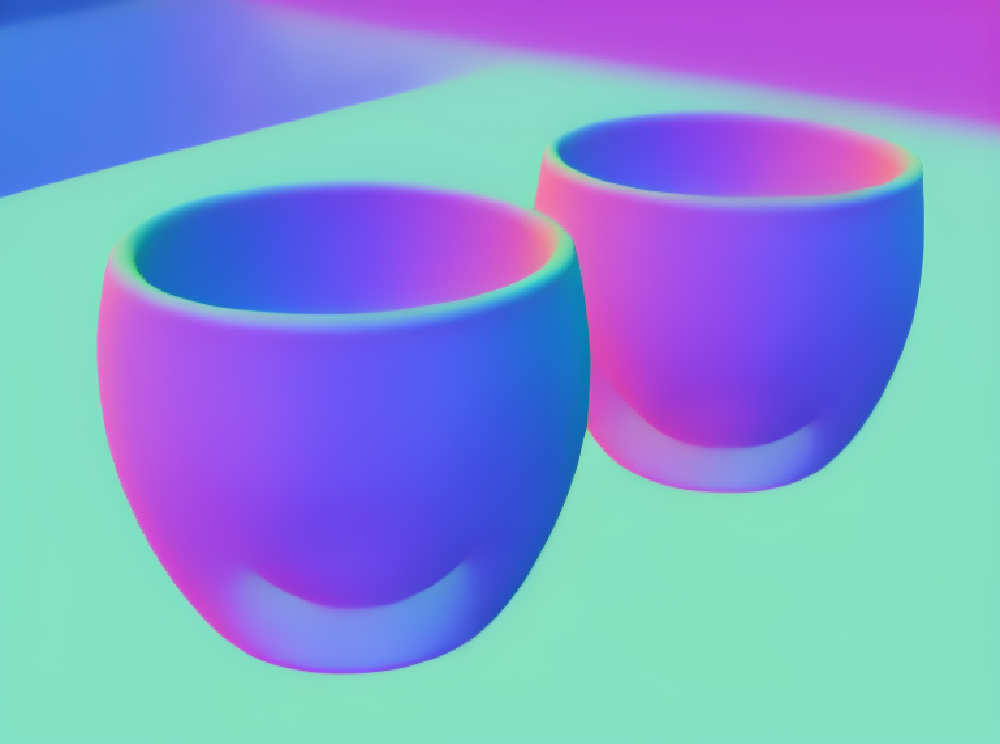}{-0.3cm}{-0.25cm} &
        \teaserbox{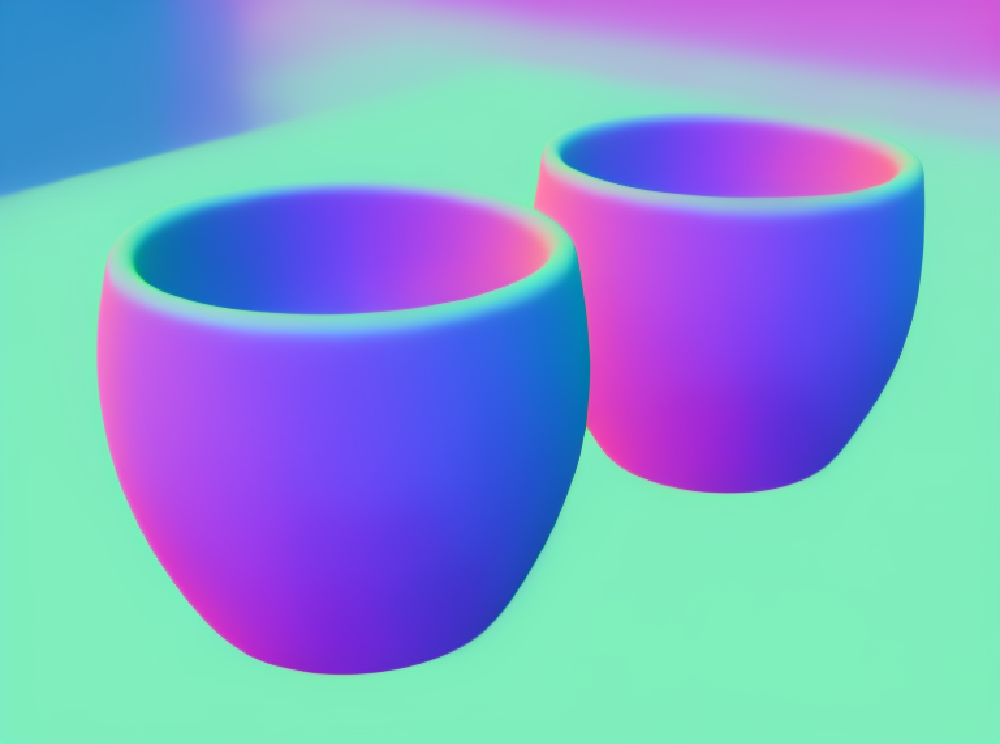}{-0.3cm}{-0.25cm} \\
        {\scriptsize Input} & {\scriptsize MoGe-2} & {\scriptsize E2E-FT} & {\scriptsize Marigold} & {\scriptsize GeoWizard} & {\scriptsize Lotus-D} & {\scriptsize \textbf{Ours}} \\
    \end{tabular}
    \caption{\textbf{In-the-wild qualitative results.} \textbf{TransNormal} (ours) recovers accurate surface normals on transparent objects.
    \textbf{Row 1:} Safety goggles with vent holes behind the lens---only our method recovers the flat lens surface.
    \textbf{Row 2:} Prism under extreme lighting---only ours recovers correct triangular geometry.
    \textbf{Row 3:} Double-walled glass---ours correctly estimates outer surface normals.}
    \label{fig:teaser}
    \vspace{-3mm}
\end{figure}
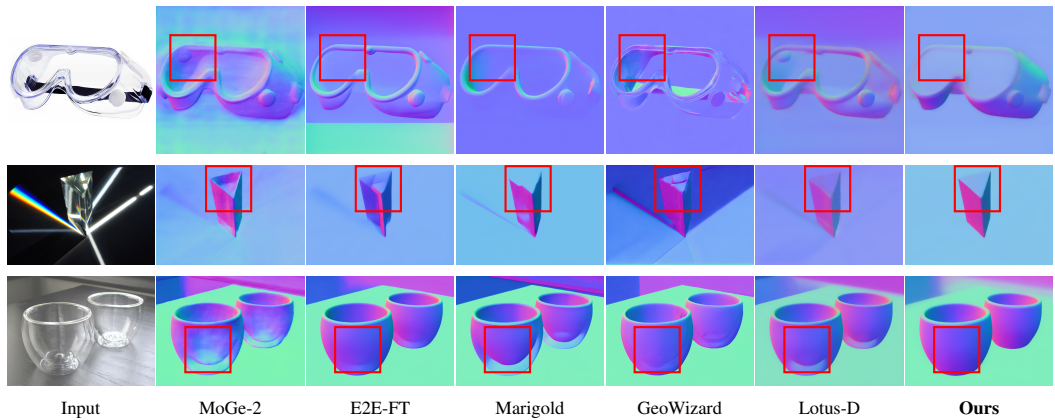

\begin{abstract}
  Monocular normal estimation for transparent objects is critical for laboratory automation, yet it remains challenging due to complex light refraction and reflection. These optical properties often lead to catastrophic failures in conventional depth and normal sensors, hindering the deployment of embodied AI in scientific environments. We propose \textbf{TransNormal}, a novel framework that adapts pre-trained diffusion priors for single-step normal regression. To handle the lack of texture in transparent surfaces, TransNormal integrates dense visual semantics from DINOv3 via a cross-attention mechanism, providing strong geometric cues. Furthermore, we employ a multi-task learning objective and wavelet-based regularization to ensure the preservation of fine-grained structural details. To support this task, we introduce \textbf{TransNormal-Synthetic}, a physics-based dataset with high-fidelity normal maps for transparent labware. Extensive experiments demonstrate that TransNormal significantly outperforms state-of-the-art methods: on the ClearGrasp benchmark, it reduces mean error by 24.4\% and improves $11.25^\circ$ accuracy by 22.8\%; on ClearPose, it achieves a 15.2\% reduction in mean error. The code and dataset will be made publicly available at \href{https://longxiang-ai.github.io/TransNormal}{https://longxiang-ai.github.io/TransNormal}.
\end{abstract}
\section{Introduction}
\label{sec:introduction}
\paragraph{Motivation.}\label{par:intro_motivation}
Embodied AI agents hold the potential to significantly accelerate scientific discovery in autonomous laboratory environments~\cite{tao2024maniskill3, fang2023anygrasp, wen2024vidman}. However, a primary barrier to their practical deployment is the perceptual instability caused by variable illumination~\cite{tobin2017domain, james2019sim, lind2024making}: variations in lighting and shadows can induce significant appearance shifts, leading to unstable detection and degraded manipulation performance. Unlike intensity-based features, surface normal maps provide a lighting-invariant geometric representation, offering a stable cue for perception and manipulation in real-world laboratories where lighting is often uncontrolled. 

Despite the maturity of normal estimation for opaque objects, transparent labware—such as beakers, pipettes, and culture dishes—presents a unique and formidable challenge. The difficulty of estimating normals for these objects is three-fold: 
\ding{172} \textbf{Geometrically}, transparent surfaces often lack discriminative textures and exhibit indistinct boundaries. Furthermore, multi-layered interfaces (\emph{e.g.}, glass-air-liquid) introduce significant structural ambiguities that are absent in opaque surfaces.
\ding{173} \textbf{Optically}, the dominance of refraction and reflection makes the visual appearance of labware highly dependent on the surrounding environment, often rendering the objects nearly invisible to standard sensors.
\ding{174} \textbf{Perceptually}, accurate reconstruction requires high-level reasoning, including object-level shape priors (\emph{e.g.}, the canonical geometry of a beaker) and contextual inference to distinguish between transparent and opaque material regions. 
Consequently, traditional geometric cues such as shading, texture gradients, and edge detection, which are the cornerstones of normal estimation for opaque objects, become unreliable or entirely absent. This necessitates a more robust approach that can leverage deep priors to resolve the inherent ambiguities of transparent surfaces.

Given the physical complexities of light transport, monocular normal estimation for transparent objects necessitates high-level reasoning about materials and global shapes. However, most existing frameworks~\citep{bae2024dsine} predominantly treat this as a localized regression task, relying on local image or photometric cues. While effective for textured opaque surfaces, these inductive biases are fundamentally ill-suited for transparent labware, where refraction and reflection decouple local appearance from underlying geometry. Furthermore, current research in the transparent domain has focused largely on 3D shape estimation, depth completion or 6D pose estimation~\cite{sajjan2020clear, chen2022clearpose,fang2022transcg,kim2024transpose}, leaving the estimation of dense surface normals, which is a critical representation for fine-grained robotic manipulation and liquid handling, relatively under-explored. The lack of high-quality benchmarks with dense normal annotations further hinders progress.

Our key insight is that the inherent ambiguities of transparent surfaces can be resolved by leveraging high-level scene understanding and physical priors encoded in large-scale vision models. Unlike local discriminative kernels, generative models trained on diverse web-scale data may already capture the ``canonical'' geometry and material properties of objects. This motivates the use of model families whose conditioning pathways allow for task-specific guidance to bridge the gap between low-level appearance and high-level geometric structure. 
Diffusion-based dense prediction provides such a model family. Recent advances~\cite{ke2024repurposing, fu2024geowizard} demonstrate that pre-trained text-to-image models, such as Stable Diffusion, possess rich geometric and material priors. However, we observe a significant gap in current practice: many methods~\cite{he2024lotus,ke2024repurposing,zhao2025diception} utilize these models with empty or generic text prompts, leaving the cross-attention conditioning pathway, which is originally designed for complex semantic alignment, largely underutilized for geometric tasks. 

We propose \textbf{TransNormal}, a framework that repurposes Stable Diffusion's conditioning mechanism for dense semantic injection. Rather than relying on sparse text, we inject dense visual semantics from DINOv3~\cite{siméoni2025dinov3} into the diffusion backbone. 
By transforming cross-attention into a semantic-geometric guidance channel, TransNormal effectively resolves the geometric ambiguities of transparent labware using global context. To facilitate robust training and evaluation, we introduce TransNormal-Synthetic, a physics-based dataset with high-fidelity normal maps for transparent labware. Despite being trained on only $\sim$122K synthetic samples, TransNormal achieves state-of-the-art performance on ClearGrasp~\cite{sajjan2020clear} and ClearPose~\cite{chen2022clearpose} (\tabref{tab:transparent}), reducing mean error on real-world data by significant margins. Our key contributions are as follows:
\begin{itemize}
    \item \textbf{Semantic-Geometric Conditioning:} We identify a critical underutilization in diffusion-based dense prediction and propose to replace sparse text conditioning with dense DINOv3 visual semantics to provide material-aware geometric guidance. 
    \item \textbf{TransNormal Framework:} We present a novel architecture that adapts Stable Diffusion for single-step normal regression. TransNormal achieves superior generalization to transparent surfaces with significantly fewer training samples than traditional Transformer-based discriminative baselines.
    \item \textbf{Physics-Based Dataset:} We introduce TransNormal-Synthetic, a high-quality benchmark providing physically accurate normal maps rendered from 3D labware meshes, enabling controlled and systematic evaluation of transparent object perception.
    \item \textbf{State-of-the-Art Performance:} Our method sets new performance standards across multiple benchmarks. On ClearGrasp, TransNormal reduces mean angular error by 24.4\% and improves $11.25^\circ$ accuracy by 22.8\%; on the real-world ClearPose dataset, it achieves a 15.2\% error reduction. These results demonstrate robust zero-shot transfer from synthetic training to complex, real-world laboratory environments.
\end{itemize}

\section{Related Work}
\label{sec:related_work}

\subsection{Geometric Dense Prediction and Generative Priors}
Recovering geometric properties such as depth and surface normals has evolved through three paradigms. Early physics-based methods relied on Structure from Motion (SfM)~\cite{tomasi1992shape}, photometric stereo~\cite{woodham1980photometric}, and multi-view geometry~\cite{scharstein2002taxonomy}, but were brittle under real-world conditions. The discriminative learning paradigm~\cite{eigen2014depth, eftekhar2021omnidata, ranftl2020towards} and recent large-scale models like MoGe~\cite{wang2025moge,wang2025moge2} and Depth Anything~\cite{yang2024depth1,yang2024depth2} achieved remarkable success, yet struggle with out-of-distribution scenarios such as transparent or reflective surfaces.

Most recently, a \textbf{generative paradigm} has emerged, reframing dense prediction as conditional generation. Models like Marigold~\cite{ke2024repurposing} and GeoWizard~\cite{fu2024geowizard} leverage world priors from large-scale diffusion models~\cite{rombach2022high} for strong zero-shot generalization. The adaptation of these priors follows three trajectories: (a)~\emph{stochastic generative} methods (\emph{e.g.}, Marigold, DepthFM~\cite{gui2025depthfm}) use multi-step diffusion but suffer from inference inefficiency and structural variance; (b)~\emph{deterministic feed-forward} approaches (\emph{e.g.}, Diffusion-E2E-FT~\cite{martingarcia2024diffusione2eft}, Lotus~\cite{he2024lotus}) fine-tune backbones for speed but often lose fine-grained details; (c)~\emph{coarse-to-fine} strategies (\emph{e.g.}, StableNormal~\cite{ye2024stablenormal}) bridge this gap but often reintroduce stochasticity in refinement. 
A critical limitation across these methods is their underutilization of semantic conditioning---they typically use empty text prompts or simple category labels, leaving rich semantic priors largely unexploited. This overlooks the potential of dense visual semantics: recent self-supervised encoders like DINOv2~\cite{oquab2023dinov2} and DINOv3~\cite{siméoni2025dinov3} capture robust object-centric representations that persist even under refractive distortions, offering a more suitable guidance signal for geometry estimation. Our work builds upon this generative paradigm, integrating such dense semantic guidance to address the challenges of transparent materials. 

\subsection{Geometry Estimation for Transparent Objects}
Perception of transparent objects is uniquely challenging due to refraction and reflections that cause commodity depth sensors to produce large holes or distortions. Related tasks include transparent object segmentation~\cite{xie2020trans10k, xie2021trans10kv2, sun2023trosd} and 6D pose estimation~\cite{zhang2022transnet, jiang2024ebfa6d}, which share similar optical challenges. Early methods like ClearGrasp~\cite{sajjan2020clear} and DREDS~\cite{dai2022domain} pioneered learning-based depth completion, while subsequent work~\cite{zhu2021lidf, xu2022transparenet, hong2022cluedepth, cai2023consistent} recovered true depth from corrupted RGB-D inputs, enabled by benchmarks like ClearPose~\cite{chen2022clearpose} and TransCG~\cite{fang2022transcg}. Physics-based approaches have also explored monocular shape from refraction~\cite{sulc2021refraction} and refractive flow for normal estimation~\cite{tang2024rftrans}, while polarization cameras offer complementary normal cues~\cite{shao2023polarization}. When multi-view RGB is accessible, neural implicit representations enable full geometry recovery~\cite{ichnowski2021dexnerf, li2023neto, zhou2023nvs4glass, deng2024tnsr, sun2024nunerf, li2025tsgs}, though requiring dense viewpoints. Very recently, video diffusion models~\cite{hu2025depthcrafter} have been adapted for geometry estimation, with DKT~\cite{xu2025dkt} extending this paradigm to transparent object depth; however, these methods require temporal sequences as input, limiting their applicability to single-image scenarios. Accurate geometry also underpins robotic manipulation~\cite{Bai_Zhang_Tao_Wu_Wang_Xu_2023, bai2025star, bai2025retrieval}. Training data has evolved from Physics-Based Rendering (PBR)~\cite{sajjan2020clear} to generative synthesis~\cite{zhang2024transparent, agrawal2024clear}. Our approach fine-tunes a generative backbone on a curated synthetic dataset that disentangles geometry from material appearance, internalizing a rich prior of transparent phenomena.

\section{Preliminaries}
\label{sec:preliminaries}
\begin{figure*}[t]
    \centering
    \includegraphics[width=\textwidth]{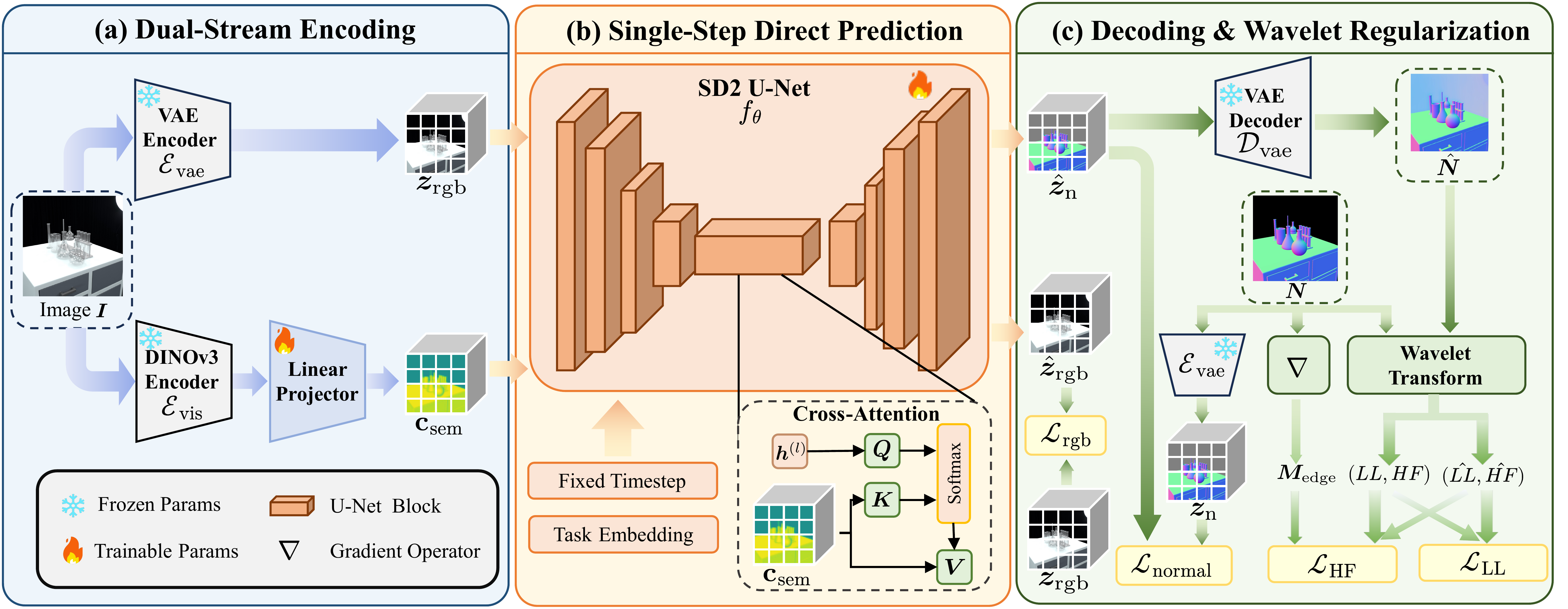}
    \caption{\textbf{Overview of the TransNormal framework.}
    (a) \textbf{Dual-Stream Encoding}: the frozen VAE encoder $\mathcal{E}_{\text{vae}}$ extracts RGB latent $\bm{z}_{\text{rgb}}$, while a frozen DINOv3 encoder $\mathcal{E}_{\text{vis}}$ with a trainable linear projector produces semantic conditioning $\bm{c}_{\text{sem}}$;
    (b) \textbf{Single-Step Direct Prediction}: the fine-tuned SD2 U-Net $f_{\vtheta}$ directly regresses the normal latent $\hat{\bm{z}}_{\text{n}}$ from $\bm{z}_{\text{rgb}}$ at a fixed timestep $T$, where the spatial feature $\bm{h}^{(l)}$ at each layer $l$ provides queries $\bm{Q}$, and $\bm{c}_{\text{sem}}$ provides keys $\bm{K}$ and values $\bm{V}$ for cross-attention;
    (c) \textbf{Decoding \& Wavelet Regularization}: the frozen VAE decoder $\mathcal{D}_{\text{vae}}$ reconstructs the predicted normal map $\hat{\bm{N}}$, supervised by latent-space losses ($\Ls_{\text{rgb}}$, $\Ls_{\text{normal}}$) and wavelet-domain losses ($\Ls_{\text{HF}}$, $\Ls_{\text{LL}}$) that separately penalize high-frequency details and low-frequency structure. (\S~\ref{par:method_overview})}
    \label{fig:pipeline}
\end{figure*}

\textbf{Latent Diffusion Models.}
Our framework is built upon Stable Diffusion~\citep{rombach2022high}, which performs the diffusion process in a compressed latent space for computational efficiency. This is enabled by a pre-trained Variational Auto-Encoder (VAE) consisting of an encoder $\mathcal{E}(\cdot)$ and a decoder $\mathcal{D}(\cdot)$, which maps between RGB space and latent space, \emph{i.e.}, $\mathcal{E}(\bm{x})=\bm{z}^x$, $\mathcal{D}(\bm{z}^x)\approx \bm{x}$. Following recent dense prediction works~\citep{ke2024repurposing, fu2024geowizard,xu2024matters,ye2024stablenormal}, we also map dense annotations into this latent space: $\mathcal{E}(\bm{y})=\bm{z}^y$, $\mathcal{D}(\bm{z}^y)\approx \bm{y}$.

\textbf{Diffusion Process.}
Stable Diffusion establishes a probabilistic model through a \emph{forward} noising process and a \emph{reversal} denoising process. In the \emph{forward} process, Gaussian noise is gradually added to the latent $\bm{z}^y$ over time steps $t\in [1,T]$:
\begin{equation}
    \bm{z}^y_t = \sqrt{\bar{\alpha}_t}\bm{z}^y + \sqrt{1-\bar{\alpha}_t}\rvepsilon, \quad \rvepsilon \sim \gN(0, \mI),
\end{equation}
where $\bar{\alpha}_t := \prod_{s=1}^t (1- \beta_s)$ and $\{\beta_t\}_{t=1}^T$ is the noise schedule. At $t=T$, $\bm{z}^y_T$ approximates pure Gaussian noise. In the \emph{reversal} process, a U-Net $f_{\vtheta}$~\citep{ronneberger2015u} iteratively removes noise to recover $\bm{z}^y$.

\textbf{Single-Step Regression for Dense Prediction.}
While the standard diffusion formulation relies on iterative sampling for stochastic generation, dense prediction tasks (\emph{e.g.}, normal estimation) are inherently deterministic. Recent studies~\citep{ke2024repurposing, he2024lotus,xu2024matters} demonstrate that the pre-trained U-Net can be effectively repurposed for direct regression. Adopting this strategy, we simplify the inference process: instead of multi-step denoising, we fix the timestep at $T$ and train the network to directly predict the clean annotation latent $\bm{z}^y$ from the input image latent $\bm{z}^x$ in a single forward pass:
\begin{equation}
    \hat{\bm{z}}^y = f_{\vtheta}(\bm{z}^x, T).
\end{equation}
This approach leverages the strong priors of Stable Diffusion while ensuring deterministic and efficient prediction.

\textbf{Notation.} For clarity in the subsequent method description (\S~\ref{sec:method}), we adopt more descriptive subscripts: $\bm{z}^x \equiv \bm{z}_{\text{rgb}}$ denotes the RGB image latent and $\bm{z}^y \equiv \bm{z}_{\text{n}}$ denotes the normal map latent.

\section{Method}
\label{sec:method}
\paragraph{Method Overview.}\label{par:method_overview}
Given an input RGB image $\bm{I} \in \R^{H \times W \times 3}$, our goal is to predict the surface normal map $\bm{N} \in \R^{H \times W \times 3}$. We build \textbf{TransNormal} by repurposing Stable Diffusion 2 (SD2) as a single-step normal predictor with semantic conditioning. This section follows SD2's data flow: (a) encoders and semantic conditioning; (b) the U-Net prediction module; and (c) VAE decoding and training objectives.

\subsection{Dual-Stream Encoding}\label{ssec:encoders_semantic}
\paragraph{Semantic Guidance via Visual Prompting.}
Previous diffusion-based methods often rely on CLIP~\cite{radford2021learning} text encoders with generic or empty prompts, leaving the powerful cross-attention mechanism underutilized. For transparent objects, where refraction corrupts local textures, such sparse conditioning is insufficient. We therefore replace the text encoder with a frozen DINOv3 visual encoder $\mathcal{E}_{\text{vis}}$ to extract dense, object-level semantic features:
\begin{equation}
    \bm{F}_{\text{sem}} = \mathcal{E}_{\text{vis}}(\bm{I}) \in \R^{N_p \times d_{\text{dino}}},
\end{equation}
where $N_p = \lfloor H/p \rfloor \times \lfloor W/p \rfloor$ is the number of patch tokens with patch size $p$ and feature dimension $d_{\text{dino}}$. These features are then projected into the U-Net's cross-attention dimension via a trainable linear projector $\bm{W}_{\text{proj}} \in \R^{d_{\text{dino}} \times d_{\text{unet}}}$:
\begin{equation}
    \bm{c}_{\text{sem}} = \bm{F}_{\text{sem}} \bm{W}_{\text{proj}} \in \R^{N_p \times d_{\text{unet}}}.
\end{equation}
This stream effectively acts as a dense ``visual prompt'', injecting robust semantic priors that persist even under refractive distortions. The DINOv3 encoder is kept frozen and only the lightweight projector $\bm{W}_{\text{proj}}$ is trained. 

\paragraph{Latent Content Encoding.}
To leverage the generative priors of Stable Diffusion, the second stream maps the input image into the model's native latent space using the frozen VAE encoder $\mathcal{E}_{\text{vae}}$:
\begin{equation}
    \bm{z}_{\text{rgb}} = \mathcal{E}_{\text{vae}}(\bm{I}) \in \R^{h \times w \times 4},
\end{equation}
where $(h, w) = (\lfloor H/8 \rfloor, \lfloor W/8 \rfloor)$. This latent representation $\bm{z}_{\text{rgb}}$ serves as the direct input to the U-Net, preserving spatial structure and fine-grained details for the regression task. Similarly, during training, the ground truth normal map $\bm{N}$ is also encoded into the latent space:
\begin{equation}
    \bm{z}_{\text{n}} = \mathcal{E}_{\text{vae}}(\bm{N}) \in \R^{h \times w \times 4}.
\end{equation}

\subsection{Single-Step Prediction with Semantic Injection}
\label{sec:notation}
\paragraph{Detail Preserver via Dual-Task Learning.} To avoid catastrophic forgetting when fine-tuning a pre-trained diffusion model~\citep{zhai2023investigating}, we follow~\citet{he2024lotus} and \emph{adopt} their task switcher with two \emph{fixed} task embeddings $s \in \{s_{\text{n}}, s_{\text{rgb}}\}$, added to the time embedding as class-label conditions. The same U-Net $f_{\vtheta}$ serves both tasks: $s_{\text{n}}$ triggers normal prediction and $s_{\text{rgb}}$ triggers RGB reconstruction. These embeddings are kept fixed during training. This switch preserves fine detail while adapting the model to geometry.

\paragraph{Single-Step Normal Prediction.} 
Unlike prior diffusion-based methods that inject noise and recover clean latents through iterative denoising, we directly input clean RGB latents and predict normal latents in a single forward pass. We initialize the predictor $f_{\vtheta}$ from the SD2 U-Net and fully fine-tune it for single-step normal regression. The model predicts the clean normal latent conditioned on the RGB latent, semantic features, and the normal task embedding $s_{\text{n}}$:
\begin{equation}
    \hat{\bm{z}}_{\text{n}} = f_{\vtheta}(\bm{z}_{\text{rgb}}, T, \bm{c}_{\text{sem}}, s_{\text{n}}),
\end{equation}
where $T$ is a fixed timestep embedding. Similarly, the RGB reconstruction task predicts $\hat{\bm{z}}_{\text{rgb}} = f_{\vtheta}(\bm{z}_{\text{rgb}}, T, \bm{c}_{\text{sem}}, s_{\text{rgb}})$. As illustrated in \figref{fig:pipeline}, the U-Net follows a standard encoder-decoder structure with skip connections. At each layer $l$, we flatten the spatial feature map into tokens $\bm{h}^{(l)} \in \R^{n \times d_l}$, where $n$ is the number of spatial tokens and $d_l$ is the feature channel dimension at layer $l$. We then apply cross-attention with $\bm{c}_{\text{sem}}$ as keys and values:

\begin{equation}
\begin{aligned}
    &\bm{Q} = \bm{h}^{(l)} \bm{W}_Q, \quad \bm{K} = \bm{c}_{\text{sem}} \bm{W}_K, \quad \bm{V} = \bm{c}_{\text{sem}} \bm{W}_V, \\
    &\text{CrossAttn}(\bm{h}^{(l)}, \bm{c}_{\text{sem}}) = \text{Softmax}\left(\frac{\bm{Q}\bm{K}^\top}{\sqrt{d_k}}\right)\bm{V},
\end{aligned}
\end{equation}
where $d_k$ is the attention head dimension. This design allows spatial features to query high-level semantics for geometry inference under ambiguous transparency cues.

\subsection{Decoding \& Wavelet Regularization}
\paragraph{Decoding.} The predicted latent is decoded to the pixel space in a single step:
\begin{equation}
    \hat{\bm{N}} = \mathcal{D}_{\text{vae}}(\hat{\bm{z}}_{\text{n}}).
\end{equation}


\paragraph{Training Losses.} We use three losses: $\Ls_{\text{normal}}$, $\Ls_{\text{rgb}}$, and $\Ls_{\text{wavelet}}$. We define the latent reconstruction losses as:
\begin{equation}
    \Ls_{\text{normal}} = \| \hat{\bm{z}}_{\text{n}} - \bm{z}_{\text{n}} \|^2_2, \quad 
    \Ls_{\text{rgb}} = \| \hat{\bm{z}}_{\text{rgb}} - \bm{z}_{\text{rgb}} \|^2_2.
\end{equation}

\begin{wrapfigure}{r}{0.5\columnwidth}
    \centering
    \vspace{-3mm}
    \includegraphics[width=0.48\columnwidth]{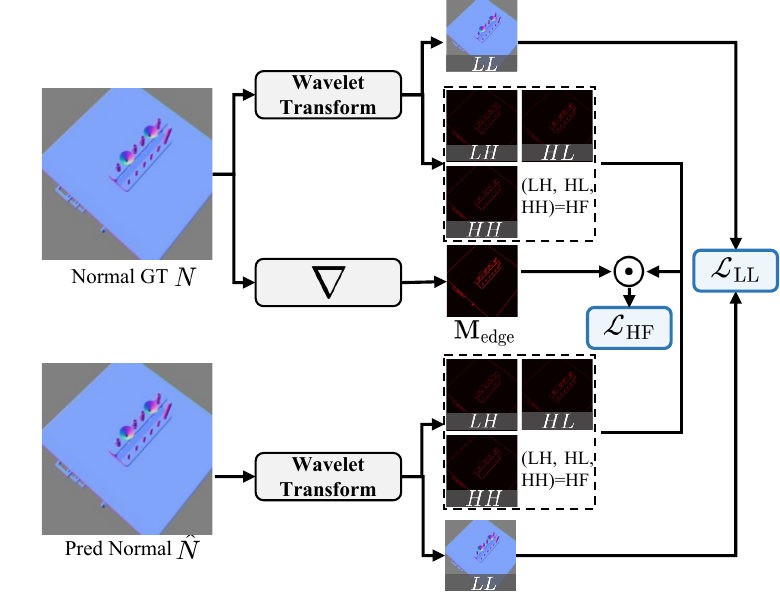}
    \vspace{-2mm}
    \caption{\textbf{Wavelet Edge-Aware Regularization.} Haar wavelet decomposes normals into low-frequency (LL) and high-frequency (LH, HL, HH) sub-bands. An edge mask $\bm{M}_{\text{edge}}$ enables: \ding{172} LL fidelity for overall shape; \ding{173} edge-aligned HF supervision for sharp boundaries.}
    \vspace{-7mm}
    \label{fig:wavelet}
\end{wrapfigure}

\paragraph{Wavelet Edge-Aware Regularization.}\label{par:wavelet_regularization} Laboratory glassware exhibits a distinctive geometric prior: sharp normal discontinuities occur primarily at object boundaries and structural edges (\emph{e.g.}, rims, bases, and liquid-glass interfaces), while interior regions exhibit smooth, continuous surfaces. Standard pixel-wise losses treat all regions uniformly, often over-smoothing edges to minimize global error. We address this through a wavelet-based regularization that provides edge-selective frequency supervision (\figref{fig:wavelet}).

Using the 2D Haar wavelet transform $\mathcal{W}$, we decompose both the predicted normal $\hat{\bm{N}}$ and ground truth $\bm{N}$:
\begin{equation}
\begin{aligned}
    \mathcal{W}(\hat{\bm{N}}) &= \{\hat{LL}, \hat{LH}, \hat{HL}, \hat{HH}\}, \\
    \mathcal{W}(\bm{N}) &= \{LL, LH, HL, HH\},
\end{aligned}
\end{equation}
where $LL \in \R^{3 \times \frac{H}{2} \times \frac{W}{2}}$ is the low-frequency approximation and $HF = [LH; HL; HH] \in \R^{9 \times \frac{H}{2} \times \frac{W}{2}}$ denotes the channel-wise concatenation of the three high-frequency sub-bands; predicted sub-bands use a hat, \emph{e.g.}, $\hat{LL}$ and $\hat{HF}$. We define the edge mask $\bm{M}_{\text{edge}} = \frac{1}{2}\left(\|\nabla_x \bm{N}\|_2 + \|\nabla_y \bm{N}\|_2\right)$, normalized to $[0,1]$ and downsampled to match the sub-band resolution, where $\nabla_x$ and $\nabla_y$ denote finite differences. The wavelet loss is then:
\begin{equation}
\begin{aligned}
    \Ls_{\text{LL}} &= \|\hat{LL} - LL\|_1, \\
    \Ls_{\text{HF}} &= \| \bm{M}_{\text{edge}} \odot (\hat{HF} - HF) \|_1, \\
    \Ls_{\text{wavelet}} &= \Ls_{\text{LL}} + \Ls_{\text{HF}}.
\end{aligned}
\end{equation}

The two terms target complementary geometric aspects: \ding{172} \textbf{Low-frequency fidelity}: supervises the $LL$ sub-band to ensure correct overall shape and smooth curvature alignment with the ground truth. \ding{173} \textbf{Edge-selective high-frequency alignment}: enforces $HF$ fidelity only at edges (weighted by $\bm{M}_{\text{edge}}$), preserving sharp boundary reconstruction without introducing constraints on interior regions. 

\paragraph{Total Loss.}
The final objective combines the normal/RGB losses with the wavelet regularization:
\begin{equation}
    \Ls_{\text{total}} = \Ls_{\text{normal}} + \lambda_{\text{rgb}} \Ls_{\text{rgb}} + \lambda_{\text{wv}} \Ls_{\text{wavelet}}.
\end{equation}
\section{Experiments}
\label{sec:experiments}

\setlength{\tabcolsep}{5pt}
\newcommand{\methodcite}[2]{\makebox[\linewidth][l]{#1\hfill(\citeauthor{#2})}}
\begin{table*}[!t]
\scriptsize
\caption{\textbf{Quantitative comparison on transparent object normal estimation}. We evaluate on ClearGrasp, our proposed TransNormal-Synthetic, and ClearPose datasets.
Metrics: mean angular error (Mean$\downarrow$, lower is better) and percentage of pixels within $11.25^\circ$ and $30^\circ$ thresholds ($\uparrow$, higher is better).
TransNormal achieves the best results across all three datasets.
The \colorbox{best}{best}, \colorbox{best2}{second best}, and \colorbox{best3}{third best} results are highlighted. $^\star$: diffusion-based; $^\dagger$: transformer-based. SA: SIGGRAPH Asia. (\S~\ref{ssec:transparent_exp})
}
\label{tab:transparent}
\scriptsize
\centering
\resizebox{\textwidth}{!}{
\begin{tabular}{>{\raggedright\arraybackslash}p{0.21\textwidth}|c|ccc|ccc|ccc|c}
\toprule

\multirow{2}{*}{Method} 
& \multirow{2}{*}{Venue}
& \multicolumn{3}{c|}{ClearGrasp (Synthetic)} 
& \multicolumn{3}{c|}{TransNormal-Synthetic} 
& \multicolumn{3}{c|}{ClearPose (Real-World)} 
& \textcolor{black}{Avg.} \\
& 
& Mean$\downarrow$ & $11.25^\circ\uparrow$ & $30^\circ\uparrow$
& Mean$\downarrow$ & $11.25^\circ\uparrow$ & $30^\circ\uparrow$
& Mean$\downarrow$ & $11.25^\circ\uparrow$ & $30^\circ\uparrow$
& Rank            \\
\midrule

\methodcite{Omnidata}{eftekhar2021omnidata}
& ICCV 21
& 36.9 & 15.1 & 49.1
& 11.3 & 80.9 & 89.3
& 48.3 & 10.8 & 33.8
& 12.3  \\

\methodcite{Omnidata V2$^\dagger$}{kar20223d}
& CVPR 22
& 33.8 & 18.3 & 55.9
& 8.2 & 87.0 & 92.6
& 51.7 & 13.8 & 33.2
& 10.9  \\

\methodcite{GeoWizard$^{\star}$}{fu2024geowizard}
& ECCV 24
& 31.3 & 20.8 & 59.5
& 9.4 & 78.9 & 95.0
& 36.8 & 14.2 & 49.7
& 10.1  \\

\methodcite{StableNormal$^{\star}$}{ye2024stablenormal}
& SA 24
& 32.0 & 17.5 & 65.3
& 7.6 & 86.8 & 96.3
& 37.1 & 14.1 & 57.5
& 8.9  \\

\methodcite{Marigold$^{\star}$}{ke2024repurposing}
& CVPR 24
& 27.6 & 31.0 & 65.3
& \cellcolor{best3}6.2 & \cellcolor{best3}90.4 & 96.3
& 33.0 & 25.5 & 57.5
& 6.3  \\

\methodcite{DSINE}{bae2024dsine}
& CVPR 24
& 25.7 & 26.4 & 68.6
& 13.2 & 70.3 & 90.7
& 40.2 & 15.9 & 46.3
& 9.6  \\

\methodcite{Diff-E2E-FT$^{\star}$}{martingarcia2024diffusione2eft}
& WACV 25
& 22.6 & \cellcolor{best2}42.1 & 73.3
& \cellcolor{best2}5.2 & \cellcolor{best2}91.9 & 97.0
& 32.0 & \cellcolor{best3}32.5 & 59.4
& \cellcolor{best2}3.3  \\

\methodcite{GenPercept$^{\star}$}{xu2024matters}
& ICLR 25
& 25.8 & 30.3 & 70.9
& 6.9 & 87.6 & 97.0
& 31.6 & 31.2 & \cellcolor{best3}63.0
& \cellcolor{best3}4.2  \\

\methodcite{Lotus-G$^\star$}{he2024lotus}
& ICLR 25
& \cellcolor{best2}21.7 & \cellcolor{best3}39.7 & \cellcolor{best3}75.4
& 8.2 & 82.3 & 96.7
& 31.8 & 28.8 & 60.4
& 5.2  \\

\methodcite{Lotus-D$^\star$}{he2024lotus}
& ICLR 25
& \cellcolor{best3}21.9 & 37.0 & \cellcolor{best2}75.7
& 9.0 & 80.9 & \cellcolor{best3}97.1
& \cellcolor{best3}31.3 & 23.2 & 59.5
& 5.3  \\

\methodcite{MoGe-2$^{\dagger}$}{wang2025moge2}
& NeurIPS 25
& 26.6 & 17.0 & 64.2
& \cellcolor{best3}6.2 & 90.1 & 96.8
& 36.2 & 14.3 & 48.3
& 7.8  \\

\methodcite{Diception$^\star$}{zhao2025diception}
& NeurIPS 25
& 29.5 & 25.8 & 65.3
& 7.1 & 88.3 & \cellcolor{best2}97.3
& \cellcolor{best2}31.0 & \cellcolor{best2}33.8 & \cellcolor{best2}63.5
& 5.0  \\

\midrule

\makebox[\linewidth][l]{\textbf{TransNormal}\hfill\textbf{(Ours)}}
& -
& \cellcolor{best}16.4 & \cellcolor{best}51.7 & \cellcolor{best}85.0
& \cellcolor{best}4.1 & \cellcolor{best}93.5 & \cellcolor{best}98.2
& \cellcolor{best}26.3 & \cellcolor{best}35.9 & \cellcolor{best}69.8
& \cellcolor{best}1.0  \\

\bottomrule
\end{tabular}
}
\vspace{-2mm}
\end{table*}

\subsection{Implementation Details}
We implement the proposed TransNormal by fine-tuning Stable Diffusion 2~\cite{rombach2022high}.
During training, the VAE encoder and decoder are kept frozen, while the U-Net parameters and the linear projector are updated. The task embeddings $s_{\text{n}}$ and $s_{\text{rgb}}$ remain fixed. For the DINOv3 encoder, we use patch size $p=16$.
For optimization, we use the AdamW~\cite{loshchilov2019decoupled} optimizer with a learning rate of $3\times10^{-5}$. 
We apply random horizontal flipping for data augmentation during training.
All models are trained on 8 NVIDIA A100 GPUs (80G) with a total batch size of 32 for 15,000 steps. 
During inference, we directly predict the normal map in a single inference step.
For loss weights, we set $\lambda_{\text{rgb}} = 1.0$ and $\lambda_{\text{wv}} = 0.1$, with equal weights for the $LL$ and edge high-frequency terms in the wavelet regularization.
\subsection{Environment Setup}
\paragraph{Training Data.}
This work aims to achieve strong performance using relatively limited supervised data. The normal estimation task is trained solely on a collection of synthetic data. During training, we sample from the following datasets with a ratio of \textbf{35:15:45:5}:
\ding{172}~\emph{ClearGrasp}~\cite{sajjan2020clear} (35\%): a dataset for transparent objects containing 45,454 synthetic normal images;
\ding{173}~\emph{TransNormal-Synthetic} (15\%): a Blender-rendered dataset of laboratory scenes with transparent glassware introduced in this work, providing 3,555 training and 395 testing samples with pixel-accurate normals, depth, and segmentation masks (details in Appendix~\ref{sec:dataset});
\ding{174}~\emph{Hypersim}~\cite{roberts2021hypersim} (45\%): a photorealistic synthetic dataset of 461 indoor scenes, from which we utilize the official training split retaining 39,648 samples after filtering, resized to $576 \times 768$;
\ding{175}~\emph{Virtual KITTI}~\cite{cabon2020virtual} (5\%): a synthetic street-scene dataset covering five urban scenes, from which we use four scenes comprising 33,580 samples, cropped to $352 \times 1216$. 
\begin{figure*}[t]
    \centering
    \setlength{\tabcolsep}{1pt}
    \renewcommand{\arraystretch}{0.6}
    \newcommand{\imgw}{0.132\textwidth}
    %
    %
    \begin{tabular}{@{}c@{\hspace{1pt}}cc|ccccc@{}}
        & \small Input/Mask & \small GT & \small Lotus & \small MoGe-2 & \small E2E-FT & \small GenPercept & \small \textbf{Ours} \\[2pt]
        %
        \multirow{2}{*}[8ex]{\rotatebox{90}{\small TransNormal-Synthetic}} &
        \includegraphics[width=\imgw]{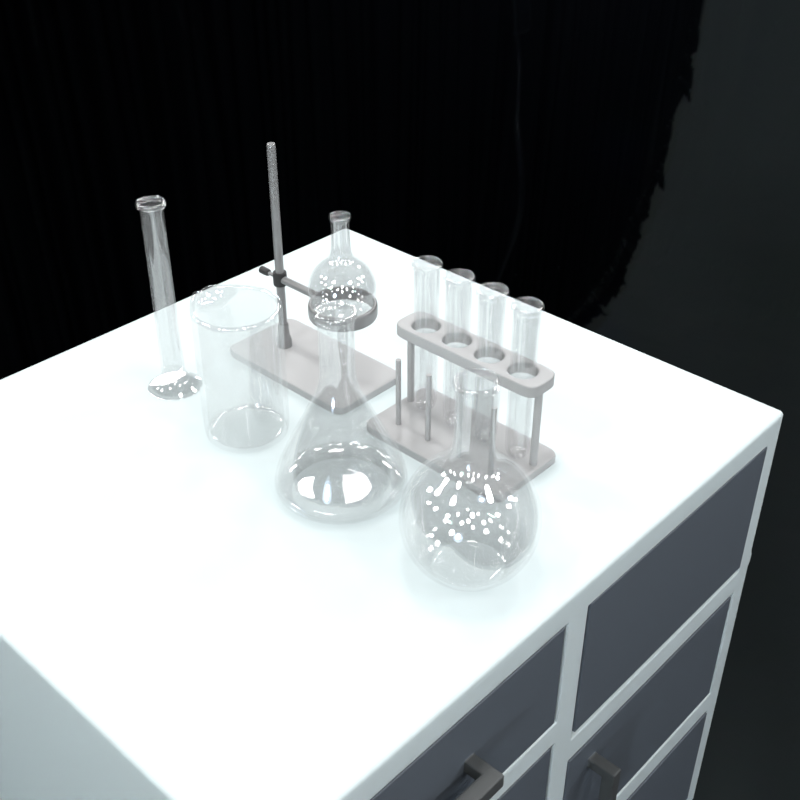} &
        \includegraphics[width=\imgw]{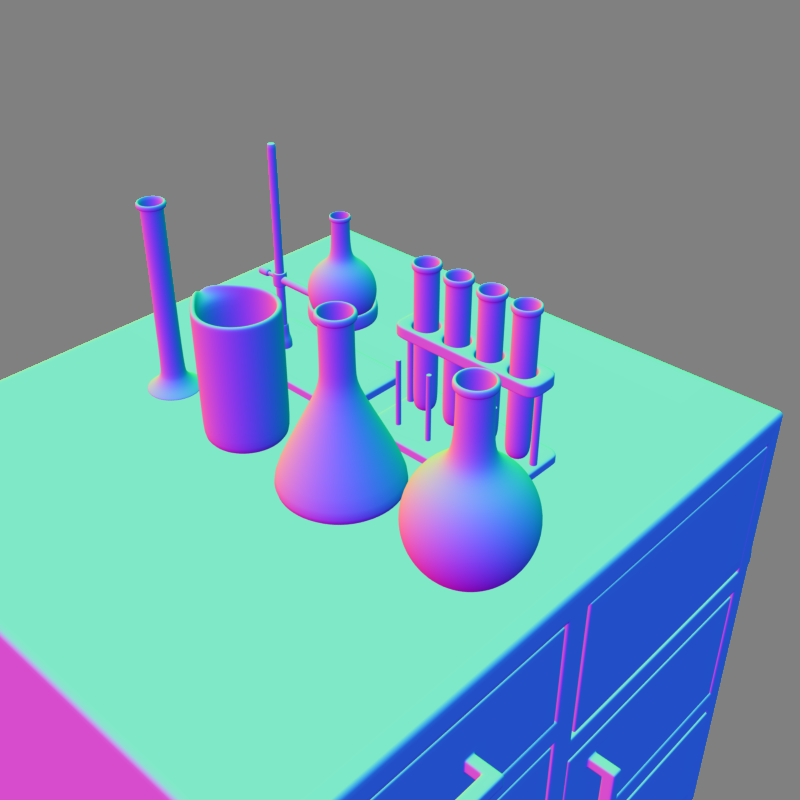} &
        \includegraphics[width=\imgw]{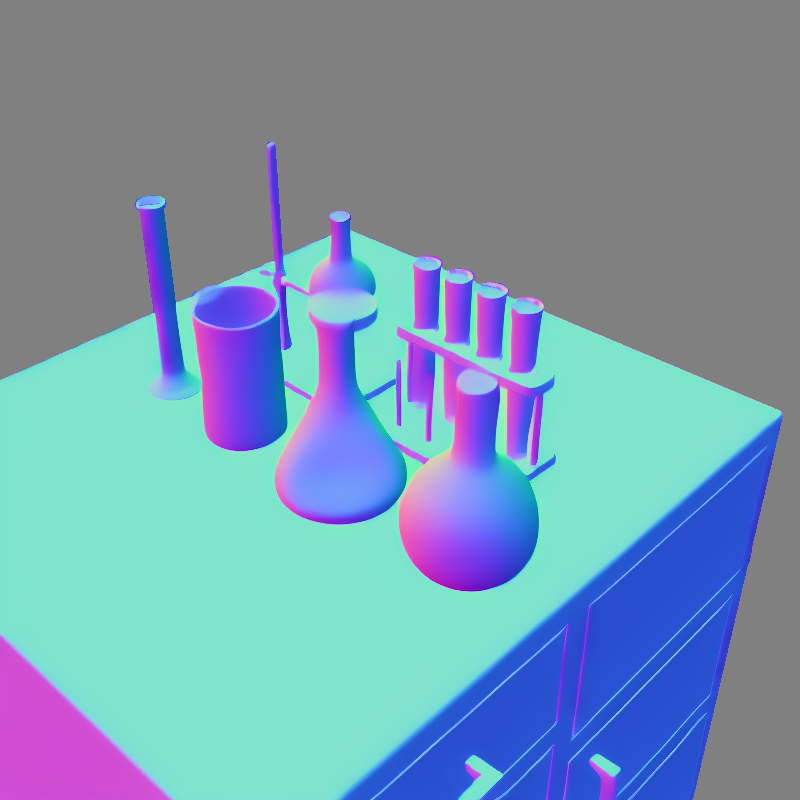} &
        \includegraphics[width=\imgw]{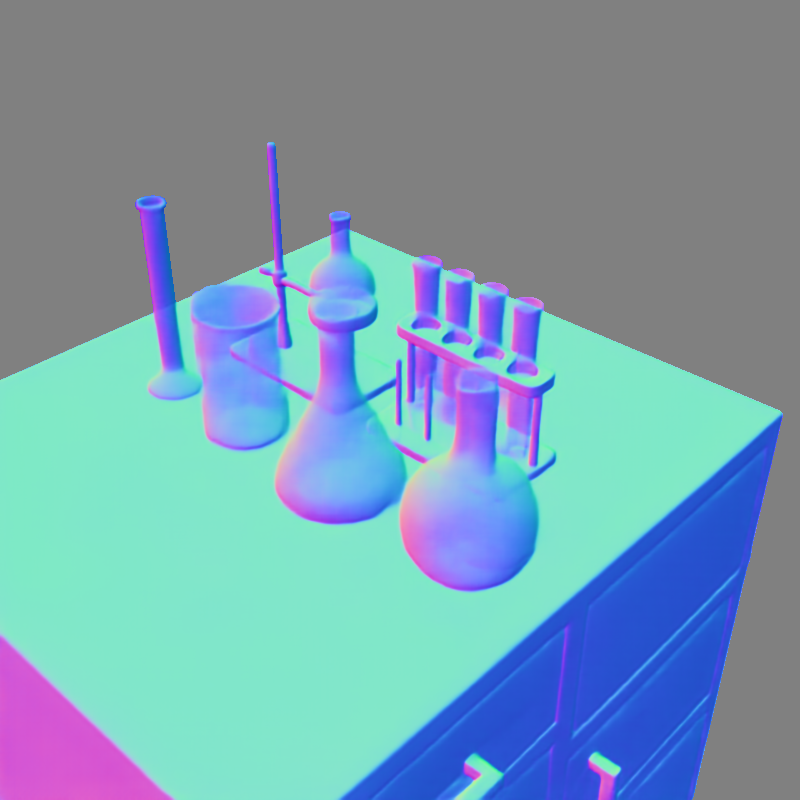} &
        \includegraphics[width=\imgw]{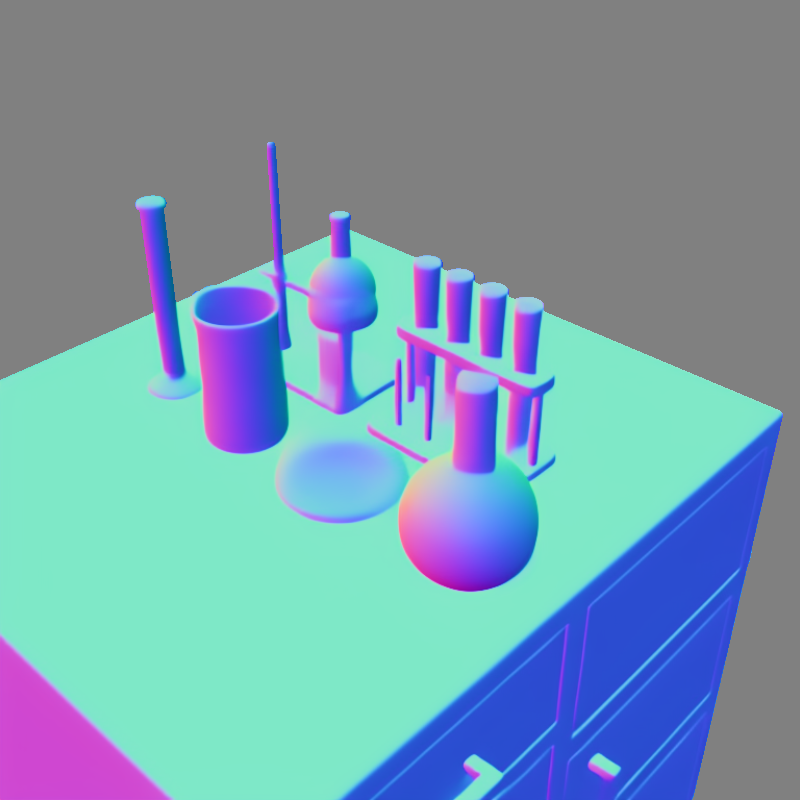} &
        \includegraphics[width=\imgw]{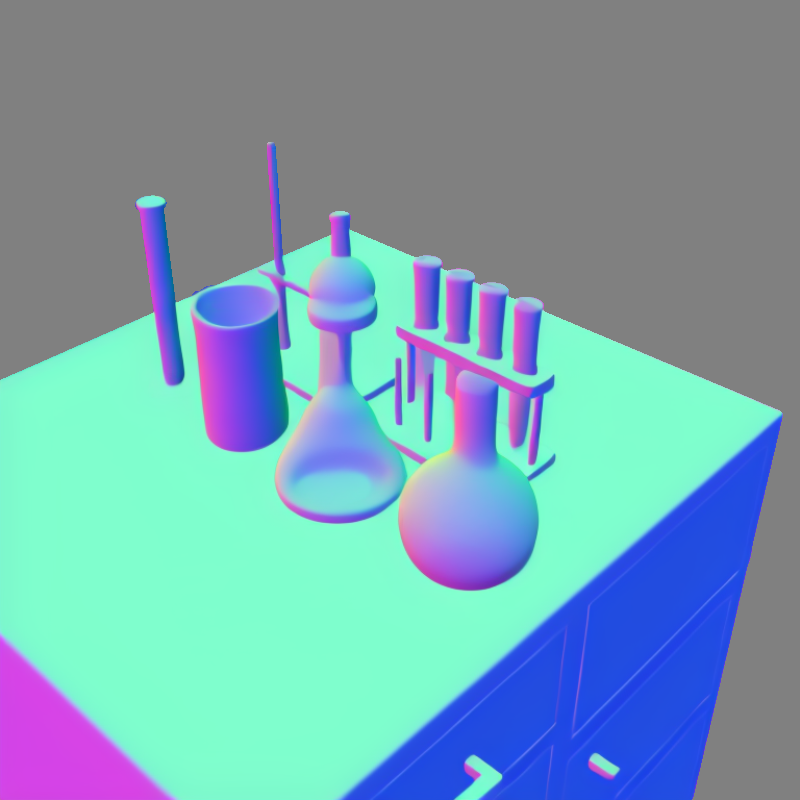} &
        \includegraphics[width=\imgw]{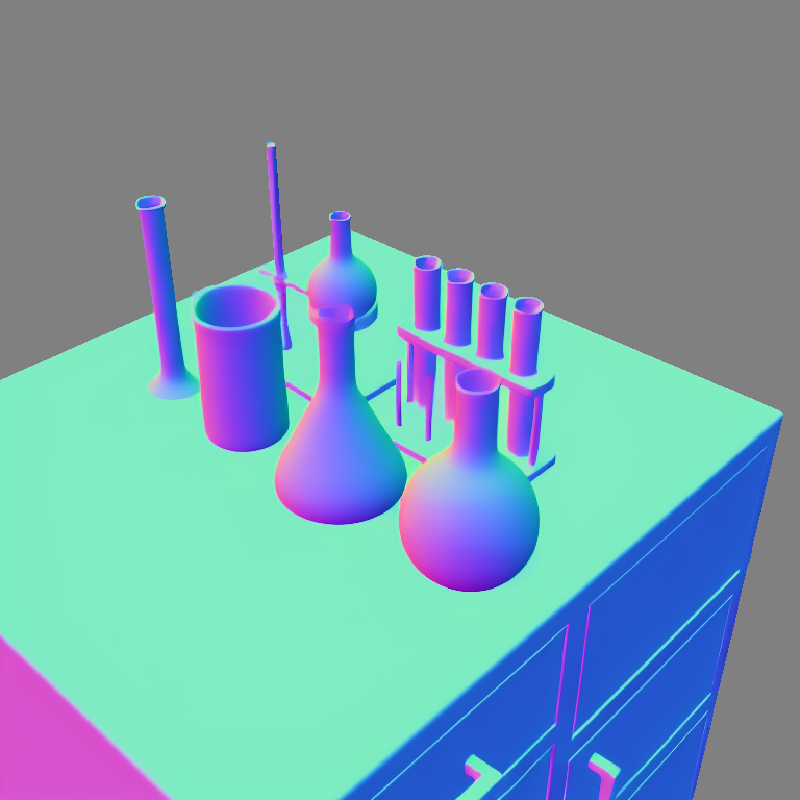} \\
        &
        \includegraphics[width=\imgw]{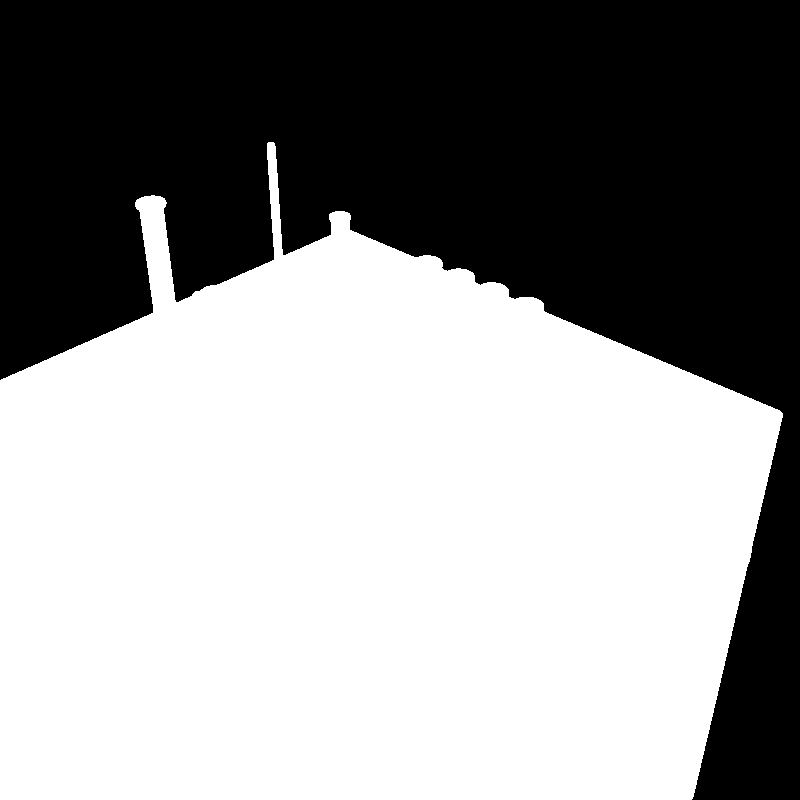} &
        \includegraphics[width=\imgw]{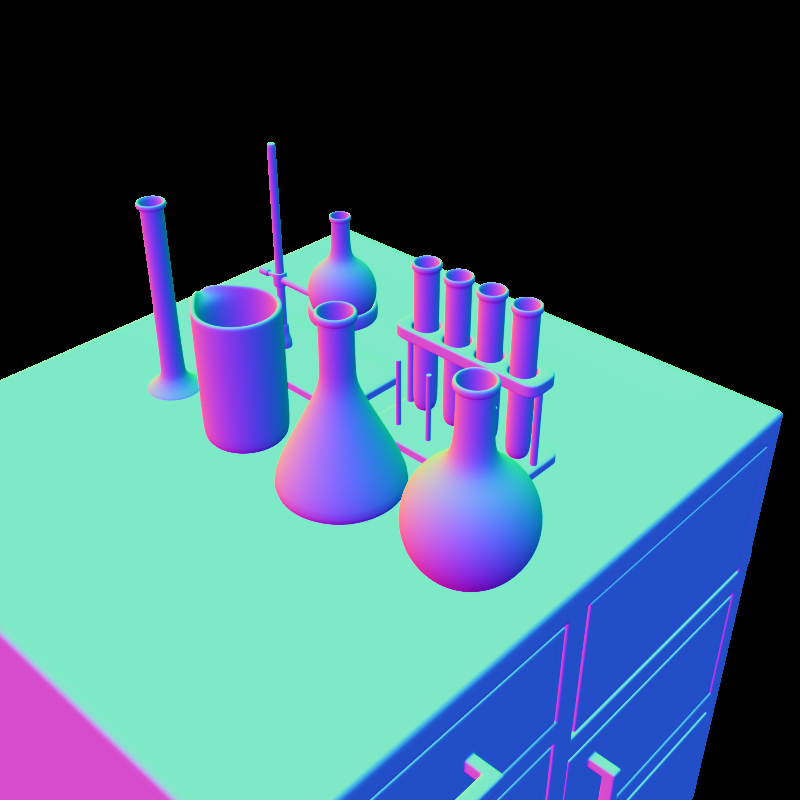} &
        \includegraphics[width=\imgw]{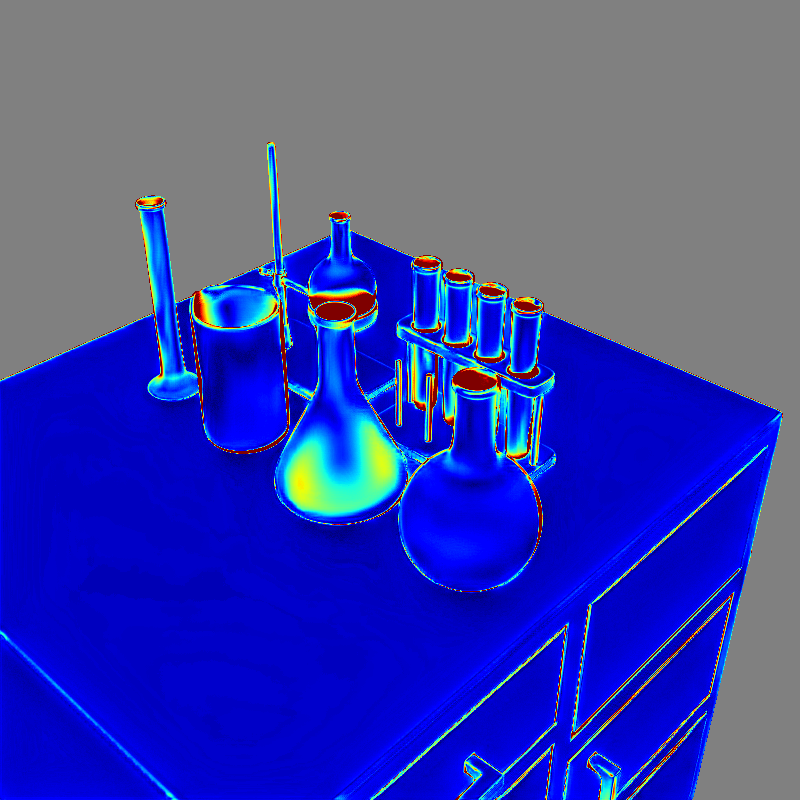} &
        \includegraphics[width=\imgw]{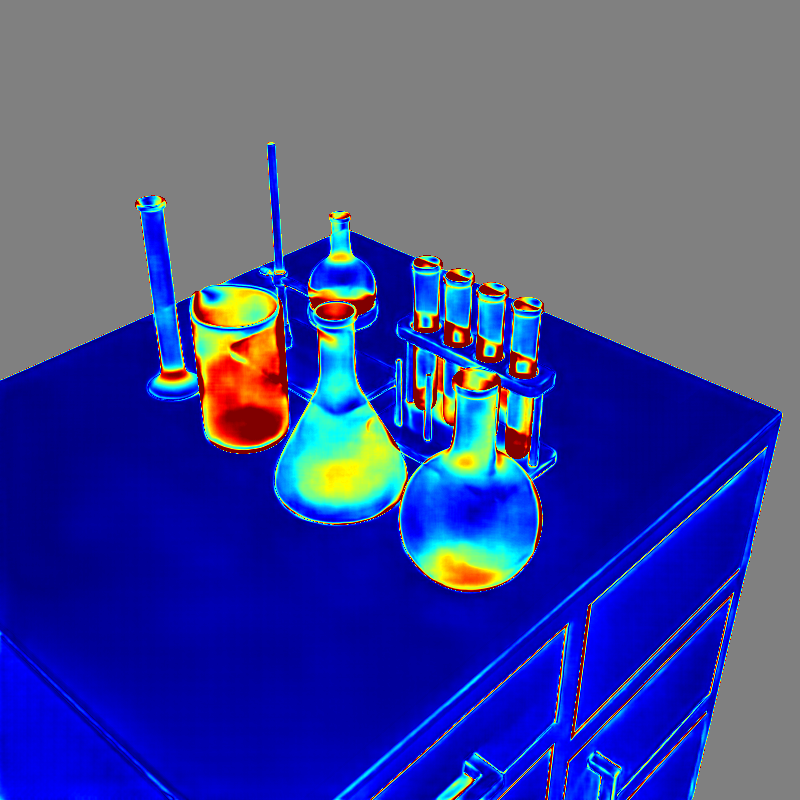} &
        \includegraphics[width=\imgw]{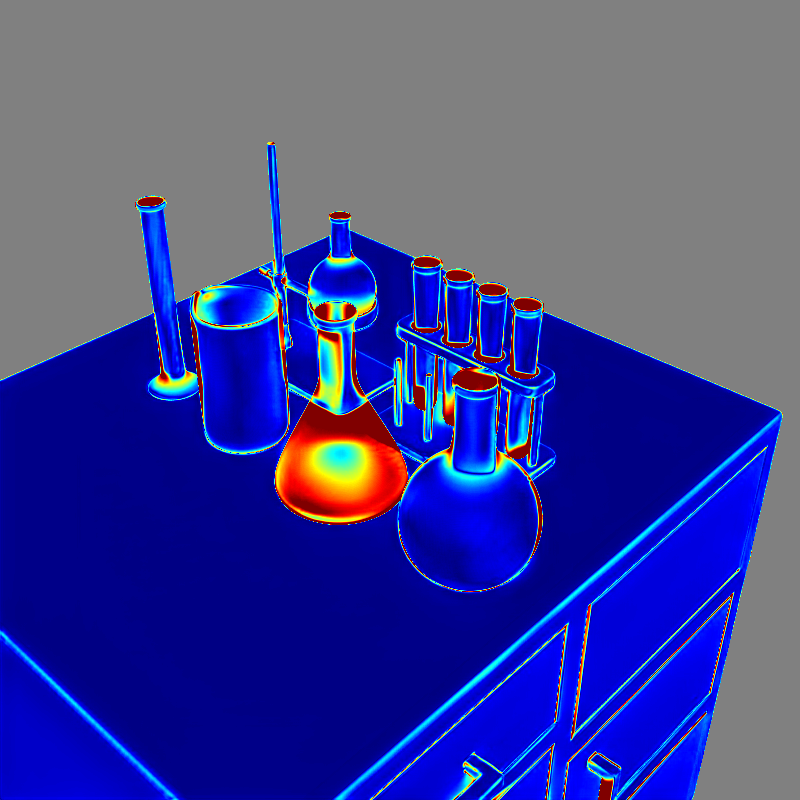} &
        \includegraphics[width=\imgw]{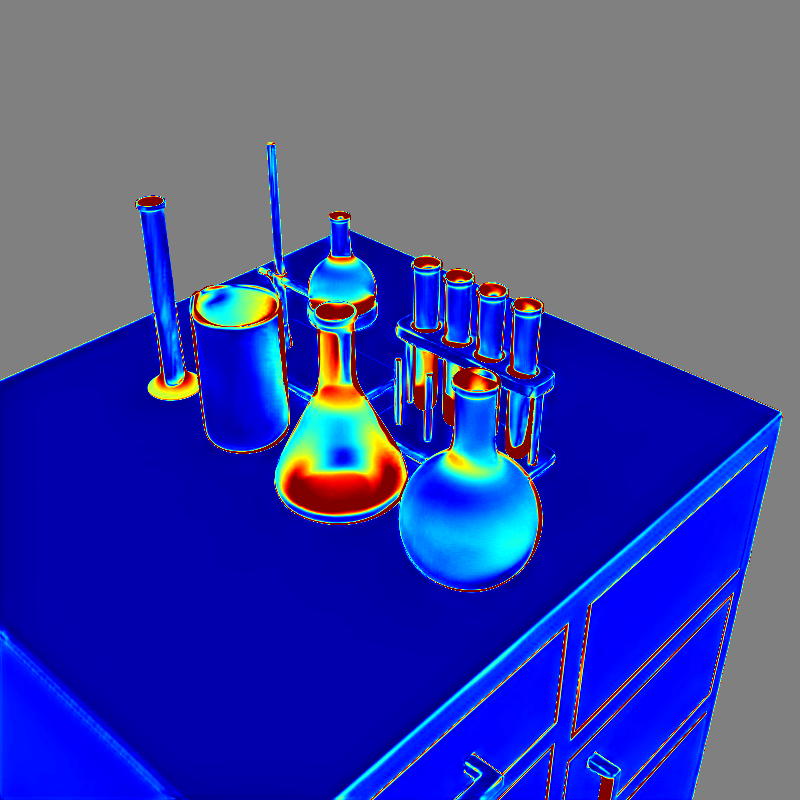} &
        \includegraphics[width=\imgw]{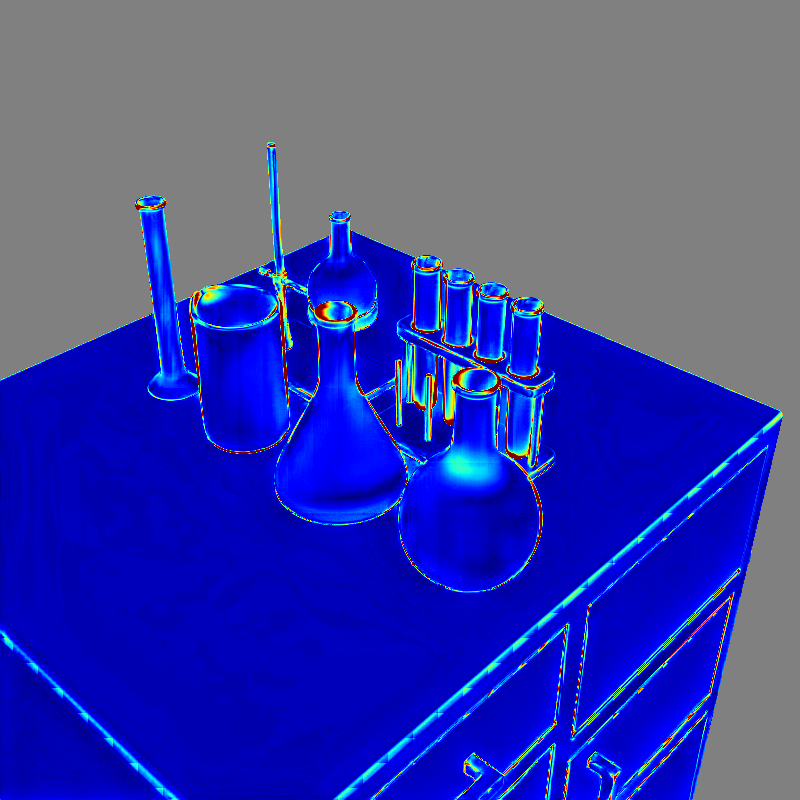} \\[2pt]
        %
        \multirow{2}{*}[3.5ex]{\rotatebox{90}{\small ClearGrasp}} &
        \includegraphics[width=\imgw]{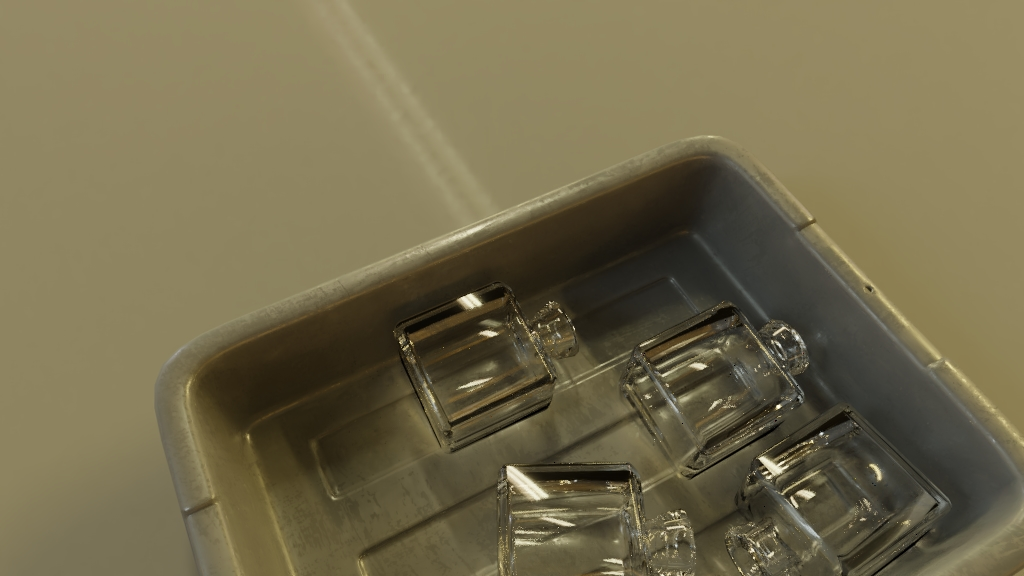} &
        \includegraphics[width=\imgw]{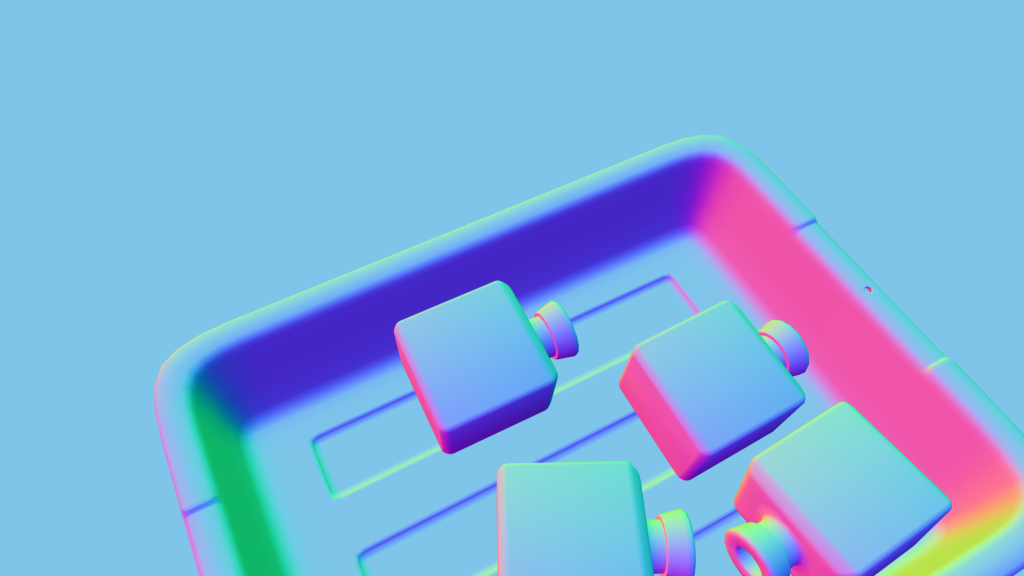} &
        \includegraphics[width=\imgw]{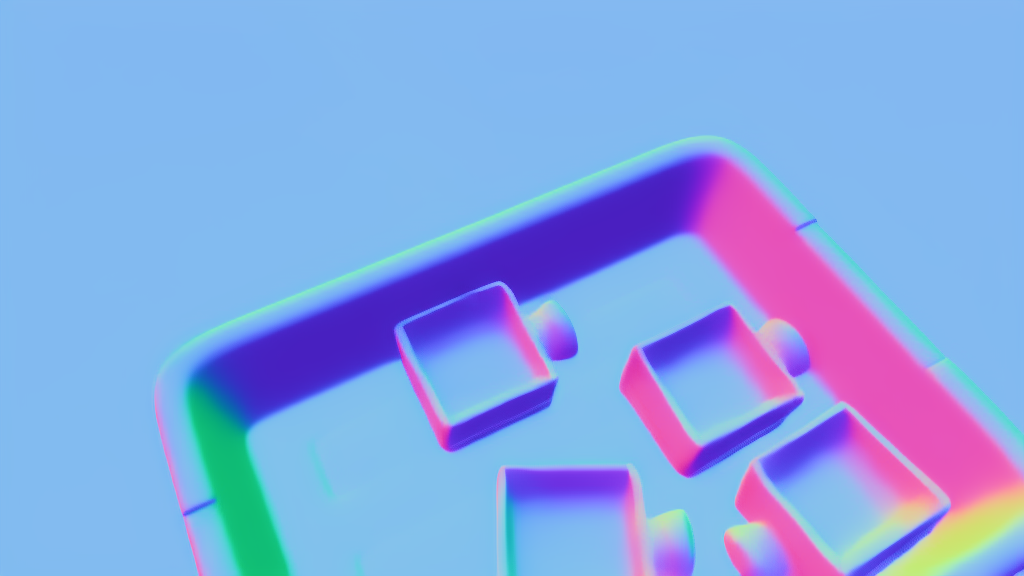} &
        \includegraphics[width=\imgw]{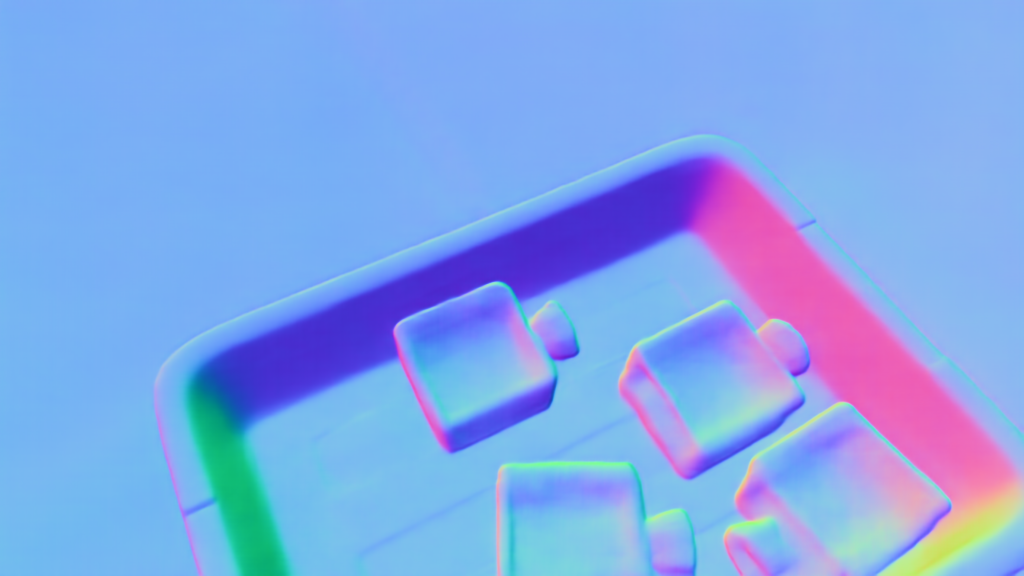} &
        \includegraphics[width=\imgw]{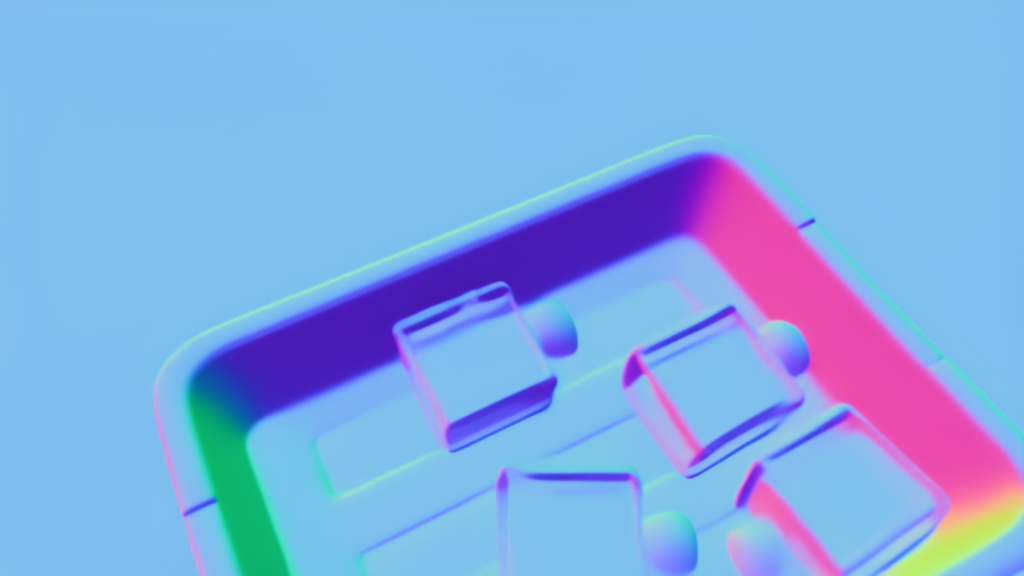} &
        \includegraphics[width=\imgw]{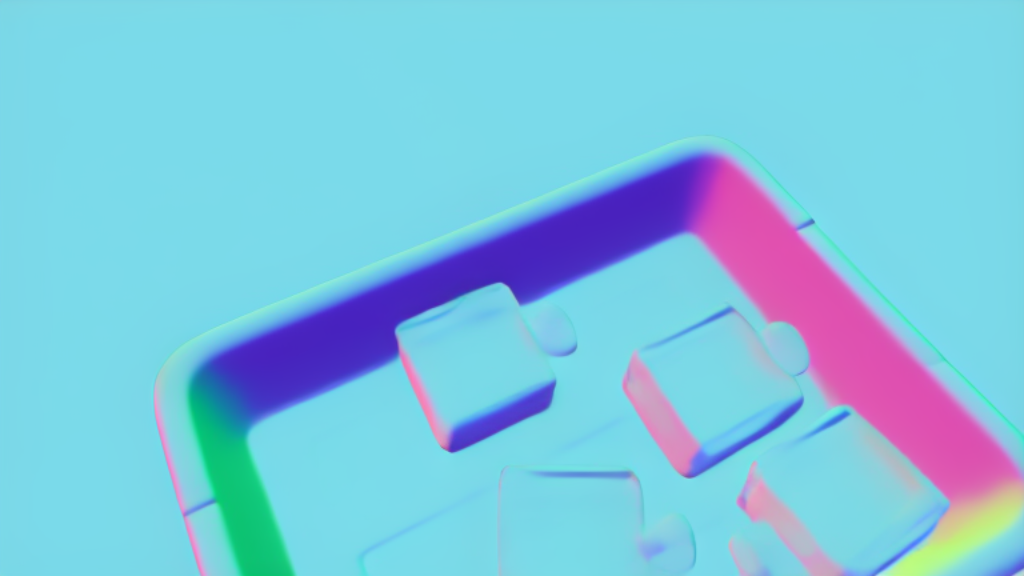} &
        \includegraphics[width=\imgw]{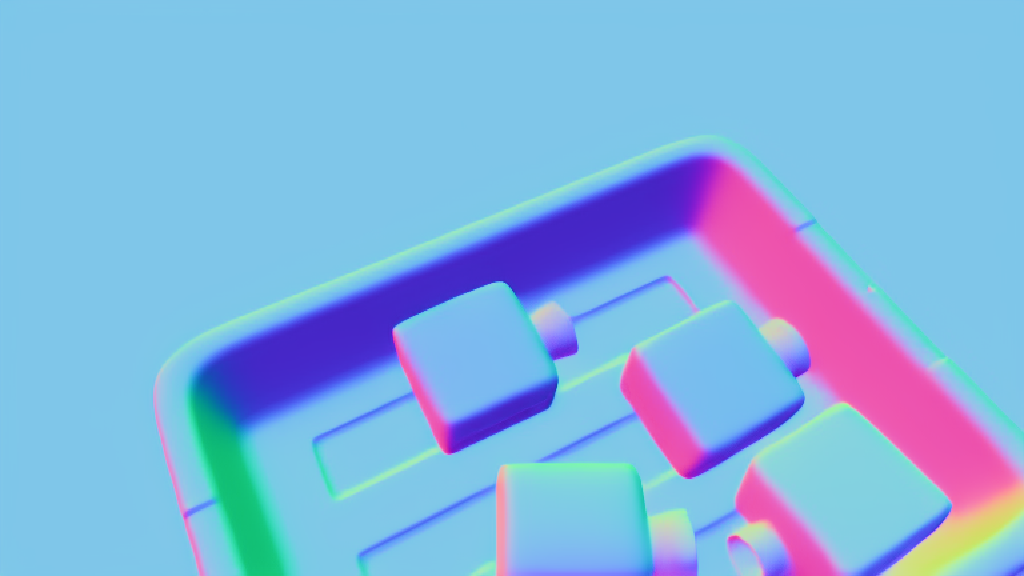} \\
        &
        \includegraphics[width=\imgw]{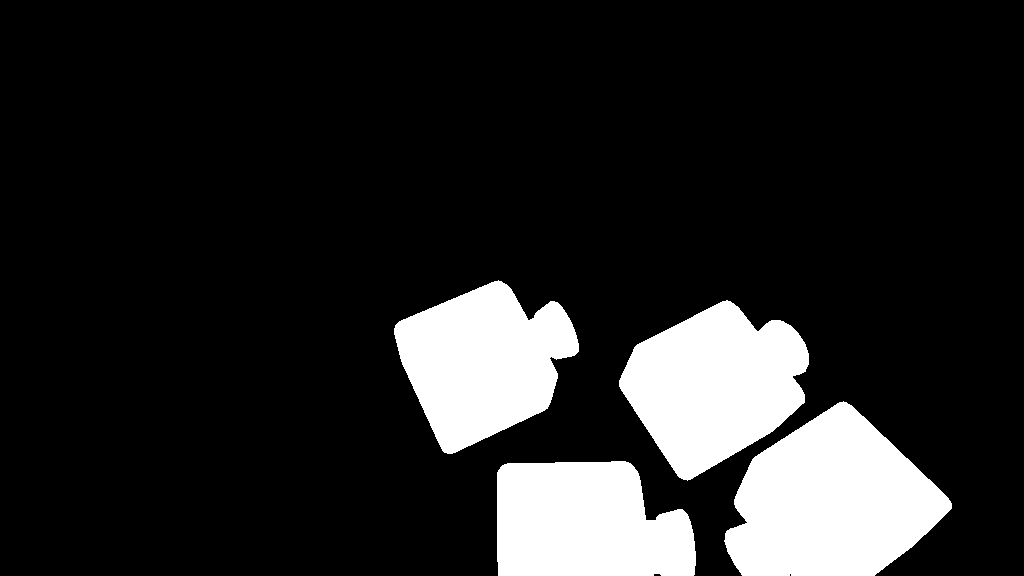} &
        \includegraphics[width=\imgw]{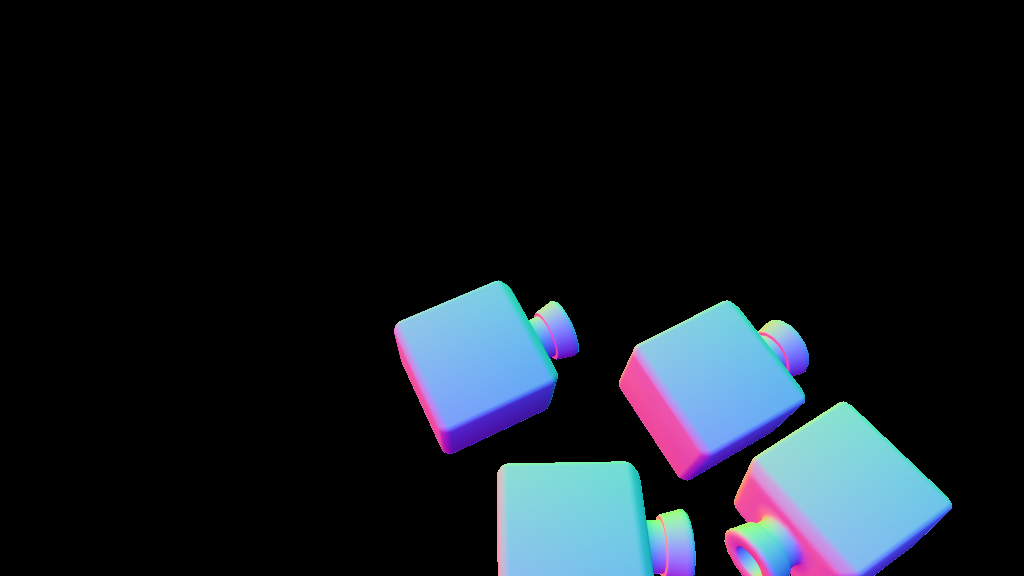} &
        \includegraphics[width=\imgw]{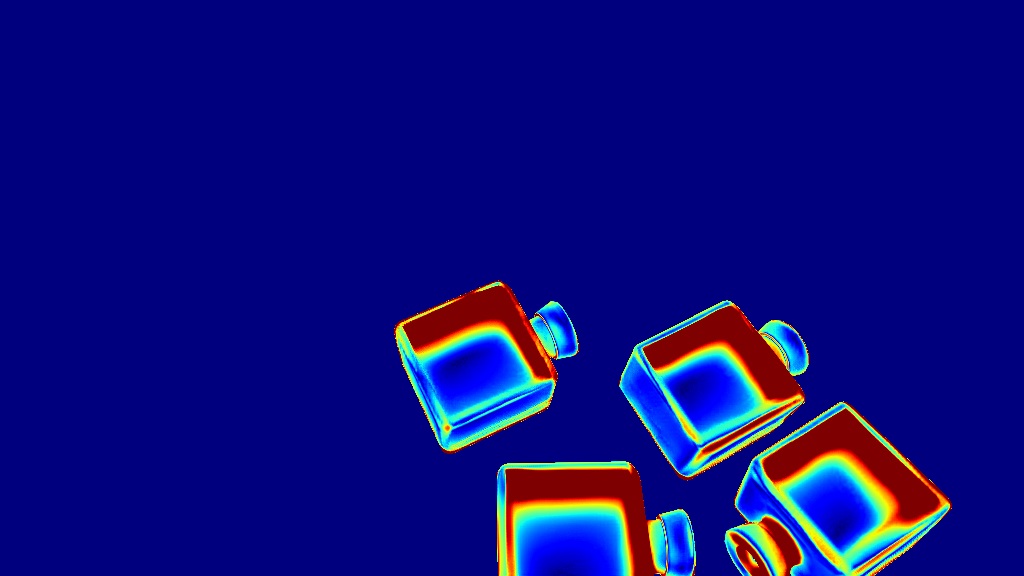} &
        \includegraphics[width=\imgw]{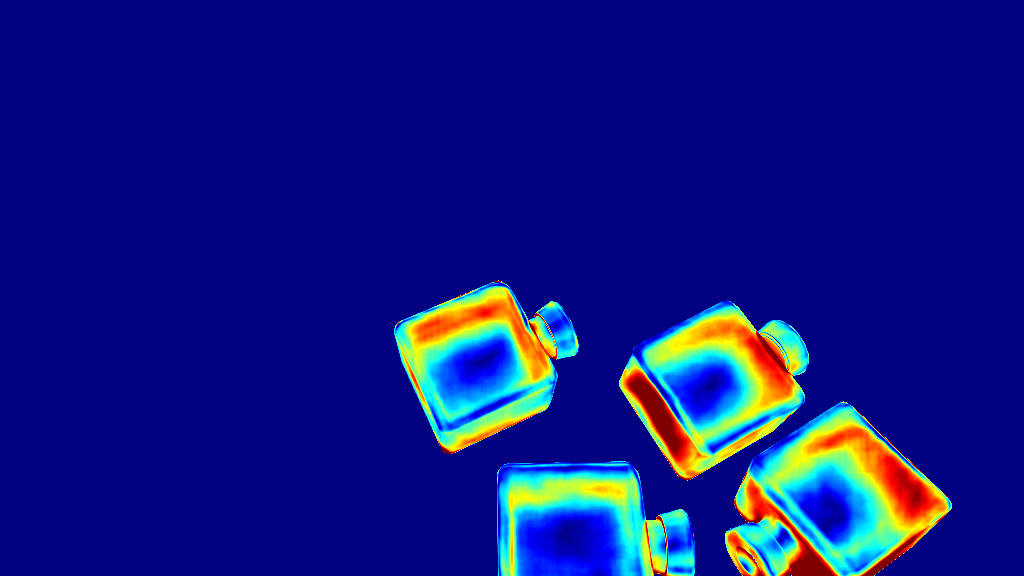} &
        \includegraphics[width=\imgw]{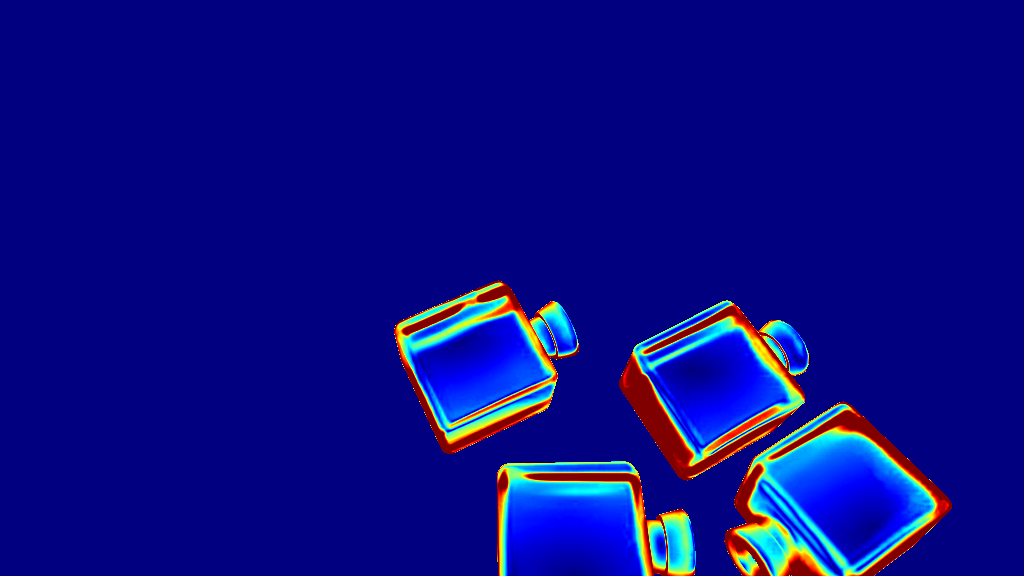} &
        \includegraphics[width=\imgw]{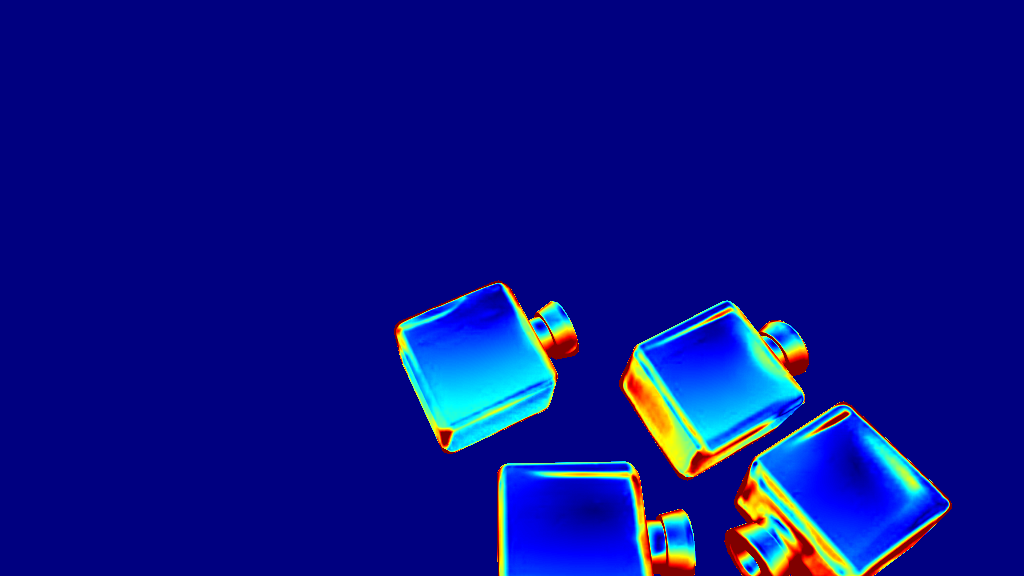} &
        \includegraphics[width=\imgw]{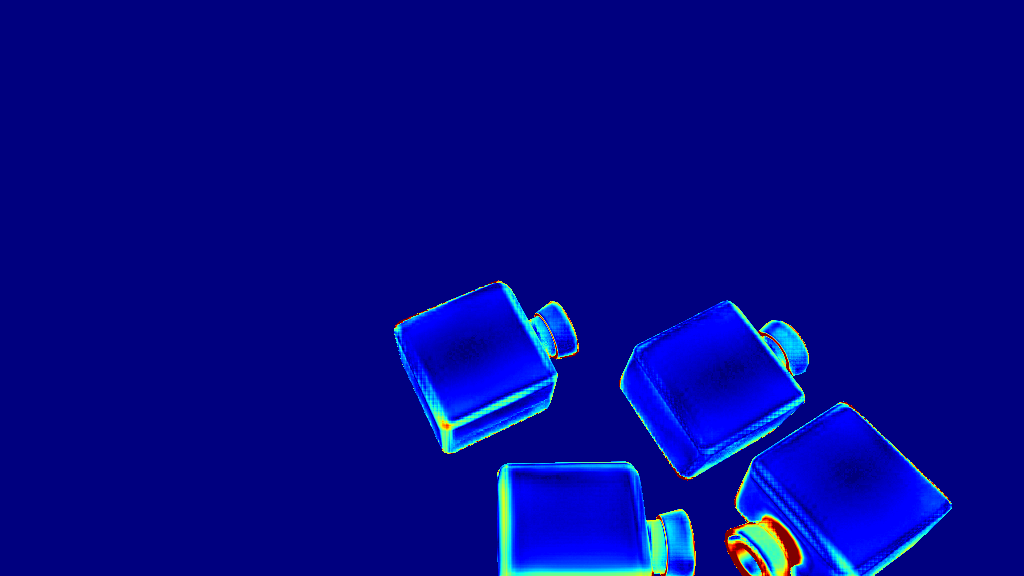} \\[2pt]
        %
        \multirow{2}{*}[2.5ex]{\rotatebox{90}{\small ClearPose}} &
        \includegraphics[width=\imgw]{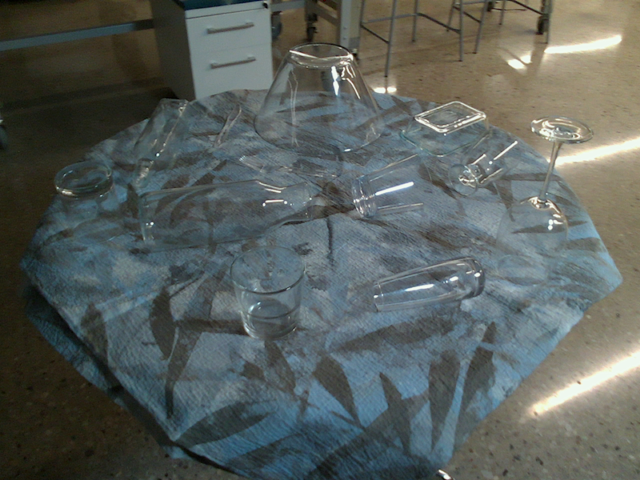} &
        \includegraphics[width=\imgw]{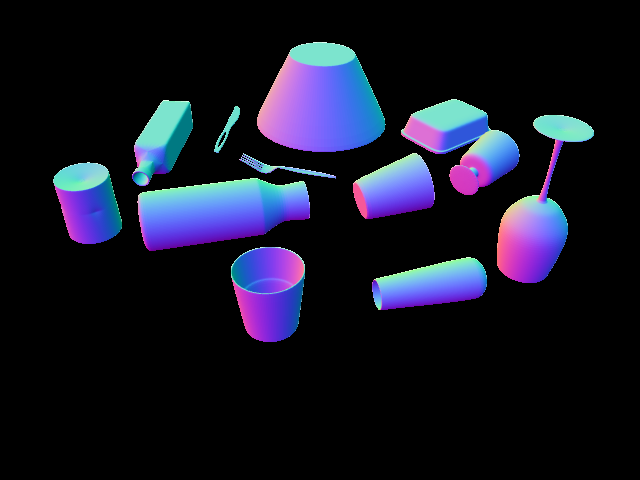} &
        \includegraphics[width=\imgw]{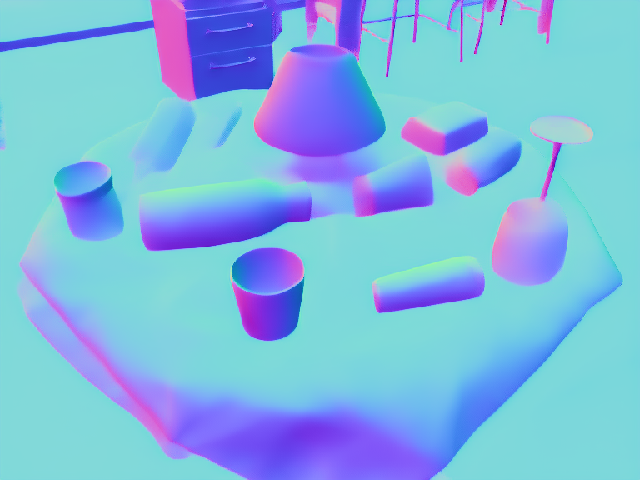} &
        \includegraphics[width=\imgw]{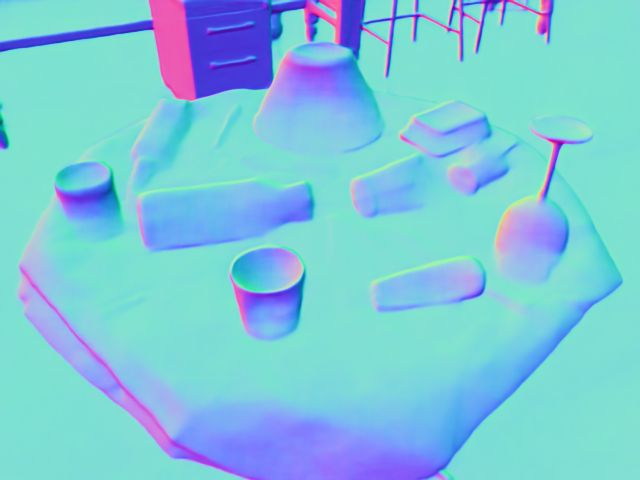} &
        \includegraphics[width=\imgw]{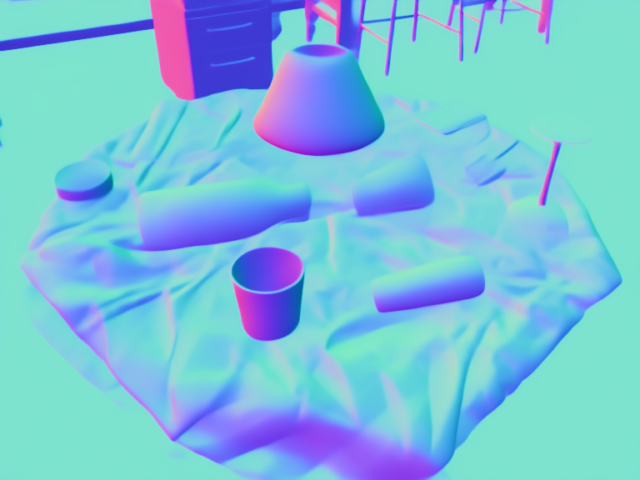} &
        \includegraphics[width=\imgw]{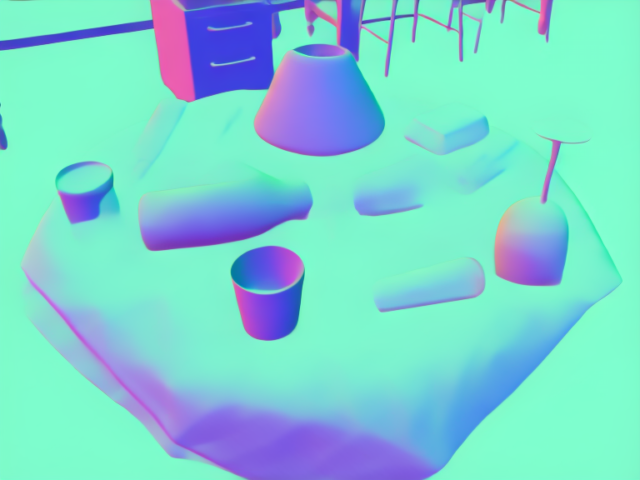} &
        \includegraphics[width=\imgw]{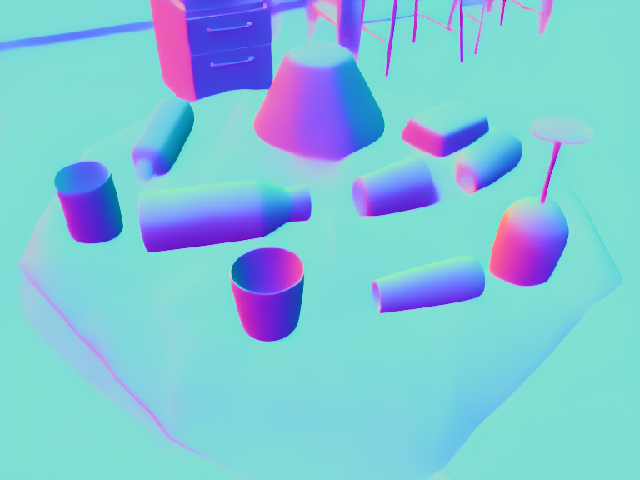} \\
        &
        \includegraphics[width=\imgw]{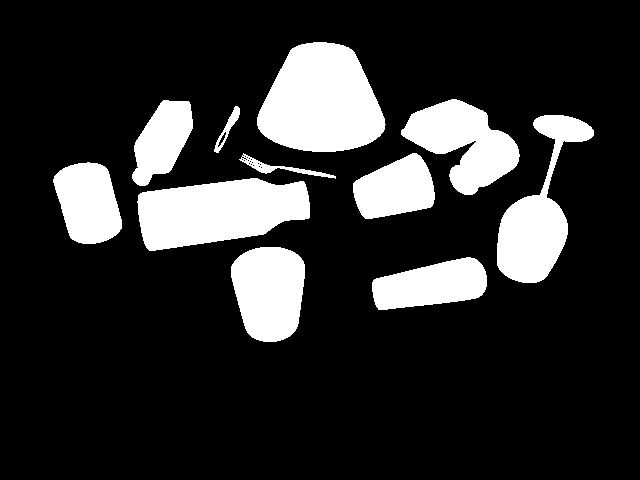} &
        \includegraphics[width=\imgw]{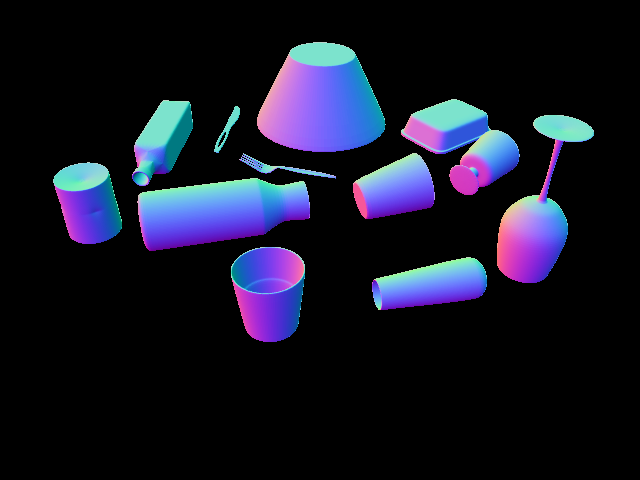} &
        \includegraphics[width=\imgw]{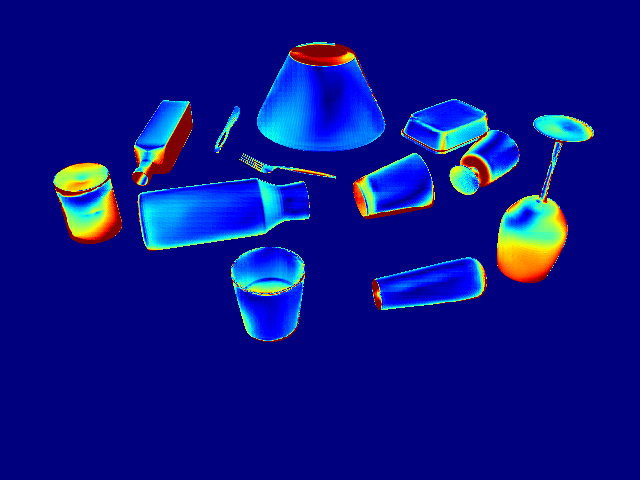} &
        \includegraphics[width=\imgw]{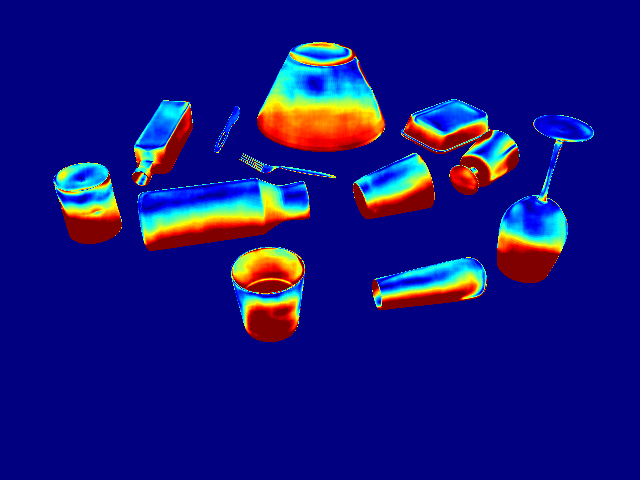} &
        \includegraphics[width=\imgw]{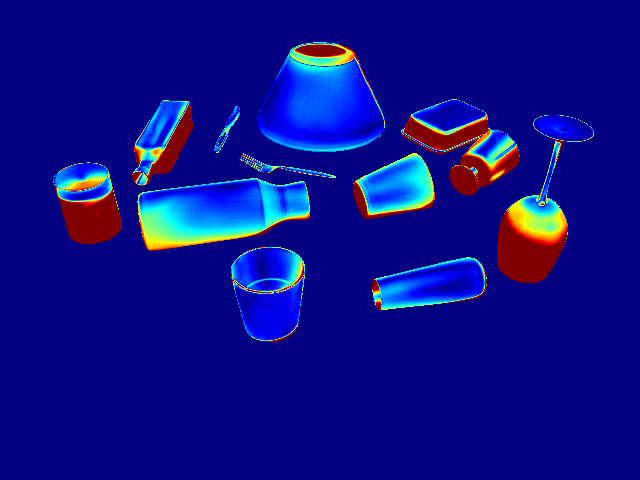} &
        \includegraphics[width=\imgw]{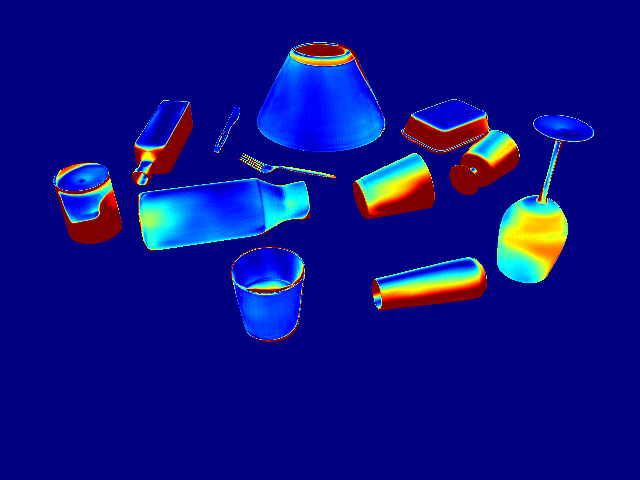} &
        \includegraphics[width=\imgw]{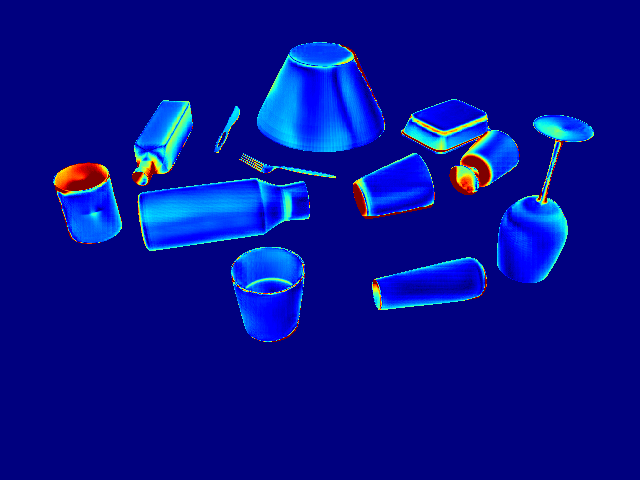} \\
    \end{tabular}
    \caption{\textbf{Qualitative comparison on transparent object normal estimation.}
    We compare our method against state-of-the-art approaches across TransNormal-Synthetic, ClearGrasp, and ClearPose datasets.
    For each dataset, the top row shows predicted normals and the bottom row shows angular error maps (blue: low, red: high).
    Notably, even on ClearPose, an extremely challenging real-world dataset with diverse transparent objects under cluttered scenes, our method achieves superior zero-shot performance compared to other approaches.
    Existing methods produce blurry or incorrect normals on transparent regions due to refraction, while our method recovers sharp and accurate surface geometry. Please zoom in \faSearch~for details. (\S~\ref{par:qual_comparison})}
    \label{fig:comparison}
\end{figure*}

\paragraph{Evaluation Data.}
\label{ssec:eval-dataset}
We evaluate TransNormal on transparent object normal estimation using: the synthetic test split of \emph{ClearGrasp}~\cite{sajjan2020clear} (408 samples), the held-out test set of \emph{TransNormal-Synthetic} (395 samples), and \emph{ClearPose}~\cite{chen2022clearpose} (120 samples). ClearPose is a challenging real-world dataset with diverse transparent objects under varying lighting conditions; we use it for zero-shot evaluation (not included in training) to assess generalization. For ClearPose, we use the subset with available meshes and recompute normals by reprojecting the ground-truth mesh, evaluating only within the transparent object mask. We apply this protocol to all compared methods.
\paragraph{Baselines.}\label{par:baselines}
We compare TransNormal against representative normal estimation methods on the task of transparent object normal reconstruction.
The baselines include models trained on opaque or general scenes (Omnidata~\cite{eftekhar2021omnidata}, Omnidata V2~\cite{kar20223d}, DSINE~\cite{bae2024dsine}) and diffusion-based dense prediction methods (GeoWizard~\cite{fu2024geowizard}, StableNormal~\cite{ye2024stablenormal}, Marigold~\cite{ke2024repurposing}, Lotus~\cite{he2024lotus}, Diffusion-E2E-FT~\cite{martingarcia2024diffusione2eft}, GenPercept~\cite{xu2024matters}, MoGe-2~\cite{wang2025moge2}, Diception~\cite{zhao2025diception}).

\subsection{Qualitative Results}\label{ssec:qual_results}

\paragraph{Comparison with Baselines.}\label{par:qual_comparison}
\Figref{fig:comparison} presents qualitative comparisons between TransNormal and state-of-the-art methods across three transparent object benchmarks. For each dataset, the first row shows predicted normal maps, and the second row displays error maps within the transparent object mask (blue: low error, red: high error). Existing methods produce severely distorted normal predictions in transparent regions, as they are misled by refracted background textures. In contrast, TransNormal leverages DINOv3 semantic guidance to provide high-level shape understanding, enabling accurate geometry recovery even under challenging refractive conditions. Additional qualitative results are provided in Appendix~\ref{ssec:extended_comparison}, and in-the-wild generalization examples are shown in Appendix~\ref{ssec:cross_category_zeroshot}.

\subsection{Quantitative Results}
\label{ssec:transparent_exp}

\paragraph{Metrics.}
\label{sec:metrics}
Following prior works~\cite{bae2024dsine,ye2024stablenormal,he2024lotus}, we measure the \emph{mean angular error} (Mean$\downarrow$) and the percentage of pixels within $11.25^\circ$ and $30^\circ$ thresholds ($\uparrow$).
The Avg.~Rank is computed by ranking each method on every metric across all three datasets, then averaging all nine per-metric ranks.

\textbf{Results on ClearGrasp.}
On the synthetic ClearGrasp benchmark, TransNormal achieves a mean angular error of 16.4°, outperforming the previous best method Lotus-G (21.7°) by 24.4\% relative improvement. Our method achieves 51.7\% accuracy at the strict 11.25° threshold and 85.0\% at 30°, indicating better fine-grained geometric accuracy. These results suggest that DINOv3 semantic guidance helps reduce ambiguities caused by refraction in transparent objects, where discriminative methods like DSINE (25.7°) and recent diffusion-based methods like Marigold (27.6°) struggle due to misleading local texture cues.

\textbf{Results on TransNormal-Synthetic.}
Our proposed synthetic benchmark provides controlled evaluation of transparent object understanding. TransNormal achieves the best performance with 4.1° mean error and 93.5\% accuracy at 11.25°, surpassing the strong baseline Diffusion-E2E-FT (5.2°, 91.9\%). The consistent gains across metrics suggest that our semantic-guided architecture helps disentangle geometry from optical appearance, which is a key design principle of the TransNormal-Synthetic dataset. 

\textbf{Results on ClearPose.}
On the large-scale ClearPose dataset, TransNormal achieves the best results among the compared methods with 26.3° mean error and 69.8\% accuracy at 30°, outperforming Diception (31.0°, 63.5\%) and Lotus-D (31.3°, 59.5\%). The 15.2\% relative improvement in mean error suggests good generalization to diverse transparent object categories and poses. Traditional methods trained on opaque objects show large degradation (Omnidata V2: 51.7°), while our approach maintains best performance by leveraging semantic understanding to infer plausible geometry under challenging refractive conditions.

\textbf{Ablation Studies.}\label{par:ablation_studies}
We conduct comprehensive ablation experiments on the ClearPose dataset to validate the effectiveness of our key design choices (\tabref{tab:ablation_loss}, \tabref{tab:ablation_encoder}, \tabref{tab:ablation}, and \figref{fig:ablation}).

\begin{wraptable}{r}{0.5\columnwidth}
    \vspace{-4mm}
    \scriptsize
    \setlength{\tabcolsep}{2pt}
    \centering
    \caption{\textbf{Ablation on loss functions.} (\S~\ref{par:ablation_studies})}
    \label{tab:ablation_loss}
    \begin{tabular}{l|ccc}
        \toprule
        \multirow{2}{*}{Loss Config.}
        & \multicolumn{3}{c}{ClearPose} \\
        & Mean$\downarrow$ & $11.25^\circ\uparrow$ & $30^\circ\uparrow$ \\
        \midrule
        w/o $\Ls_{\text{wav}}$ & \cellcolor{best3}29.1 & \cellcolor{best3}30.0 & \cellcolor{best3}64.1 \\
        LL only & 29.4 & 29.4 & \cellcolor{best3}64.1 \\
        LL + int. HF & \cellcolor{best2}27.6 & \cellcolor{best2}33.5 & \cellcolor{best2}67.1 \\
        \textbf{LL + edge (Ours)} & \cellcolor{best}26.3 & \cellcolor{best}35.9 & \cellcolor{best}69.8 \\
        \bottomrule
    \end{tabular}
    \vspace{-3mm}
\end{wraptable}

\textit{\ding{172} Loss function design (\tabref{tab:ablation_loss}, details in Appendix~\ref{ssec:ablation_loss_extended}).}
Our wavelet-based loss design is important for transparent objects.
Removing the wavelet regularization increases mean error from 26.3° to 29.1°, a 10.6\% relative degradation.
The spatially-selective frequency supervision is key: supervising only the LL sub-band lacks edge sharpness.
The ``LL + interior HF'' configuration improves upon LL-only by suppressing spurious gradients in smooth regions, but still underperforms our full design that emphasizes edge-selective HF alignment.

\begin{wraptable}{r}{0.5\columnwidth}
    \vspace{-4mm}
    \scriptsize
    \setlength{\tabcolsep}{2pt}
    \centering
    \caption{\textbf{Ablation on semantic encoder.} (\S~\ref{par:ablation_studies})}
    \label{tab:ablation_encoder}
    \begin{tabular}{l|ccc}
    \toprule
    \multirow{2}{*}{Encoder}
    & \multicolumn{3}{c}{ClearPose} \\
    & Mean$\downarrow$ & $11.25^\circ\uparrow$ & $30^\circ\uparrow$ \\
    \midrule
    DINOv2 & \cellcolor{best3}28.5 & 30.9 & \cellcolor{best3}66.1 \\
    SigLIP2 & \cellcolor{best2}27.2 & \cellcolor{best2}34.5 & \cellcolor{best2}67.8 \\
    SAM2 & \cellcolor{best3}28.5 & \cellcolor{best3}31.1 & \cellcolor{best3}66.1 \\
    \textbf{DINOv3 (Ours)} & \cellcolor{best}26.3 & \cellcolor{best}35.9 & \cellcolor{best}69.8 \\
    \bottomrule
    \end{tabular}
    \vspace{-3mm}
\end{wraptable}

\textit{\ding{173} Semantic encoder choice (\tabref{tab:ablation_encoder}, details in Appendix~\ref{ssec:ablation_encoder_extended}).}
We compare four vision encoders for semantic guidance.
DINOv3 achieves the best results across all metrics, outperforming DINOv2~\citep{oquab2023dinov2} (28.5°), SigLIP2~\citep{tschannen2025siglip} (27.2°), and Segment Anything Model 2 (SAM2)~\citep{ravi2024sam2} (28.5°).
The superior performance of DINOv3 can be attributed to its stronger object-level semantic understanding, which is critical for inferring geometry from misleading optical cues.

\begin{wraptable}{r}{0.55\columnwidth}
    \vspace{-4mm}
    \scriptsize
    \setlength{\tabcolsep}{2pt}
    \centering
    \caption{\textbf{Ablation on fine-tuning strategies.} (\S~\ref{par:ablation_studies})}
    \label{tab:ablation}
    \begin{tabular}{l|cc|ccc}
    \toprule
    \multirow{2}{*}{Method}
    & \multicolumn{2}{c|}{Fine-tune}
    & \multicolumn{3}{c}{ClearPose} \\
    & DINOv3 & U-Net
    & Mean$\downarrow$ & $11.25^\circ\uparrow$ & $30^\circ\uparrow$ \\
    \midrule
    w/o DINOv3 & -- & Full & \cellcolor{best3}27.7 & \cellcolor{best3}33.4 & \cellcolor{best3}67.2 \\
    U-Net LoRA & Frozen & LoRA & 29.8 & 26.2 & 63.4 \\
    DINOv3 LoRA & LoRA & Full & \cellcolor{best2}27.5 & \cellcolor{best2}34.7 & \cellcolor{best2}67.5 \\
    \textbf{Ours} & Frozen & Full & \cellcolor{best}26.3 & \cellcolor{best}35.9 & \cellcolor{best}69.8 \\
    \bottomrule
    \end{tabular}
    \vspace{-3mm}
\end{wraptable}

\textit{\ding{174} Fine-tuning strategies (\tabref{tab:ablation}, details in Appendix~\ref{ssec:ablation_finetune_extended}).}
Full fine-tuning (Full FT) of the U-Net with frozen DINOv3 encoder achieves the best performance (26.3° mean error).
Removing DINOv3 guidance degrades performance to 27.7°, confirming the importance of semantic features.
LoRA-based adaptation hurts performance for both U-Net and DINOv3, suggesting that bridging the domain gap requires sufficient model capacity and that fine-tuning the encoder on limited data risks overfitting.

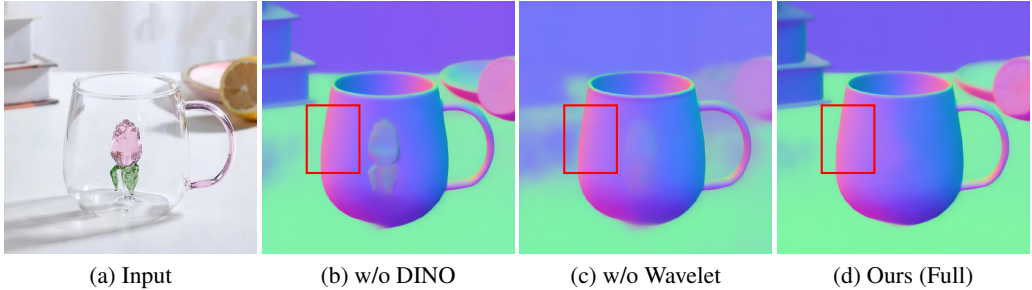
\begin{figure}[t]
    \centering
    \setlength{\tabcolsep}{1pt}
    \newcommand{\ablationbox}[1]{%
        \begin{tikzpicture}
            \node[inner sep=0pt] (img) {\includegraphics[width=0.24\columnwidth]{#1}};
            \draw[red, line width=0.8pt] ([xshift=0.6cm, yshift=0.3cm] img.west) rectangle ([xshift=1.3cm, yshift=-0.6cm] img.west);
        \end{tikzpicture}%
    }
    \begin{tabular}{cccc}
        \includegraphics[width=0.24\columnwidth]{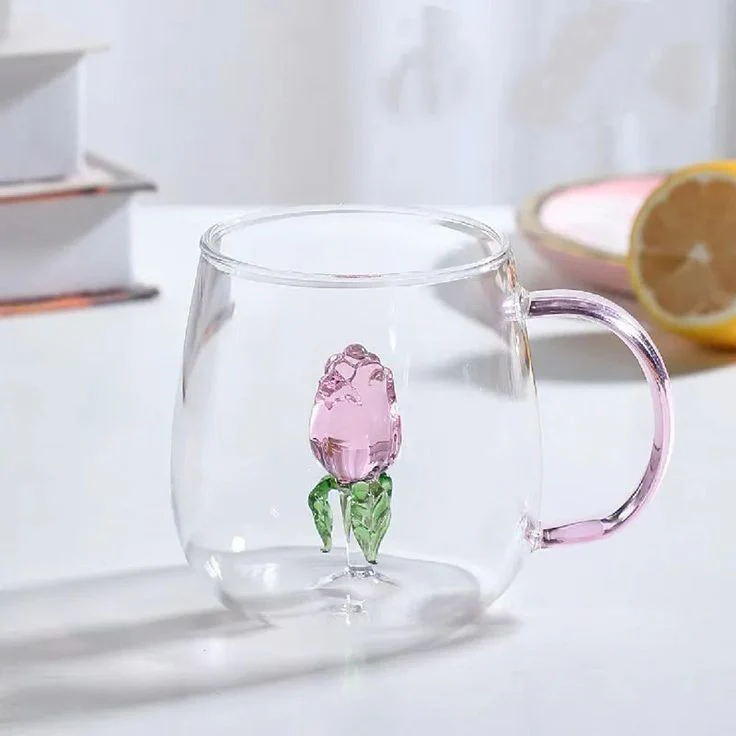} &
        \ablationbox{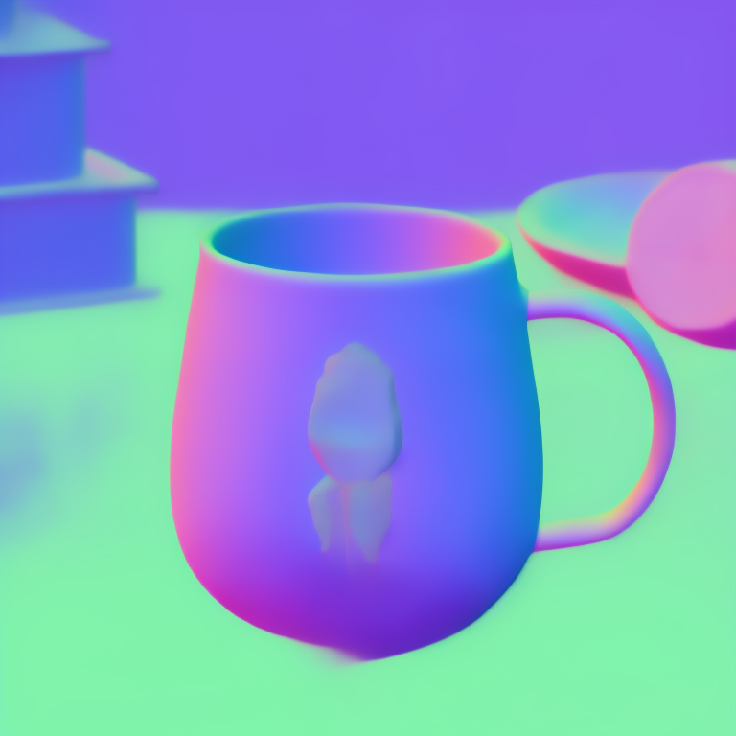} &
        \ablationbox{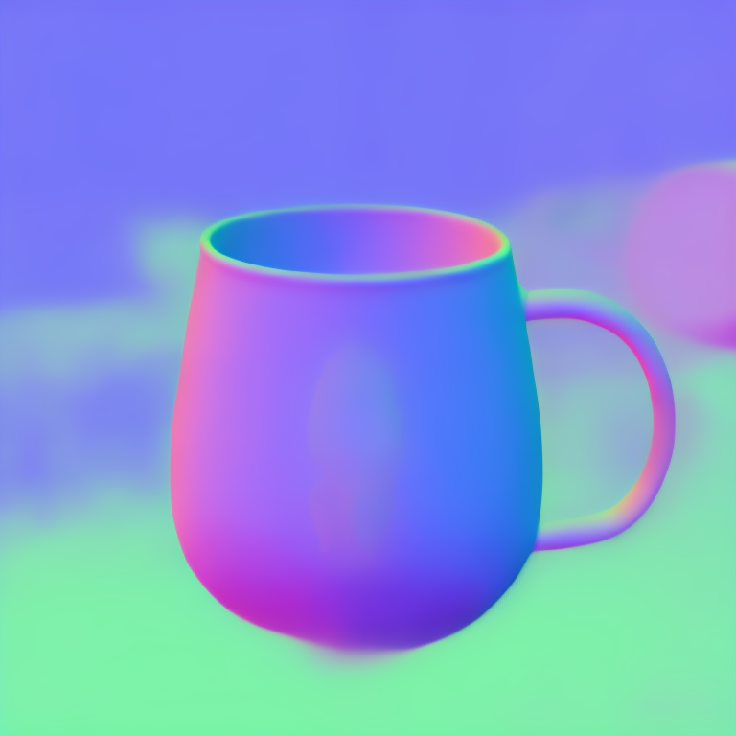} &
        \ablationbox{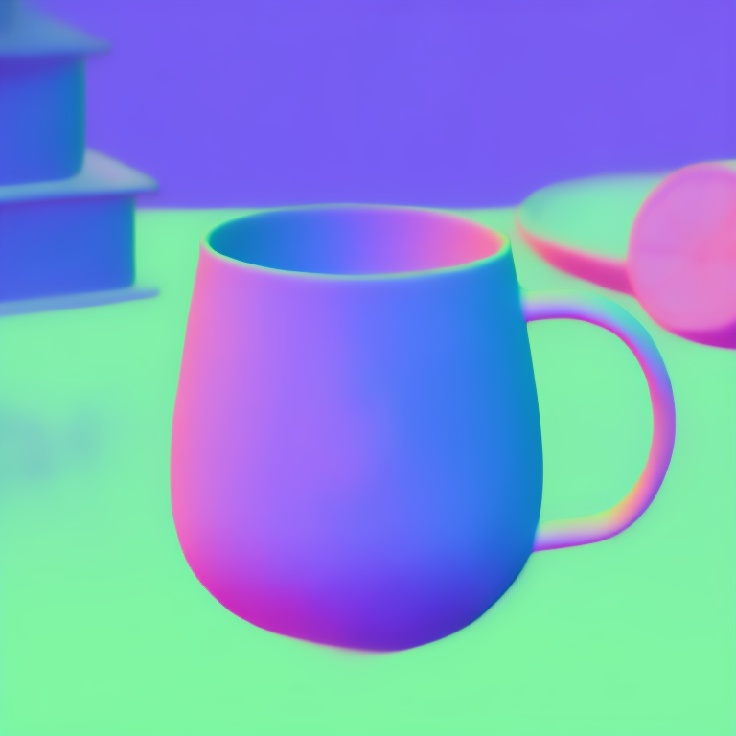} \\
        \small (a) Input & \small (b) w/o DINO & \small (c) w/o Wavelet & \small (d) Ours (Full) \\
    \end{tabular}
    \caption{\textbf{Qualitative ablation study on in-the-wild objects.}
    (a) In-the-wild input RGB image, a transparent cup with a flower inside.
    (b) Without DINOv3 semantic guidance, the model fails to recognize that the cup is transparent, incorrectly predicting the internal flower as surface geometry.
    (c) Without wavelet loss, the output exhibits discontinuous artifacts on smooth surfaces.
    (d) Our full model achieves both correct transparency understanding and smooth, continuous predictions.
    (\S~\ref{par:ablation_studies})}
    \vspace{-5mm}
    \label{fig:ablation}
\end{figure}

\section{Conclusion}
\label{sec:conclusion}
We present TransNormal, a framework for transparent object normal estimation that elevates the task from low-level feature extraction to high-level scene understanding. By replacing the underutilized text conditioning in Stable Diffusion with dense DINOv3 visual semantics, we transform the cross-attention mechanism into a powerful semantic-injection channel that resolves geometric ambiguities caused by refraction and reflection.
TransNormal achieves the best results among the compared methods across three transparent object benchmarks with an average rank of 1.0, using only $\sim$122K synthetic training samples ($\sim$1.4\% of MoGe-2's 8.9M). This supports the effectiveness of adapting generative priors with semantic guidance for specialized geometric tasks, and suggests a path toward more reliable embodied AI systems in laboratory automation.



\bibliographystyle{plainnat}
\bibliography{main}

\appendix
\newpage
\section{The TransNormal-Synthetic Dataset}
\label{sec:dataset}

To address the scarcity of high-quality surface normal annotations for transparent objects, we introduce \textbf{TransNormal-Synthetic}, a curated synthetic dataset specifically designed for robust geometric perception. Leveraging the advanced physics-based rendering capabilities of Blender, we generate a diverse set of laboratory-style scenes containing ubiquitous transparent glassware such as beakers, test tubes, and pipettes. We will release the Blender scripts and \texttt{.blend} files (including various material presets), enabling users to construct custom datasets through simple scene composition.

\begin{figure*}[t]
    \centering
    \includegraphics[width=0.95\textwidth]{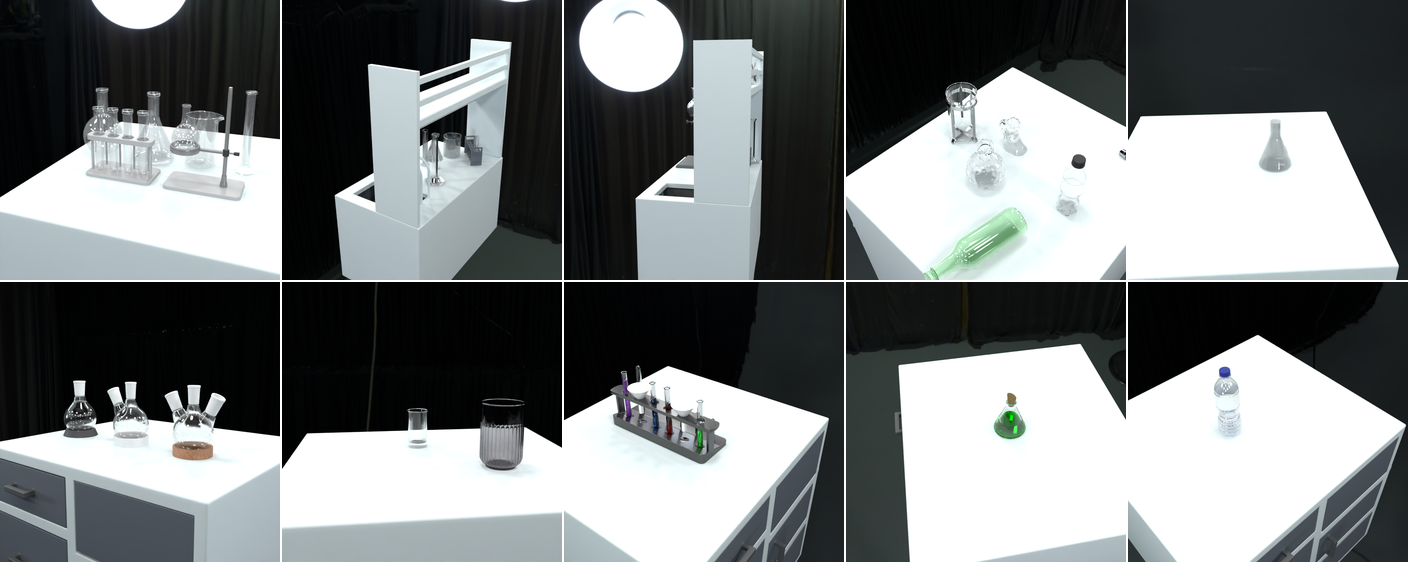}
    \caption{\textbf{Scene gallery of TransNormal-Synthetic.} 
    Representative RGB renderings from different laboratory scenes, showcasing the diversity of transparent glassware configurations, lighting conditions, and background setups. (\S~\ref{sec:dataset})}
    \label{fig:dataset_gallery}
\end{figure*}

\begin{figure*}[t]
    \centering
    \makebox[0.158\textwidth]{\small RGB}%
    \makebox[0.158\textwidth]{\small Normal}%
    \makebox[0.158\textwidth]{\small Depth}%
    \makebox[0.158\textwidth]{\small Mask}%
    \makebox[0.158\textwidth]{\small RGB w/o trans}%
    \makebox[0.158\textwidth]{\small Normal w/o trans}\\[-2pt]
    \includegraphics[width=0.95\textwidth]{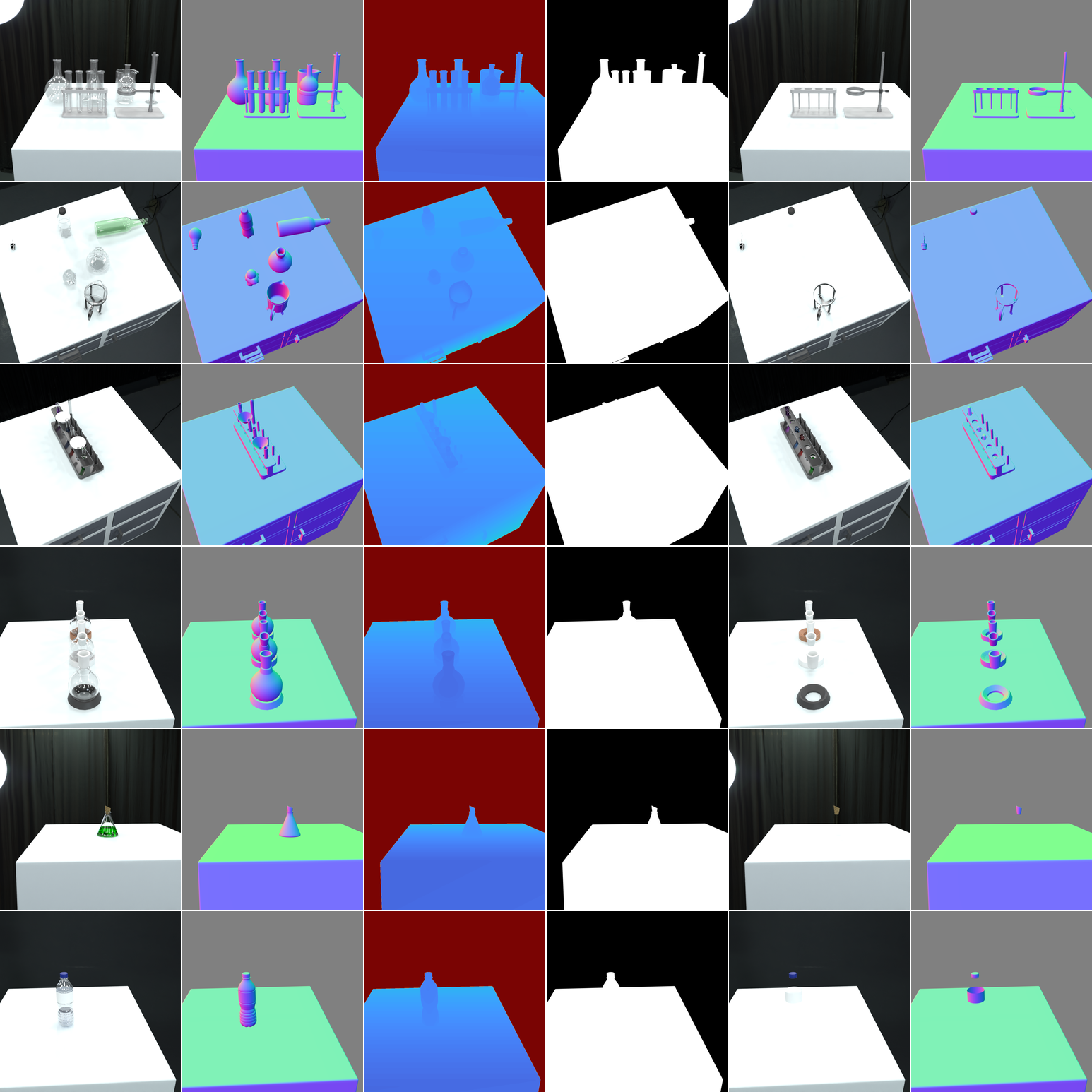}
    \caption{\textbf{Multi-modal annotations in TransNormal-Synthetic.} 
    Each row shows a different scene with six annotation types. The material-decoupled design (with/without transparent objects) enables the model to learn geometry invariant to optical appearance. (\S~\ref{ssec:dataset_generation})}
    \label{fig:dataset_overview}
\end{figure*}

\begin{figure}[t]
    \centering
    \small
    \makebox[0.317\columnwidth]{RGB}%
    \makebox[0.317\columnwidth]{Normal}%
    \makebox[0.317\columnwidth]{Depth}\\[-2pt]
    \includegraphics[width=0.95\columnwidth]{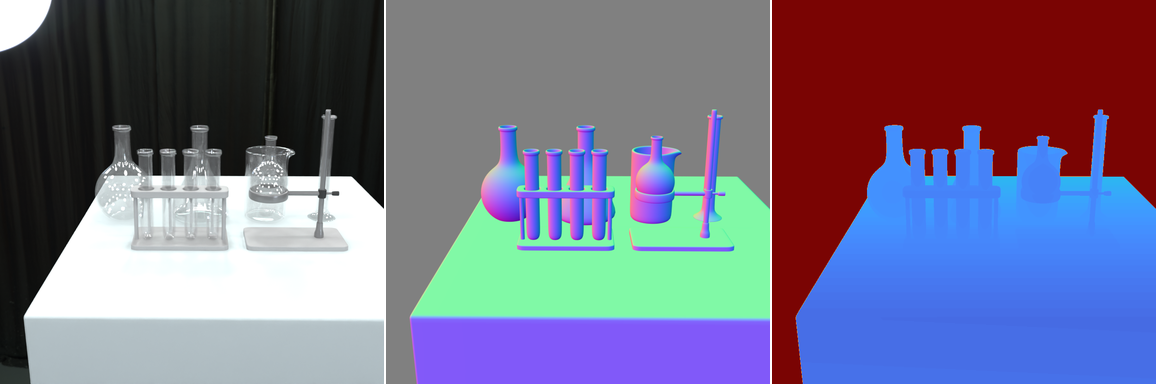}\\[2pt]
    \makebox[0.317\columnwidth]{Foreground Mask}%
    \makebox[0.317\columnwidth]{Transparent Mask}%
    \makebox[0.317\columnwidth]{RGB (randomized material)}\\[-2pt]
    \includegraphics[width=0.95\columnwidth]{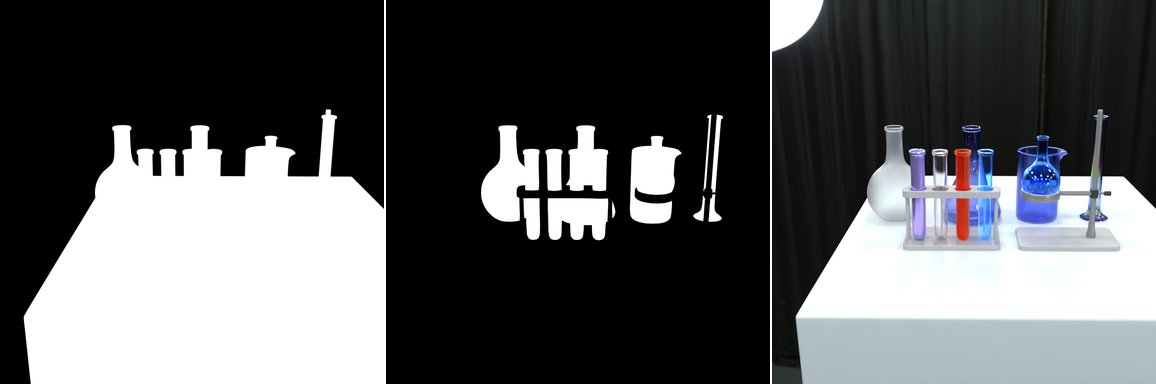}\\[2pt]
    \makebox[0.317\columnwidth]{RGB w/o Transparent Objects}%
    \makebox[0.317\columnwidth]{Normal w/o Transparent Objects}%
    \makebox[0.317\columnwidth]{Depth w/o Transparent Objects}\\[-2pt]
    \includegraphics[width=0.95\columnwidth]{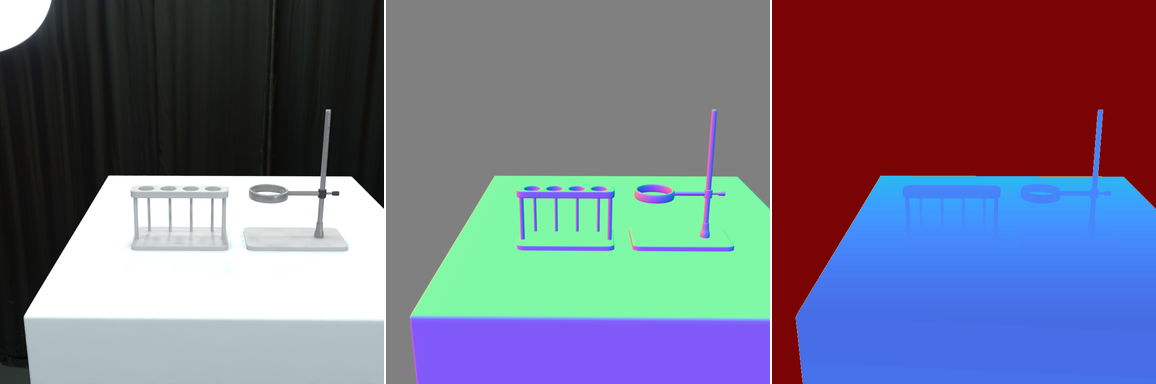}
    \caption{\textbf{Annotation detail visualization.} 
    (Row 1) Standard rendering with transparent objects;
    (Row 2) Foreground mask, transparent mask, and RGB with randomized transparent material;
    (Row 3) Reference rendering without transparent objects.
    This triplet structure enables geometry-appearance disentanglement. (\S~\ref{ssec:dataset_generation})}
    \label{fig:dataset_annotation}
\end{figure}

\begin{figure*}[t]
    \centering
    \makebox[0.155\textwidth]{\small RGB}%
    \makebox[0.155\textwidth]{\small Normal}%
    \makebox[0.155\textwidth]{\small Depth}%
    \hspace{0.012\textwidth}%
    \makebox[0.155\textwidth]{\small RGB}%
    \makebox[0.155\textwidth]{\small Normal}%
    \makebox[0.155\textwidth]{\small Depth}\\[-2pt]
    \includegraphics[width=0.95\textwidth]{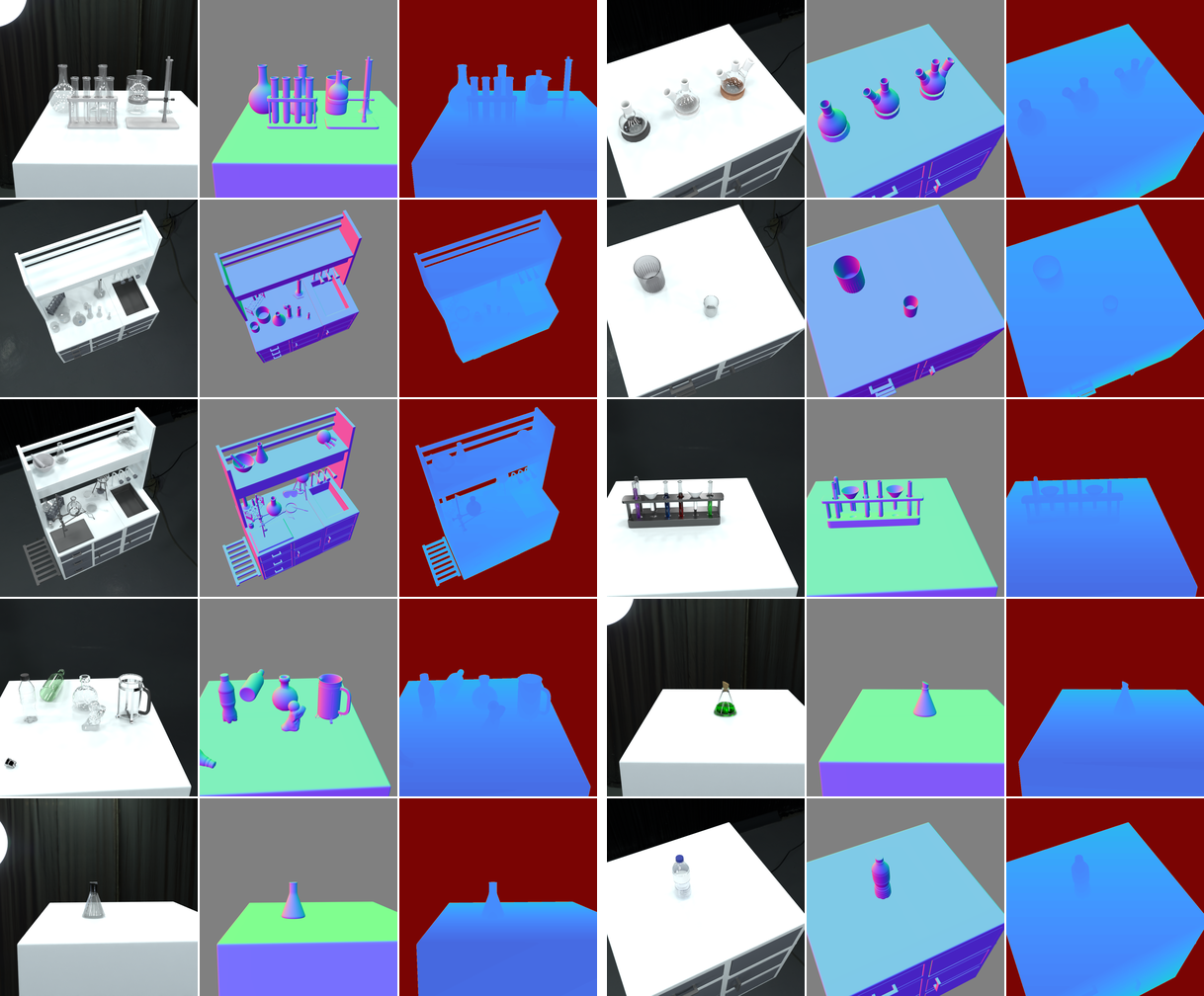}
    \caption{\textbf{Comprehensive scene coverage in TransNormal-Synthetic.} 
    RGB images, surface normals, and depth maps across 10 representative scenes, demonstrating the dataset's coverage of diverse transparent object arrangements, viewpoints, and lighting conditions. (\S~\ref{sec:dataset})}
    \label{fig:dataset_comprehensive}
\end{figure*}

\subsection{Data Generation and Composition}\label{ssec:dataset_generation}
TransNormal-Synthetic provides comprehensive multi-modal labels across 10 scenes, with 3,950 images in total. Each sample consists of the following components:

\begin{itemize}
    \item \textbf{RGB Image Sequences}: To encourage invariance to optical appearance, each viewpoint includes three versions: (1) \emph{RGB}, the standard rendering containing transparent objects; (2) \emph{RGB with randomized material}, rendered by randomizing the transparent material parameters while keeping geometry fixed; and (3) \emph{RGB background-only}, rendered by removing transparent objects to provide a clean reference.
    \item \textbf{Diverse Material Presets}: We provide multiple material options including translucent, fully transparent, and specular/glossy materials, enabling systematic evaluation under varying optical properties.
    \item \textbf{High-Precision Ground Truth}: We export pixel-accurate surface normal maps and 16-bit depth maps directly from the rendering engine. The depth maps are normalized following a 10m maximum distance protocol, consistent with laboratory-scale sensing.
    \item \textbf{Comprehensive Masks}: Each sample includes detailed segmentation masks, specifically identifying all objects (\emph{foreground mask}) and specifically isolating transparent surfaces (\emph{mask\_transparent}).
    \item \textbf{Camera Parameters}: Full intrinsic matrices and 6D camera poses are provided to support potential downstream geometric reasoning tasks.
\end{itemize}

\subsection{Material-Decoupled Design for Future Research}
Beyond standard RGB-normal pairs, TransNormal-Synthetic provides a \emph{material-decoupled} structure that enables future research on appearance-invariant geometry learning. By providing paired renderings that randomize transparent material parameters while keeping geometry fixed, this design can force a model to recognize that while the RGB appearance changes drastically with material variations, the underlying surface normal remains constant. 

The inclusion of \emph{RGB background-only} reference images further enables auxiliary tasks such as background inpainting, potentially leading to deeper understanding of light transport in refractive and scattering regions. While our current method uses only the standard RGB renderings, we release these additional modalities to support future exploration of material-invariant training strategies. 

\section{More Quantitative Results}
\label{sec:more_quantitative}

\subsection{Inference Efficiency}
\label{ssec:inference_efficiency}

We benchmark TransNormal on a single NVIDIA A100 GPU, reporting average latency, FPS, and memory usage over 10 runs (\tabref{tab:inference}). Mixed precision (BF16/FP16) yields $\sim$2.5$\times$ speedup over FP32, achieving 4.03 FPS. Peak memory is $\sim$11 GB, fitting within 16GB consumer GPUs.

\begin{table}[!t]
\centering
\caption{Inference efficiency of TransNormal (averaged over runs).(\S~\ref{ssec:inference_efficiency})}
\label{tab:inference}
\scriptsize
\begin{tabular}{l|c|c|c|c|c}
\toprule
\textbf{Precision} & \textbf{Time (ms)} & \textbf{FPS} & \textbf{Peak Mem (MB)} & \textbf{Delta Mem (MB)} & \textbf{Model Load (MB)} \\
\midrule
BF16 & 247.98 & 4.03 & 11098.4 & 3642.2 & 7447.0 \\
FP16 & 247.63 & 4.03 & 11098.0 & 3642.0 & 7447.0 \\
FP32 & 615.43 & 1.63 & 10467.6 & 2200.1 & 8255.8 \\
\bottomrule
\end{tabular}
\end{table}

\subsection{Loss Function Ablation Across Datasets}
\label{ssec:ablation_loss_extended}

\Tabref{tab:ablation_loss_extended} evaluates our loss design across all benchmarks. We compare: (1) removing RGB reconstruction loss, (2) removing wavelet loss entirely, (3) supervising only LL sub-band, (4) LL + interior HF suppression, and (5) our full design with LL + edge-selective HF. Interior regions are defined as $(1 - \bm{M}_{\text{edge}})$, where $\bm{M}_{\text{edge}}$ is the normalized GT normal gradient.

\begin{table*}[h]
\centering
\caption{\textbf{Extended ablation on loss function design across three datasets.} We evaluate the contribution of each wavelet regularization component. The edge-selective high-frequency supervision (LL + edge HF) consistently outperforms alternatives. (\S~\ref{ssec:ablation_loss_extended})}
\label{tab:ablation_loss_extended}
\resizebox{\textwidth}{!}{%
\begin{tabular}{l|cccccc|cccccc|cccccc}
\toprule
\multirow{2}{*}{Loss Configuration}
& \multicolumn{6}{c|}{ClearGrasp}
& \multicolumn{6}{c|}{TransNormal-Synthetic}
& \multicolumn{6}{c}{ClearPose} \\
& Mean$\downarrow$ & $5^\circ\uparrow$ & $7.5^\circ\uparrow$ & $11.25^\circ\uparrow$ & $22.5^\circ\uparrow$ & $30^\circ\uparrow$
& Mean$\downarrow$ & $5^\circ\uparrow$ & $7.5^\circ\uparrow$ & $11.25^\circ\uparrow$ & $22.5^\circ\uparrow$ & $30^\circ\uparrow$
& Mean$\downarrow$ & $5^\circ\uparrow$ & $7.5^\circ\uparrow$ & $11.25^\circ\uparrow$ & $22.5^\circ\uparrow$ & $30^\circ\uparrow$ \\
\midrule
w/o $\mathcal{L}_{\text{rgb}}$
& 16.7 & 17.9 & 32.2 & 50.2 & \cellcolor{best3}77.2 & \cellcolor{best2}85.1
& 4.7 & 76.2 & 88.9 & \cellcolor{best3}93.2 & \cellcolor{best3}97.0 & \cellcolor{best2}98.1
& \cellcolor{best2}26.7 & \cellcolor{best3}10.1 & \cellcolor{best3}19.1 & \cellcolor{best3}32.5 & \cellcolor{best2}59.4 & \cellcolor{best2}69.1 \\
w/o $\mathcal{L}_{\text{wavelet}}$
& 17.3 & 17.0 & 30.8 & 48.3 & 75.4 & 83.9
& 5.3 & 75.9 & 88.2 & 92.9 & 96.9 & \cellcolor{best3}98.0
& 29.1 & 9.2 & 17.6 & 30.0 & 54.6 & 64.1 \\
LL only
& \cellcolor{best2}16.5 & \cellcolor{best2}18.9 & \cellcolor{best2}33.5 & \cellcolor{best2}50.9 & \cellcolor{best2}77.3 & \cellcolor{best}85.3
& \cellcolor{best2}4.4 & \cellcolor{best2}80.9 & \cellcolor{best3}89.3 & \cellcolor{best2}93.4 & \cellcolor{best}97.2 & \cellcolor{best}98.2
& 29.4 & 9.0 & 17.2 & 29.4 & 54.5 & 64.1 \\
LL + interior HF
& \cellcolor{best3}16.6 & \cellcolor{best3}18.4 & \cellcolor{best3}33.0 & \cellcolor{best3}50.8 & \cellcolor{best}77.4 & \cellcolor{best}85.3
& \cellcolor{best3}4.5 & \cellcolor{best3}80.8 & \cellcolor{best2}89.4 & \cellcolor{best2}93.4 & \cellcolor{best}97.2 & \cellcolor{best}98.2
& \cellcolor{best3}27.6 & \cellcolor{best}11.1 & \cellcolor{best2}20.6 & \cellcolor{best2}33.5 & \cellcolor{best3}57.6 & \cellcolor{best3}67.1 \\
LL + edge HF (Ours)
& \cellcolor{best}16.4 & \cellcolor{best}19.7 & \cellcolor{best}34.4 & \cellcolor{best}51.7 & \cellcolor{best3}77.2 & \cellcolor{best3}85.0
& \cellcolor{best}4.1 & \cellcolor{best}84.1 & \cellcolor{best}90.3 & \cellcolor{best}93.5 & \cellcolor{best2}97.1 & \cellcolor{best}98.2
& \cellcolor{best}26.3 & \cellcolor{best2}11.0 & \cellcolor{best}21.6 & \cellcolor{best}35.9 & \cellcolor{best}61.0 & \cellcolor{best}69.8 \\
\bottomrule
\end{tabular}%
}
\end{table*}

\subsection{Semantic Encoder Ablation Across Datasets}
\label{ssec:ablation_encoder_extended}

\Tabref{tab:ablation_encoder_extended} compares four visual encoders, DINOv2, SigLIP2, SAM2, and DINOv3, across all three benchmarks, extending the analysis from \tabref{tab:ablation_encoder}. \Tabref{tab:encoder_details} lists the specific model variants and their specifications, including parameter counts, patch sizes, and feature dimensions. 

\begin{table}[h]
\centering
\caption{\textbf{Visual encoder specifications.} Model variants, parameter counts, patch sizes, and feature dimensions for the four encoders compared in the semantic encoder ablation (\S~\ref{ssec:ablation_encoder_extended}).}
\label{tab:encoder_details}
\scriptsize
\setlength{\tabcolsep}{3pt}
\begin{tabular}{l|l|c|c|c}
\toprule
\textbf{Encoder} & \textbf{Model} & \textbf{Params} & \textbf{Patch Size} & \textbf{Feature Dim} \\
\midrule
DINOv2 & dinov2-vitl14 & 304M & 14 & 1024 \\
SigLIP2 & siglip2-large-patch16-384 & 304M & 16 & 1024 \\
SAM2 & sam2-hiera-large & 224M & 16 & 256 \\
DINOv3 (Ours) & dinov3-vith16plus & 840M & 16 & 1280 \\
\bottomrule
\end{tabular}
\end{table}

\begin{table*}[h]
\centering
\caption{\textbf{Extended ablation on semantic encoder choice across three datasets.} We evaluate DINOv2, SigLIP2, SAM2, and DINOv3 (ours) as visual semantic guidance. DINOv3 consistently outperforms alternatives across both synthetic and real-world benchmarks. (\S~\ref{ssec:ablation_encoder_extended})}
\label{tab:ablation_encoder_extended}
\resizebox{\textwidth}{!}{%
\begin{tabular}{l|cccccc|cccccc|cccccc}
\toprule
\multirow{2}{*}{Encoder}
& \multicolumn{6}{c|}{ClearGrasp}
& \multicolumn{6}{c|}{TransNormal-Synthetic}
& \multicolumn{6}{c}{ClearPose} \\
& Mean$\downarrow$ & $5^\circ\uparrow$ & $7.5^\circ\uparrow$ & $11.25^\circ\uparrow$ & $22.5^\circ\uparrow$ & $30^\circ\uparrow$
& Mean$\downarrow$ & $5^\circ\uparrow$ & $7.5^\circ\uparrow$ & $11.25^\circ\uparrow$ & $22.5^\circ\uparrow$ & $30^\circ\uparrow$
& Mean$\downarrow$ & $5^\circ\uparrow$ & $7.5^\circ\uparrow$ & $11.25^\circ\uparrow$ & $22.5^\circ\uparrow$ & $30^\circ\uparrow$ \\
\midrule
DINOv2
& \cellcolor{best3}16.5 & 17.2 & 31.3 & 48.9 & \cellcolor{best3}77.2 & \cellcolor{best2}85.9
& \cellcolor{best}3.9 & \cellcolor{best2}83.4 & \cellcolor{best2}90.0 & \cellcolor{best2}93.7 & \cellcolor{best2}97.2 & \cellcolor{best2}98.2
& 28.5 & 8.9 & 17.8 & 30.9 & 56.7 & 66.1 \\
SigLIP2
& 16.7 & \cellcolor{best3}18.0 & \cellcolor{best3}31.8 & \cellcolor{best3}49.2 & 76.9 & 85.3
& 4.7 & 74.0 & \cellcolor{best}90.3 & \cellcolor{best}93.8 & \cellcolor{best}97.3 & \cellcolor{best}98.3
& \cellcolor{best3}27.2 & \cellcolor{best2}11.0 & \cellcolor{best3}21.3 & \cellcolor{best3}34.5 & \cellcolor{best3}58.7 & \cellcolor{best3}67.8 \\
SAM2
& 16.6 & 17.0 & 31.1 & 49.0 & \cellcolor{best2}77.6 & \cellcolor{best}86.0
& 5.0 & 77.3 & 88.7 & 93.3 & \cellcolor{best3}97.1 & \cellcolor{best3}98.1
& 28.5 & 9.7 & 18.4 & 31.1 & 56.3 & 66.1 \\
DINOv3 (Ours)
& \cellcolor{best2}16.4 & \cellcolor{best}19.7 & \cellcolor{best}34.4 & \cellcolor{best2}51.7 & \cellcolor{best3}77.2 & 85.0
& \cellcolor{best2}4.1 & \cellcolor{best}84.1 & \cellcolor{best}90.3 & \cellcolor{best3}93.5 & \cellcolor{best3}97.1 & \cellcolor{best2}98.2
& \cellcolor{best}26.3 & \cellcolor{best2}11.0 & \cellcolor{best2}21.6 & \cellcolor{best}35.9 & \cellcolor{best}61.0 & \cellcolor{best}69.8 \\
\bottomrule
\end{tabular}%
}
\end{table*}

\subsection{Fine-Tuning Strategy Ablation Across Datasets}
\label{ssec:ablation_finetune_extended}

\Tabref{tab:ablation_extended} extends the fine-tuning strategy ablation from the main paper (\tabref{tab:ablation}) to all three benchmarks. We evaluate five configurations: (1) removing DINOv3 guidance entirely (using empty text prompt), (2) replacing DINOv3 with text prompt encoding (\emph{e.g.}, ``normal map''), (3) applying LoRA to the U-Net, (4) applying LoRA to DINOv3, and (5) our full model with frozen DINOv3 and fully fine-tuned U-Net.

\begin{table*}[h]
\centering
\caption{\textbf{Extended ablation on fine-tuning strategies across three datasets.} We report mean angular error (Mean$\downarrow$) and percentage of pixels within various angular thresholds ($\uparrow$). Results demonstrate consistent trends across synthetic (ClearGrasp, TransNormal-Synthetic) and real-world (ClearPose) benchmarks. (\S~\ref{ssec:ablation_finetune_extended})}
\label{tab:ablation_extended}
\resizebox{\textwidth}{!}{%
\begin{tabular}{l|cc|cccccc|cccccc|cccccc}
\toprule
\multirow{2}{*}{Method}
& \multicolumn{2}{c|}{Fine-tuning}
& \multicolumn{6}{c|}{ClearGrasp}
& \multicolumn{6}{c|}{TransNormal-Synthetic}
& \multicolumn{6}{c}{ClearPose} \\
& DINOv3 & U-Net
& Mean$\downarrow$ & $5^\circ\uparrow$ & $7.5^\circ\uparrow$ & $11.25^\circ\uparrow$ & $22.5^\circ\uparrow$ & $30^\circ\uparrow$
& Mean$\downarrow$ & $5^\circ\uparrow$ & $7.5^\circ\uparrow$ & $11.25^\circ\uparrow$ & $22.5^\circ\uparrow$ & $30^\circ\uparrow$
& Mean$\downarrow$ & $5^\circ\uparrow$ & $7.5^\circ\uparrow$ & $11.25^\circ\uparrow$ & $22.5^\circ\uparrow$ & $30^\circ\uparrow$ \\
\midrule
w/o DINOv3 & -- & Full FT
& \cellcolor{best3}16.6 & 16.8 & 31.2 & 49.2 & \cellcolor{best3}77.2 & \cellcolor{best3}85.7
& \cellcolor{best2}4.5 & \cellcolor{best2}78.8 & \cellcolor{best}90.3 & \cellcolor{best}93.8 & \cellcolor{best}97.3 & \cellcolor{best}98.2
& \cellcolor{best3}27.7 & \cellcolor{best3}10.8 & 20.1 & 33.4 & 58.1 & 67.2 \\
Text Prompt & Text & Full FT
& \cellcolor{best}16.3 & \cellcolor{best3}17.7 & \cellcolor{best3}32.2 & \cellcolor{best3}50.8 & \cellcolor{best2}77.9 & \cellcolor{best2}85.9
& 5.5 & 66.0 & 83.1 & 92.8 & \cellcolor{best2}97.2 & \cellcolor{best}98.2
& \cellcolor{best2}27.5 & \cellcolor{best}11.1 & \cellcolor{best3}20.7 & \cellcolor{best3}33.7 & \cellcolor{best3}58.5 & \cellcolor{best2}67.6 \\
U-Net LoRA & Frozen & LoRA
& \cellcolor{best}16.3 & 17.4 & 32.0 & 50.2 & \cellcolor{best}78.5 & \cellcolor{best}86.4
& 5.7 & 61.4 & \cellcolor{best3}83.7 & 92.3 & 96.8 & \cellcolor{best2}97.8
& 29.8 & 6.6 & 14.0 & 26.2 & 53.0 & 63.4 \\
DINOv3 LoRA & LoRA & Full FT
& 16.9 & \cellcolor{best2}18.3 & \cellcolor{best2}32.9 & \cellcolor{best2}51.2 & 76.9 & 84.6
& \cellcolor{best3}4.6 & \cellcolor{best3}78.6 & \cellcolor{best2}88.7 & \cellcolor{best3}93.4 & \cellcolor{best2}97.2 & \cellcolor{best}98.2
& \cellcolor{best2}27.5 & \cellcolor{best2}11.0 & \cellcolor{best2}21.0 & \cellcolor{best2}34.7 & \cellcolor{best2}59.0 & \cellcolor{best3}67.5 \\
Full model (Ours) & Frozen & Full FT
& \cellcolor{best2}16.4 & \cellcolor{best}19.7 & \cellcolor{best}34.4 & \cellcolor{best}51.7 & \cellcolor{best3}77.2 & 85.0
& \cellcolor{best}4.1 & \cellcolor{best}84.1 & \cellcolor{best}90.3 & \cellcolor{best2}93.5 & \cellcolor{best3}97.1 & \cellcolor{best}98.2
& \cellcolor{best}26.3 & \cellcolor{best2}11.0 & \cellcolor{best}21.6 & \cellcolor{best}35.9 & \cellcolor{best}61.0 & \cellcolor{best}69.8 \\
\bottomrule
\end{tabular}%
}
\end{table*}

\subsection{Training Data Ratio Ablation}
\label{ssec:ablation_data_ratio}

Our training combines ClearGrasp (CG) and TransNormal-Synthetic (TN)---both synthetic transparent object datasets with different object diversity and rendering characteristics. We ablate the CG:TN sampling ratio while keeping other data sources (Hypersim, Virtual-KITTI) fixed, evaluating five configurations from CG-dominant (45:5) to TN-dominant (20:30). \Tabref{tab:ablation_data_ratio} shows that our default 35:15 ratio achieves strong overall performance, particularly on ClearPose---a zero-shot evaluation benchmark---indicating better generalization to unseen real-world scenarios.

\begin{table*}[h]
\centering
\caption{\textbf{Ablation on training data sampling ratio.} We vary the balance between ClearGrasp (CG) and TransNormal-Synthetic (TN)---both synthetic transparent object datasets---while keeping other data sources fixed. Results show that our default ratio (35:15) achieves strong performance, while the sensitivity to exact ratios is relatively low. (\S~\ref{ssec:ablation_data_ratio})}
\label{tab:ablation_data_ratio}
\resizebox{\textwidth}{!}{%
\begin{tabular}{l|cccccc|cccccc|cccccc}
\toprule
\multirow{2}{*}{CG:TN}
& \multicolumn{6}{c|}{ClearGrasp}
& \multicolumn{6}{c|}{TransNormal-Synthetic}
& \multicolumn{6}{c}{ClearPose} \\
& Mean$\downarrow$ & $5^\circ\uparrow$ & $7.5^\circ\uparrow$ & $11.25^\circ\uparrow$ & $22.5^\circ\uparrow$ & $30^\circ\uparrow$
& Mean$\downarrow$ & $5^\circ\uparrow$ & $7.5^\circ\uparrow$ & $11.25^\circ\uparrow$ & $22.5^\circ\uparrow$ & $30^\circ\uparrow$
& Mean$\downarrow$ & $5^\circ\uparrow$ & $7.5^\circ\uparrow$ & $11.25^\circ\uparrow$ & $22.5^\circ\uparrow$ & $30^\circ\uparrow$ \\
\midrule
40:10
& \cellcolor{best}16.4 & 18.1 & 32.3 & 50.5 & \cellcolor{best2}77.8 & \cellcolor{best}85.6
& \cellcolor{best3}4.4 & \cellcolor{best3}81.6 & 89.1 & 93.2 & 97.0 & \cellcolor{best3}98.1
& \cellcolor{best3}26.8 & \cellcolor{best3}10.8 & \cellcolor{best2}20.9 & \cellcolor{best2}34.5 & \cellcolor{best2}59.7 & \cellcolor{best2}68.6 \\
25:25
& \cellcolor{best2}16.5 & \cellcolor{best3}18.4 & 33.0 & \cellcolor{best2}51.4 & \cellcolor{best}77.9 & \cellcolor{best2}85.4
& \cellcolor{best2}4.2 & \cellcolor{best2}82.2 & \cellcolor{best3}89.8 & \cellcolor{best}93.7 & \cellcolor{best}97.3 & \cellcolor{best}98.3
& \cellcolor{best2}26.7 & \cellcolor{best2}11.0 & 20.6 & 33.9 & \cellcolor{best3}59.2 & \cellcolor{best3}68.5 \\
20:30
& \cellcolor{best3}16.9 & 18.2 & \cellcolor{best2}33.3 & \cellcolor{best3}51.3 & 77.1 & 84.6
& 5.0 & 73.0 & 88.4 & 93.4 & \cellcolor{best2}97.2 & \cellcolor{best2}98.2
& 28.5 & 10.5 & 19.5 & 32.5 & 57.0 & 66.2 \\
45:5
& \cellcolor{best3}16.9 & \cellcolor{best2}19.1 & \cellcolor{best3}33.2 & 49.9 & 76.0 & 84.4
& 4.5 & 80.1 & \cellcolor{best2}90.1 & \cellcolor{best2}93.6 & \cellcolor{best2}97.2 & \cellcolor{best3}98.1
& 27.1 & \cellcolor{best}11.1 & \cellcolor{best3}20.7 & \cellcolor{best3}34.0 & 58.7 & 67.9 \\
35:15 (Ours)
& \cellcolor{best}16.4 & \cellcolor{best}19.7 & \cellcolor{best}34.4 & \cellcolor{best}51.7 & \cellcolor{best3}77.2 & \cellcolor{best3}85.0
& \cellcolor{best}4.1 & \cellcolor{best}84.1 & \cellcolor{best}90.3 & \cellcolor{best3}93.5 & \cellcolor{best3}97.1 & \cellcolor{best2}98.2
& \cellcolor{best}26.3 & \cellcolor{best2}11.0 & \cellcolor{best}21.6 & \cellcolor{best}35.9 & \cellcolor{best}61.0 & \cellcolor{best}69.8 \\
\bottomrule
\end{tabular}%
}
\end{table*}

\section{More Qualitative Results}
\label{sec:more_qualitative}

\subsection{Extended Baseline Comparisons}
\label{ssec:extended_comparison}
\begin{figure*}[t]
    \centering
    \setlength{\tabcolsep}{1pt}
    \renewcommand{\arraystretch}{0.6}
    \newcommand{\imgwext}{0.135\textwidth}
    %
    %
    \begin{tabular}{@{}c@{\hspace{1pt}}cc|ccccc@{}}
        & \small Input/Mask & \small GT & \small Lotus & \small MoGe-2 & \small E2E-FT & \small GenPercept & \small \textbf{Ours} \\[2pt]
        %
        \multirow{2}{*}[3.5ex]{\rotatebox{90}{\small TransNormal}} &
        \includegraphics[width=\imgwext]{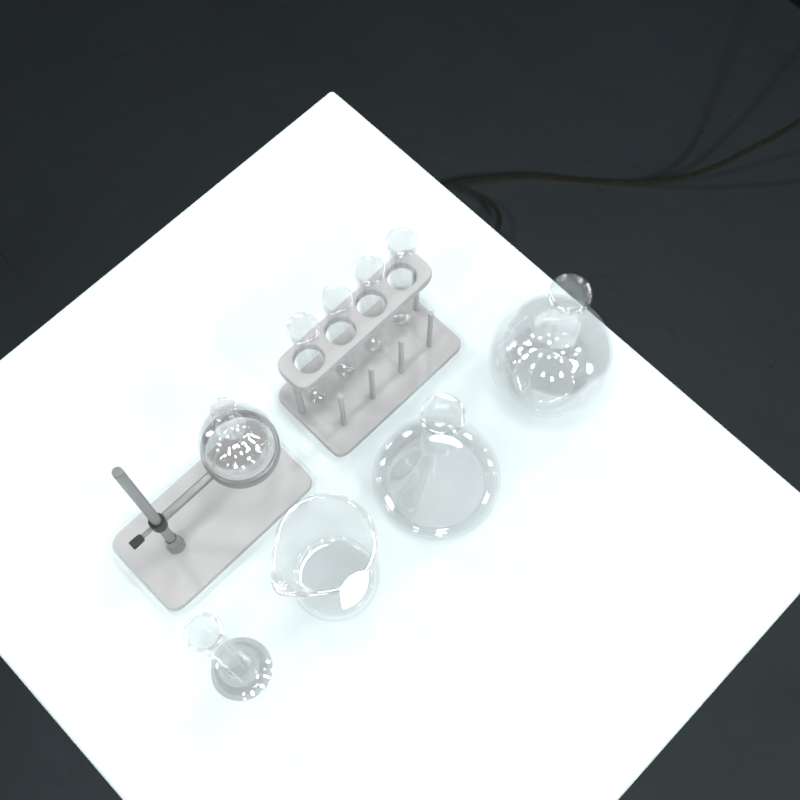} &
        \includegraphics[width=\imgwext]{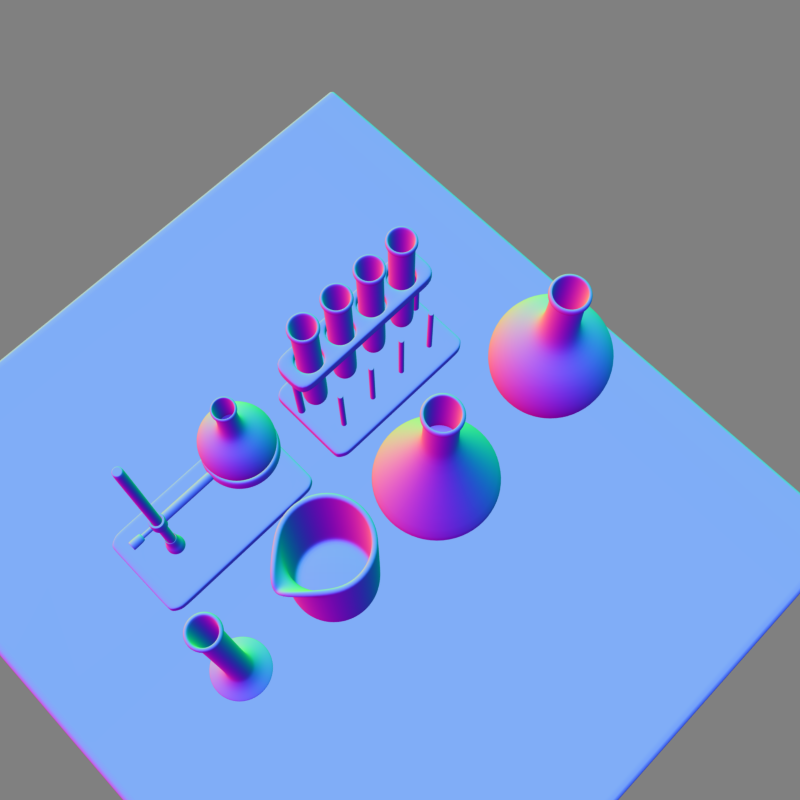} &
        \includegraphics[width=\imgwext]{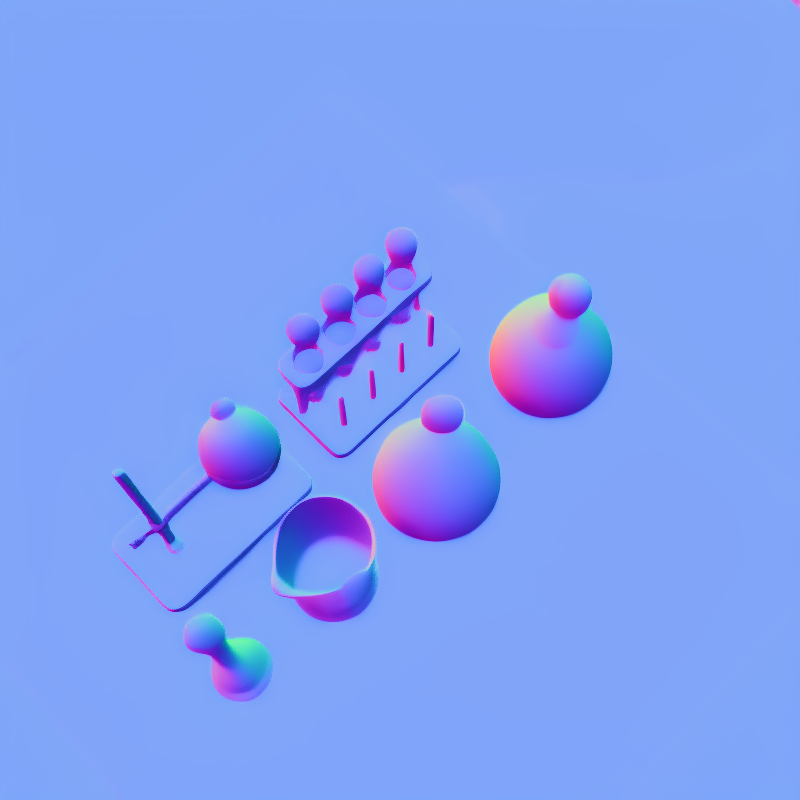} &
        \includegraphics[width=\imgwext]{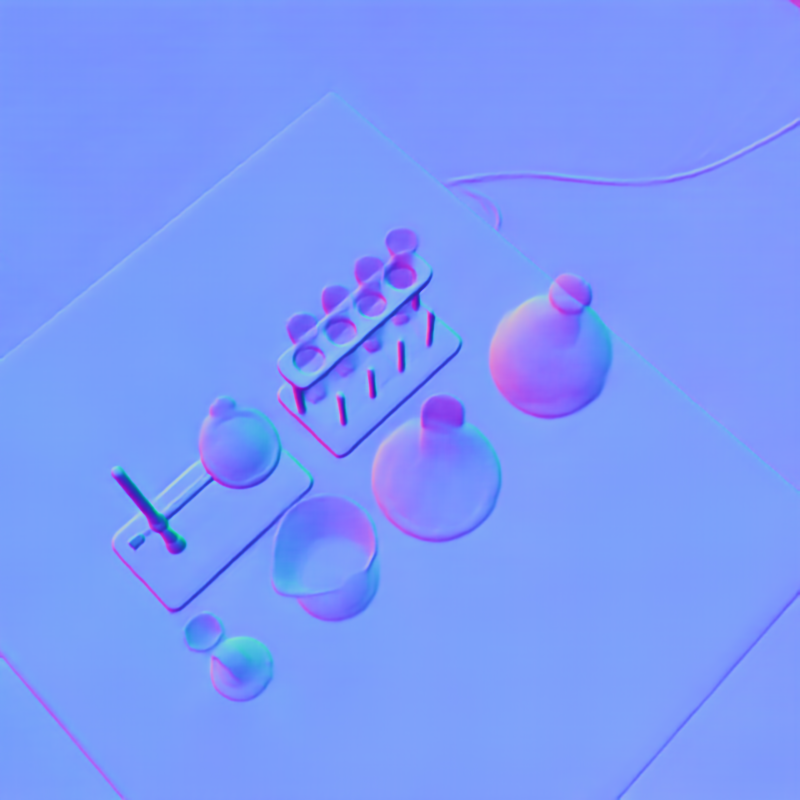} &
        \includegraphics[width=\imgwext]{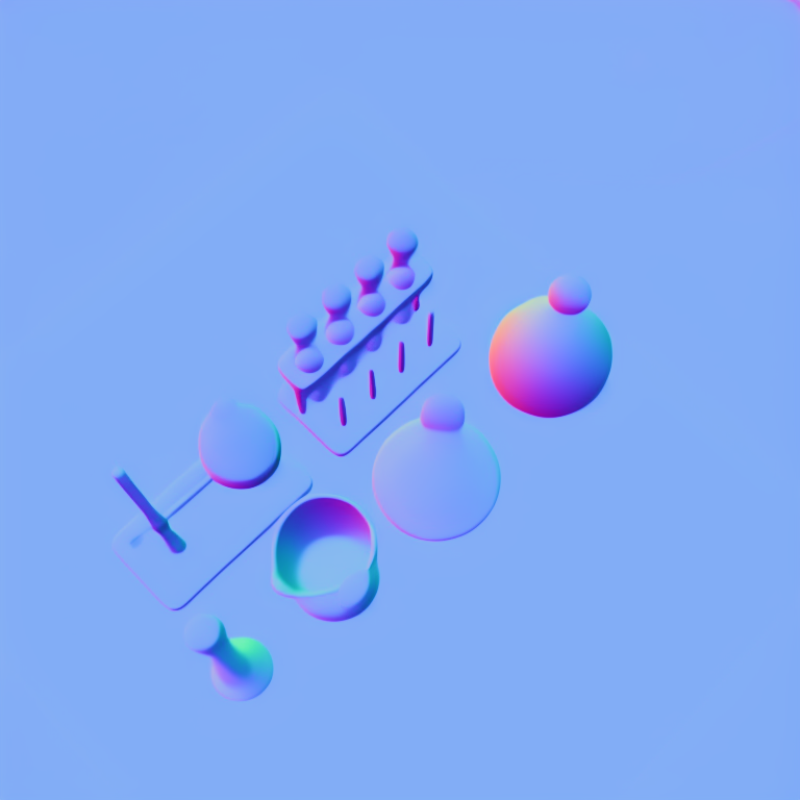} &
        \includegraphics[width=\imgwext]{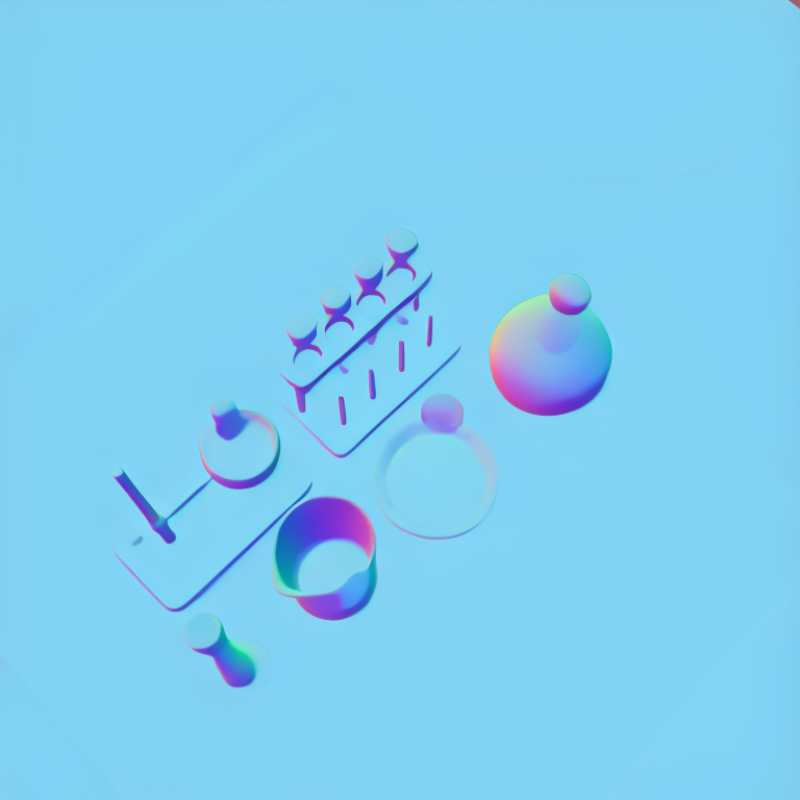} &
        \includegraphics[width=\imgwext]{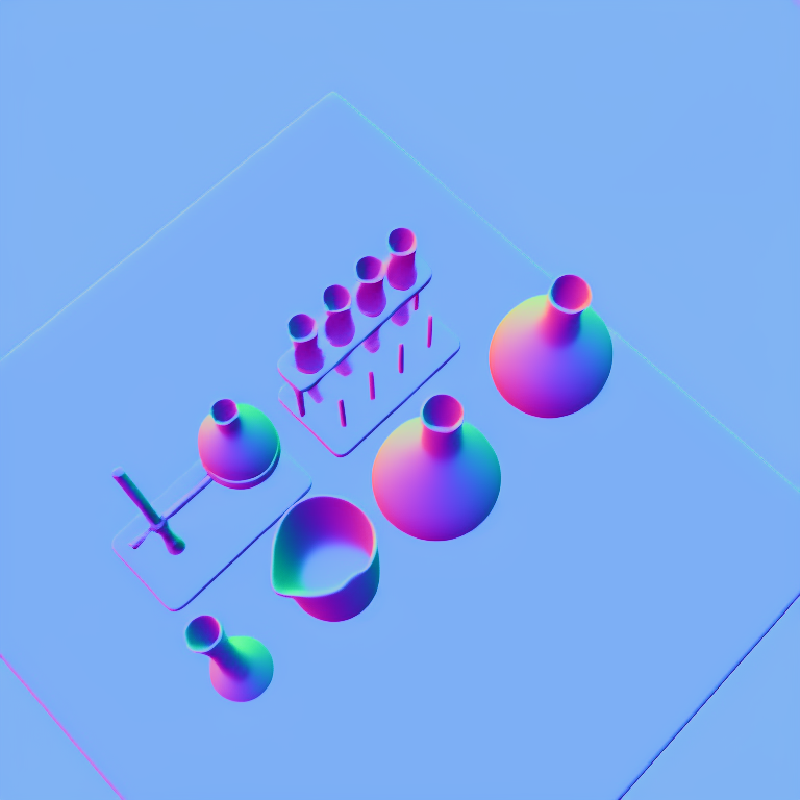} \\
        &
        \includegraphics[width=\imgwext]{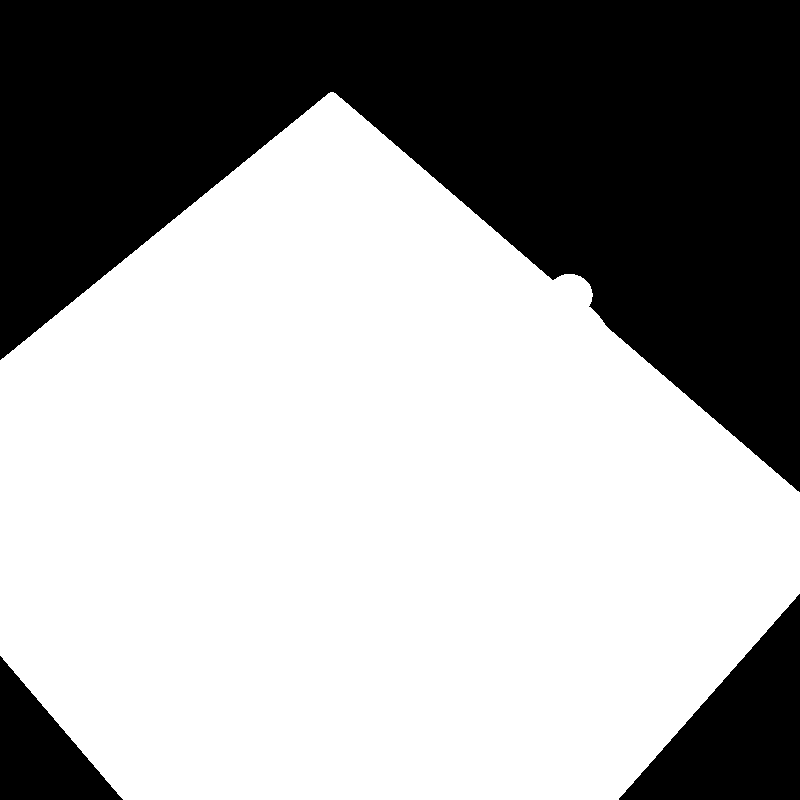} &
        \includegraphics[width=\imgwext]{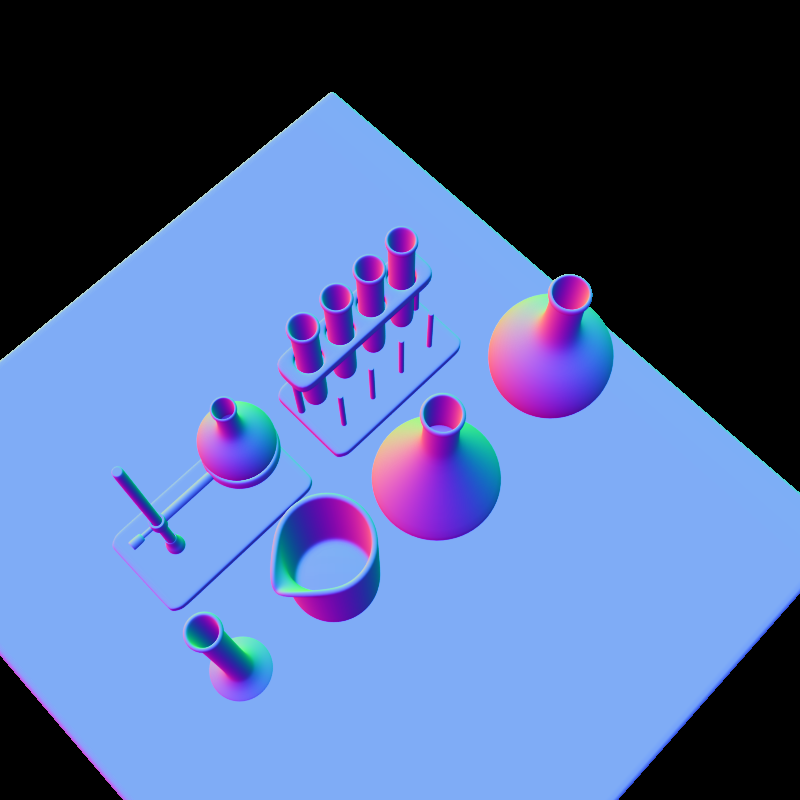} &
        \includegraphics[width=\imgwext]{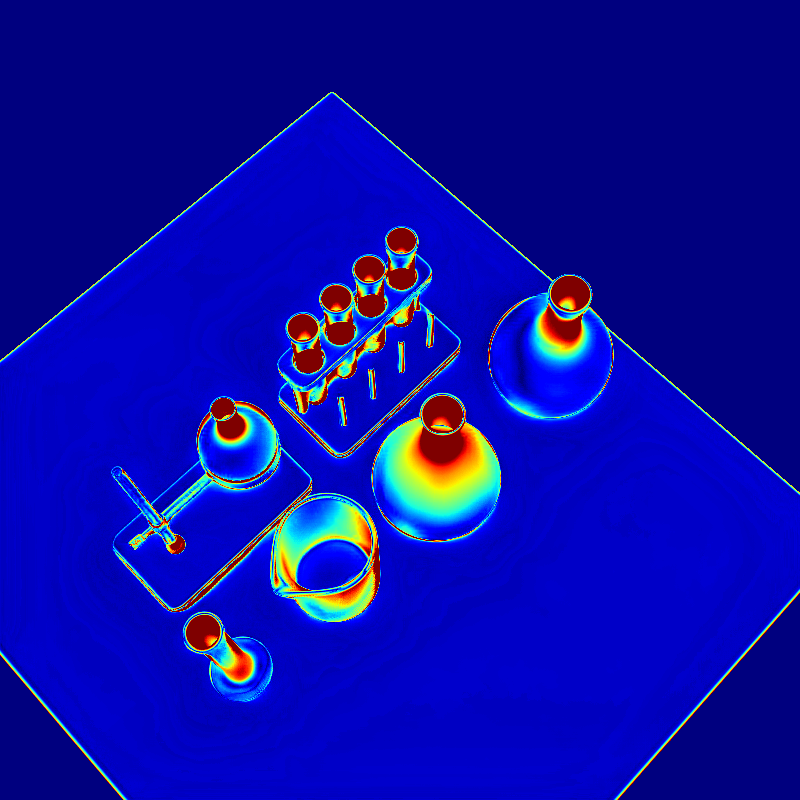} &
        \includegraphics[width=\imgwext]{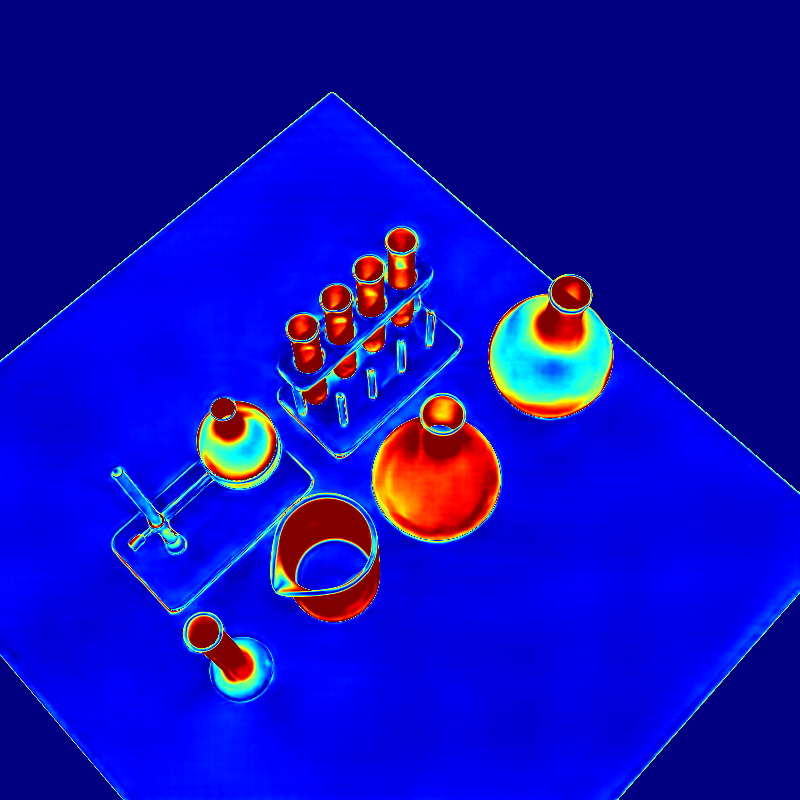} &
        \includegraphics[width=\imgwext]{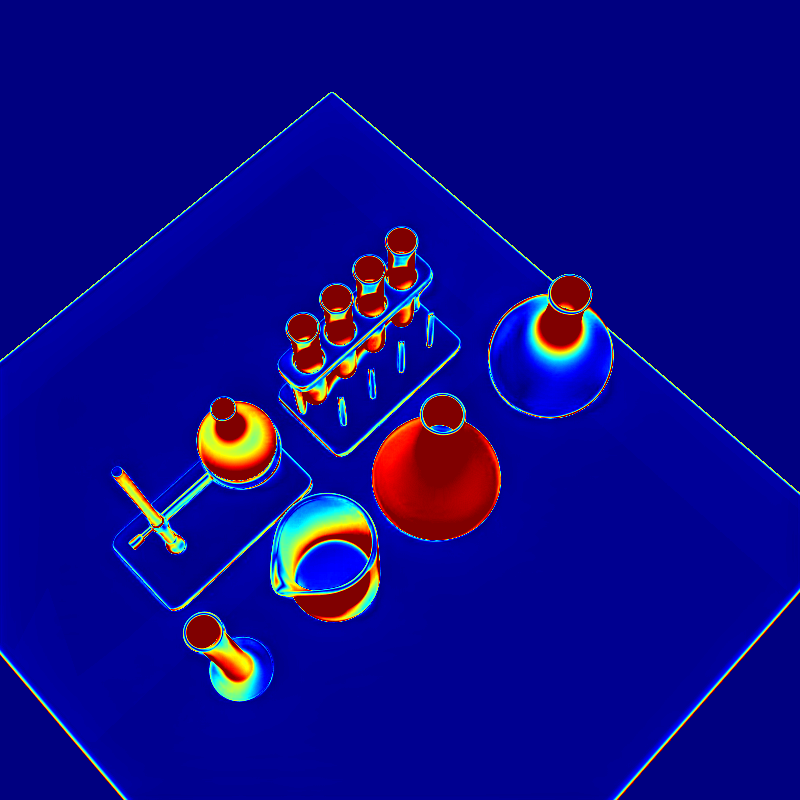} &
        \includegraphics[width=\imgwext]{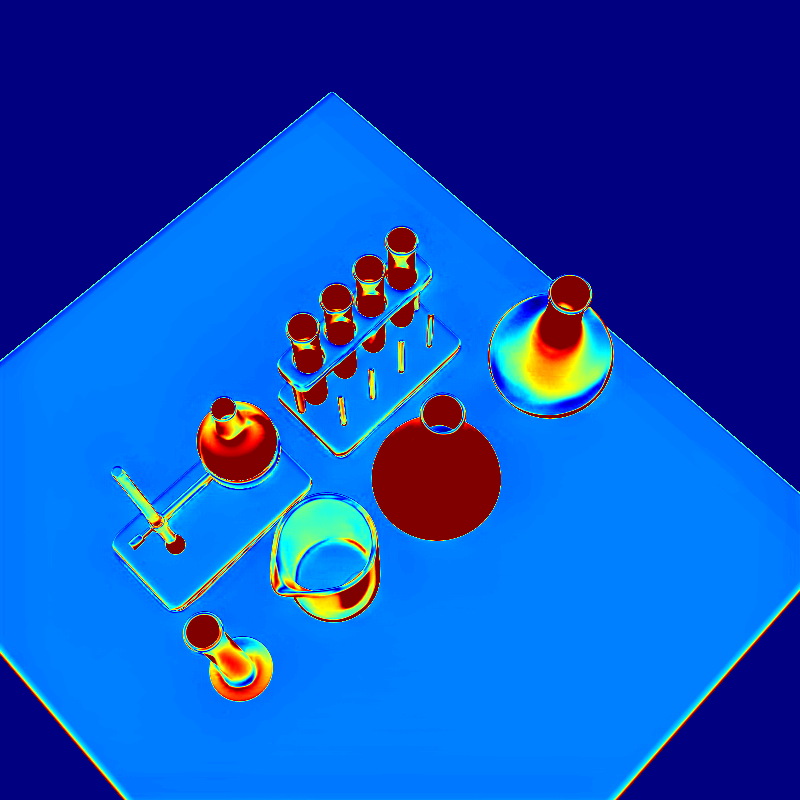} &
        \includegraphics[width=\imgwext]{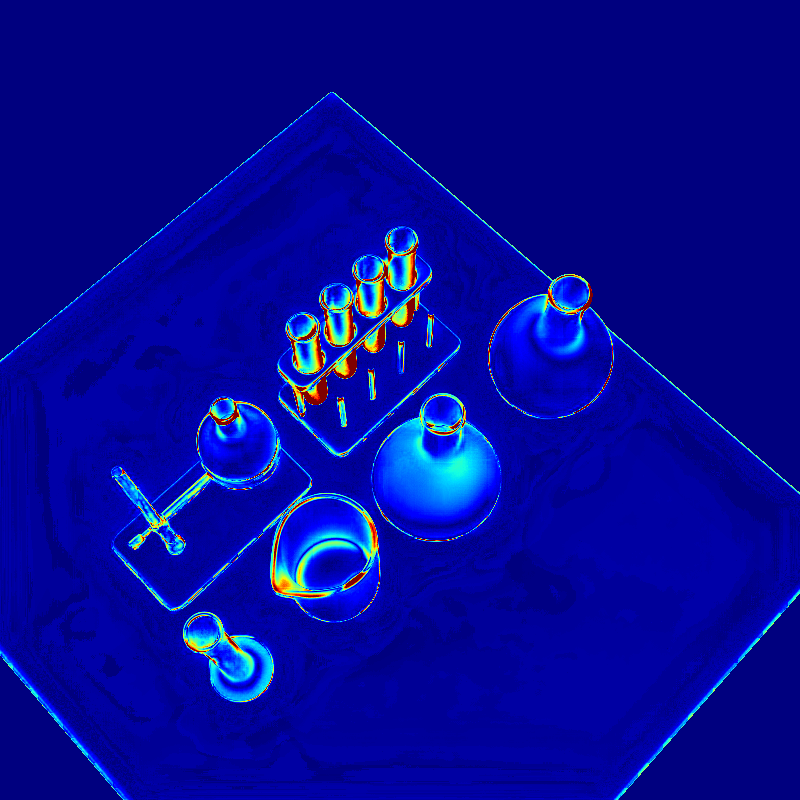} \\[6pt]
        %
        & \small Input/Mask & \small GT & \small DSINE & \small Marigold & \small StableNormal & \small GeoWizard & \small Diception \\[2pt]
        %
        \multirow{2}{*}[3.5ex]{\rotatebox{90}{\small TransNormal}} &
        \includegraphics[width=\imgwext]{sources/datasets/tranlab_02_view_0381/input.png} &
        \includegraphics[width=\imgwext]{sources/datasets/tranlab_02_view_0381/gt_normal.png} &
        \includegraphics[width=\imgwext]{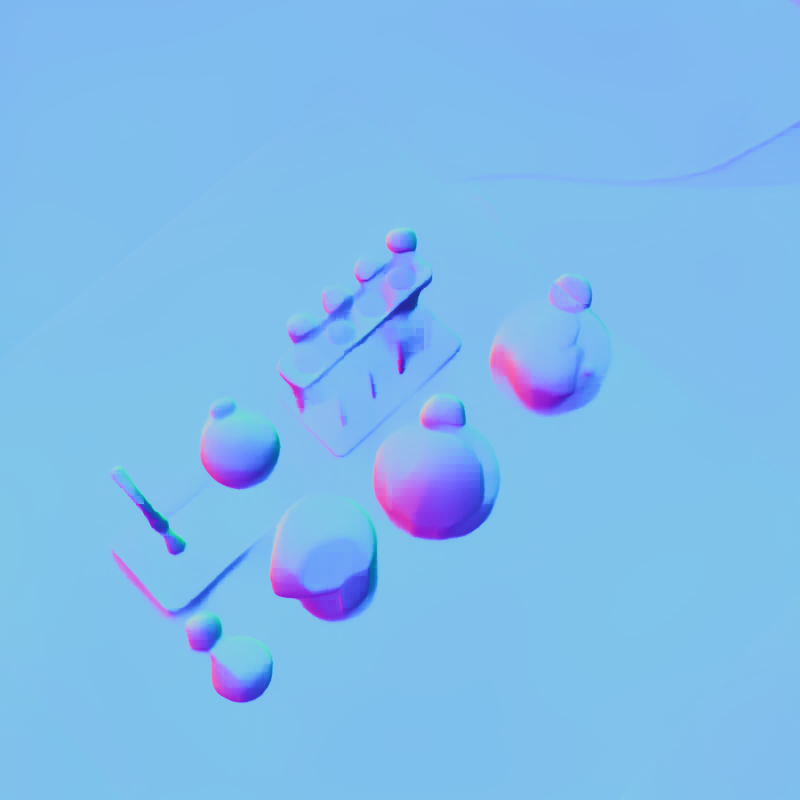} &
        \includegraphics[width=\imgwext]{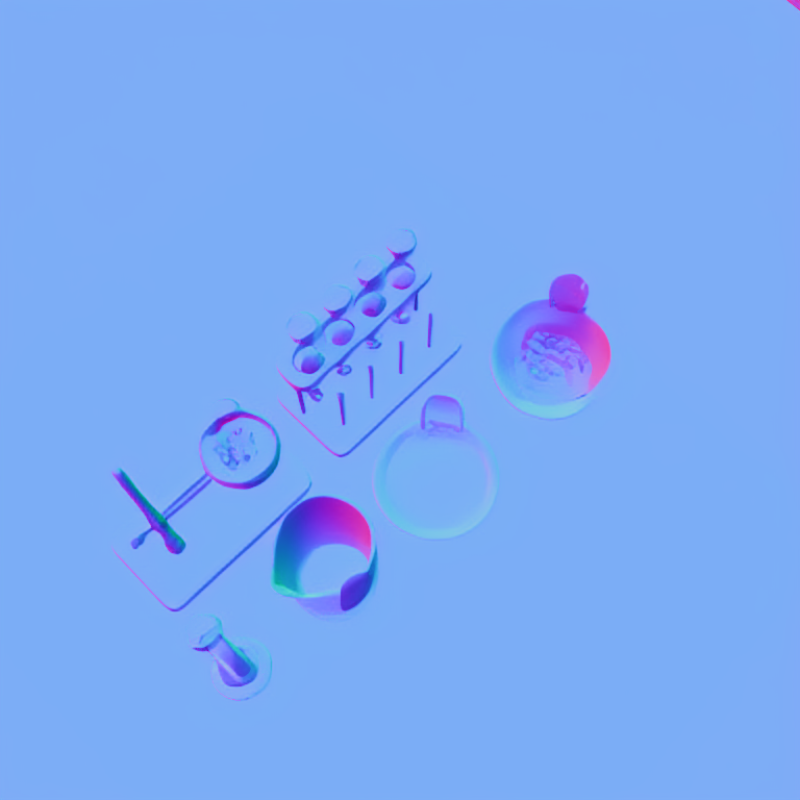} &
        \includegraphics[width=\imgwext]{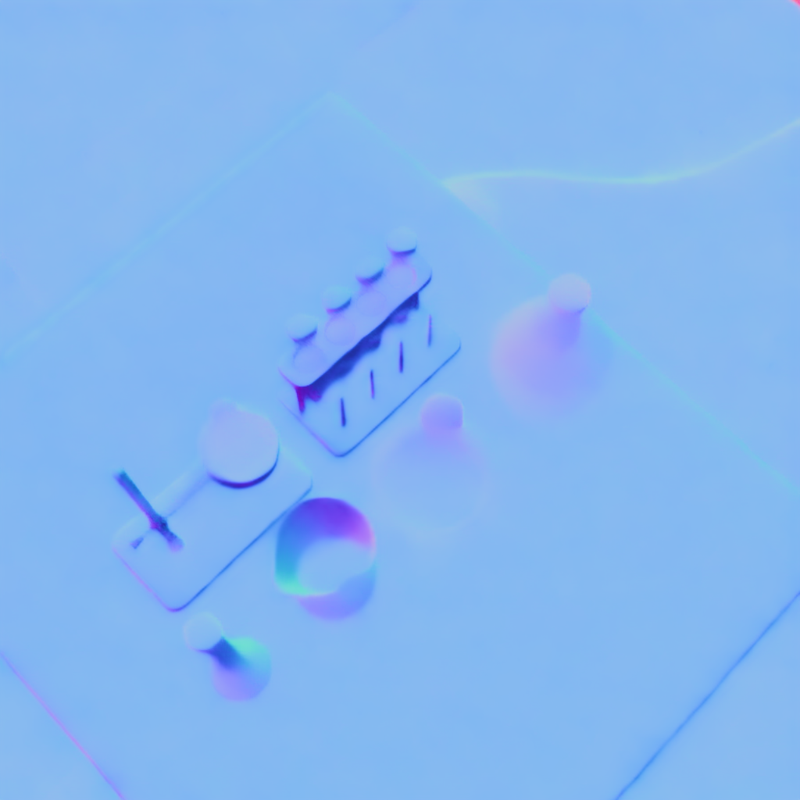} &
        \includegraphics[width=\imgwext]{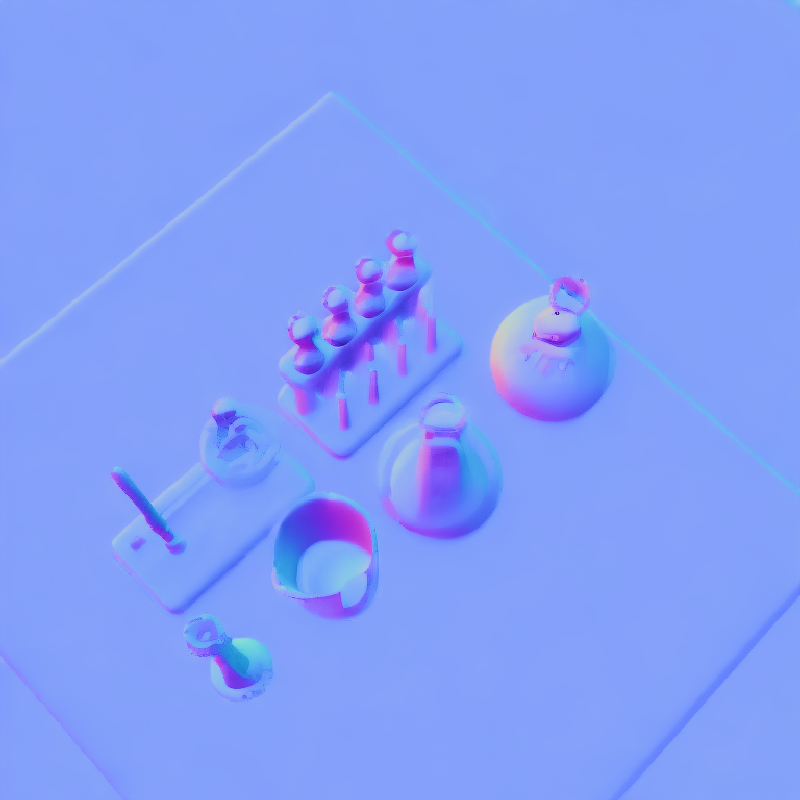} &
        \includegraphics[width=\imgwext]{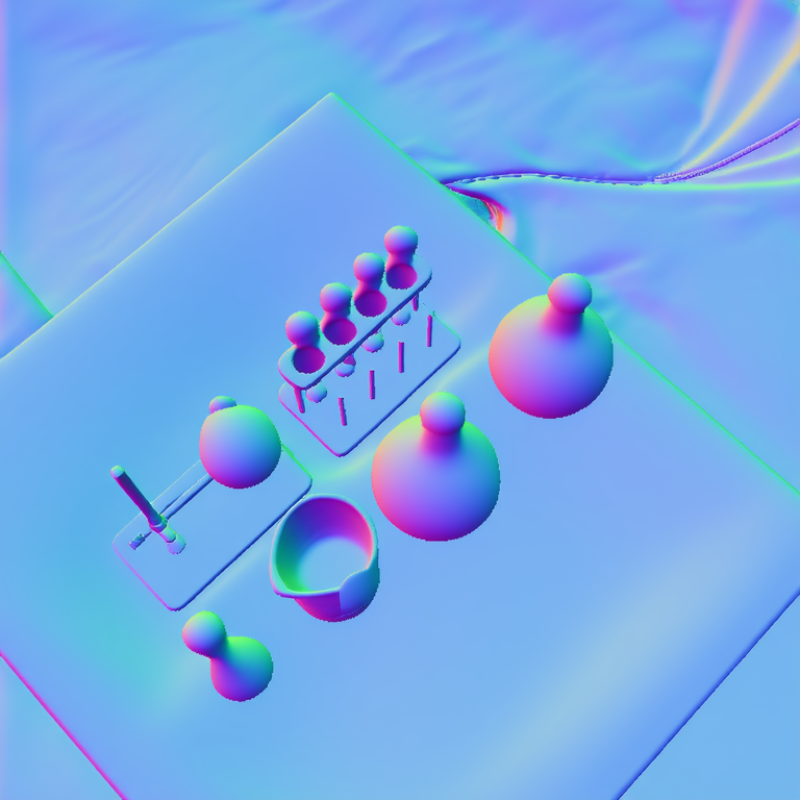} \\
        &
        \includegraphics[width=\imgwext]{sources/datasets/tranlab_02_view_0381/mask.png} &
        \includegraphics[width=\imgwext]{sources/datasets/tranlab_02_view_0381/Ours_gt_masked.png} &
        \includegraphics[width=\imgwext]{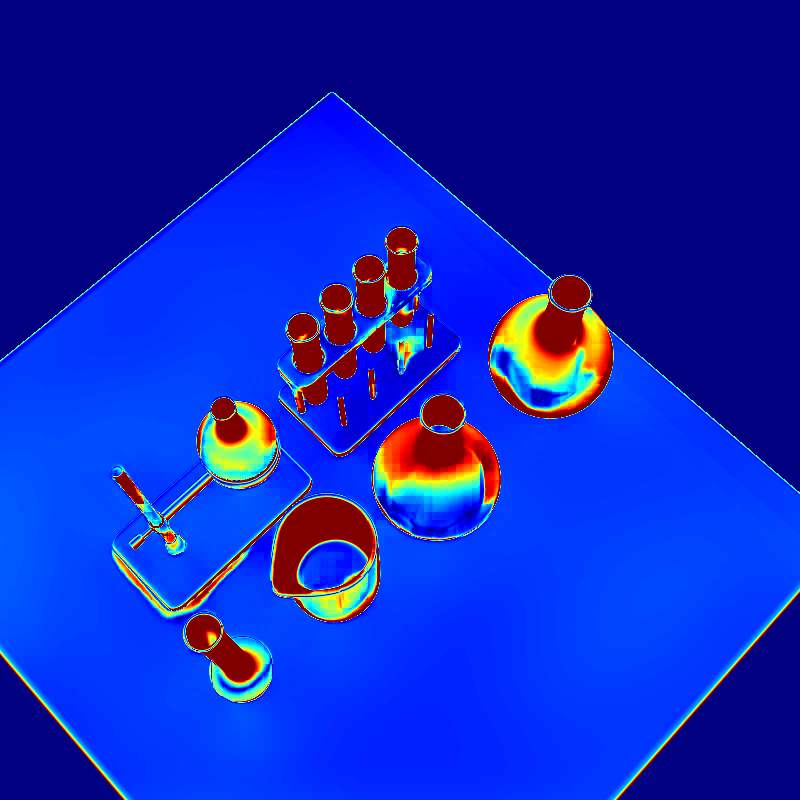} &
        \includegraphics[width=\imgwext]{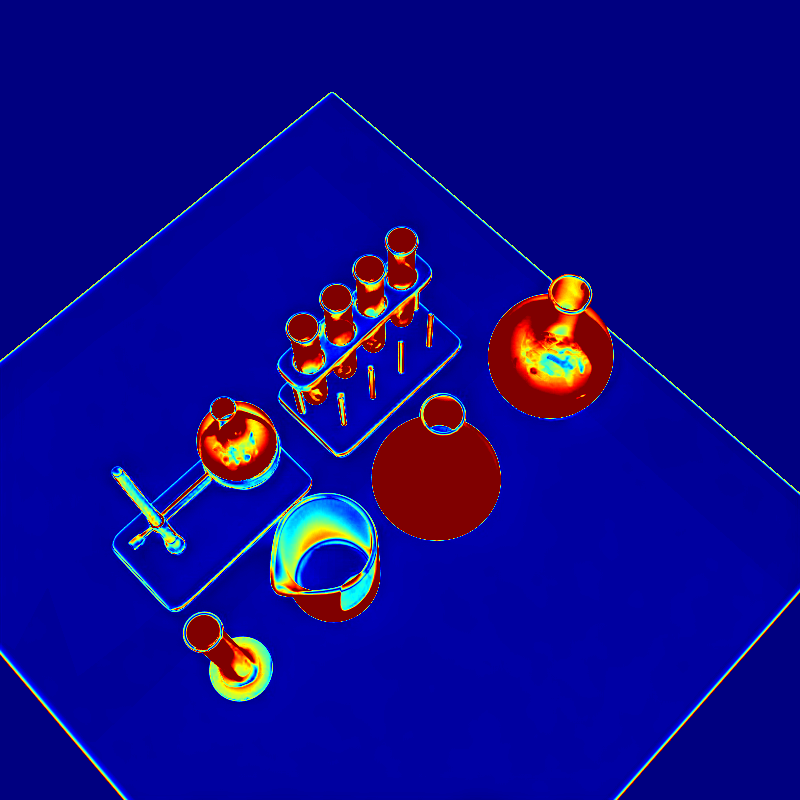} &
        \includegraphics[width=\imgwext]{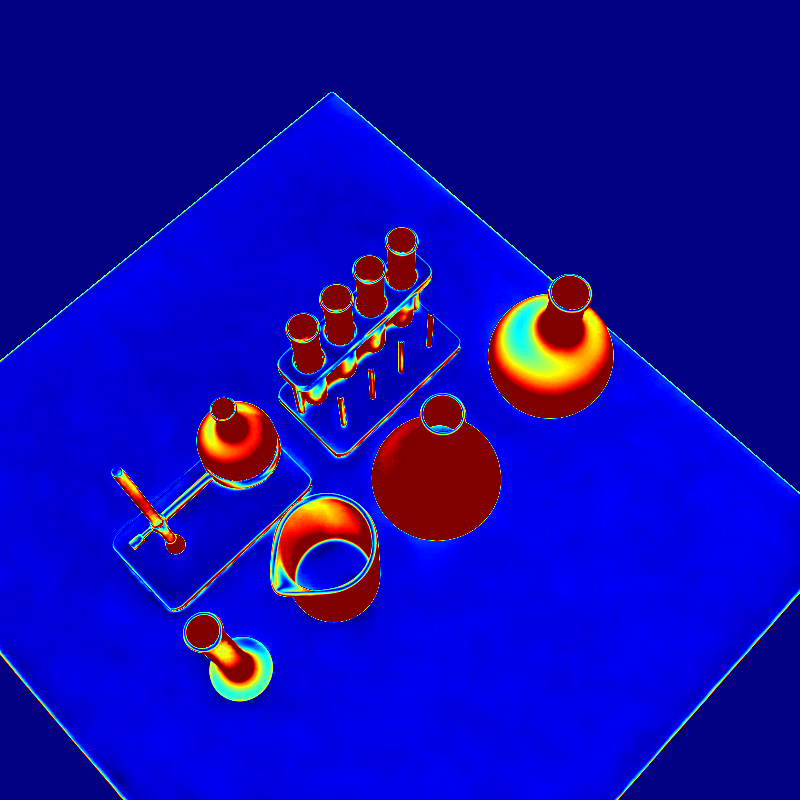} &
        \includegraphics[width=\imgwext]{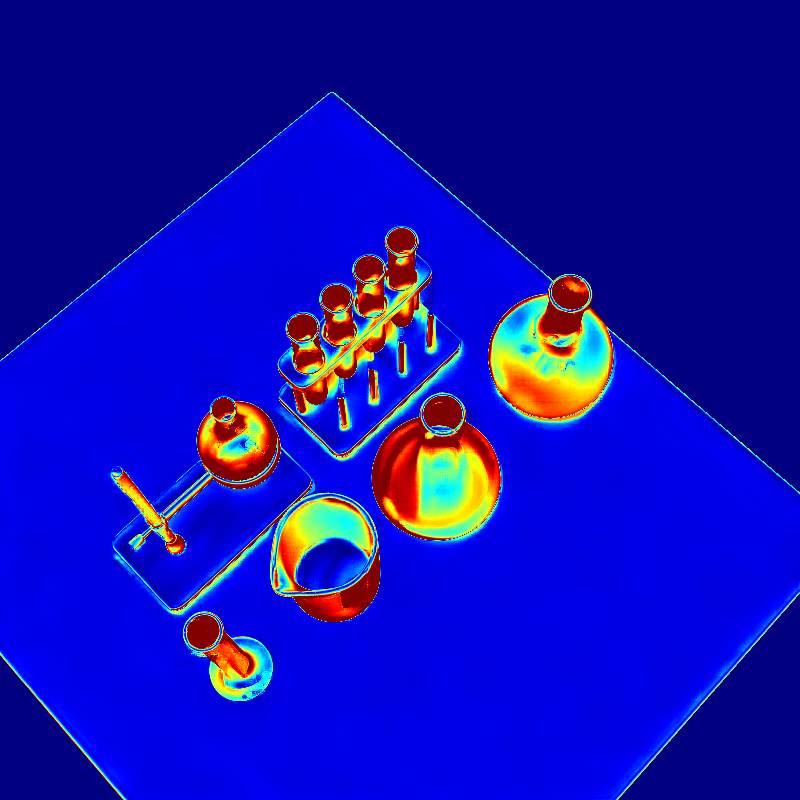} &
        \includegraphics[width=\imgwext]{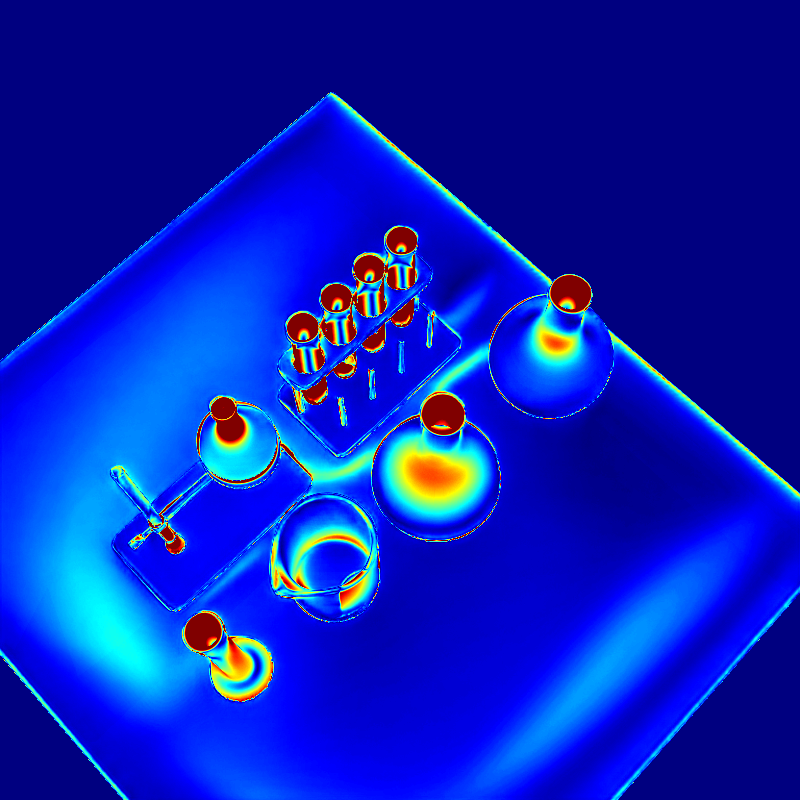} \\[12pt]
        %
        & \small Input/Mask & \small GT & \small Lotus & \small MoGe-2 & \small E2E-FT & \small GenPercept & \small \textbf{Ours} \\[2pt]
        %
        \multirow{2}{*}[3.5ex]{\rotatebox{90}{\small TransNormal}} &
        \includegraphics[width=\imgwext]{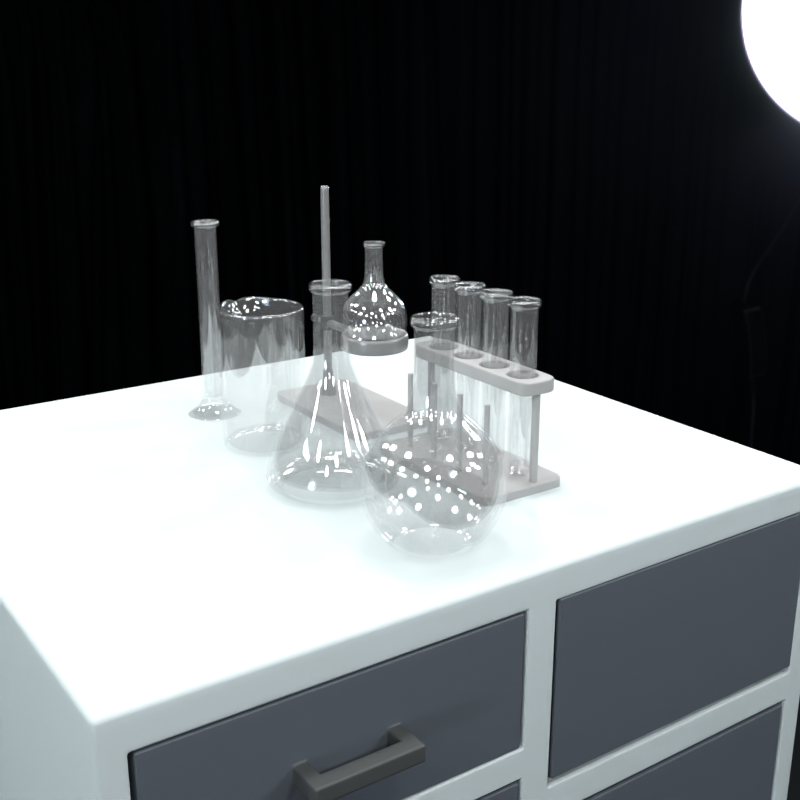} &
        \includegraphics[width=\imgwext]{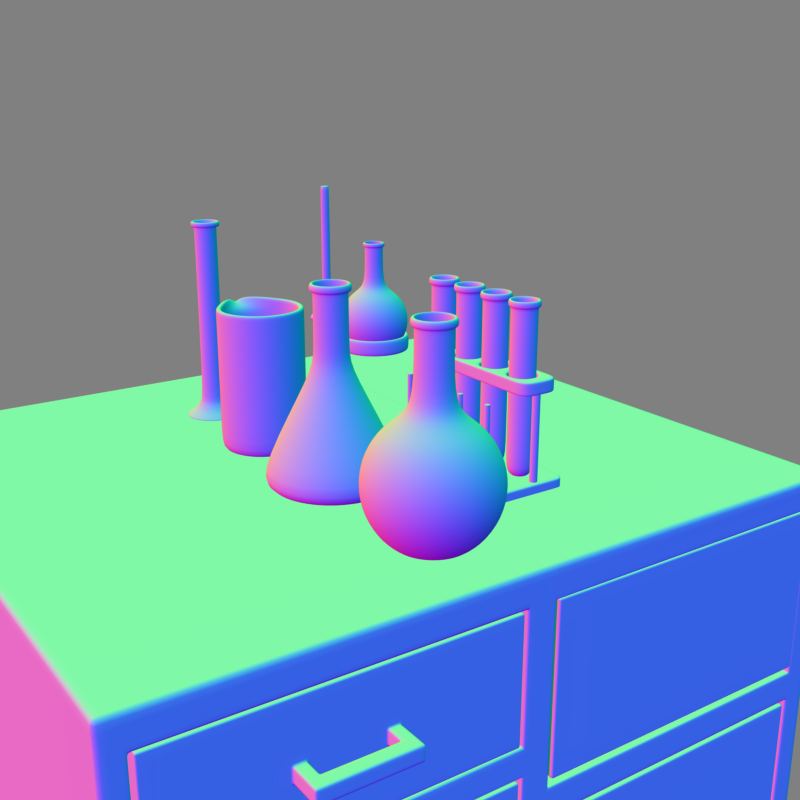} &
        \includegraphics[width=\imgwext]{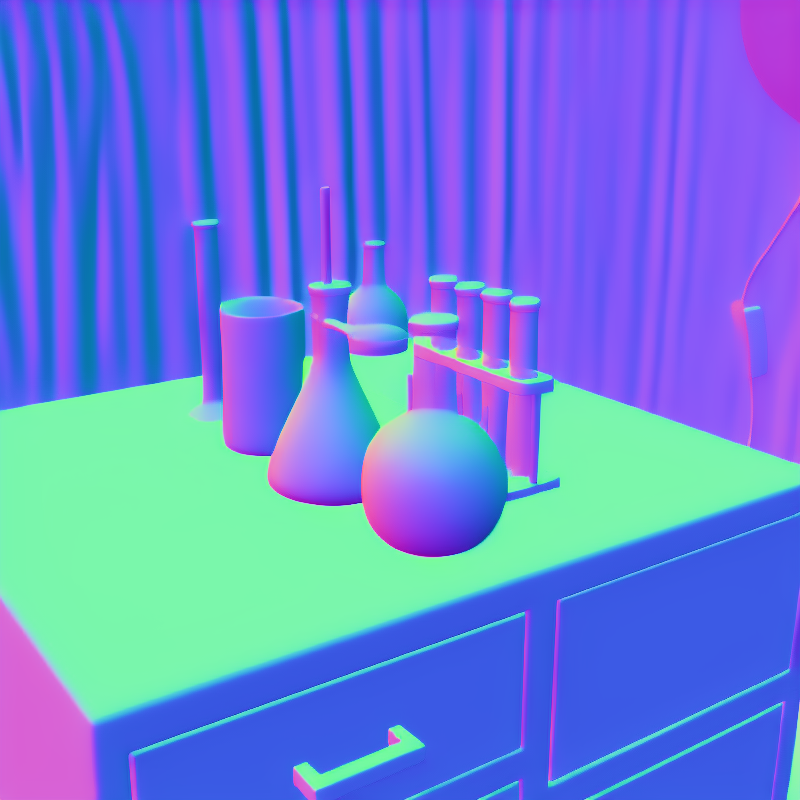} &
        \includegraphics[width=\imgwext]{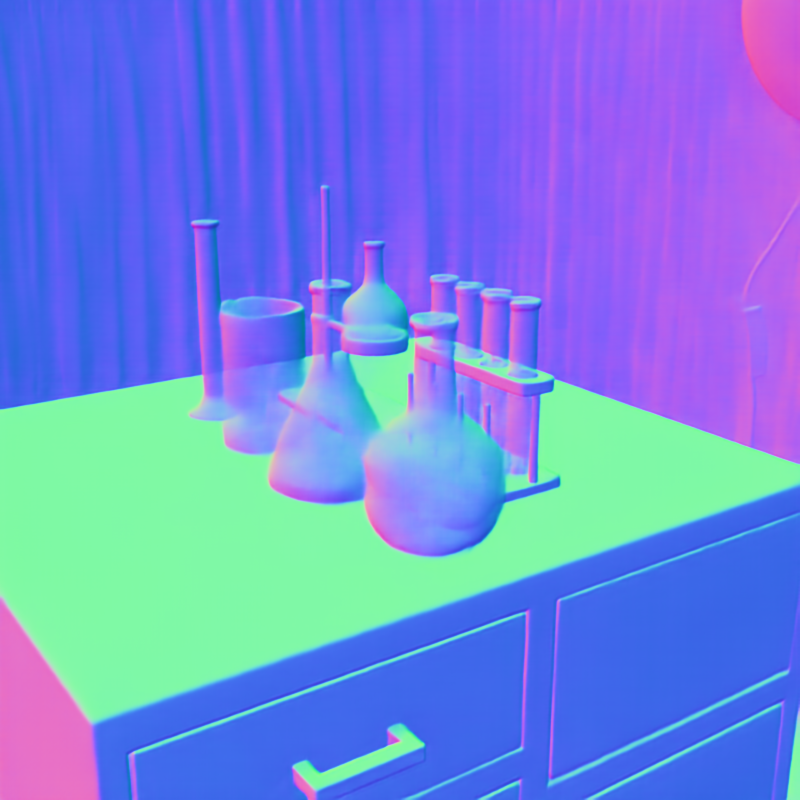} &
        \includegraphics[width=\imgwext]{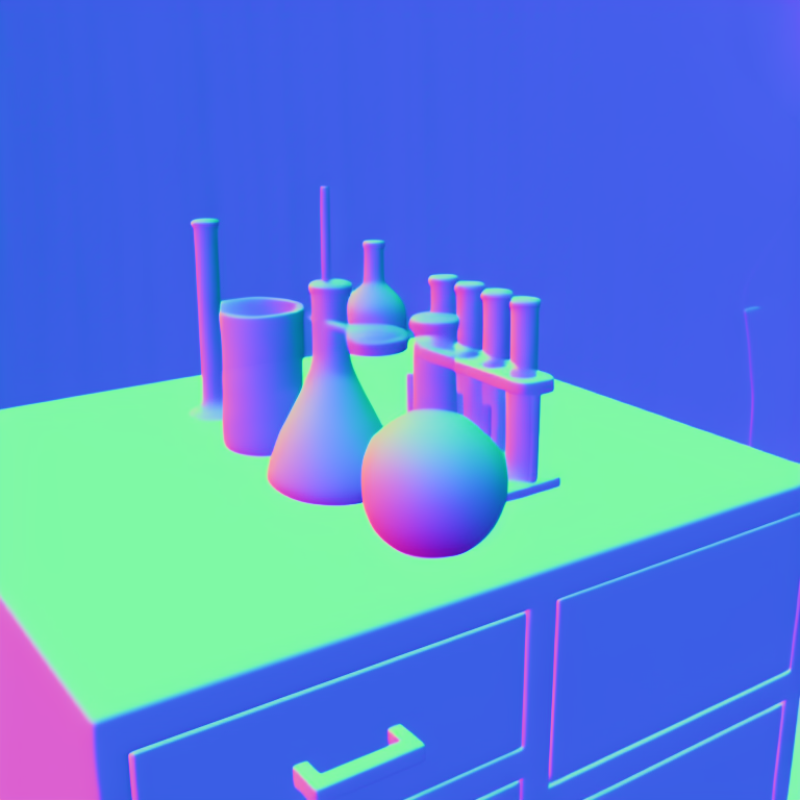} &
        \includegraphics[width=\imgwext]{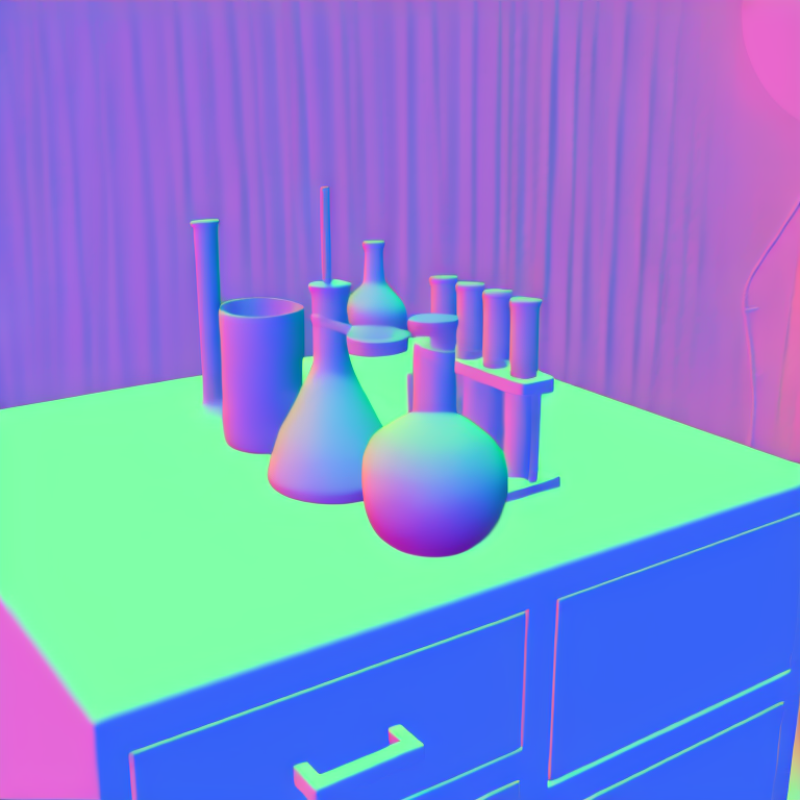} &
        \includegraphics[width=\imgwext]{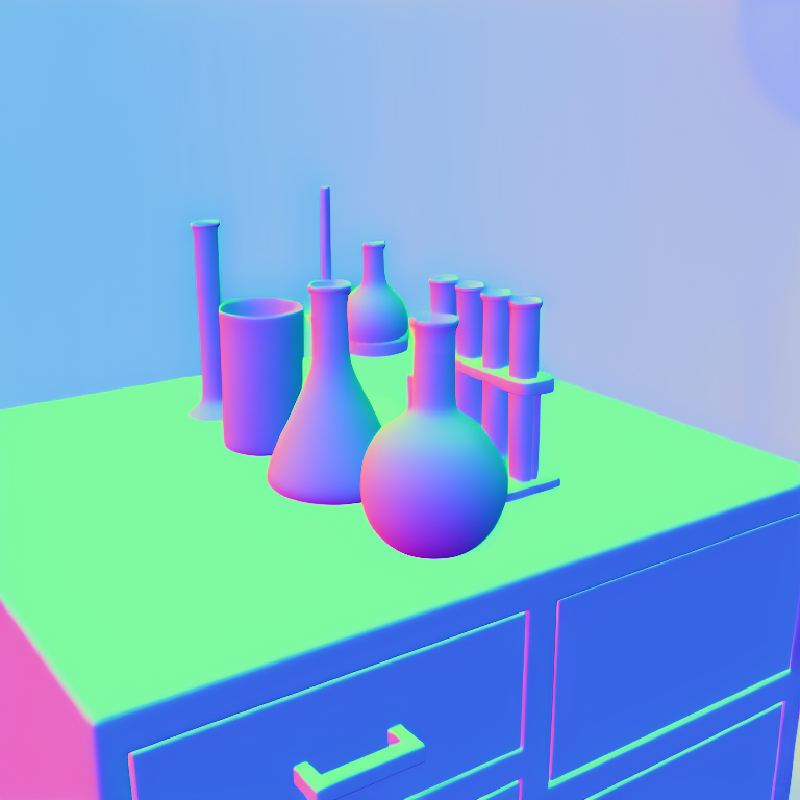} \\
        &
        \includegraphics[width=\imgwext]{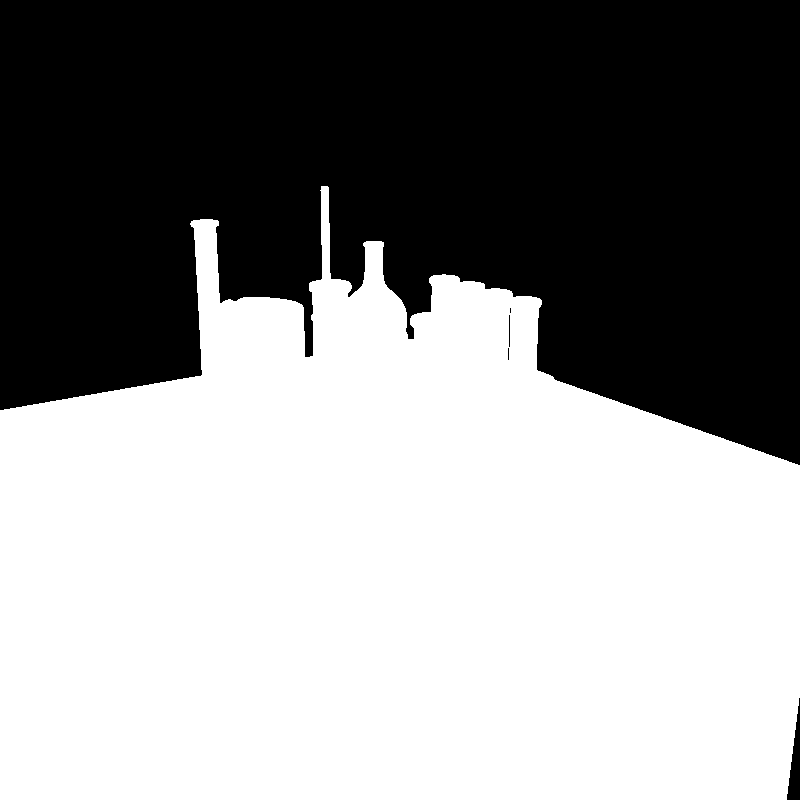} &
        \includegraphics[width=\imgwext]{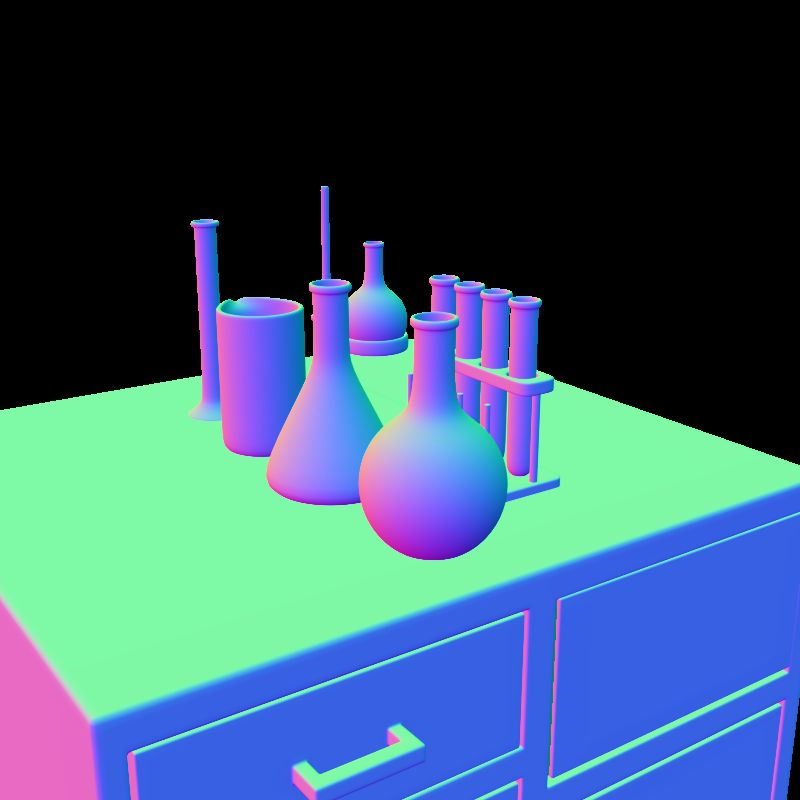} &
        \includegraphics[width=\imgwext]{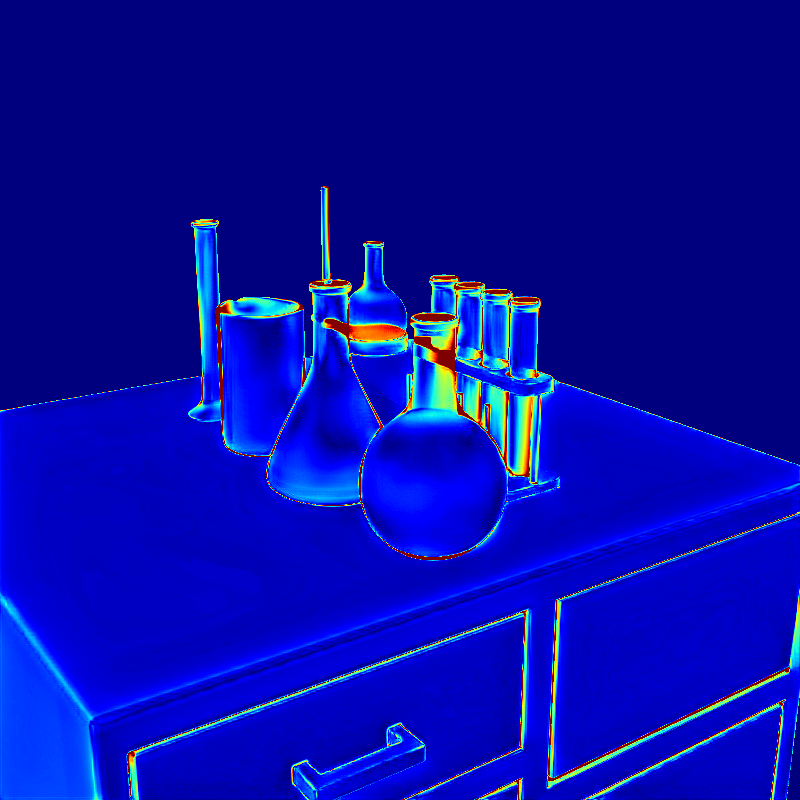} &
        \includegraphics[width=\imgwext]{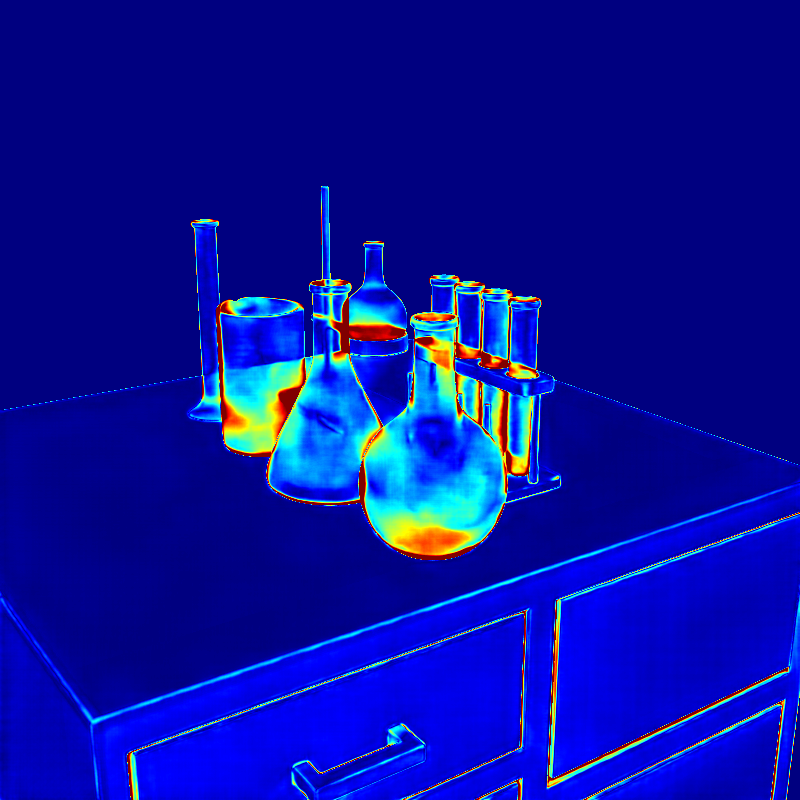} &
        \includegraphics[width=\imgwext]{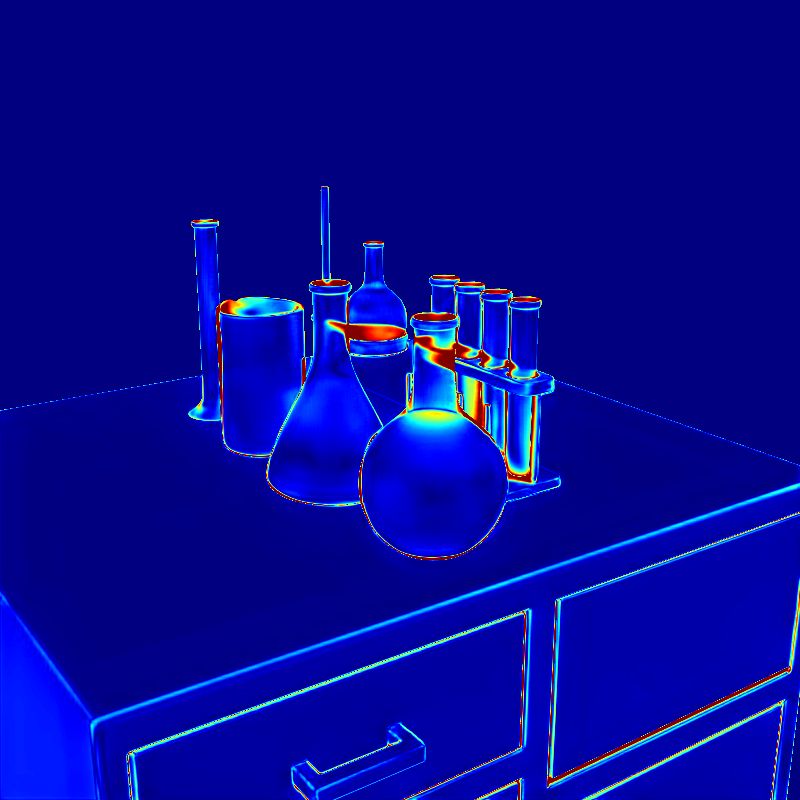} &
        \includegraphics[width=\imgwext]{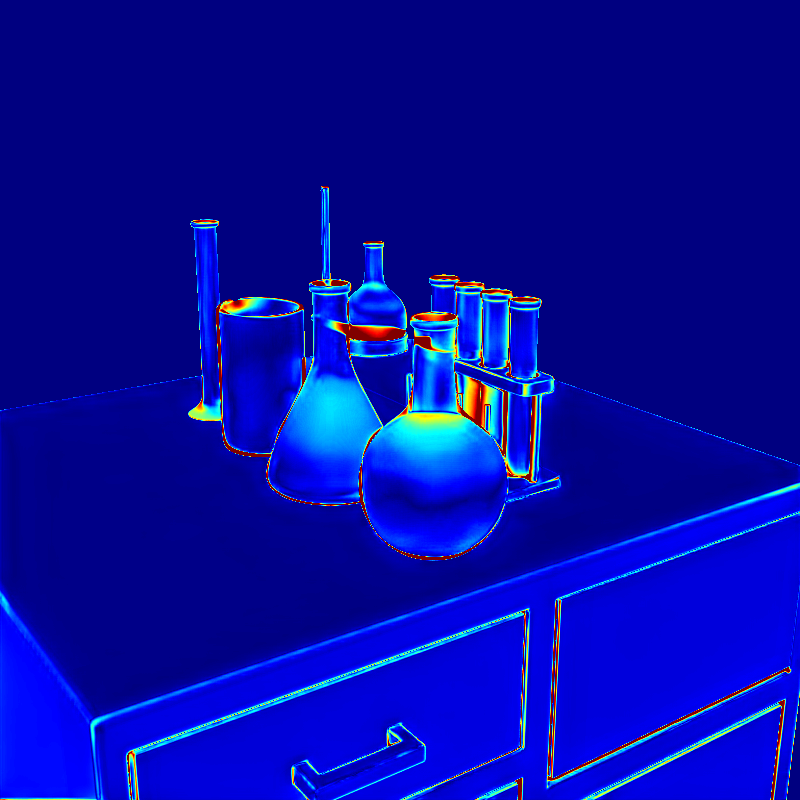} &
        \includegraphics[width=\imgwext]{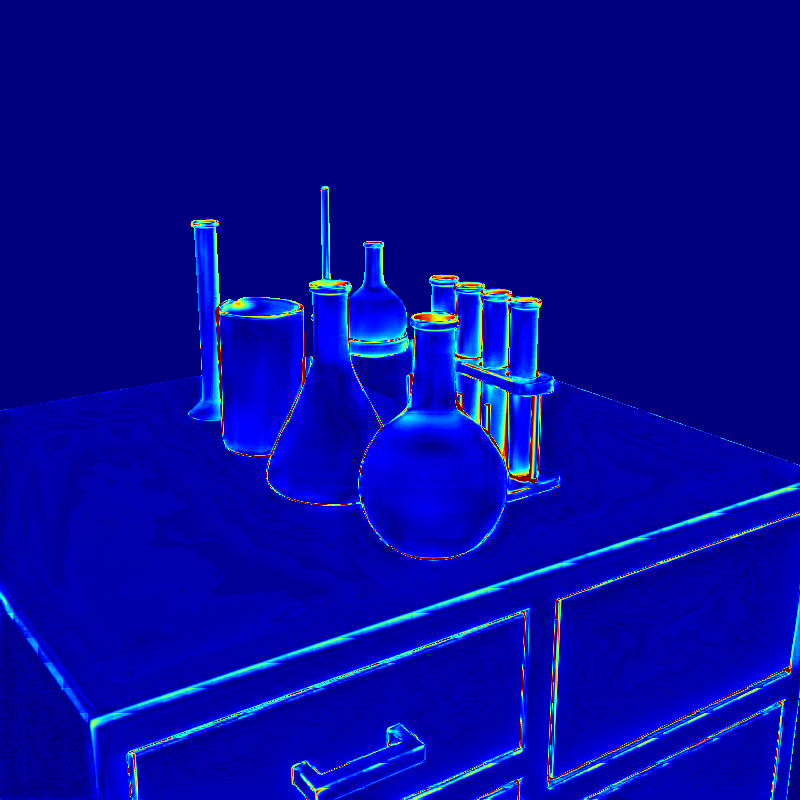} \\[6pt]
        %
        & \small Input/Mask & \small GT & \small DSINE & \small Marigold & \small StableNormal & \small GeoWizard & \small Diception \\[2pt]
        %
        \multirow{2}{*}[3.5ex]{\rotatebox{90}{\small TransNormal}} &
        \includegraphics[width=\imgwext]{sources/datasets/tranlab_02_view_0020/input.png} &
        \includegraphics[width=\imgwext]{sources/datasets/tranlab_02_view_0020/gt_normal.png} &
        \includegraphics[width=\imgwext]{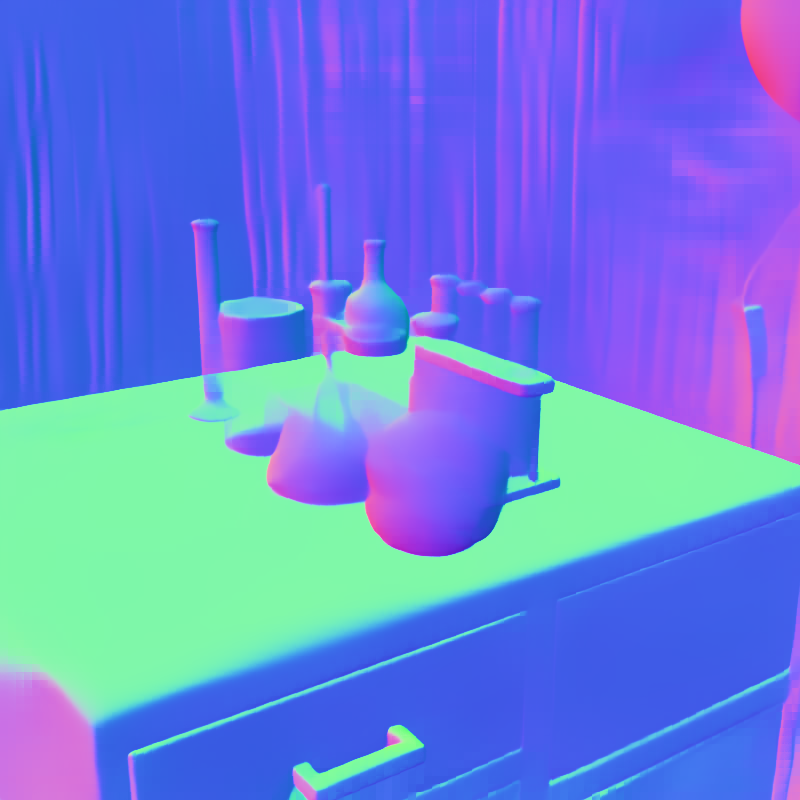} &
        \includegraphics[width=\imgwext]{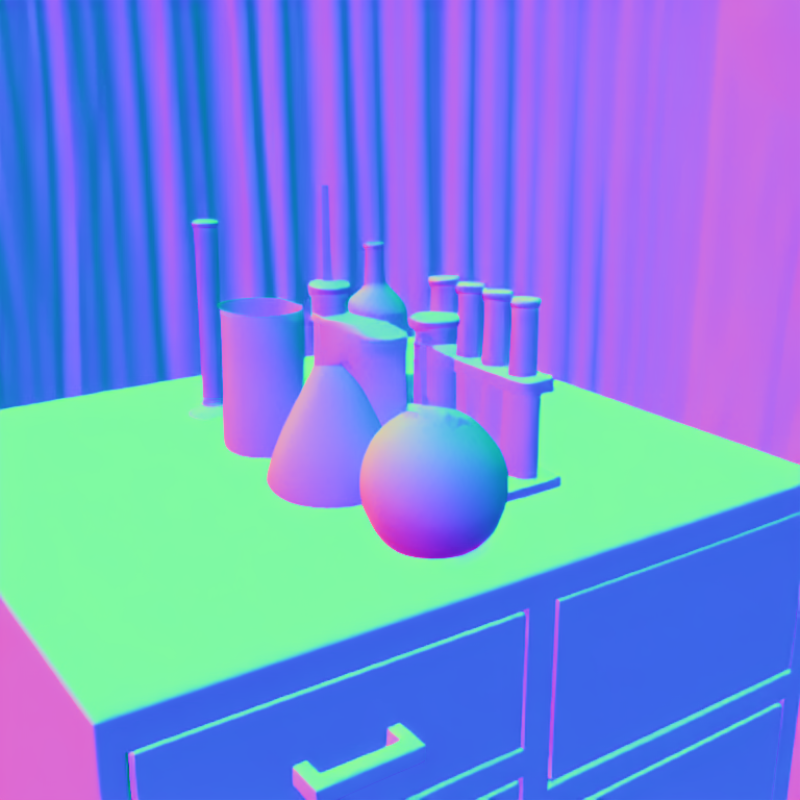} &
        \includegraphics[width=\imgwext]{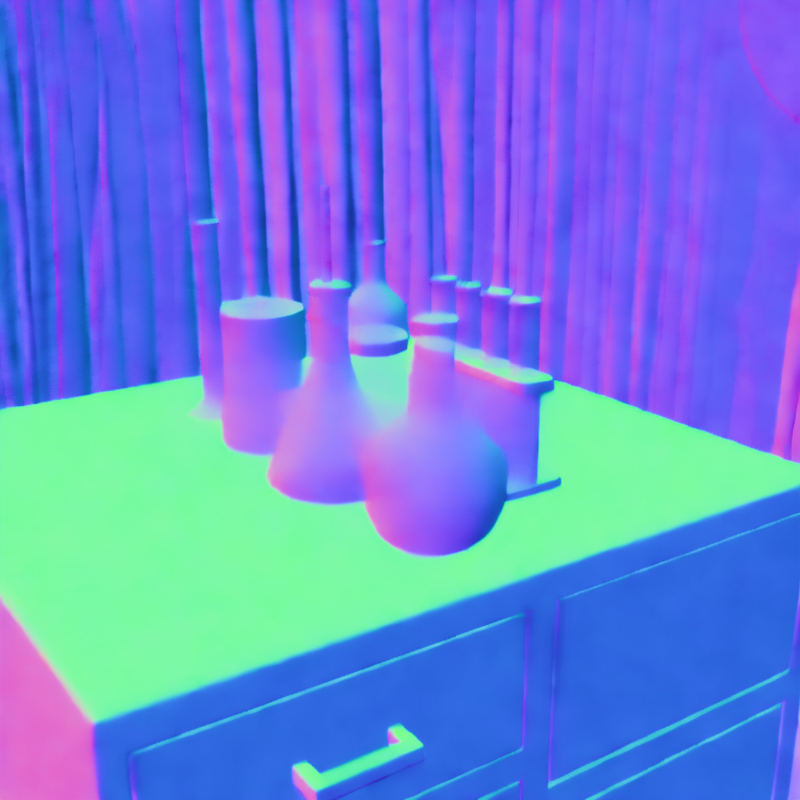} &
        \includegraphics[width=\imgwext]{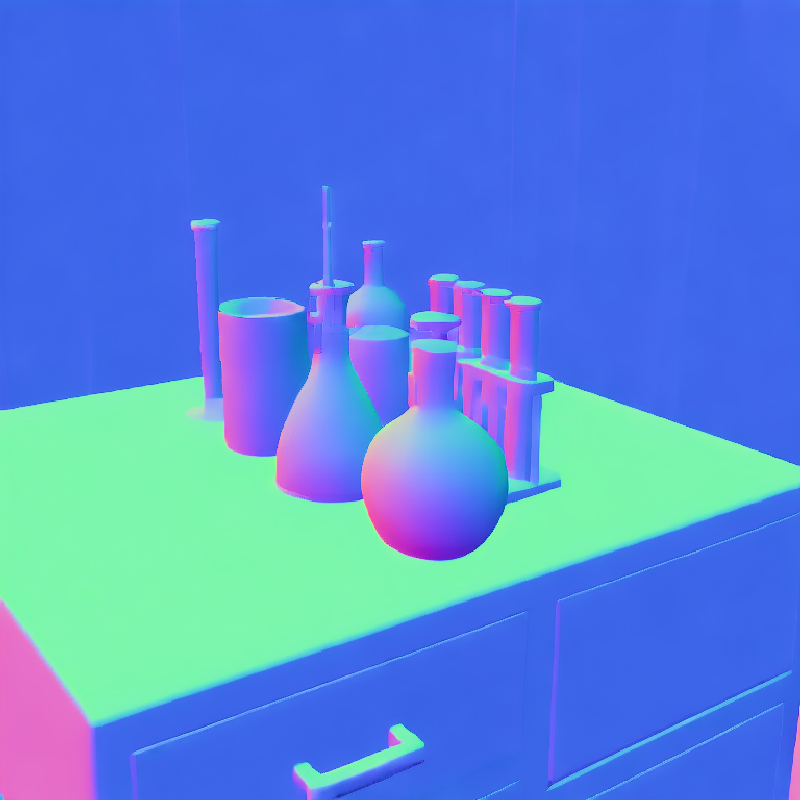} &
        \includegraphics[width=\imgwext]{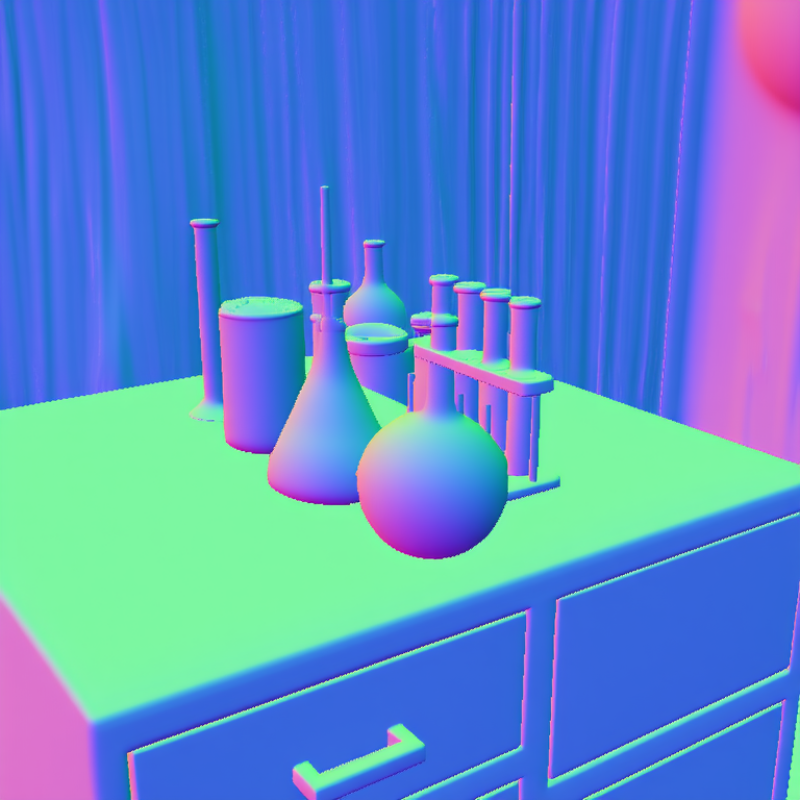} \\
        &
        \includegraphics[width=\imgwext]{sources/datasets/tranlab_02_view_0020/mask.png} &
        \includegraphics[width=\imgwext]{sources/datasets/tranlab_02_view_0020/Ours_gt_masked.png} &
        \includegraphics[width=\imgwext]{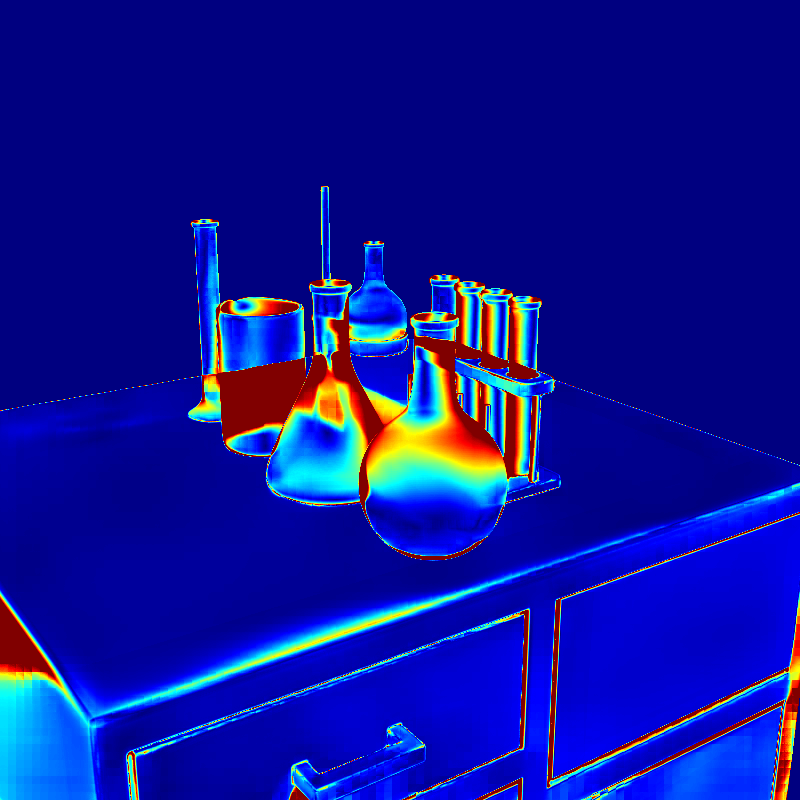} &
        \includegraphics[width=\imgwext]{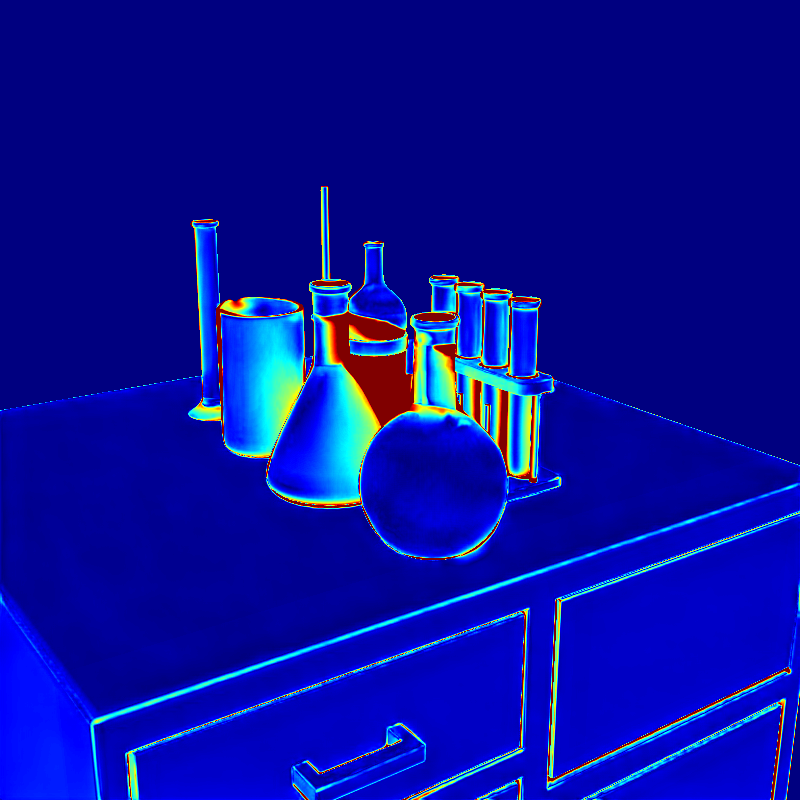} &
        \includegraphics[width=\imgwext]{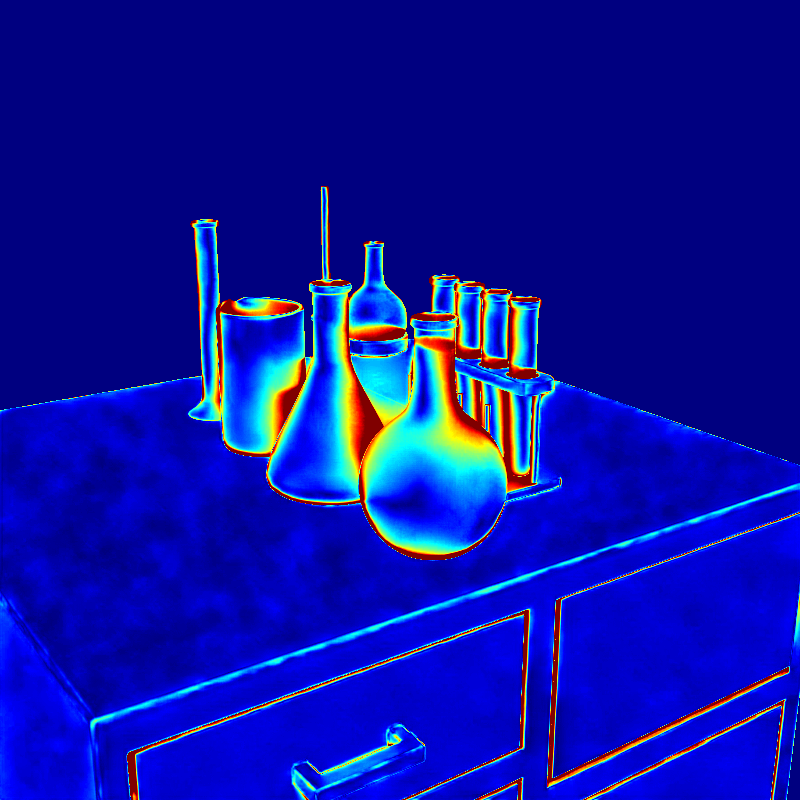} &
        \includegraphics[width=\imgwext]{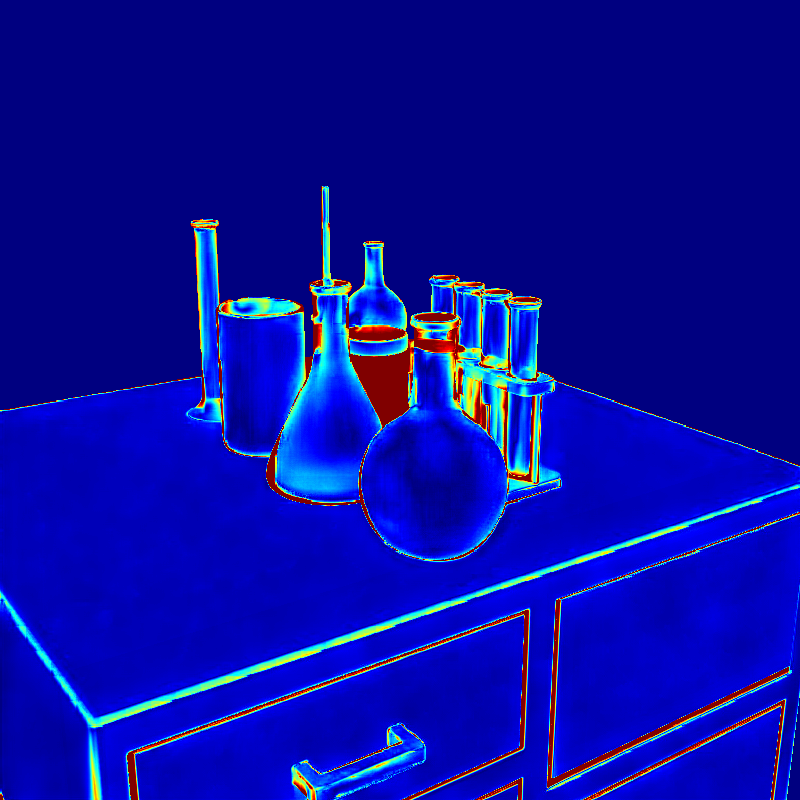} &
        \includegraphics[width=\imgwext]{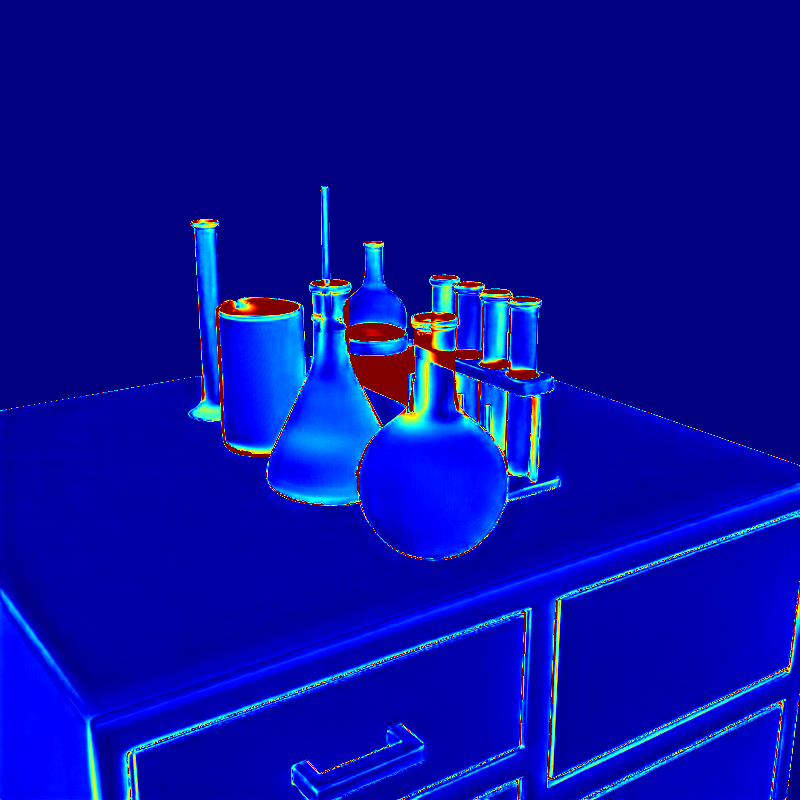} \\
    \end{tabular}
    \vspace{-1mm}
    \caption{\textbf{Extended qualitative comparison with baseline methods.}
    We compare against 9 baselines.
    Top rows show predicted normals; bottom rows show angular error maps (blue: low, red: high).
    Our method consistently produces sharper edges and lower error on transparent regions. Please zoom in \faSearch~for details. (\S~\ref{ssec:extended_comparison})}
    \vspace{-3mm}
    \label{fig:comparison_extended}
\end{figure*}

\begin{figure*}[t]
    \centering
    \setlength{\tabcolsep}{1pt}
    \renewcommand{\arraystretch}{0.6}
    \newcommand{\imgwcg}{0.135\textwidth}
    %
    %
    \begin{tabular}{@{}c@{\hspace{1pt}}cc|ccccc@{}}
        & \small Input/Mask & \small GT & \small Lotus & \small MoGe-2 & \small E2E-FT & \small GenPercept & \small \textbf{Ours} \\[2pt]
        %
        \multirow{2}{*}[3.5ex]{\rotatebox{90}{\small ClearGrasp}} &
        \includegraphics[width=\imgwcg]{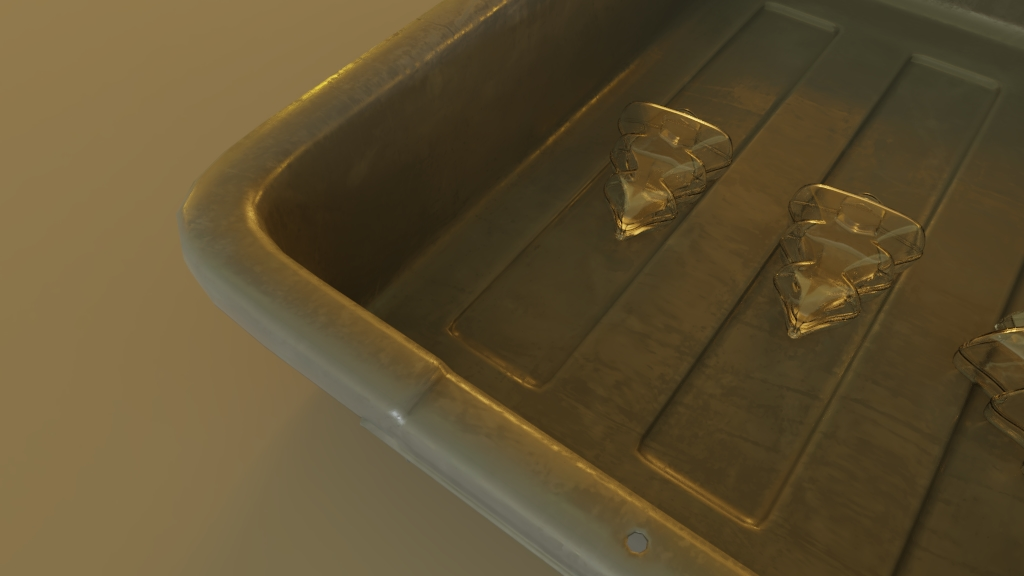} &
        \includegraphics[width=\imgwcg]{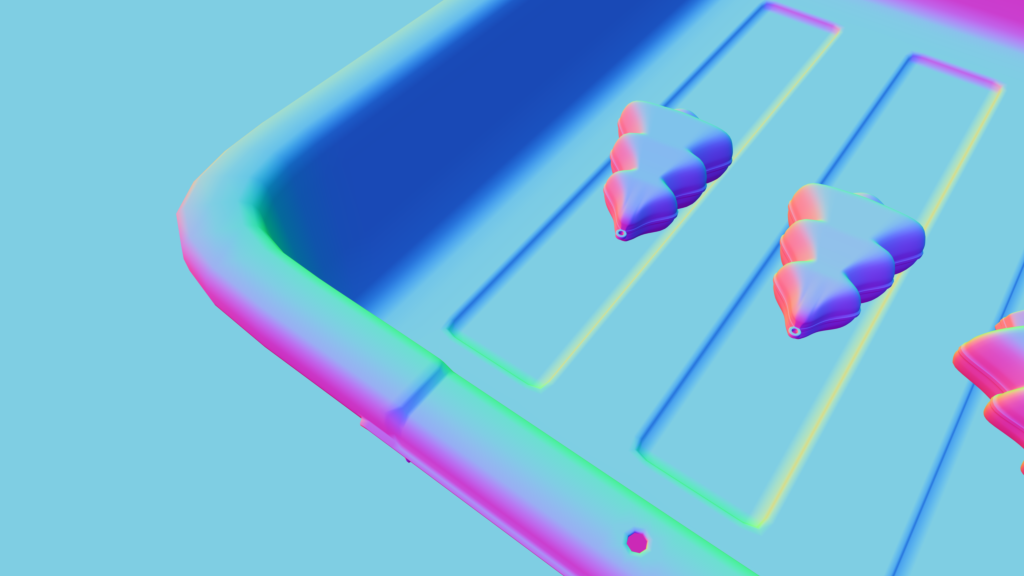} &
        \includegraphics[width=\imgwcg]{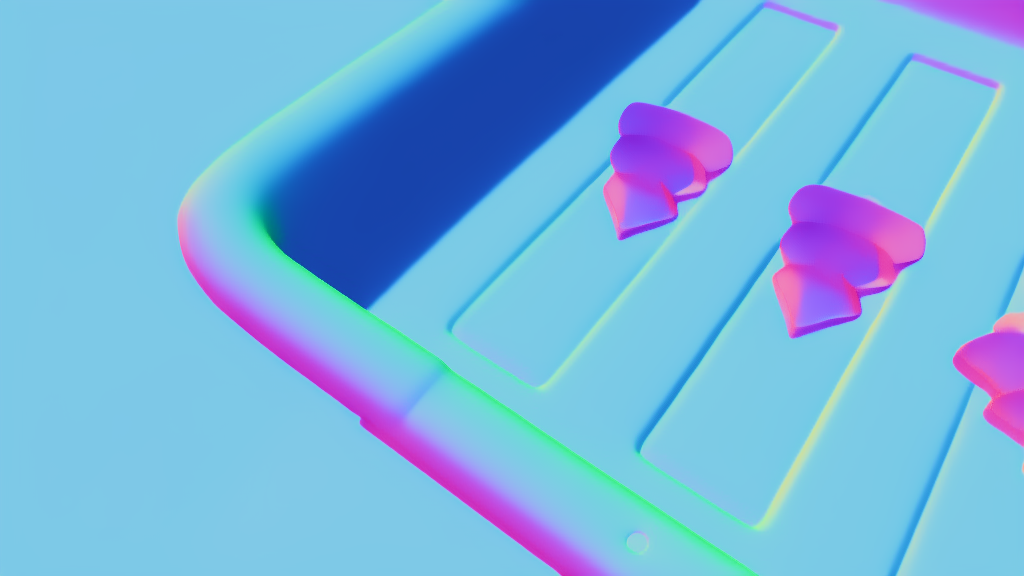} &
        \includegraphics[width=\imgwcg]{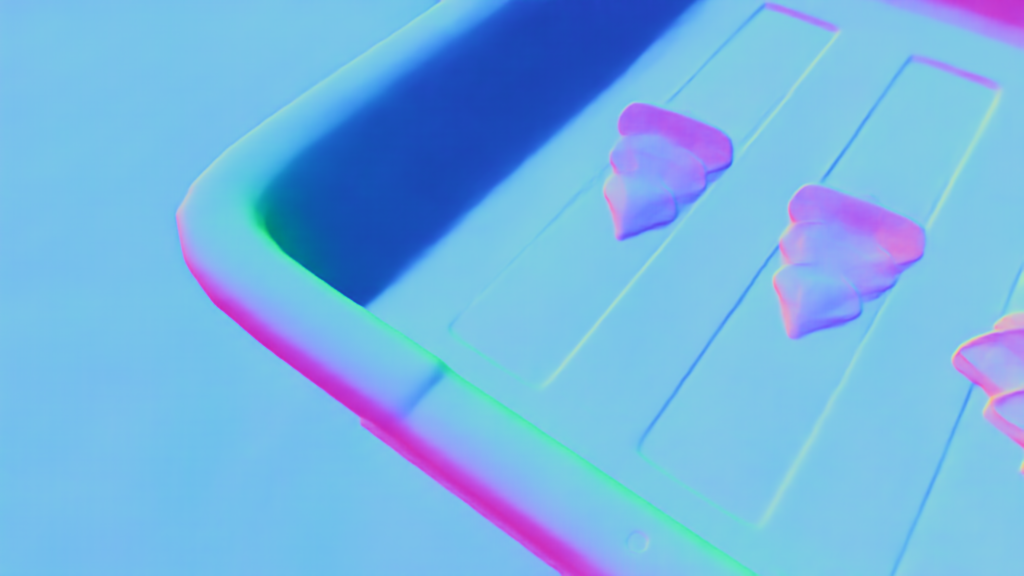} &
        \includegraphics[width=\imgwcg]{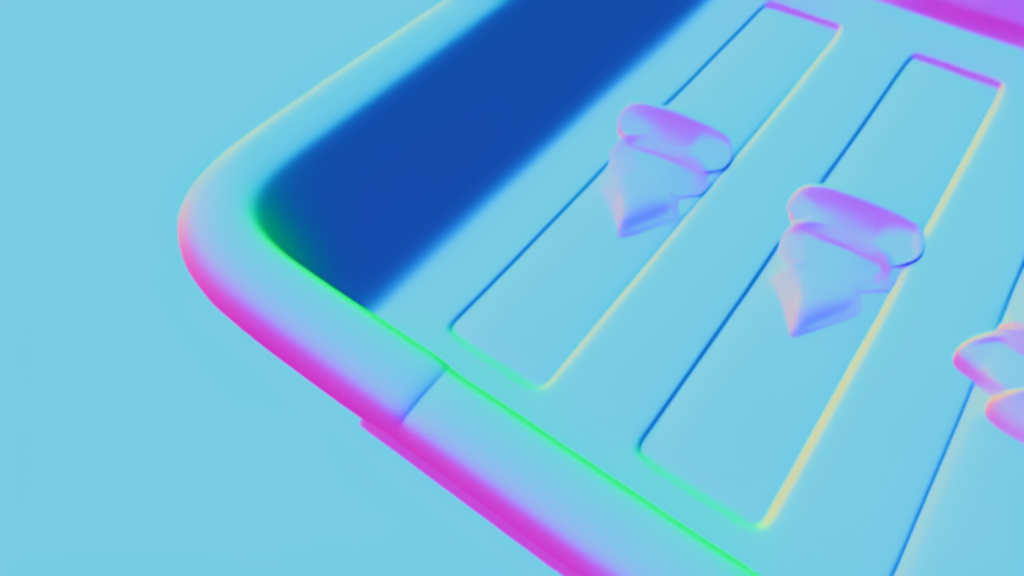} &
        \includegraphics[width=\imgwcg]{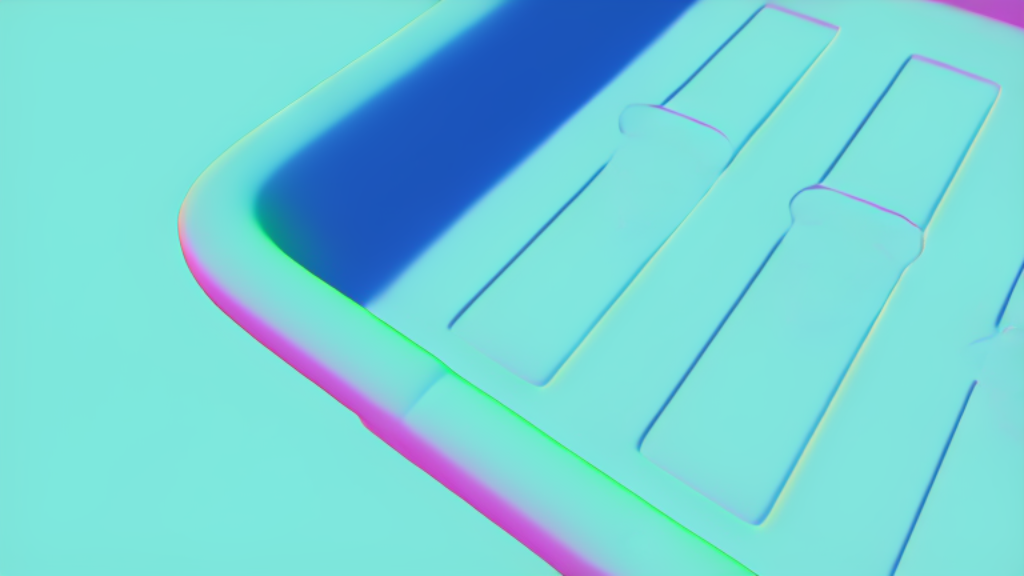} &
        \includegraphics[width=\imgwcg]{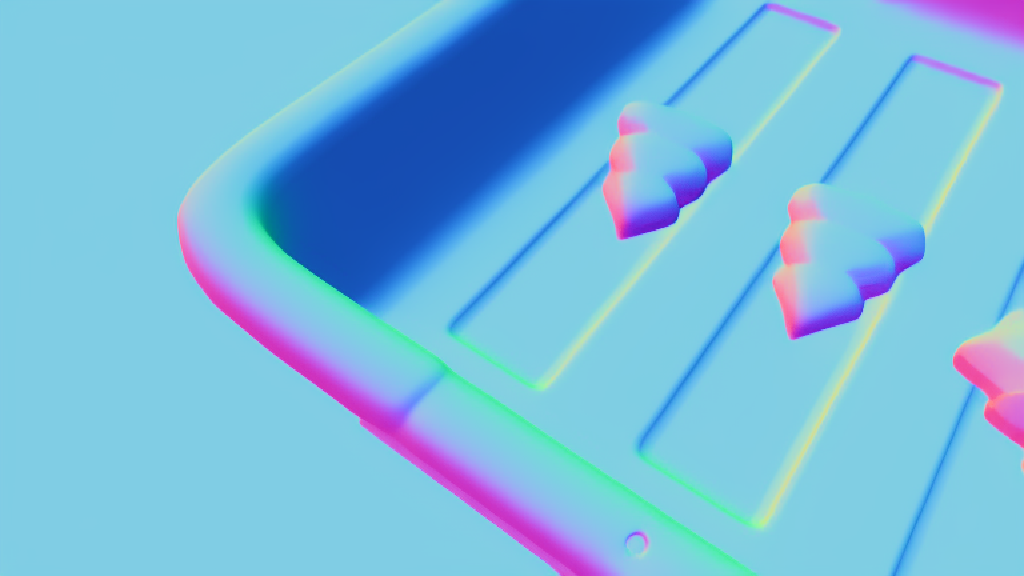} \\
        &
        \includegraphics[width=\imgwcg]{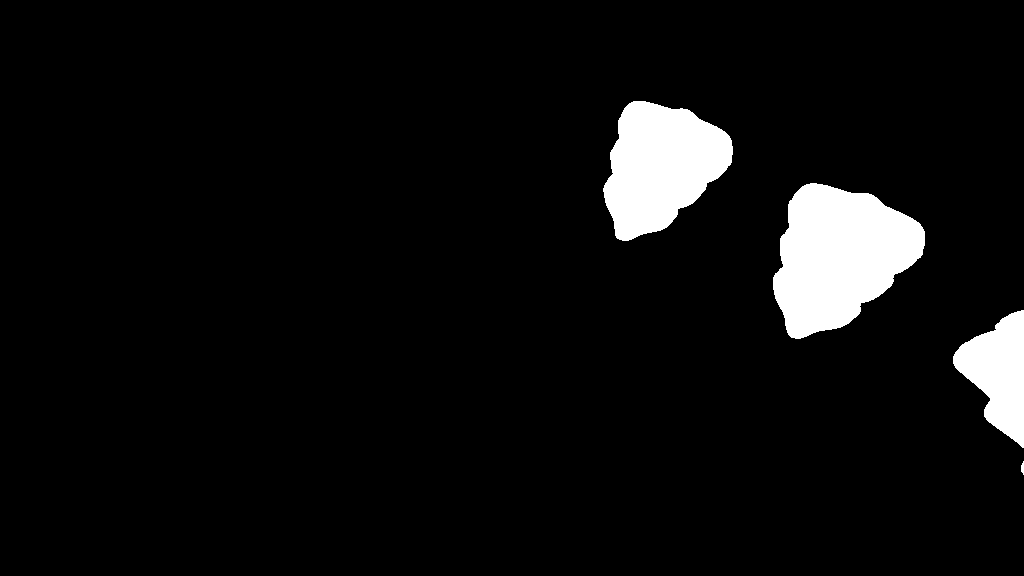} &
        \includegraphics[width=\imgwcg]{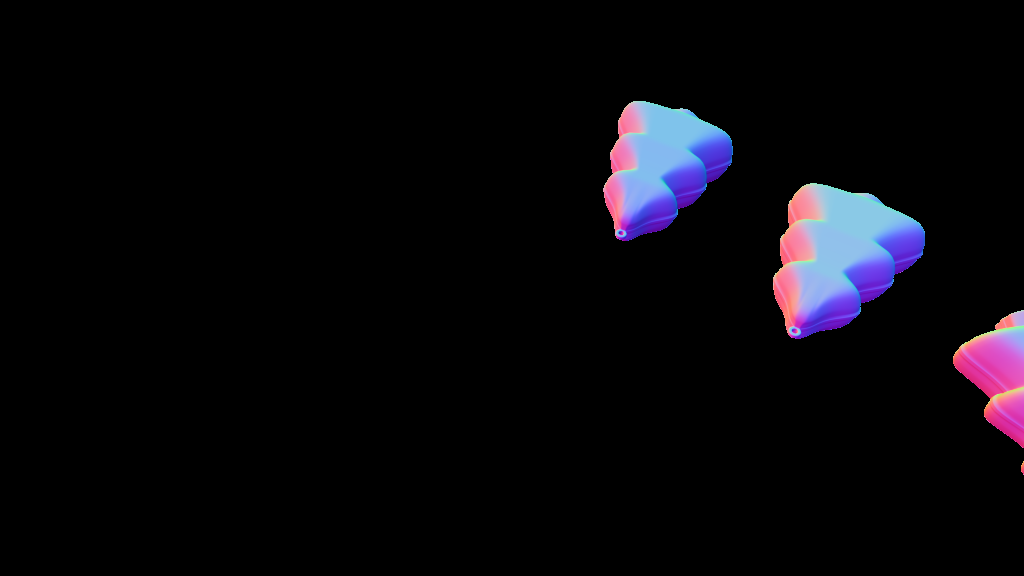} &
        \includegraphics[width=\imgwcg]{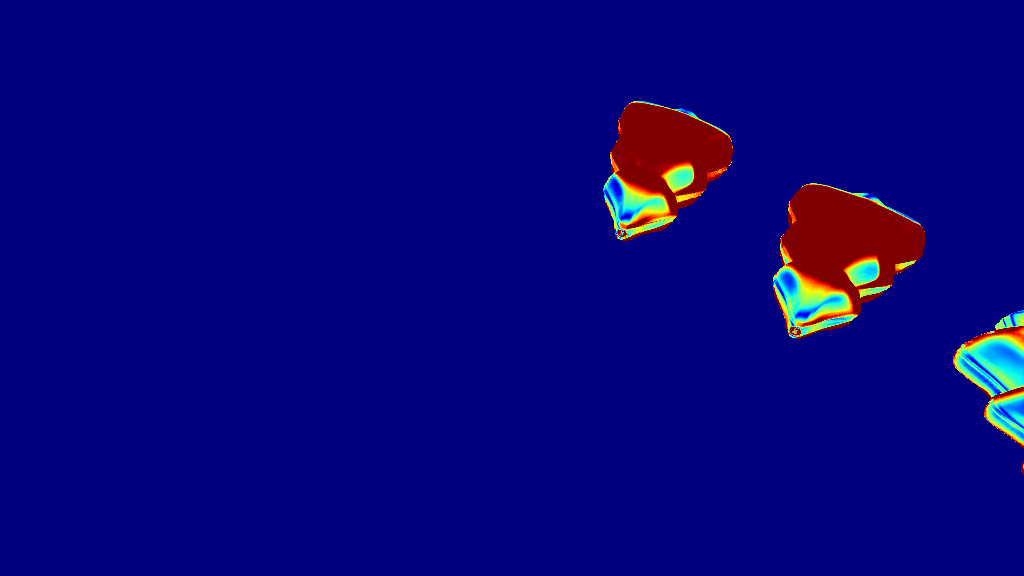} &
        \includegraphics[width=\imgwcg]{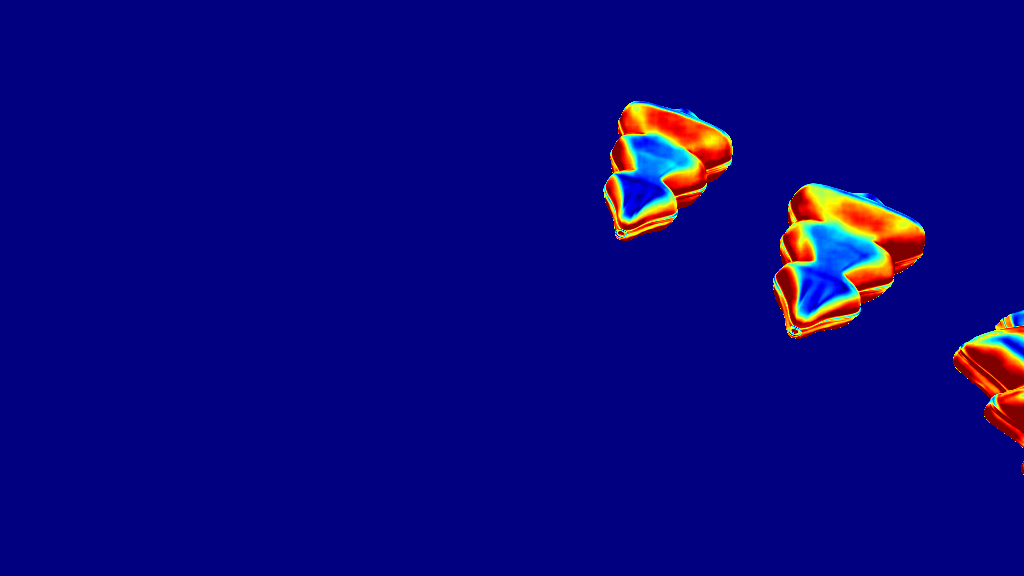} &
        \includegraphics[width=\imgwcg]{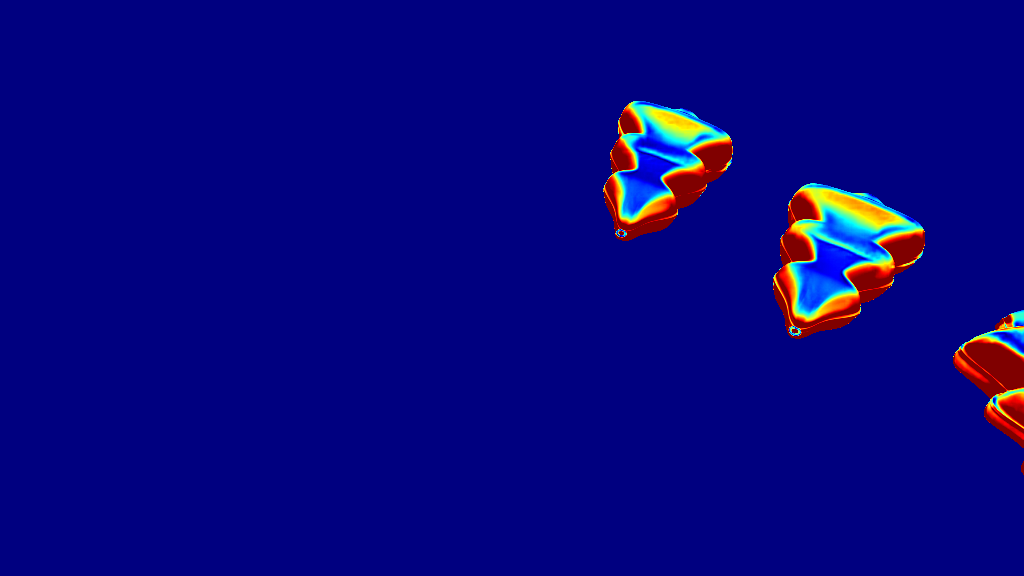} &
        \includegraphics[width=\imgwcg]{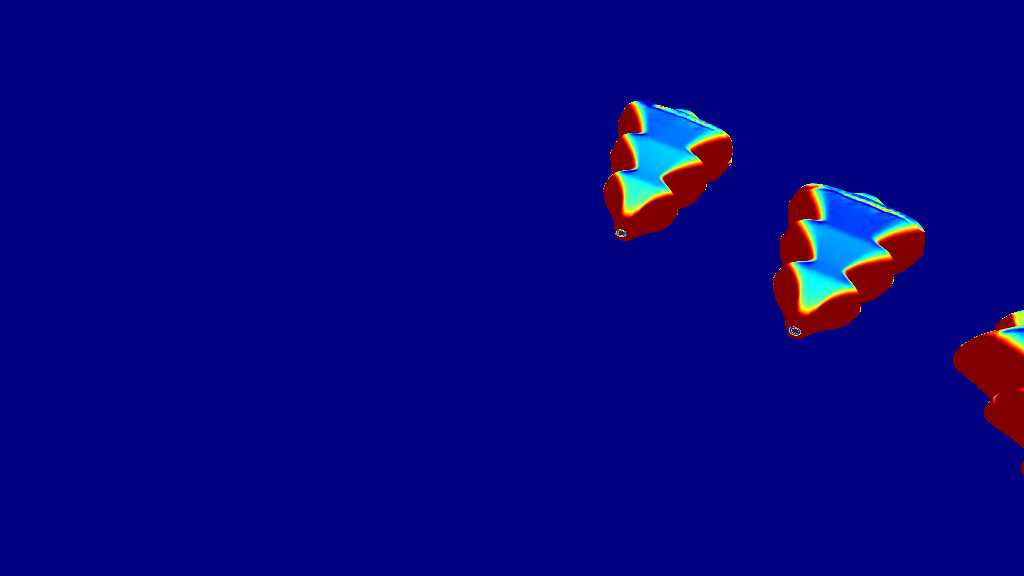} &
        \includegraphics[width=\imgwcg]{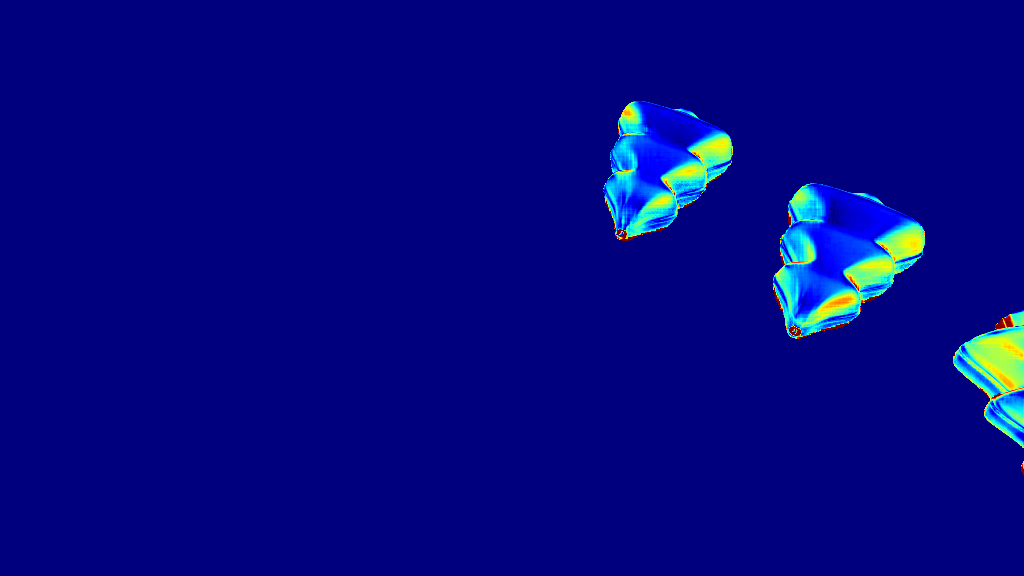} \\[6pt]
        %
        & \small Input/Mask & \small GT & \small DSINE & \small Marigold & \small StableNormal & \small GeoWizard & \small Diception \\[2pt]
        %
        \multirow{2}{*}[3.5ex]{\rotatebox{90}{\small ClearGrasp}} &
        \includegraphics[width=\imgwcg]{sources/datasets/tree-bath-bomb-test_000000102/input.png} &
        \includegraphics[width=\imgwcg]{sources/datasets/tree-bath-bomb-test_000000102/gt_normal.png} &
        \includegraphics[width=\imgwcg]{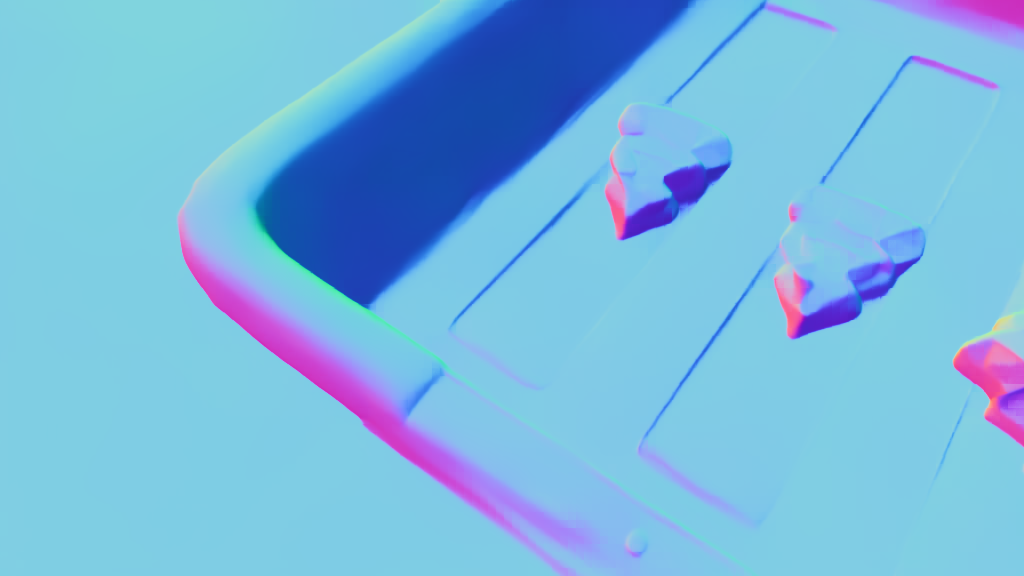} &
        \includegraphics[width=\imgwcg]{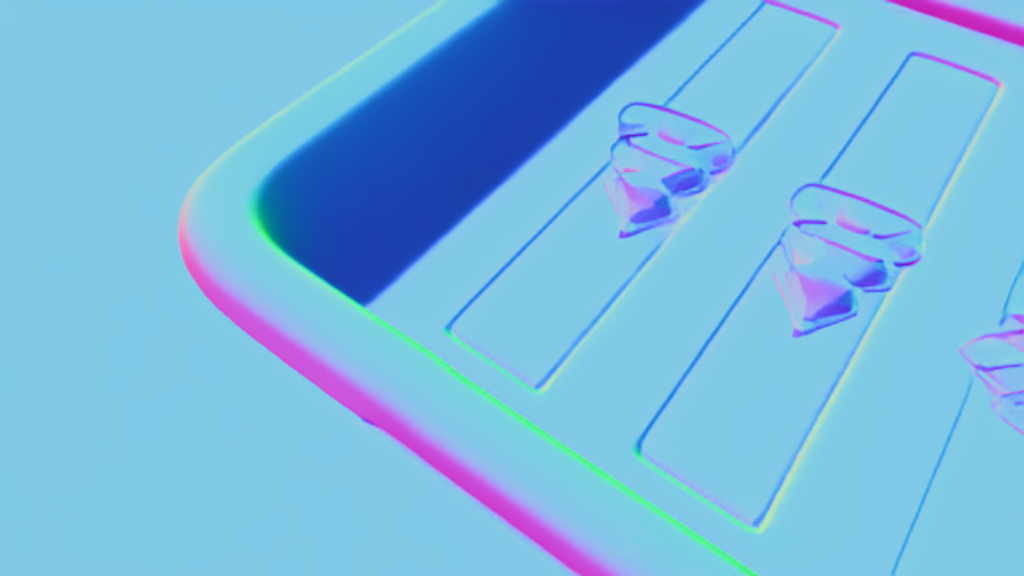} &
        \includegraphics[width=\imgwcg]{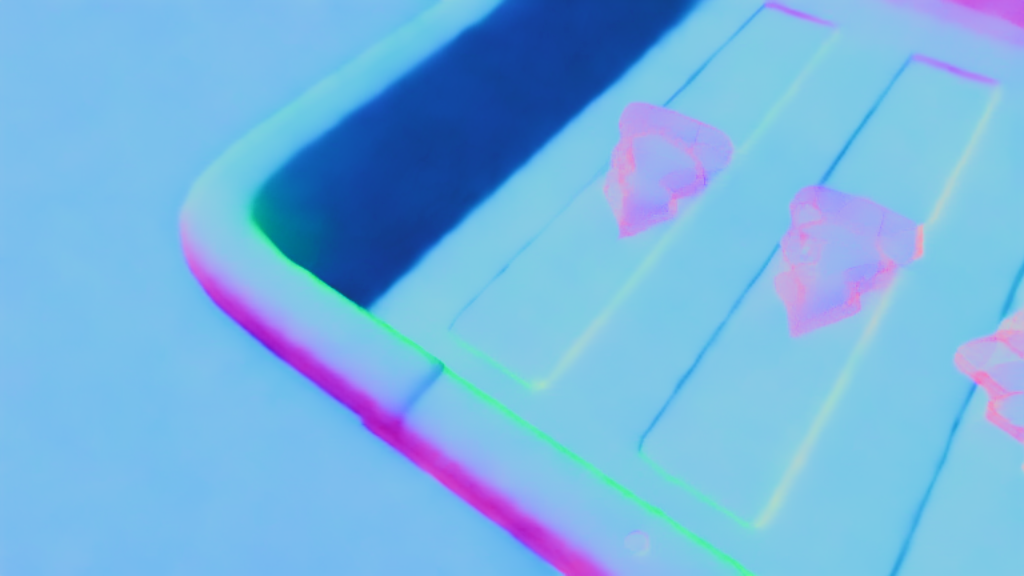} &
        \includegraphics[width=\imgwcg]{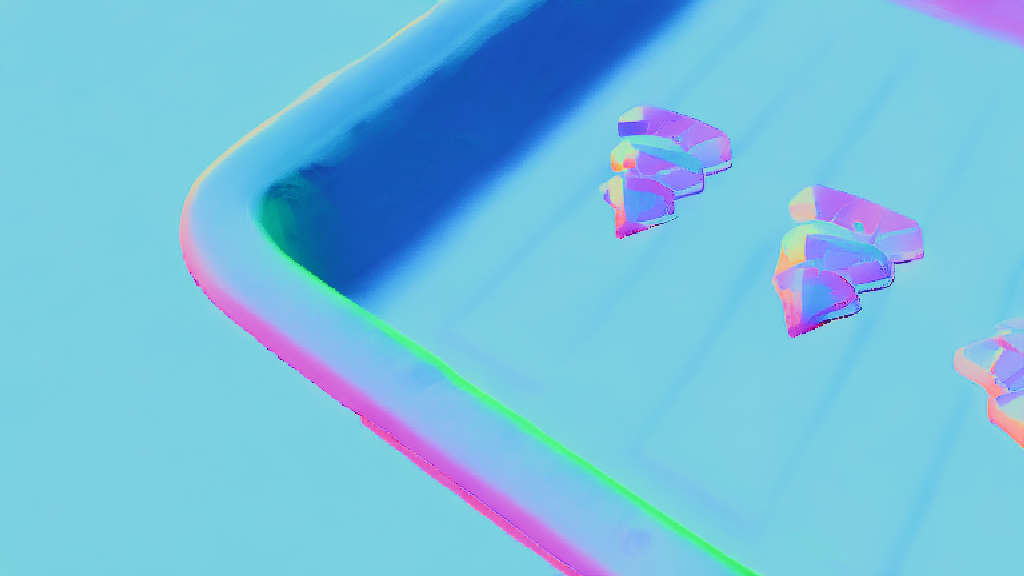} &
        \includegraphics[width=\imgwcg]{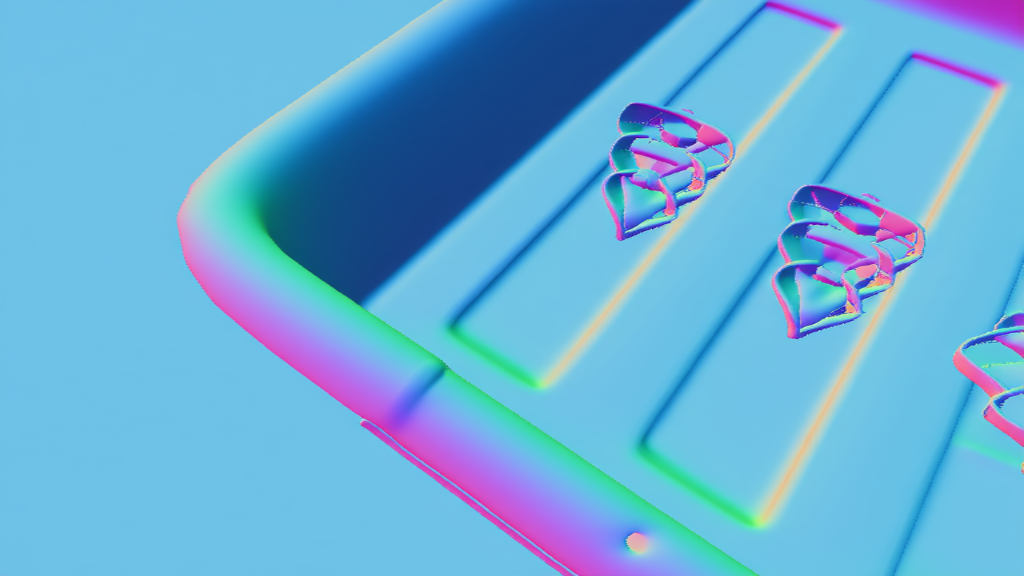} \\
        &
        \includegraphics[width=\imgwcg]{sources/datasets/tree-bath-bomb-test_000000102/mask.png} &
        \includegraphics[width=\imgwcg]{sources/datasets/tree-bath-bomb-test_000000102/Ours_gt_masked.png} &
        \includegraphics[width=\imgwcg]{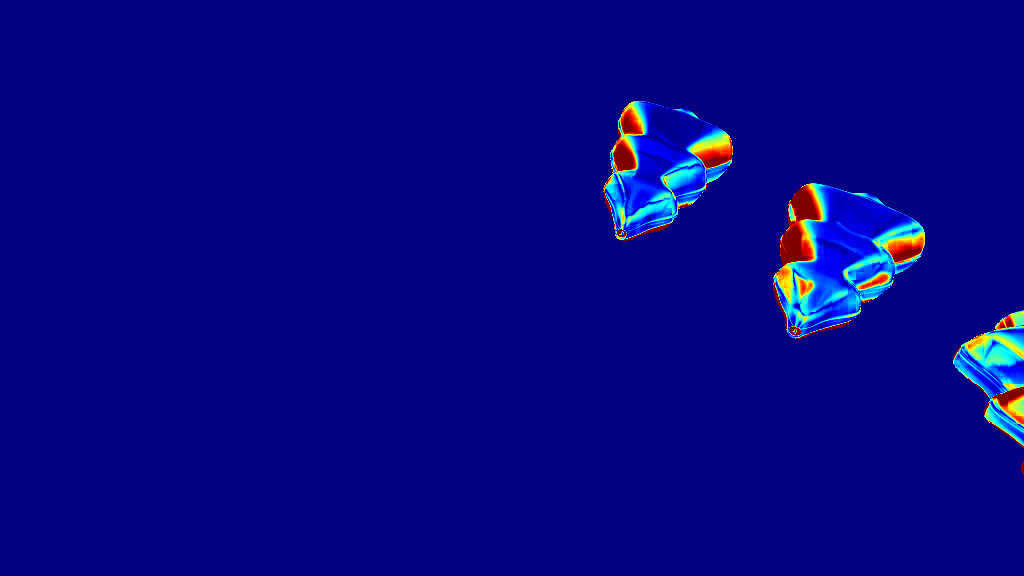} &
        \includegraphics[width=\imgwcg]{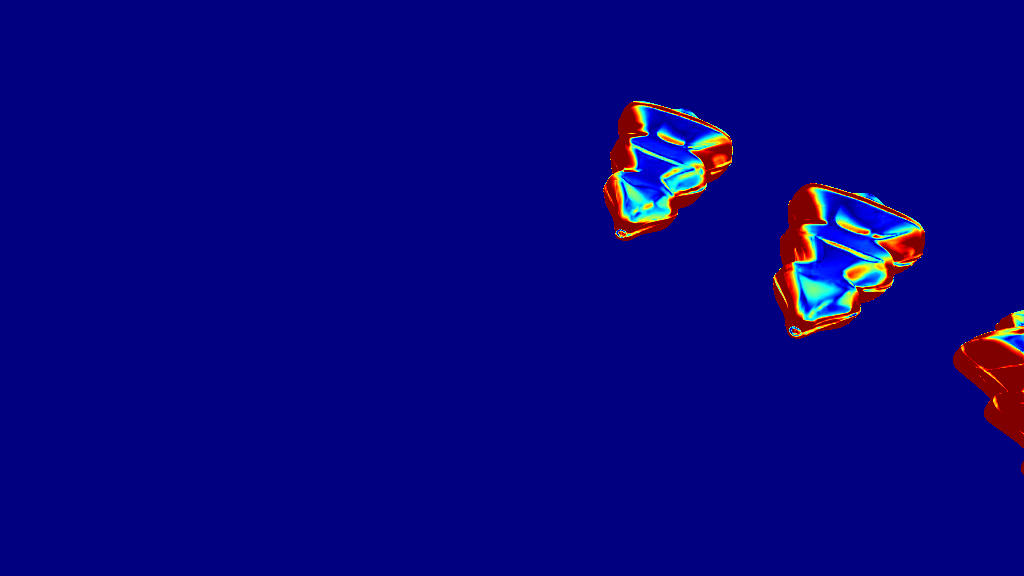} &
        \includegraphics[width=\imgwcg]{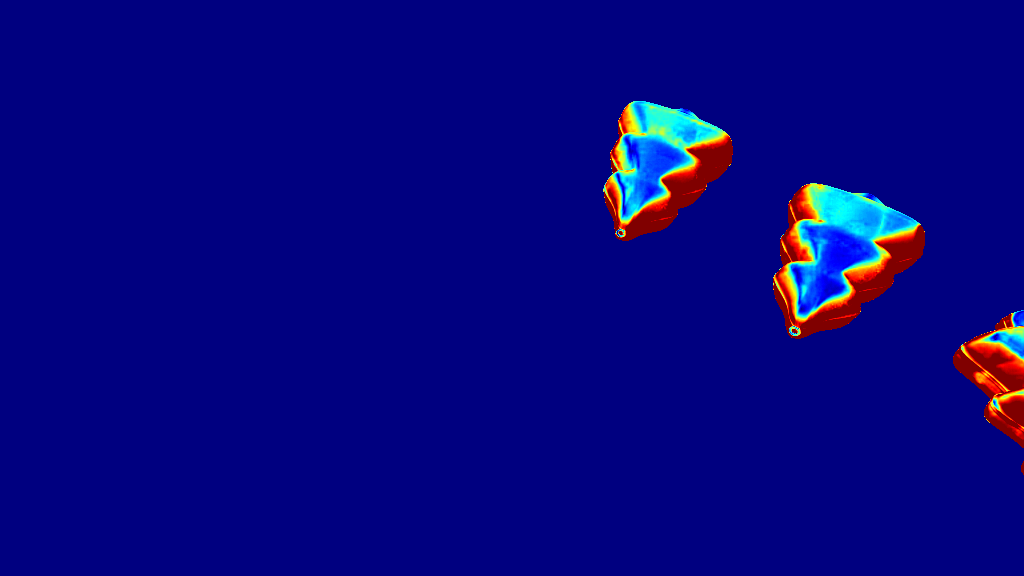} &
        \includegraphics[width=\imgwcg]{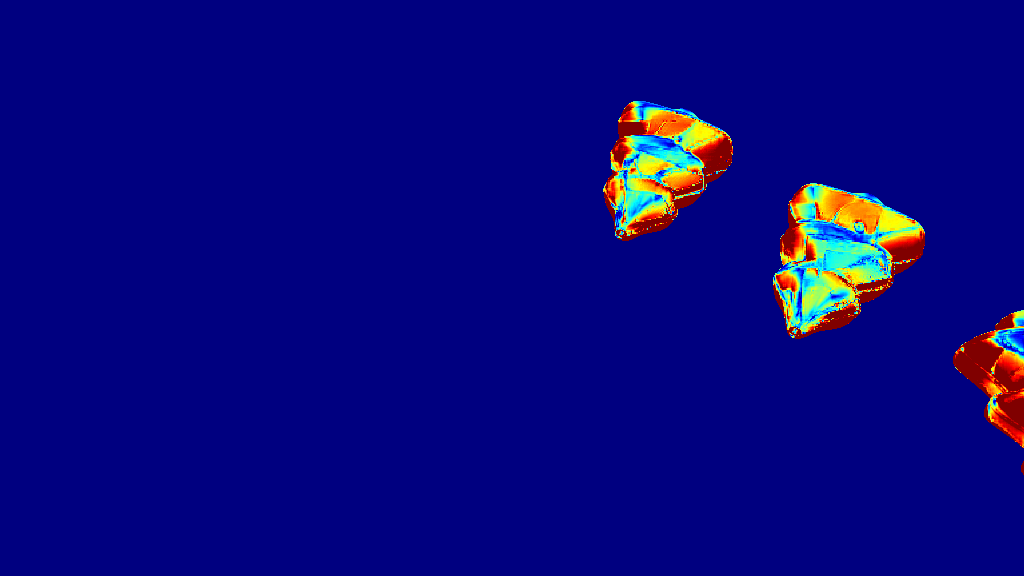} &
        \includegraphics[width=\imgwcg]{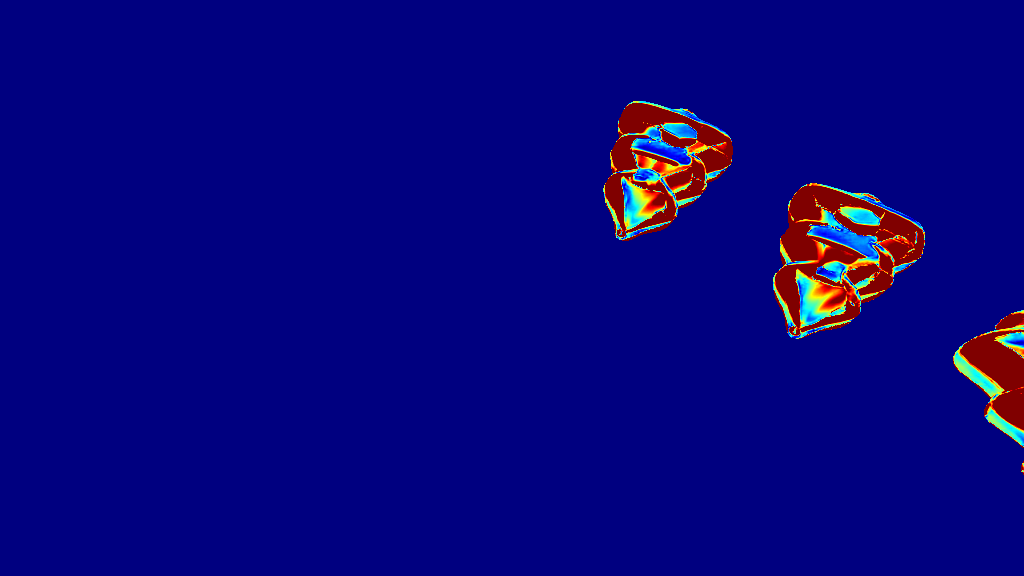} \\[12pt]
        %
        & \small Input/Mask & \small GT & \small Lotus & \small MoGe-2 & \small E2E-FT & \small GenPercept & \small \textbf{Ours} \\[2pt]
        %
        \multirow{2}{*}[3.5ex]{\rotatebox{90}{\small ClearGrasp}} &
        \includegraphics[width=\imgwcg]{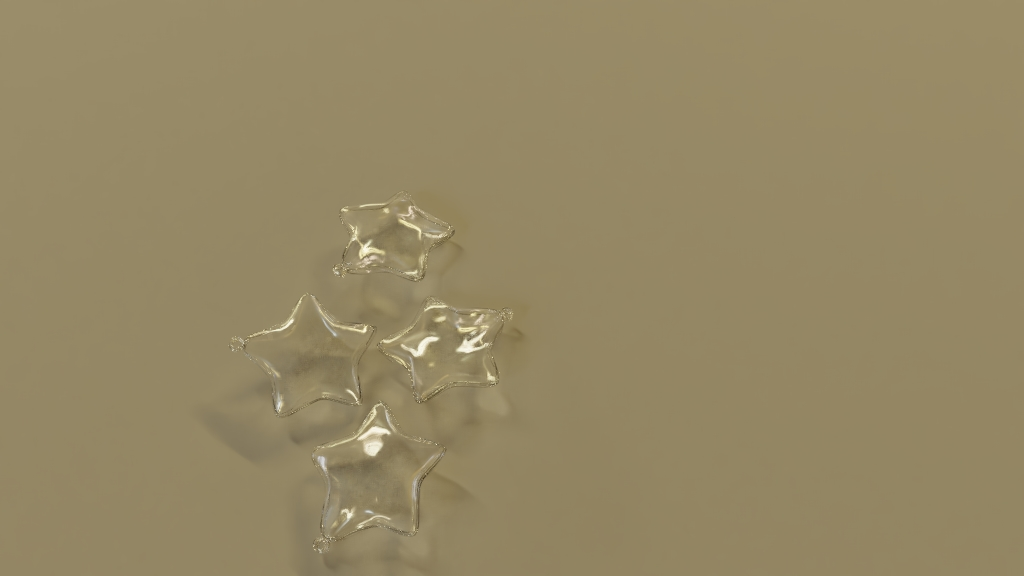} &
        \includegraphics[width=\imgwcg]{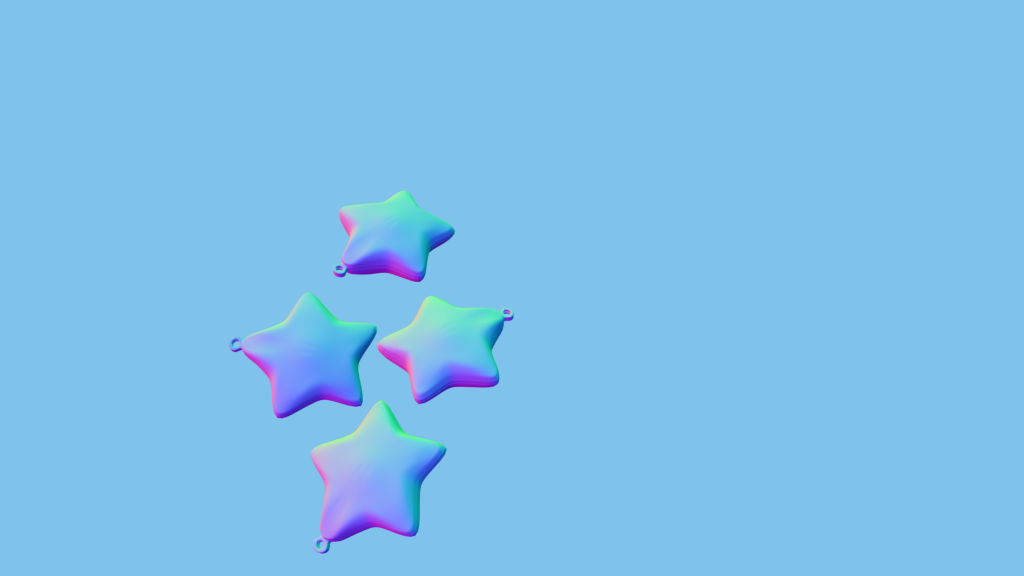} &
        \includegraphics[width=\imgwcg]{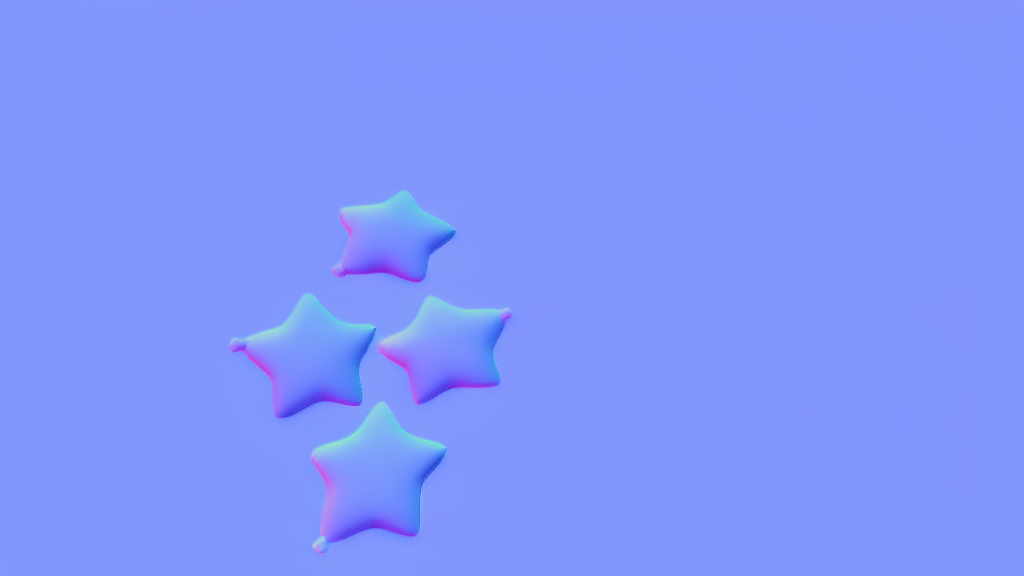} &
        \includegraphics[width=\imgwcg]{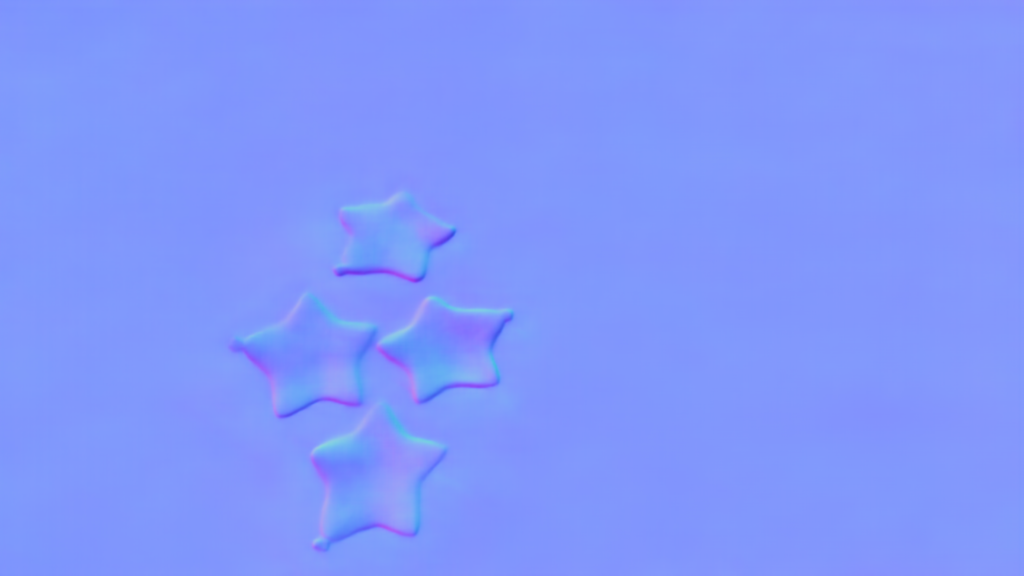} &
        \includegraphics[width=\imgwcg]{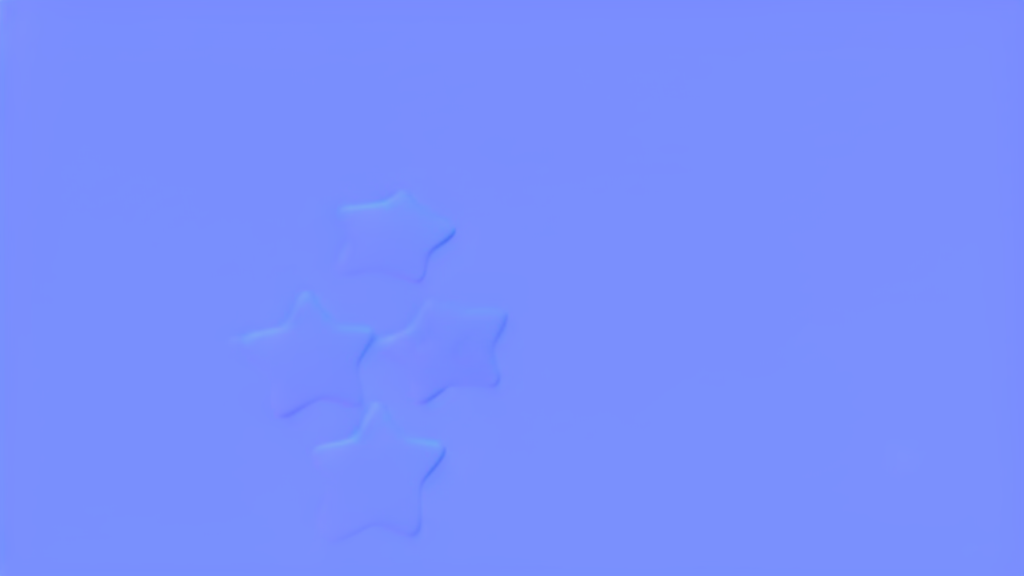} &
        \includegraphics[width=\imgwcg]{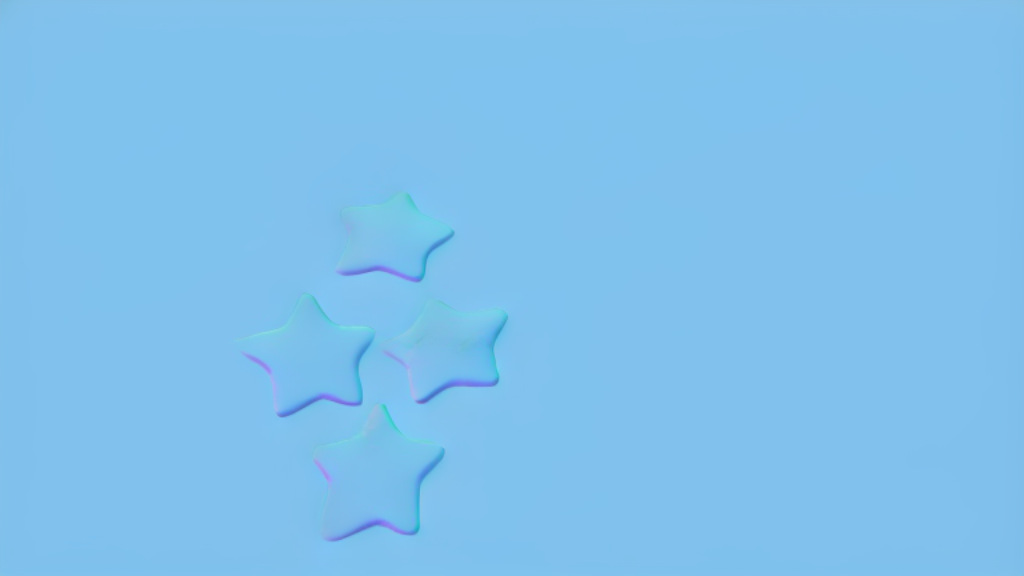} &
        \includegraphics[width=\imgwcg]{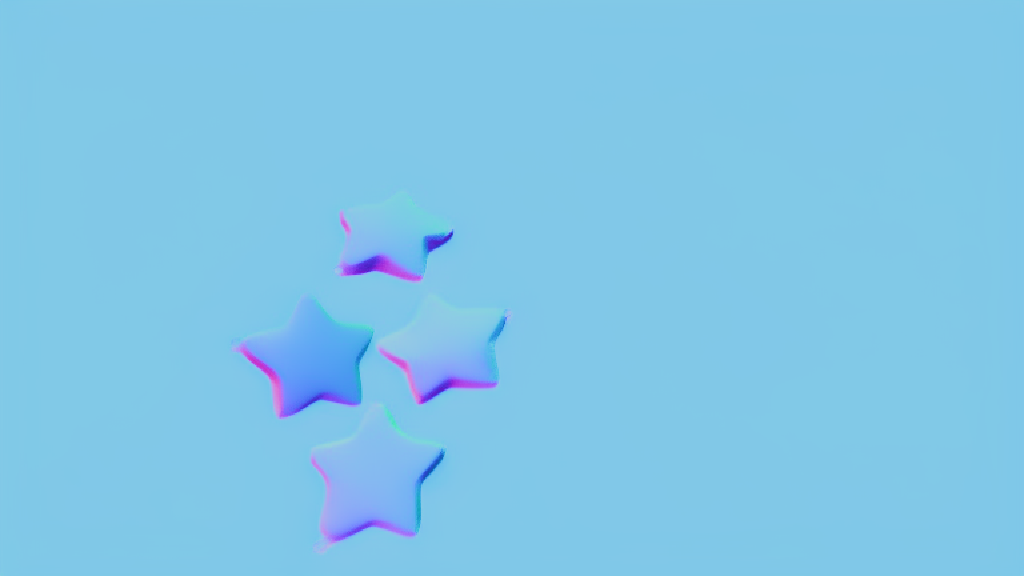} \\
        &
        \includegraphics[width=\imgwcg]{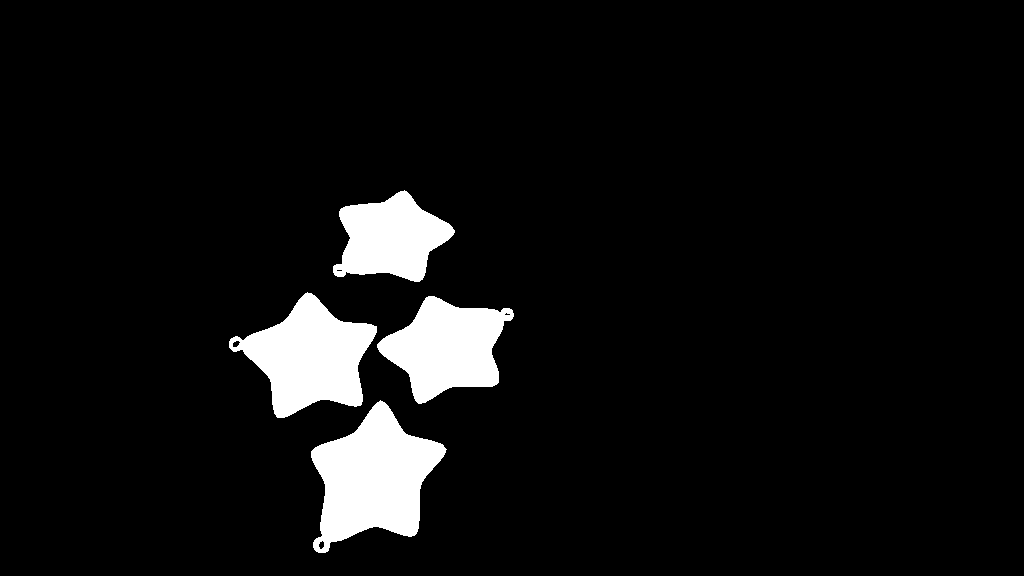} &
        \includegraphics[width=\imgwcg]{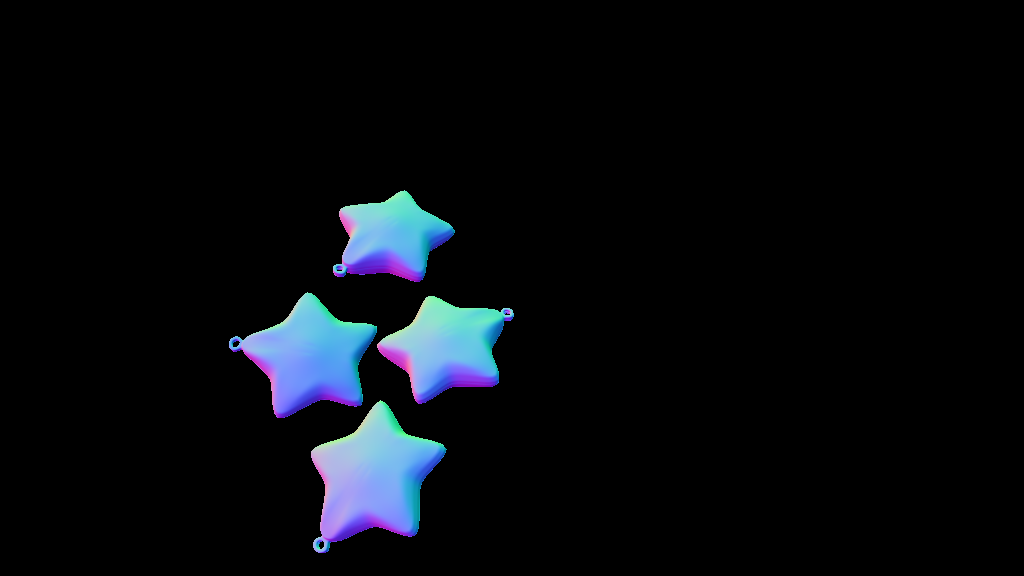} &
        \includegraphics[width=\imgwcg]{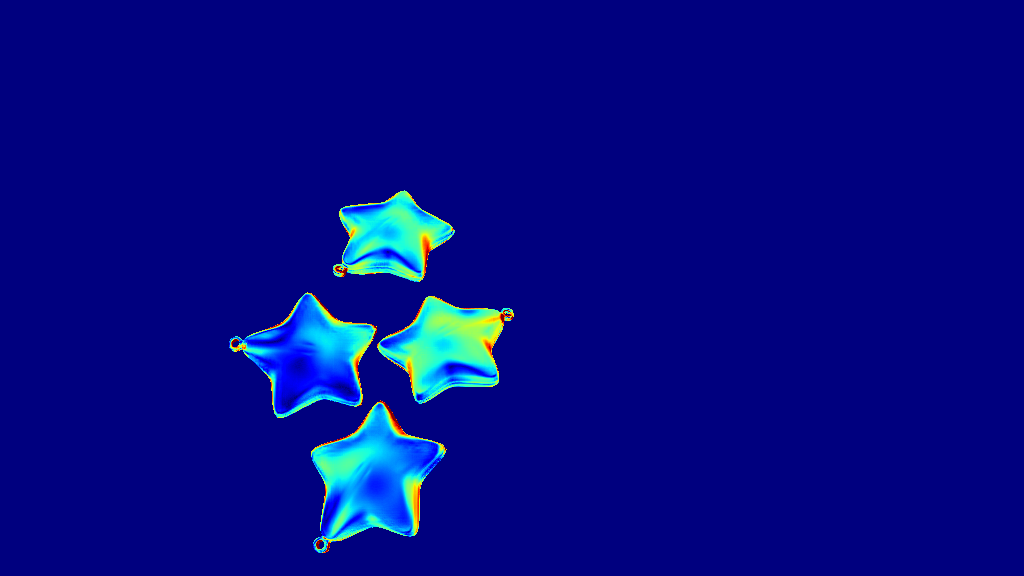} &
        \includegraphics[width=\imgwcg]{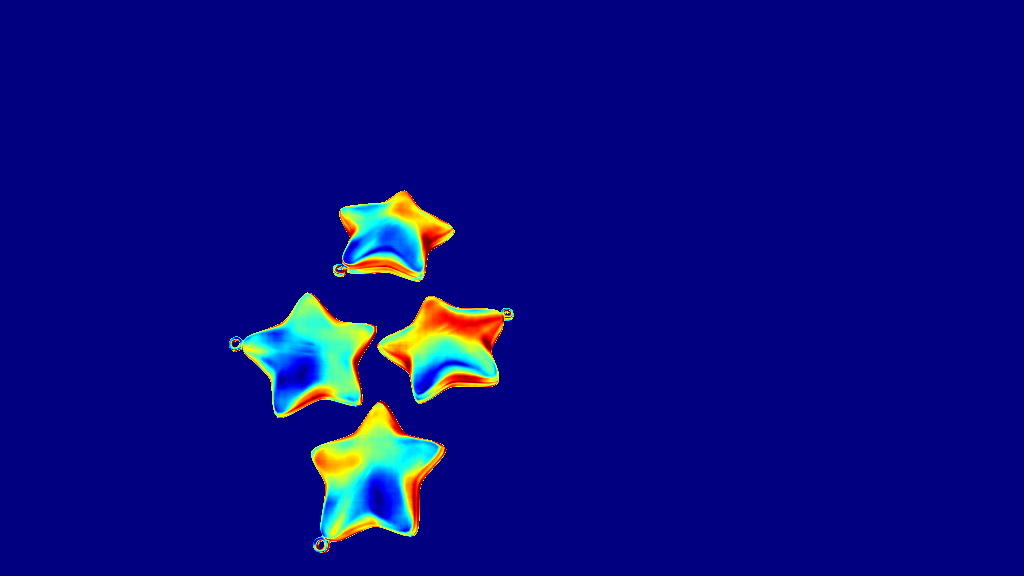} &
        \includegraphics[width=\imgwcg]{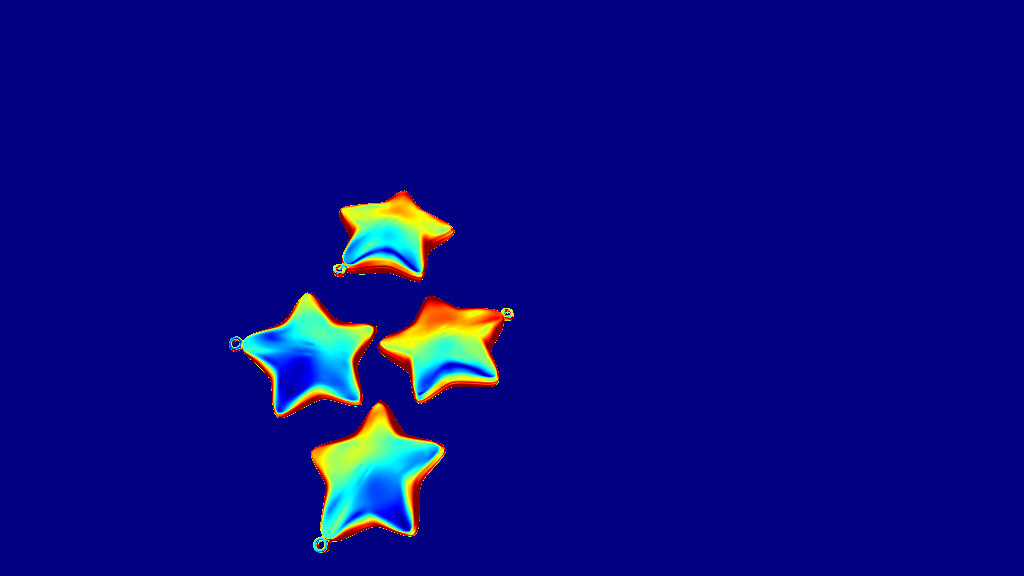} &
        \includegraphics[width=\imgwcg]{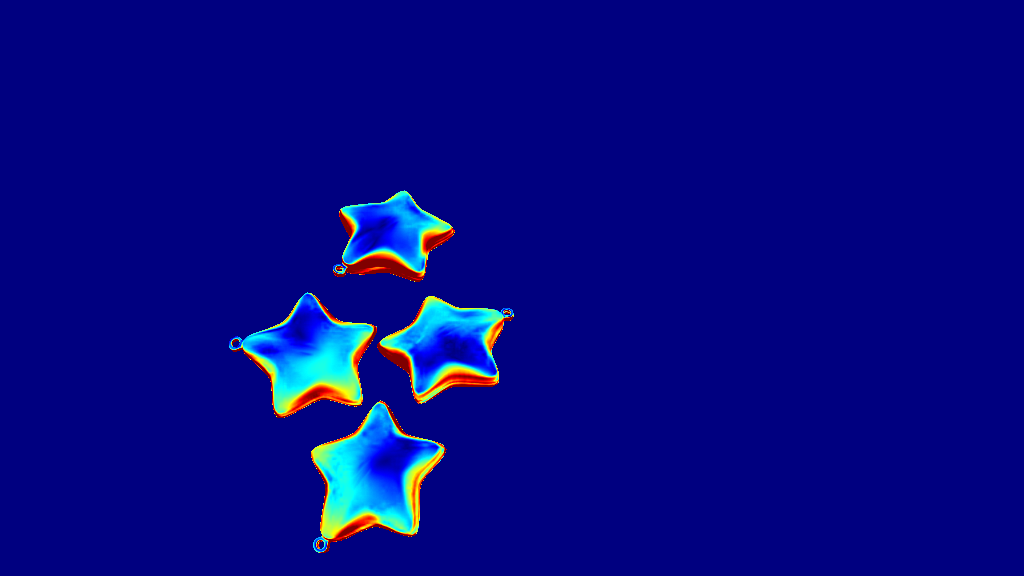} &
        \includegraphics[width=\imgwcg]{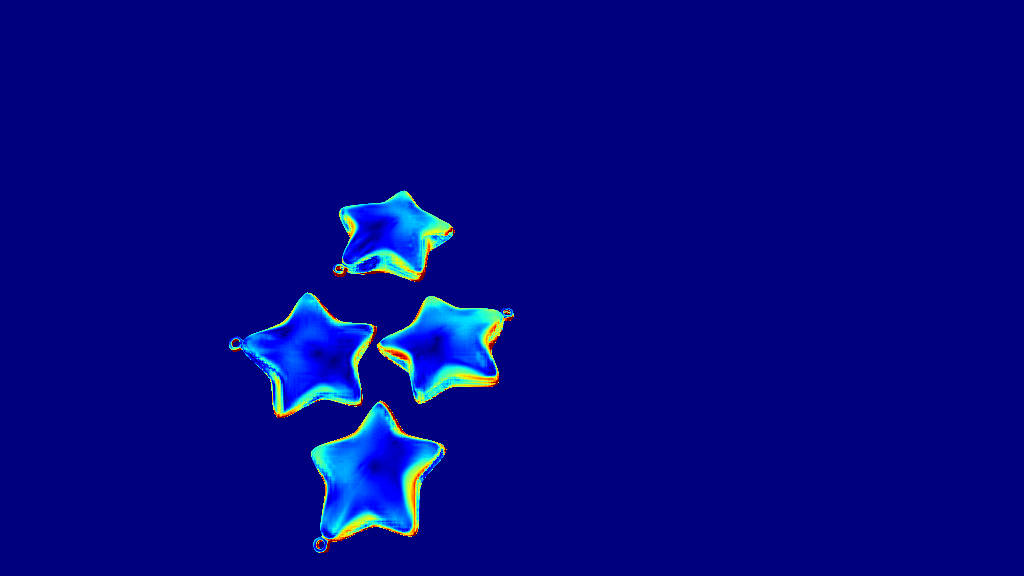} \\[6pt]
        %
        & \small Input/Mask & \small GT & \small DSINE & \small Marigold & \small StableNormal & \small GeoWizard & \small Diception \\[2pt]
        %
        \multirow{2}{*}[3.5ex]{\rotatebox{90}{\small ClearGrasp}} &
        \includegraphics[width=\imgwcg]{sources/datasets/star-bath-bomb-test_000000096/input.png} &
        \includegraphics[width=\imgwcg]{sources/datasets/star-bath-bomb-test_000000096/gt_normal.png} &
        \includegraphics[width=\imgwcg]{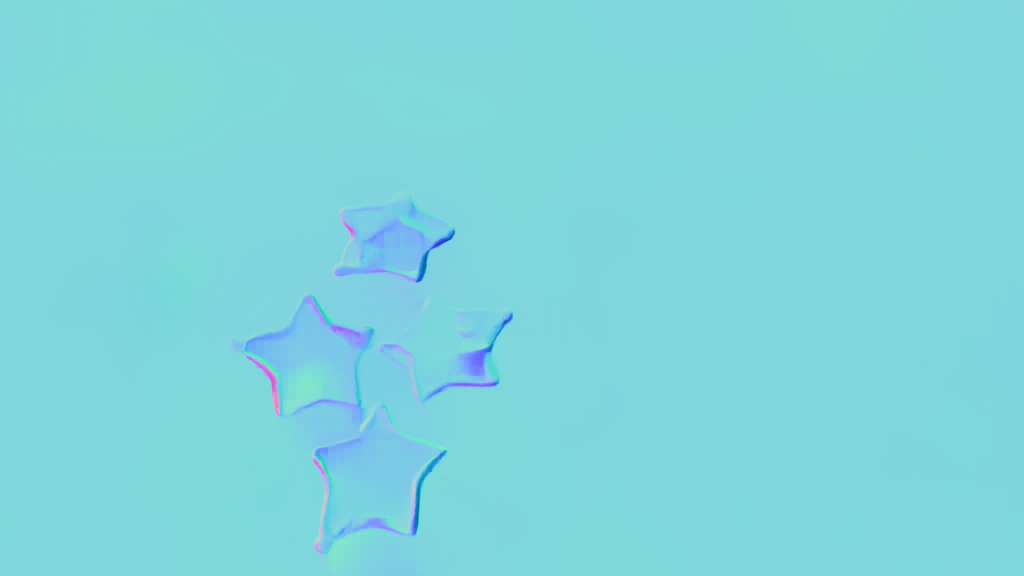} &
        \includegraphics[width=\imgwcg]{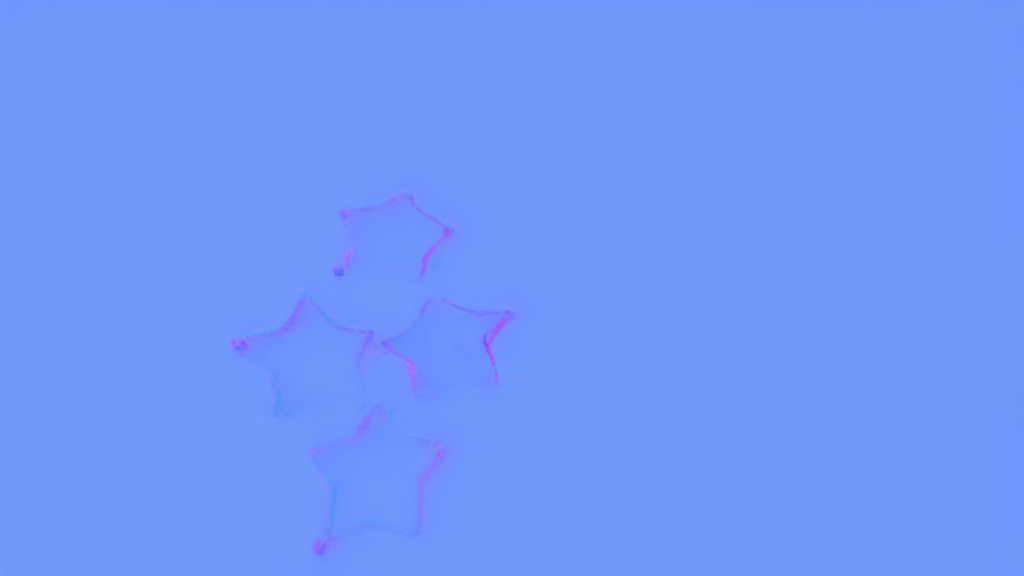} &
        \includegraphics[width=\imgwcg]{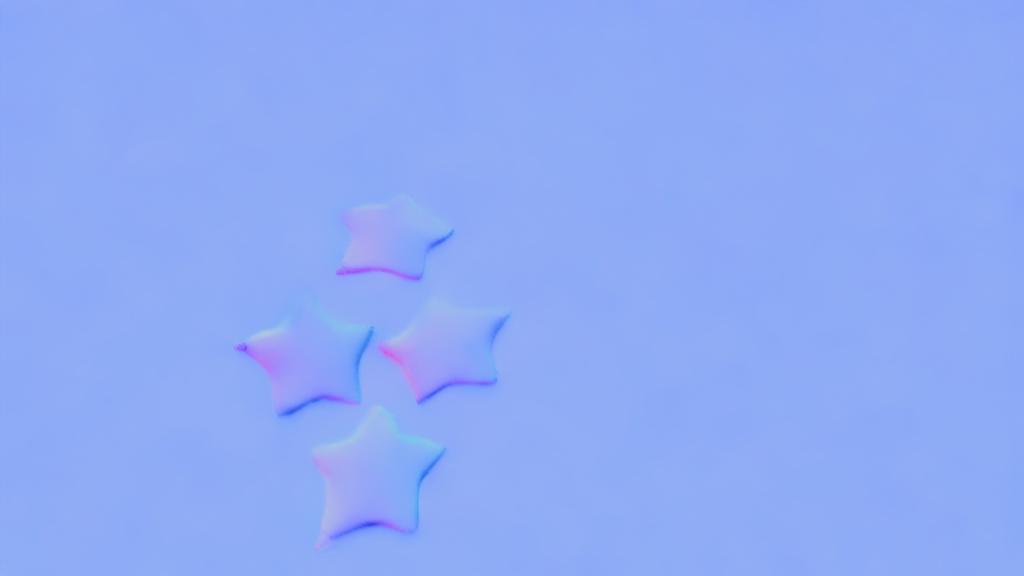} &
        \includegraphics[width=\imgwcg]{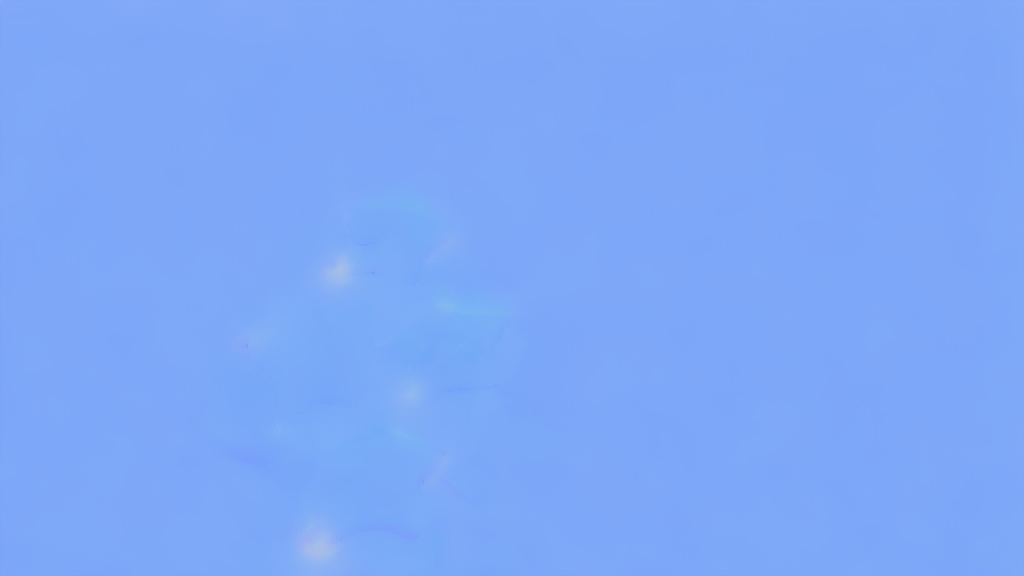} &
        \includegraphics[width=\imgwcg]{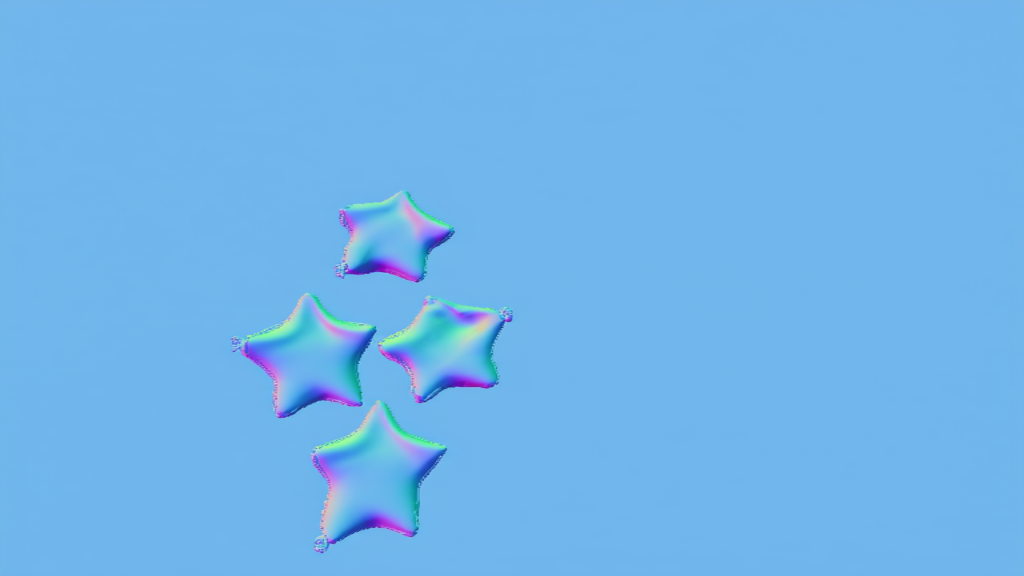} \\
        &
        \includegraphics[width=\imgwcg]{sources/datasets/star-bath-bomb-test_000000096/mask.png} &
        \includegraphics[width=\imgwcg]{sources/datasets/star-bath-bomb-test_000000096/Ours_gt_masked.png} &
        \includegraphics[width=\imgwcg]{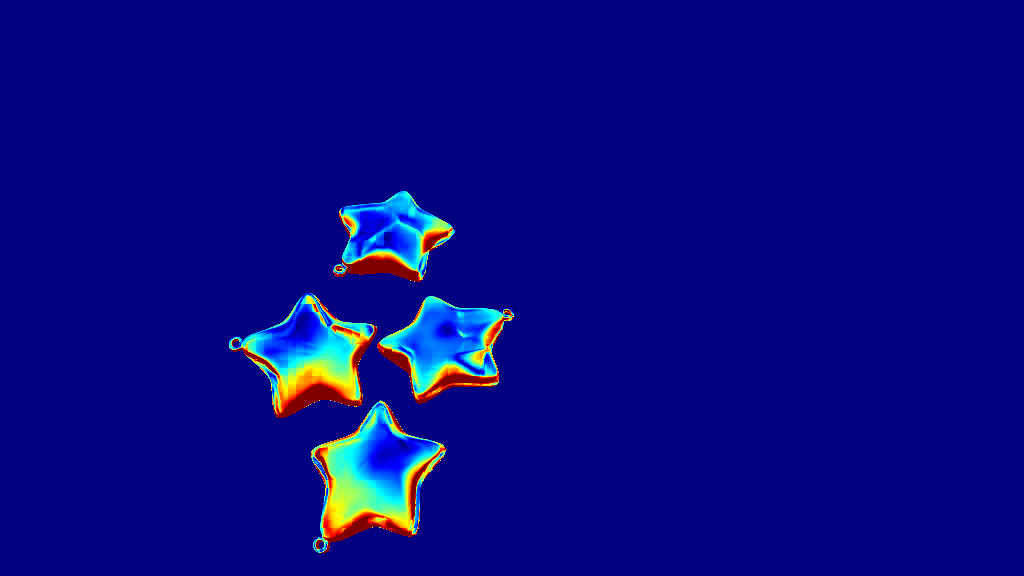} &
        \includegraphics[width=\imgwcg]{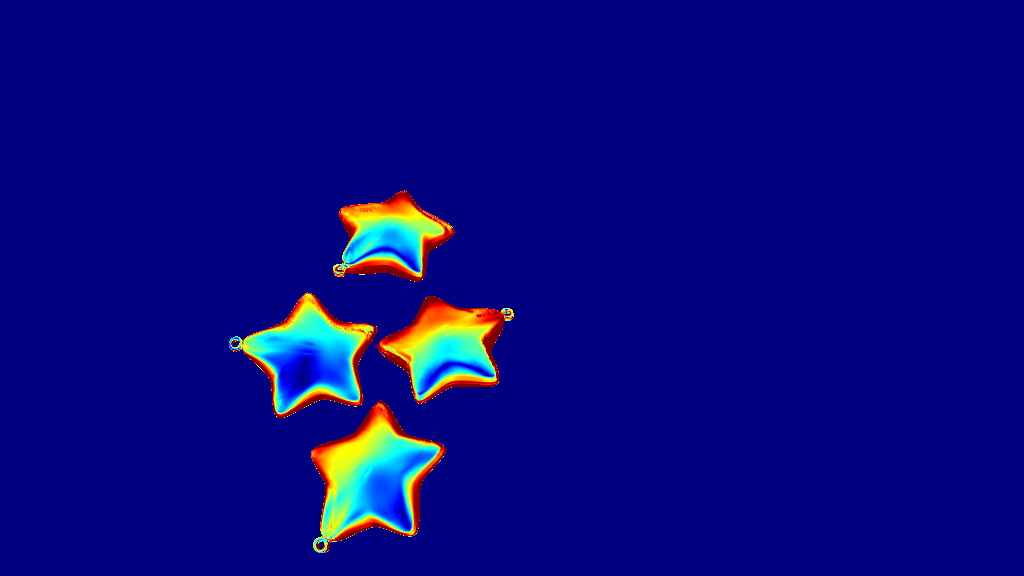} &
        \includegraphics[width=\imgwcg]{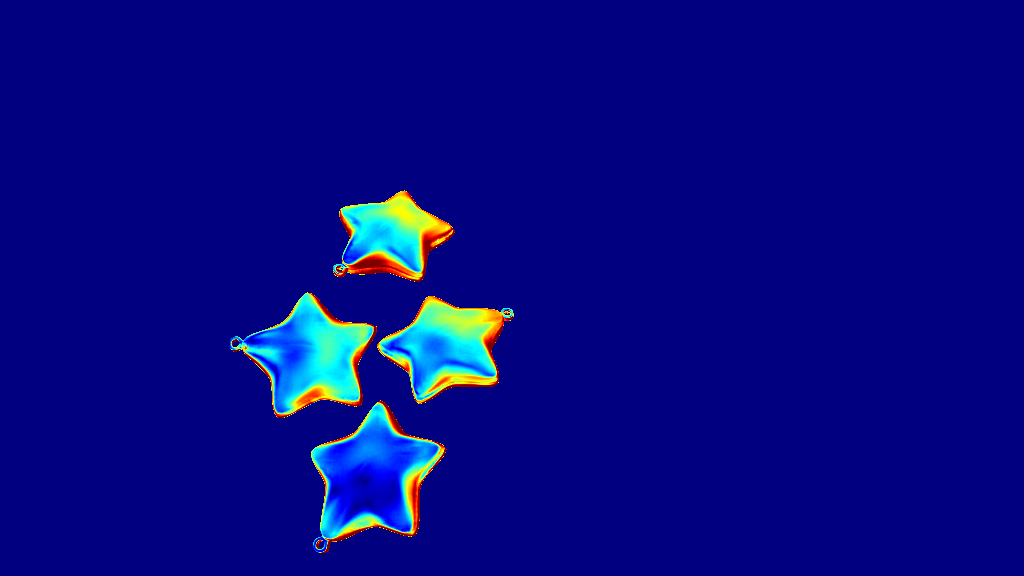} &
        \includegraphics[width=\imgwcg]{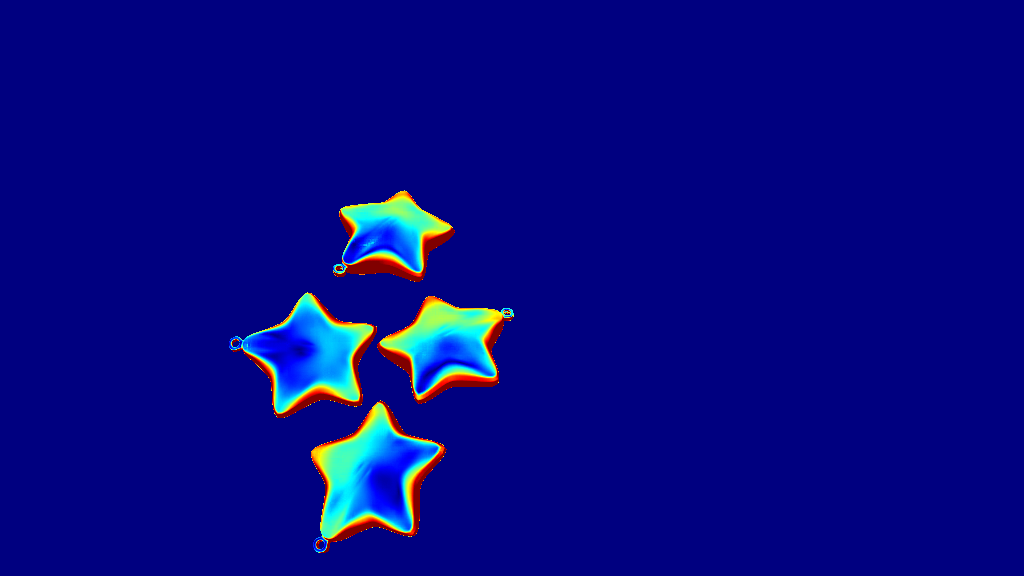} &
        \includegraphics[width=\imgwcg]{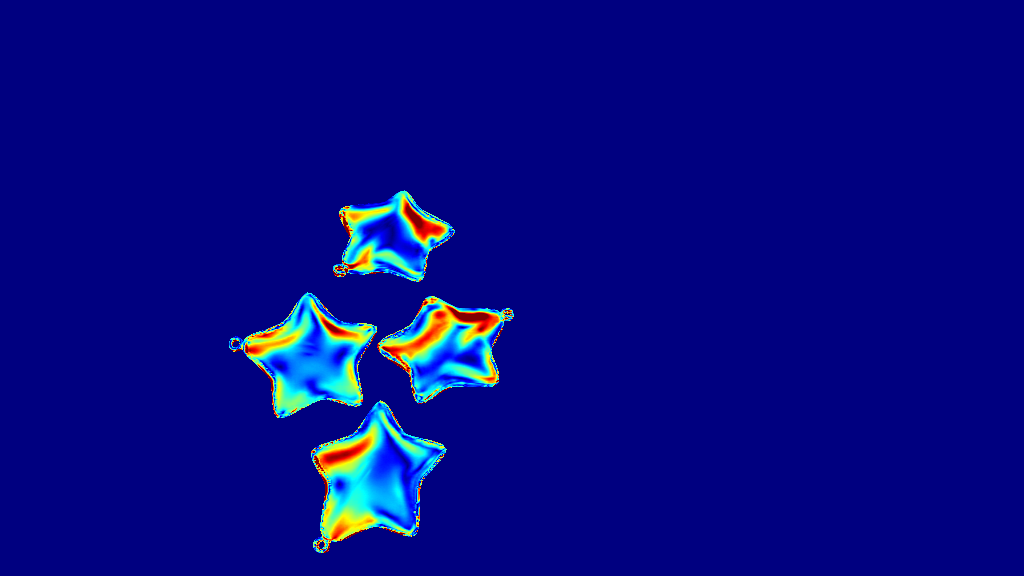} \\
    \end{tabular}
    \vspace{-1mm}
    \caption{\textbf{Extended qualitative comparison with baseline methods.}
    We compare against 9 baselines.
    Top rows show predicted normals; bottom rows show angular error maps (blue: low, red: high).
    Our method produces sharper edges and lower error on transparent regions. Please zoom in \faSearch~for details. (\S~\ref{ssec:extended_comparison})}
    \vspace{-3mm}
    \label{fig:comparison_cleargrasp}
\end{figure*}

\paragraph{Analysis.}\label{par:extended_comparison_analysis}
\Figref{fig:comparison_extended} and \Figref{fig:comparison_cleargrasp} present additional comparisons with 9 baseline methods on TransNormal-Synthetic and ClearGrasp. Across all examples, we observe consistent trends: \ding{172} baseline methods tend to over-smooth edges due to the lack of semantic guidance for distinguishing object boundaries from refracted backgrounds; \ding{173} methods without wavelet regularization produce blurred predictions on interior surfaces; \ding{174} TransNormal maintains sharp edge reconstruction while preserving smooth interior surfaces, validating our design choices.

\subsection{DINOv3 Semantic Feature Visualization}
\label{sec:cross_attention_analysis}

\begin{figure}[h!]
\centering

\begin{tabular}{@{}c@{\hspace{0.008\columnwidth}}c@{\hspace{0.008\columnwidth}}c@{\hspace{0.008\columnwidth}}c@{}}
\includegraphics[width=0.24\columnwidth]{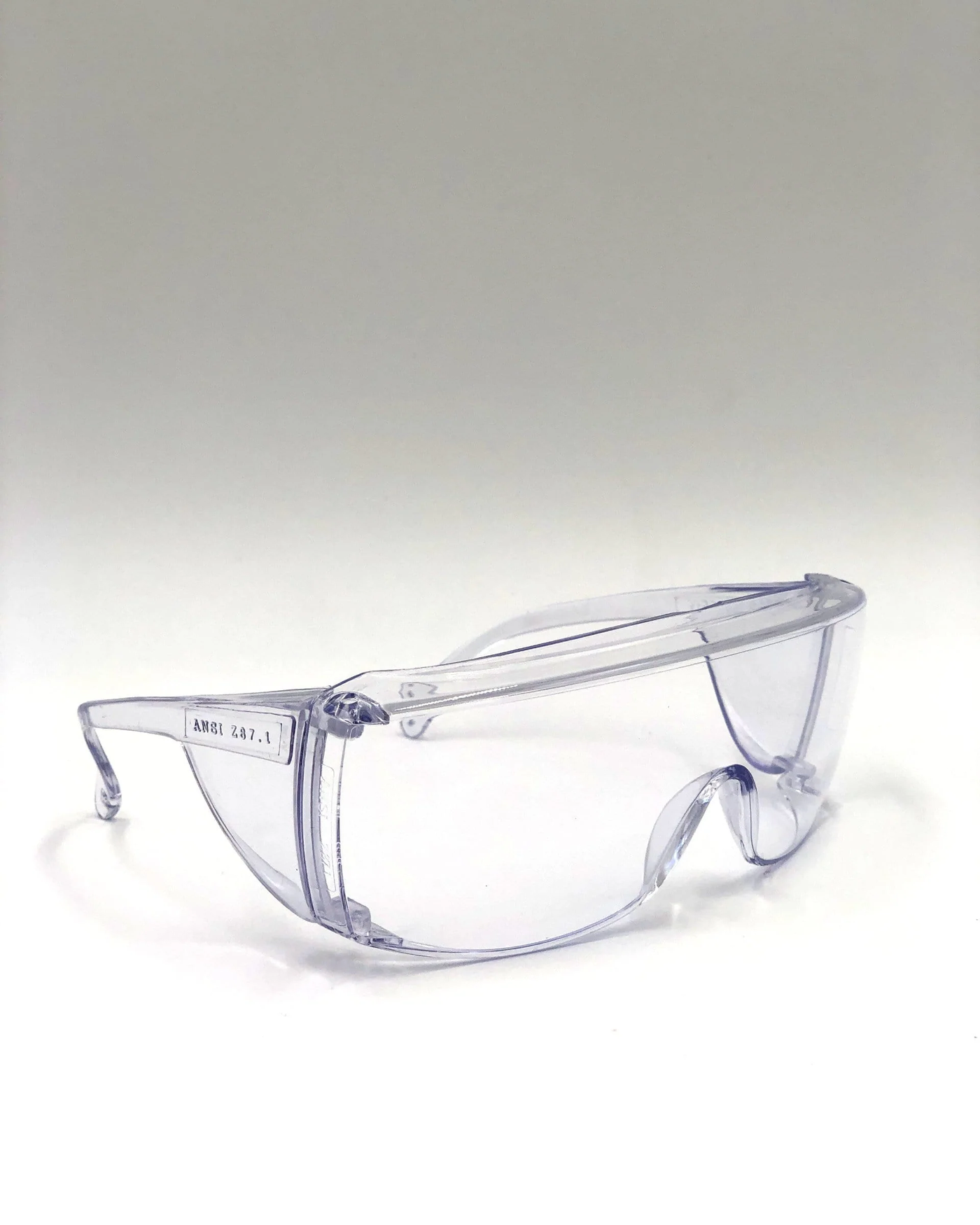} &
\includegraphics[width=0.24\columnwidth]{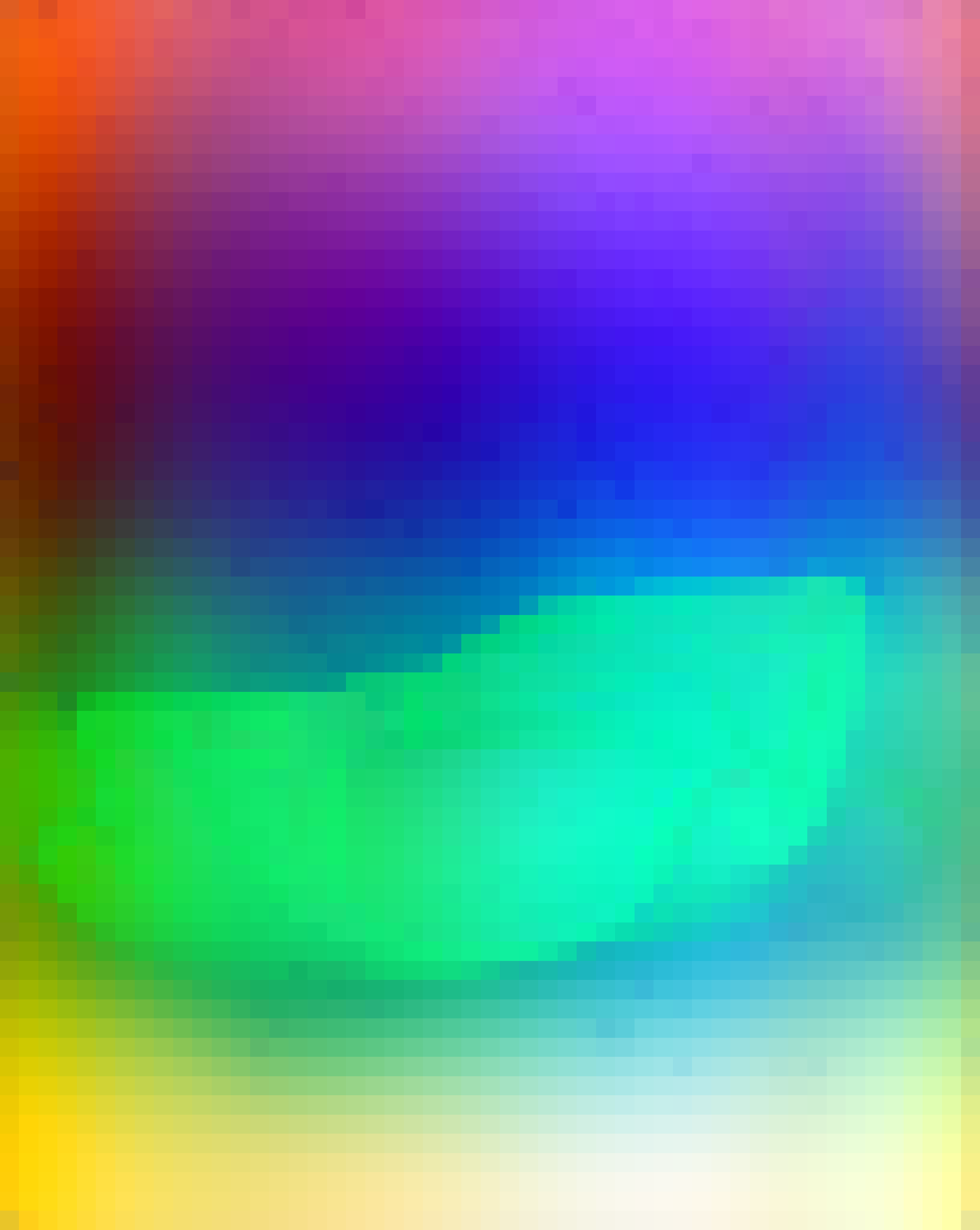} &
\includegraphics[width=0.24\columnwidth]{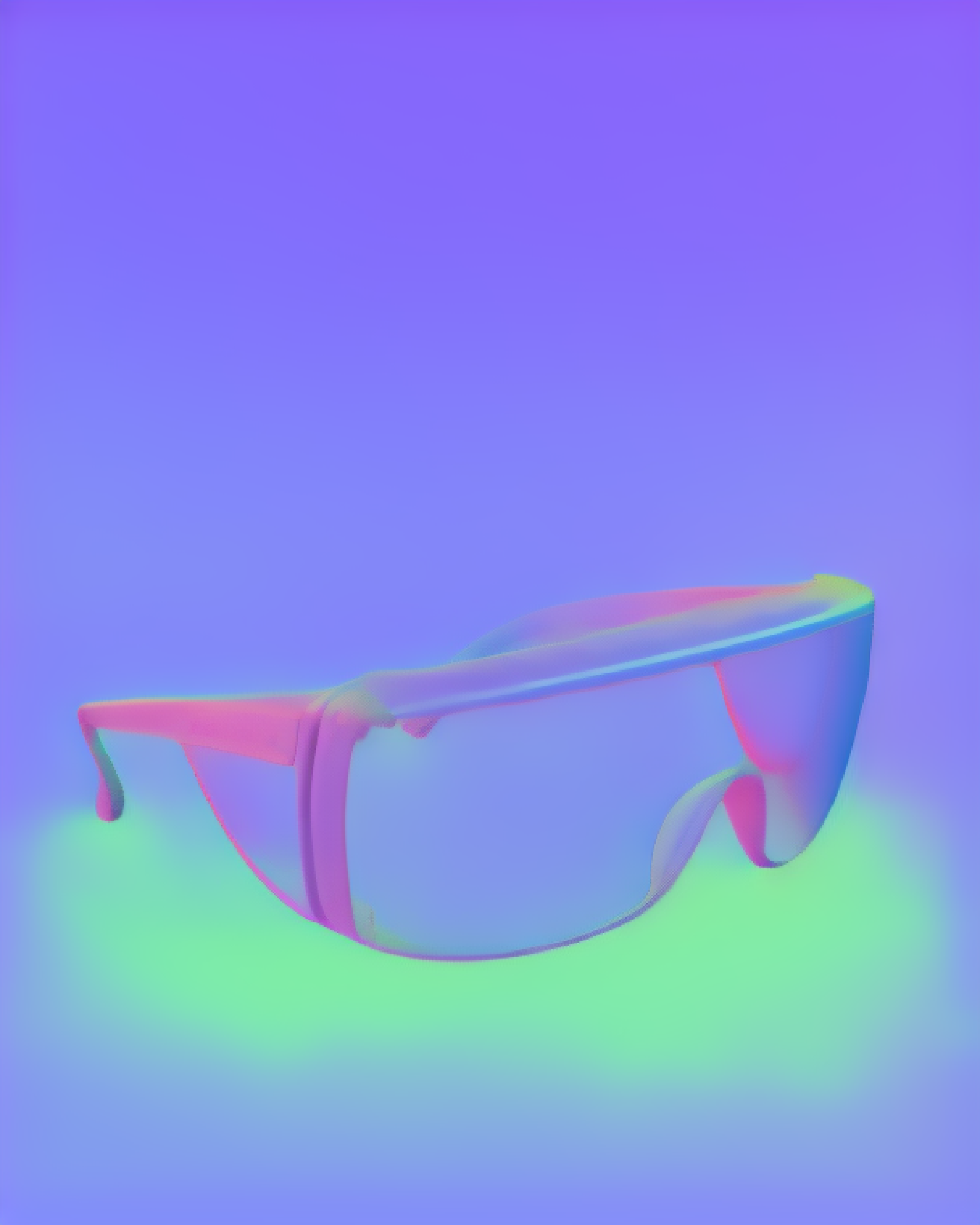} &
\includegraphics[width=0.24\columnwidth]{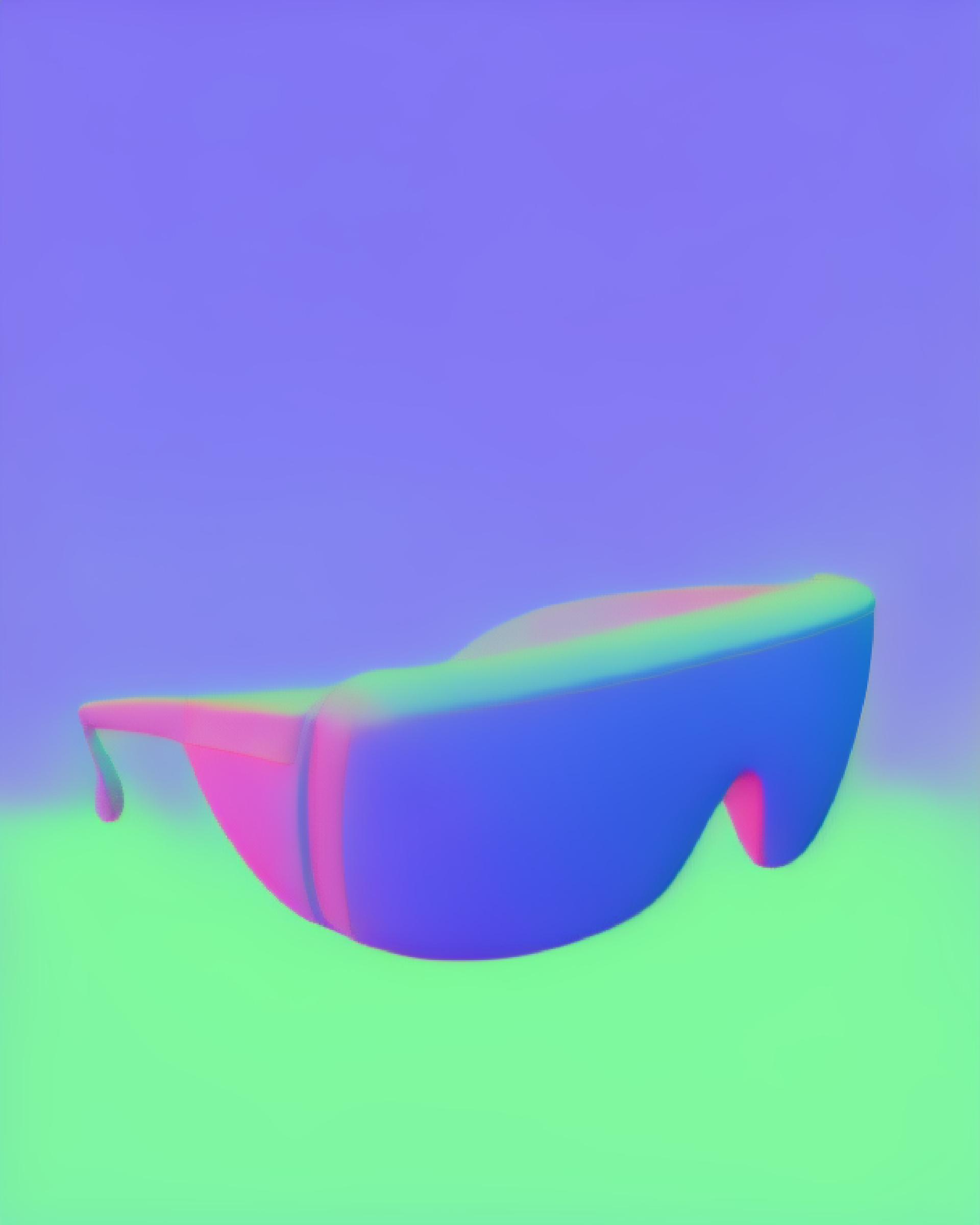} \\
\small (a) Input & \small (b) DINOv3 (PCA) & \small (c) Lotus-D & \small (d) Ours \\
\end{tabular}

\caption{\textbf{DINOv3 semantic features capture object-level geometry priors.}
(a) Input RGB image of transparent safety glasses exhibiting refraction and transmission;
(b) DINOv3 patch tokens visualized via PCA—semantic features cluster by object structure rather than local texture, encoding canonical shape priors that distinguish the eyewear from refracted background textures and transmission artifacts;
(c) Lotus-D struggles with transparent surfaces, producing noisy predictions affected by shadows and transmitted background imagery;
(d) Our method leverages DINOv3 semantics to correctly recover smooth surface geometry.
(\S~\ref{sec:cross_attention_analysis})}
\label{fig:cross_attention}
\end{figure}

A core claim of TransNormal is that DINOv3 semantic features help resolve the \emph{appearance-geometry decoupling} problem in transparent objects: refraction and transmission cause local RGB appearance to be dominated by background imagery rather than the object's intrinsic geometry. To validate this, we visualize the dense patch tokens extracted from DINOv3's final layer using Principal Component Analysis (PCA). The first three principal components are mapped to RGB channels, producing a colorized representation where similar colors indicate semantically similar regions. As shown in \Figref{fig:cross_attention}(b), DINOv3 features cluster by object structure---the eyewear forms coherent semantic groups distinct from the background---despite the transparent material causing the background to be visible through the lenses. This object-level semantic understanding enables our method to correctly infer surface geometry.

\subsection{Additional In-the-Wild Results}
\label{ssec:cross_category_zeroshot}

\begin{figure*}[t]
\centering
\setlength{\tabcolsep}{1pt}
\renewcommand{\arraystretch}{0.6}
\newcommand{\imgwzs}{0.24\textwidth}
%
%
\begin{tabular}{@{}cccc@{}}
    \small Input & \small \textbf{Ours} & \small Lotus-D~\citep{he2024lotus} & \small DSINE~\citep{bae2024dsine} \\[2pt]
    \includegraphics[width=\imgwzs]{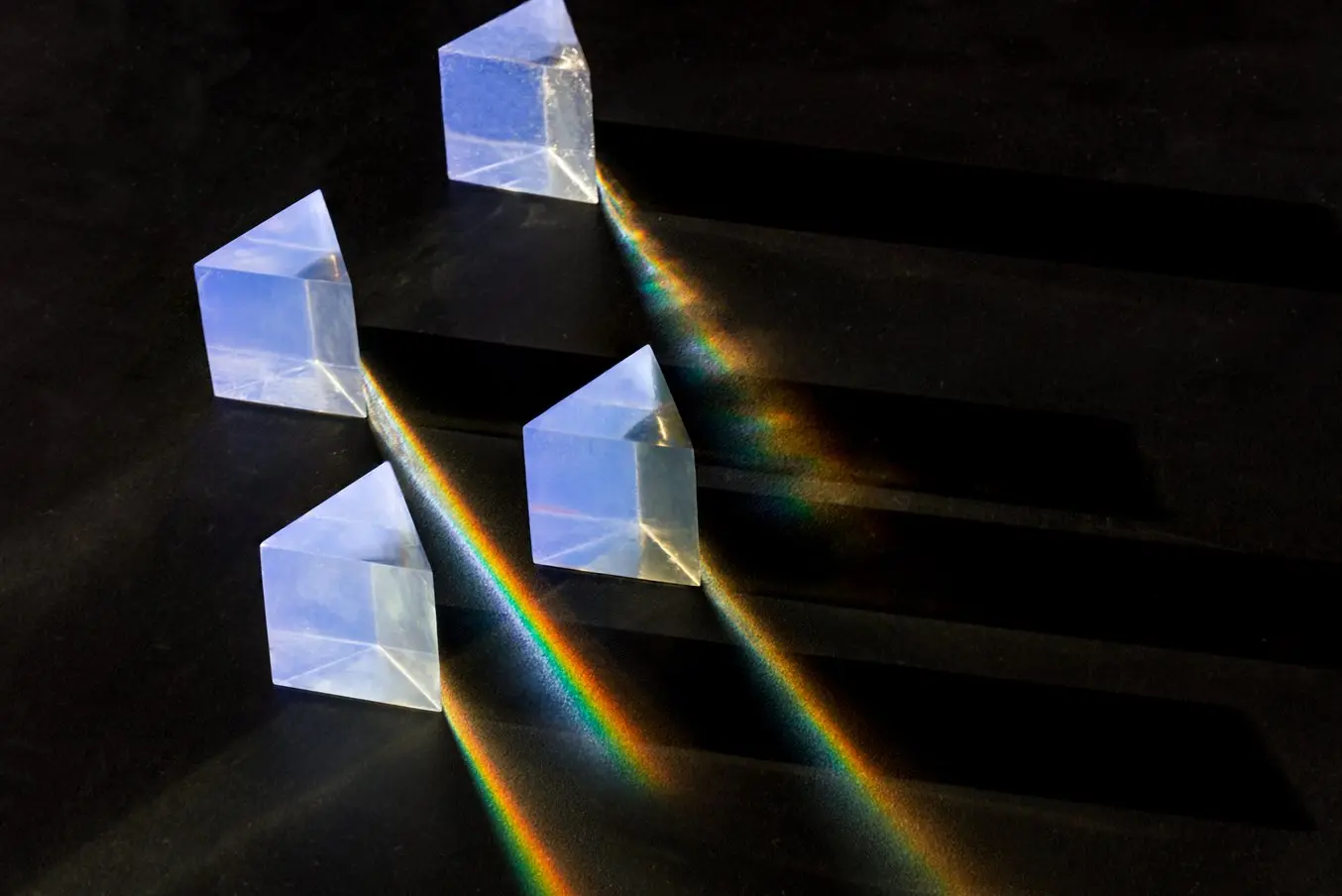} &
    \includegraphics[width=\imgwzs]{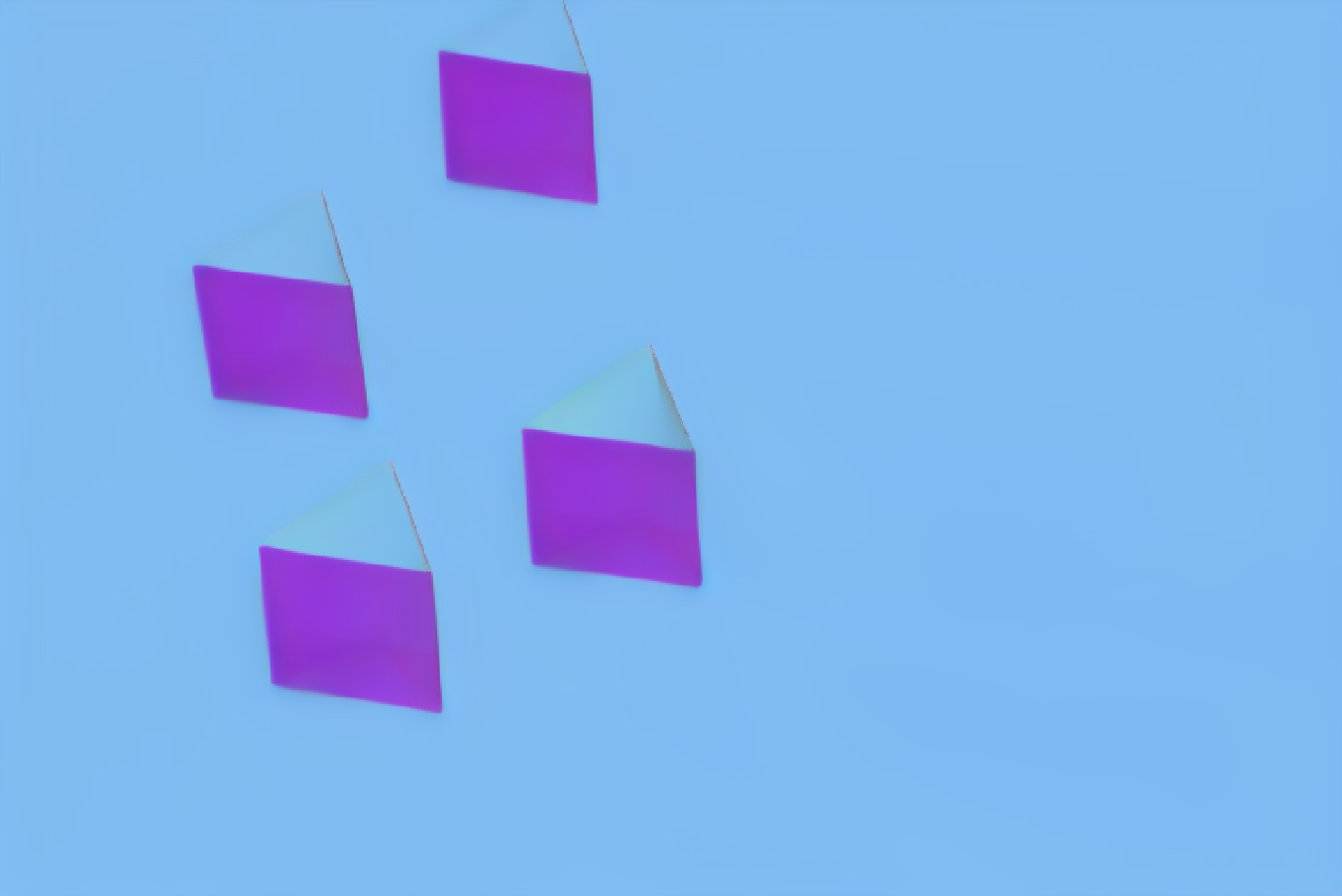} &
    \includegraphics[width=\imgwzs]{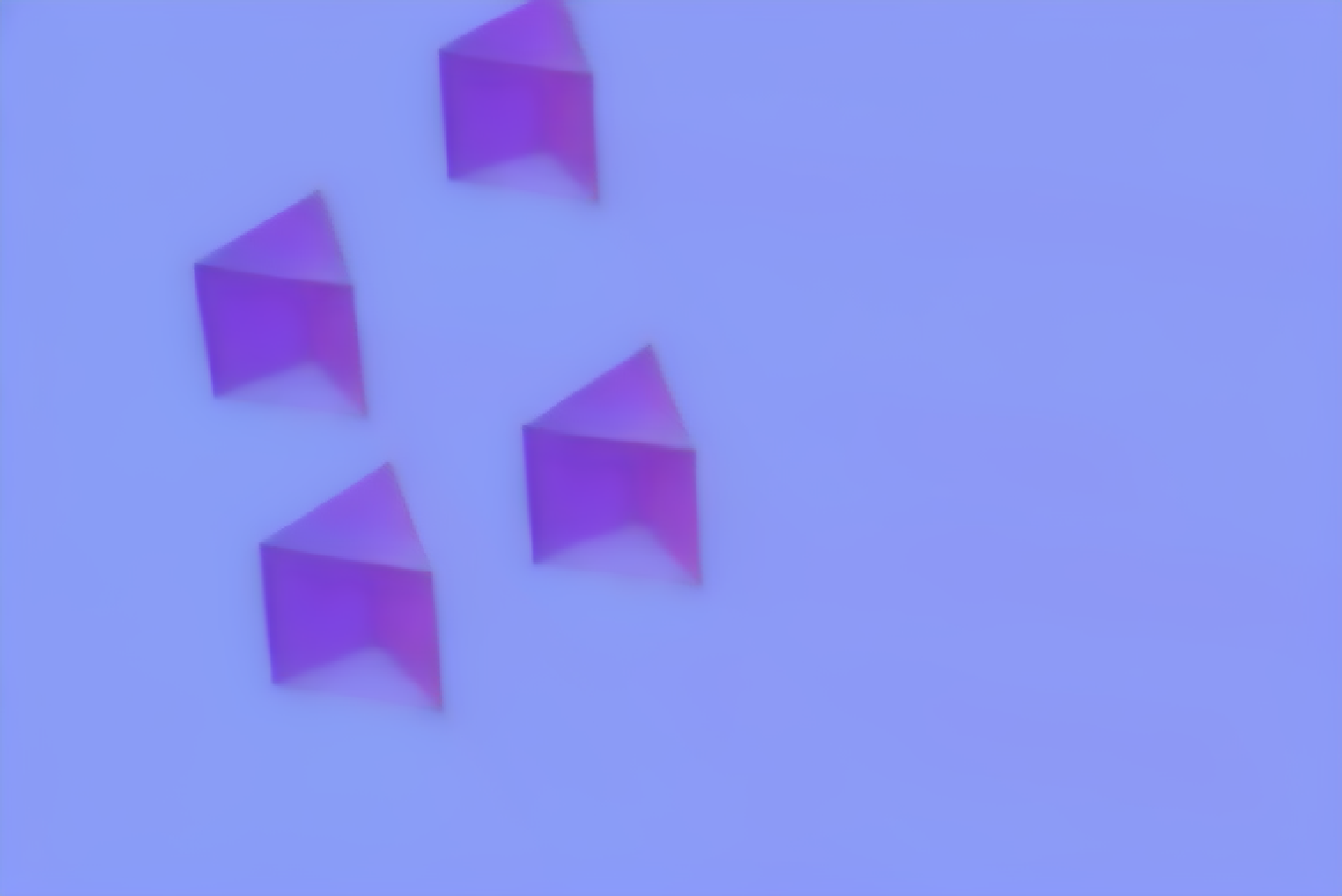} &
    \includegraphics[width=\imgwzs]{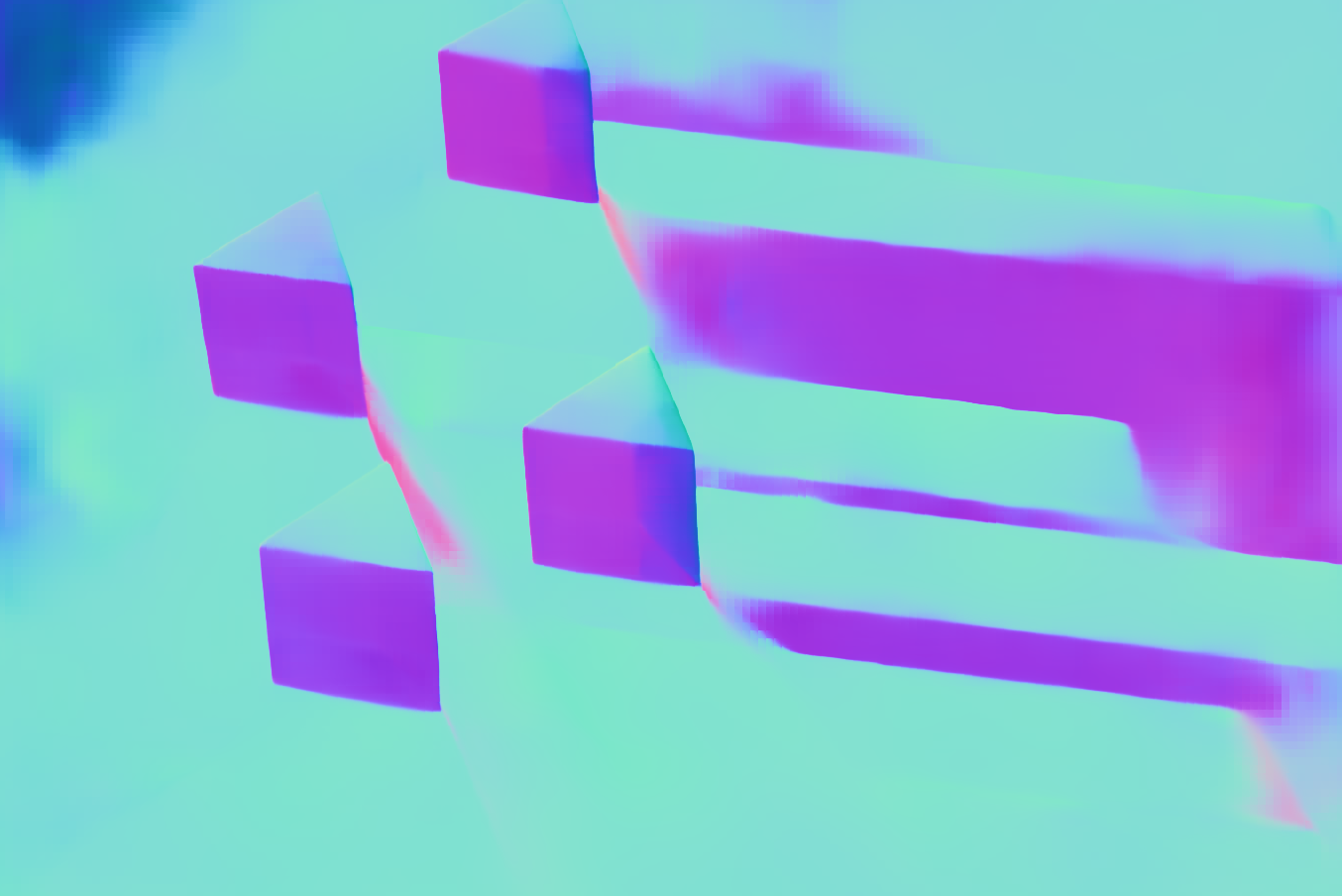} \\[2pt]
    %
    \small Diception~\citep{zhao2025diception} & \small GeoWizard~\citep{fu2024geowizard} & \small Marigold~\citep{ke2024repurposing} & \small MoGe-2~\citep{wang2025moge2} \\[2pt]
    \includegraphics[width=\imgwzs]{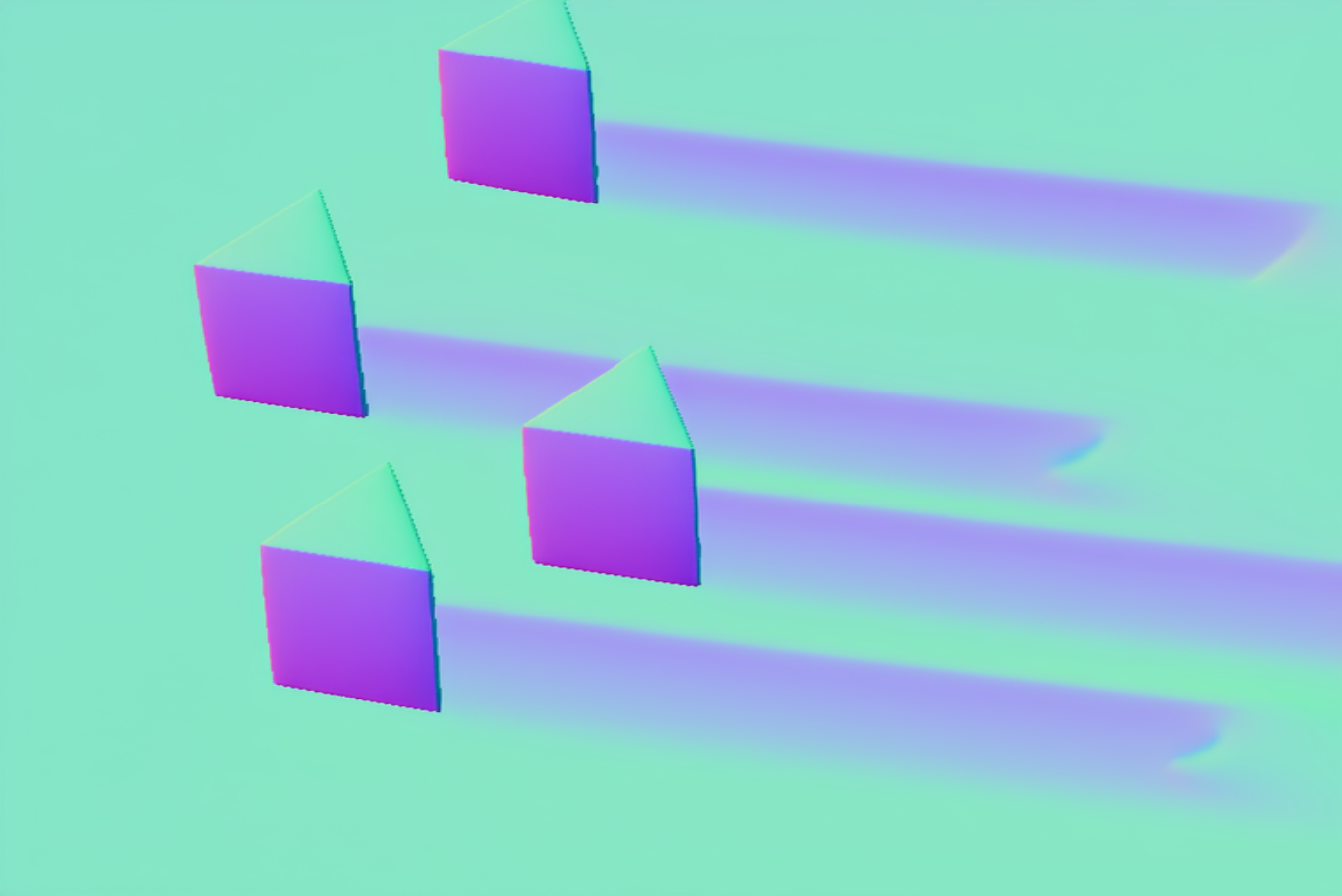} &
    \includegraphics[width=\imgwzs]{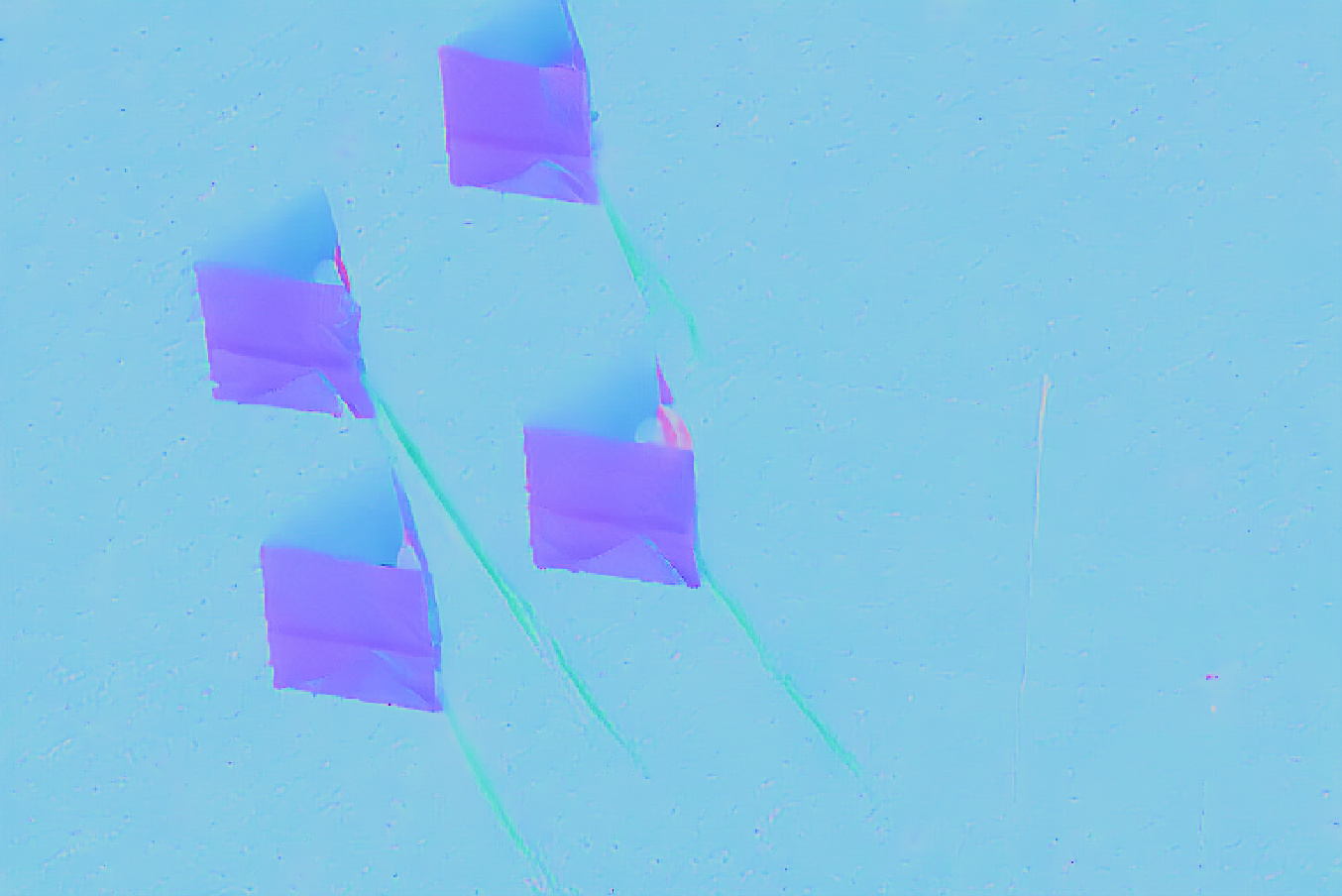} &
    \includegraphics[width=\imgwzs]{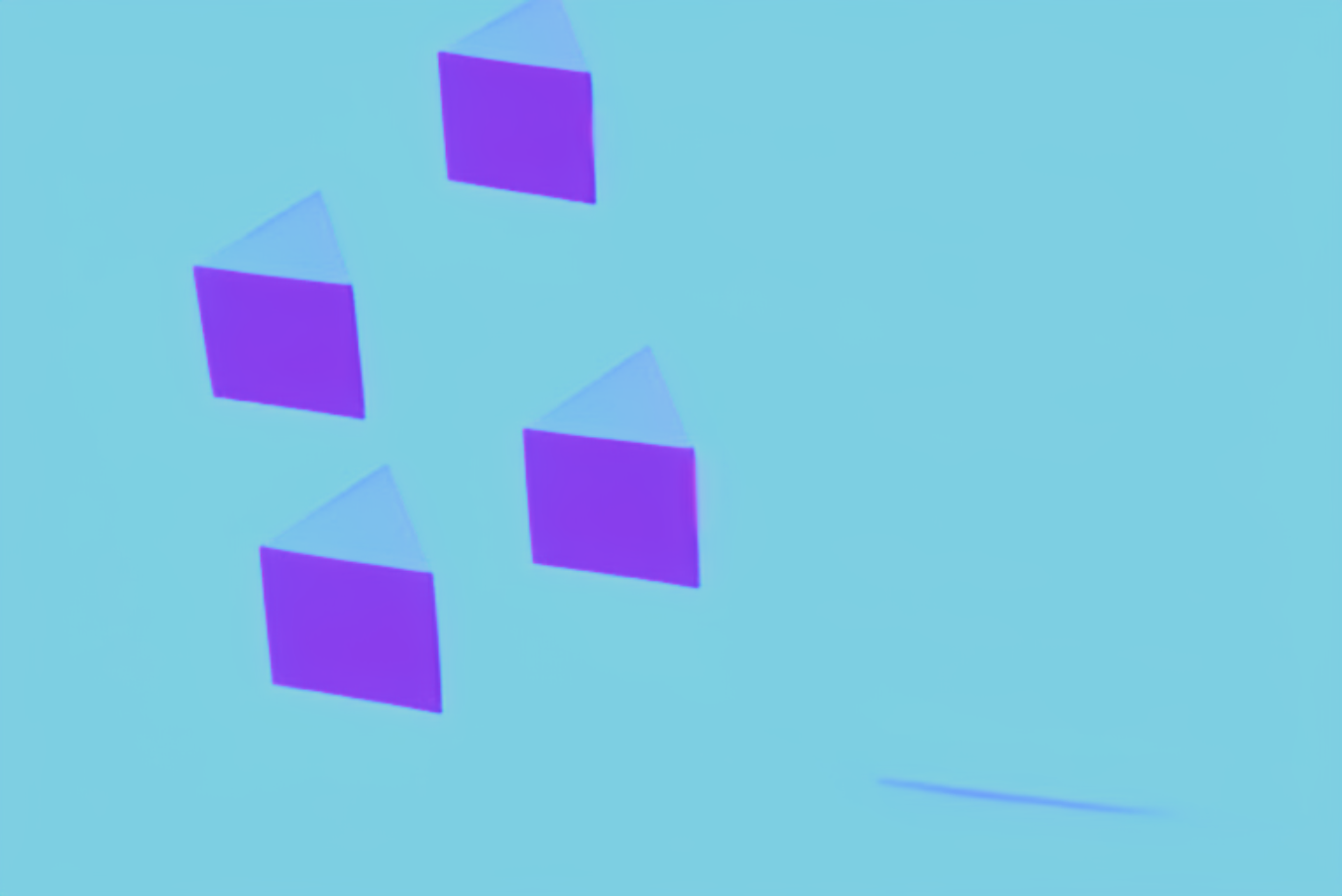} &
    \includegraphics[width=\imgwzs]{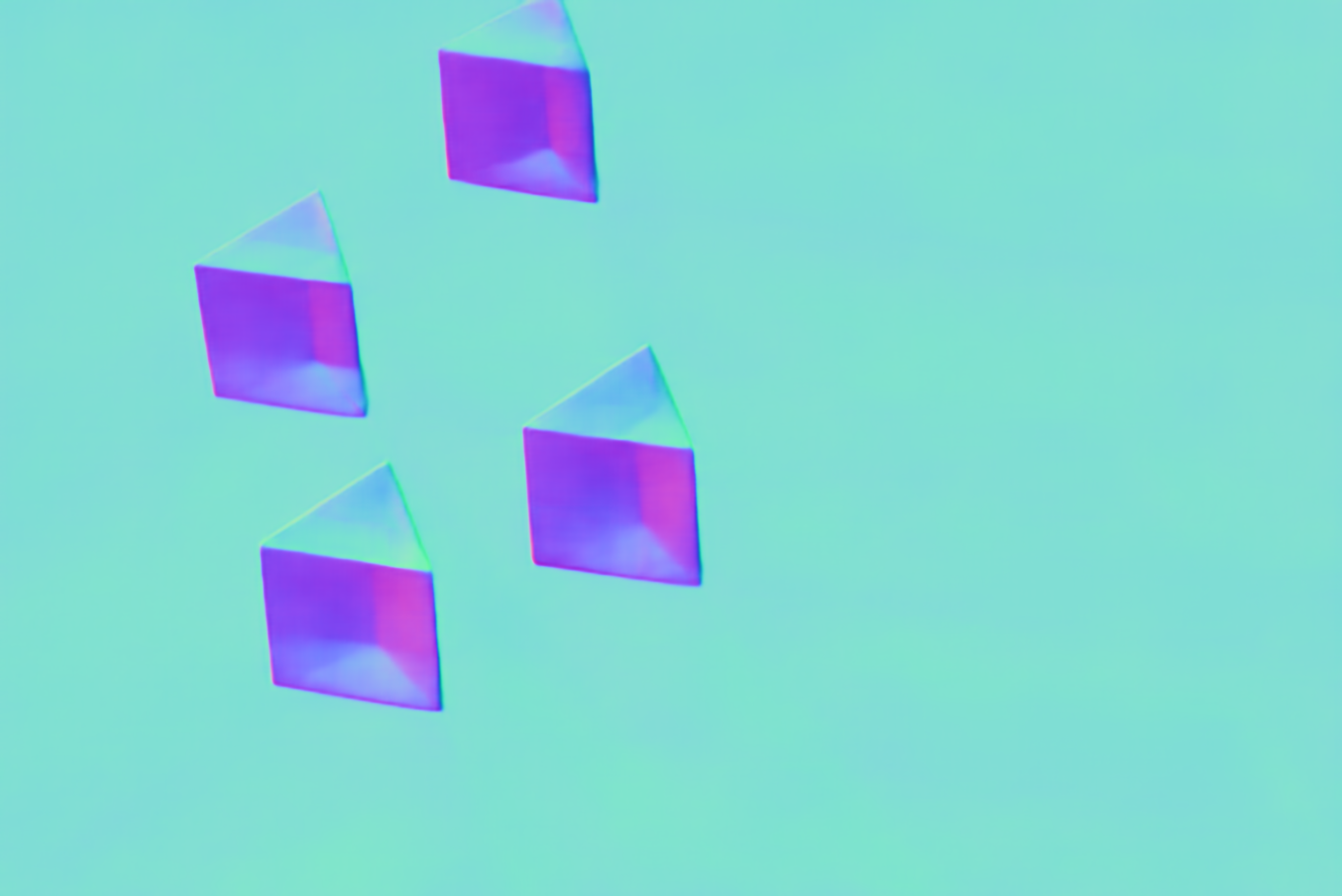} \\[8pt]
    %
    \\[8pt]
    %
    \small Input & \small \textbf{Ours} & \small Lotus-D~\citep{he2024lotus} & \small DSINE~\citep{bae2024dsine} \\[2pt]
    \includegraphics[width=\imgwzs]{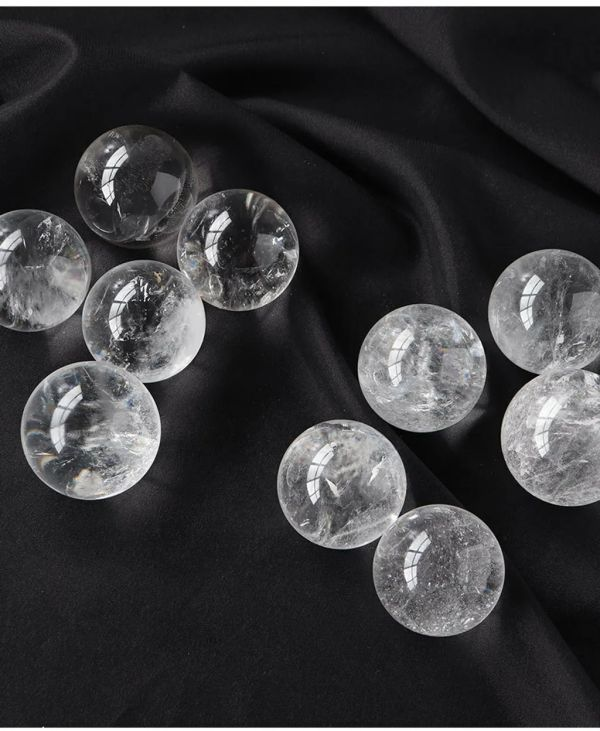} &
    \includegraphics[width=\imgwzs]{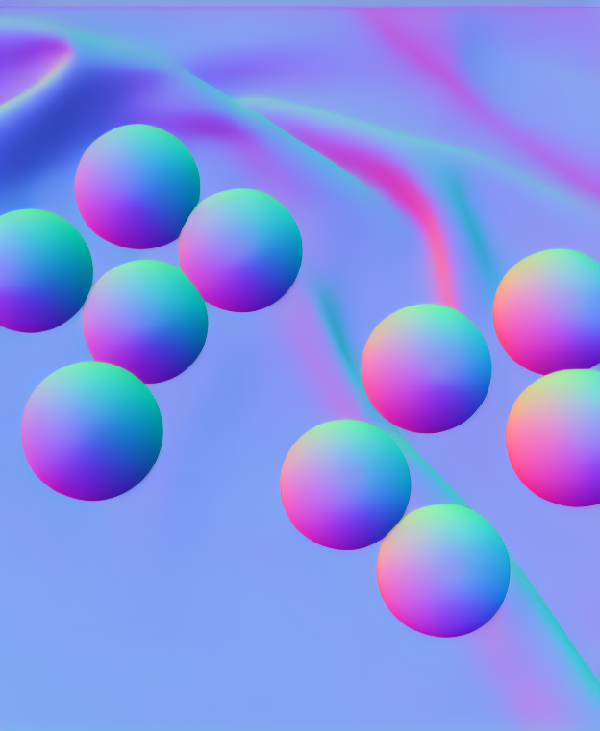} &
    \includegraphics[width=\imgwzs]{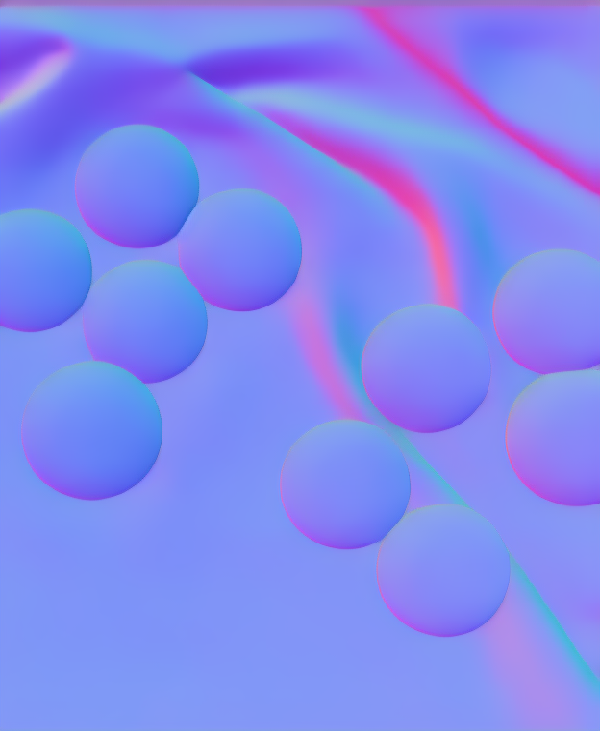} &
    \includegraphics[width=\imgwzs]{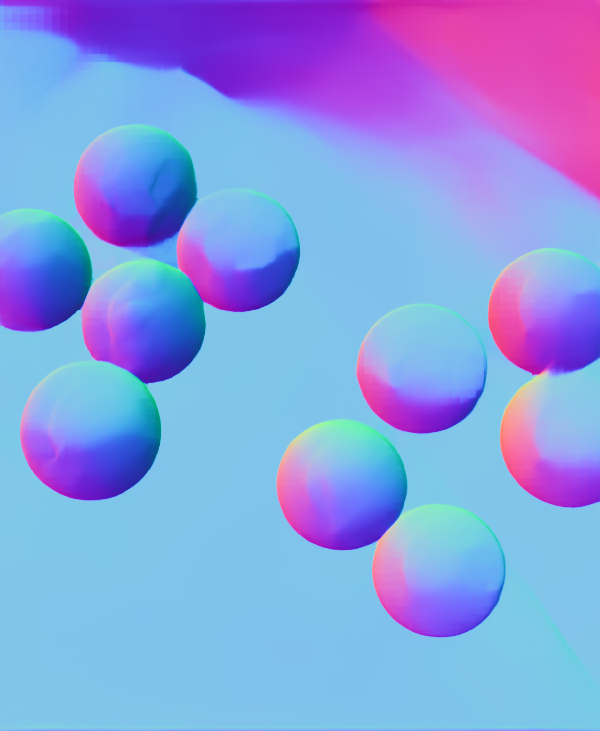} \\[2pt]
    %
    \small Diception~\citep{zhao2025diception} & \small GeoWizard~\citep{fu2024geowizard} & \small Marigold~\citep{ke2024repurposing} & \small MoGe-2~\citep{wang2025moge2} \\[2pt]
    \includegraphics[width=\imgwzs]{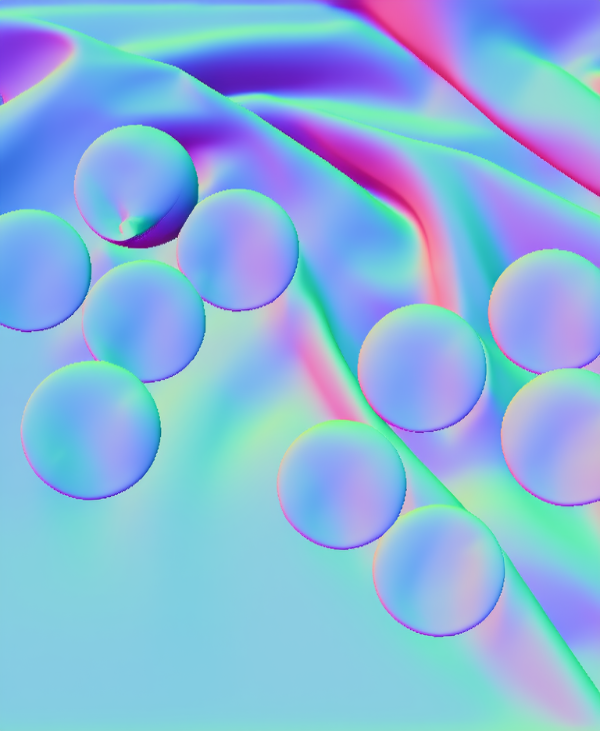} &
    \includegraphics[width=\imgwzs]{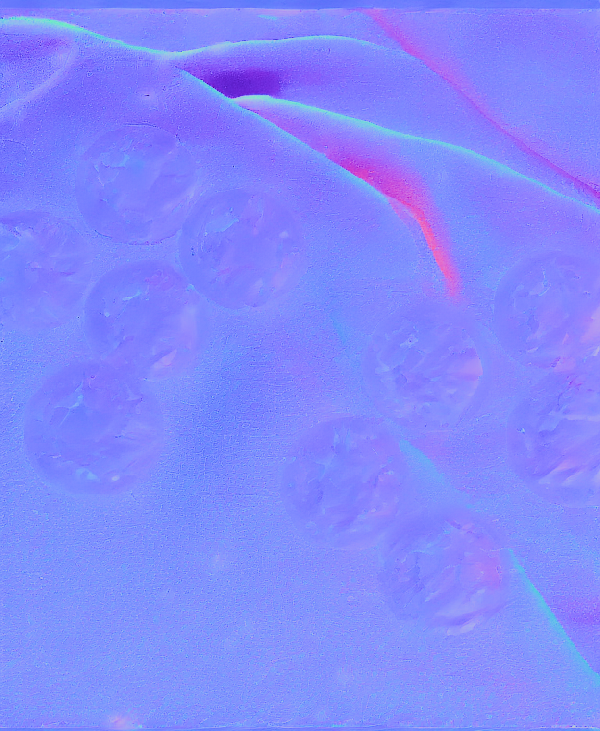} &
    \includegraphics[width=\imgwzs]{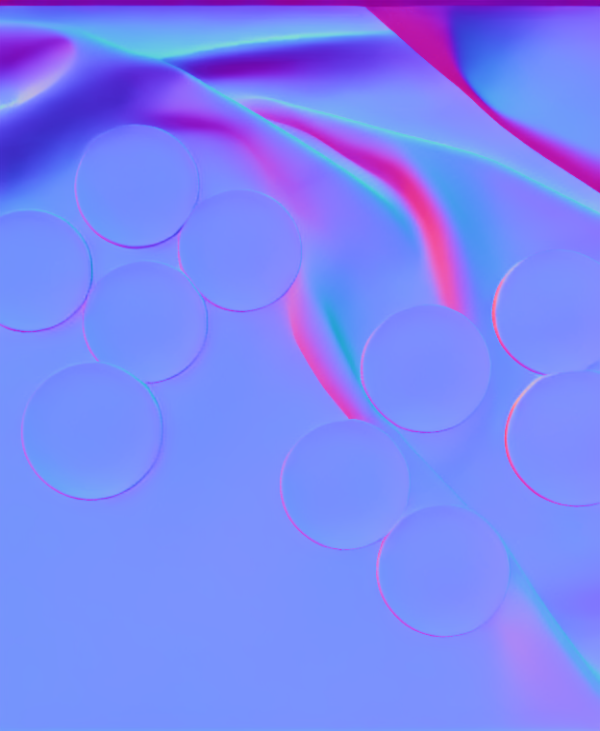} &
    \includegraphics[width=\imgwzs]{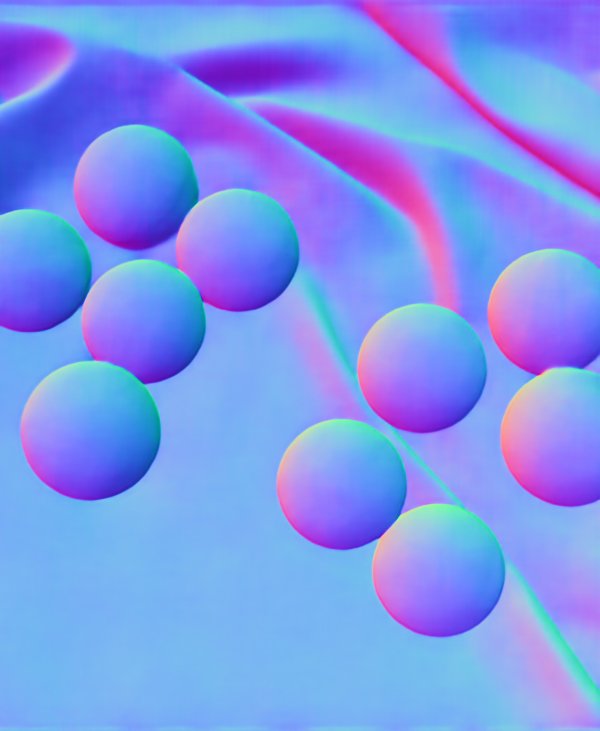} \\
\end{tabular}
\vspace{-1mm}
\caption{\textbf{Additional qualitative results on in-the-wild images.}
We evaluate TransNormal on in-the-wild transparent objects and compare with 6 baselines.
TransNormal produces more coherent surface normals on transparent regions, while baselines tend to be misled by refracted background textures or produce over-smoothed predictions.
(\S~\ref{ssec:cross_category_zeroshot})}
\vspace{-3mm}
\label{fig:cross_category}
\end{figure*}

To evaluate whether TransNormal generalizes beyond laboratory glassware, we conduct zero-shot inference on in-the-wild transparent objects. Since ground truth is unavailable for these images, we perform qualitative comparison against 6 baselines: Lotus-D, DSINE, Diception, GeoWizard, Marigold, and MoGe-2 (\Figref{fig:cross_category}).

\section{Limitations and Future Work}
\label{sec:limitations}

While TransNormal significantly advances transparent object normal estimation, several directions warrant further exploration:

\paragraph{Multi-view and Temporal Consistency.} Our current framework focuses on single-view estimation. Incorporating multi-view consistency constraints or temporal coherence for video sequences could further improve robustness and enable applications in dynamic manipulation scenarios.

\paragraph{Generalization to Other Dense Prediction Tasks.} The semantic-guided architecture demonstrates strong performance on normal estimation. Exploring its generalization to other dense prediction tasks such as depth estimation, optical flow, or material property prediction represents a promising avenue for future research.

\end{document}